\ifdefined\useorstyle
\documentclass{article}
\pdfoutput=1

\usepackage[numbers]{natbib} 
\usepackage{macros/nips_2024}

\usepackage{macros/statistics-macros}
\usepackage{macros/packages}
\usepackage{macros/editing-macros}
\usepackage{macros/formatting}
\usepackage{macros/statistics-macros}
\usepackage{caption}

\usepackage{endnotes}
\let\footnote=\endnote

\usepackage{xspace}

\usepackage{hyperref}

\usepackage{pgfplotstable}
\usepackage{graphicx}
\usepackage{float} 
\usepackage{tabularx}
\usepackage{overpic}
\usepackage{tikz}
\usepackage{rotating}
\usepackage{psfrag}
\usepackage{enumitem}
\usepackage{tikz}
\usetikzlibrary{shapes, arrows, positioning}

\title{A Planning Framework for Adaptive Labeling}
 
\date{\today}

\maketitle

\else

\documentclass[11pt]{article}
\usepackage[numbers]{natbib}
\usepackage{macros/packages}
\usepackage{macros/editing-macros}
\usepackage{macros/formatting}
\usepackage{macros/statistics-macros}

\begin{document}

\abovedisplayskip=8pt plus0pt minus3pt
\belowdisplayskip=8pt plus0pt minus3pt


\begin{center}
  {\huge A Planning Framework for Adaptive Labeling\footnote{
  A conference version of this work appeared at 
2024 Conference on Neural Information Processing Systems,
titled 
``Adaptive Labeling for Efficient Out-of-distribution Model Evaluation''.}} \\
  \vspace{.5cm}
  {\Large Daksh Mittal$^{*1}$~~  Yuanzhe Ma$^{*2}$ ~~
  Shalmali Joshi$^{3}$ ~~ Hongseok Namkoong$^{1}$
  } \\
{ Decision, Risk, and Operations Division$^{1}$,
  Department of Industrial Engineering and Operations Research$^{2}$,
  Department of Biomedical Informatics$^{3}$ \\
  Columbia University
  } \\ 
\end{center}


\fi

\def\thefootnote{*}\footnotetext{Equal contribution}

\begin{abstract}

Ground truth labels/outcomes are critical for advancing scientific and engineering applications, e.g., evaluating 
the treatment effect of an intervention or performance of a predictive model. Since randomly sampling inputs for labeling can be prohibitively expensive, we introduce an adaptive labeling framework where measurement effort can be reallocated in batches. 
We formulate this problem as a Markov decision process where
posterior beliefs evolve over time as batches of labels are collected (state transition), 
and batches (actions) are chosen to minimize uncertainty 
at the end of data collection.
We design a computational framework that is agnostic to different uncertainty quantification
approaches including those based on deep learning, and
allows a diverse array of policy gradient approaches by 
relying on continuous policy 
parameterizations.
On real and synthetic datasets, we demonstrate even a one-step lookahead policy can substantially outperform common adaptive labeling heuristics, highlighting the
virtue of planning. 
On the methodological side,
we note that standard $\mathsf{REINFORCE}$-style policy gradient estimators 
can suffer high variance since they rely only on 
zeroth order information. 
We propose a direct backpropagation-based approach, $\mathsf{Smoothed\text{-}Autodiff}$,
based on a carefully smoothed version of the original non-differentiable MDP. Our method enjoys low variance at the price of introducing bias,
and we theoretically and empirically show that this trade-off can be favorable.

\end{abstract}

 
\section{Introduction}
\label{sec:introduction}

Rigorous measurement is the foundation of scientific progress;
ground truth labels/outcomes $Y$ are essential in many applications, such as evaluating the performance of an AI model or estimating the average treatment effect (ATE) of an intervention. 
Since acquiring labels/outcomes is costly, randomly selecting  units $X$ to label can make data collection prohibitively expensive. 
Adaptive labeling can significantly improve the efficiency of data collection,
allowing measurements on more diverse units
and help assess robustness issues (tail risks) in prediction models and impact of interventions on 
underserved communities.

We illustrate this using two representative applications:

\begin{example}[Evaluating predictive accuracy under severe selection bias]
When ground truth labels are expensive, available datasets often suffer selection bias. For example, AI-based medical diagnosis systems are typically evaluated using past clinical labels, which are only available for patients who were initially screened by doctors~\citep{Institute03, Seyyed-KalantariZhMcChGh21,StrawWu22,MullainathanOb22,BalachandarGaPi24}.
Naive offline evaluation using existing data leads to unreliable assessment due to the mismatch between the overall population vs. those for whom labels are available, i.e., we cannot assess predictive accuracy on  patients who were never screened in the first place.  When there is overlap between 
the labeled and overall population, 
reweighting (importance sampling) can offer some reprieve. However, such overlap often fails to hold in many applications. Furthermore, 
due to the absence of positive pathological patterns in the available data, previously unseen patient types may never get diagnosed by the AI diagnosis system. Evaluating a model's performance on previously unscreened patients is thus a critical first step toward mitigating this vicious cycle, and can spur engineering innovations that improve the out-of-distribution robustness of AI models. 
\end{example}
    
\begin{figure}[t]
\centering
\includegraphics[scale=0.4]{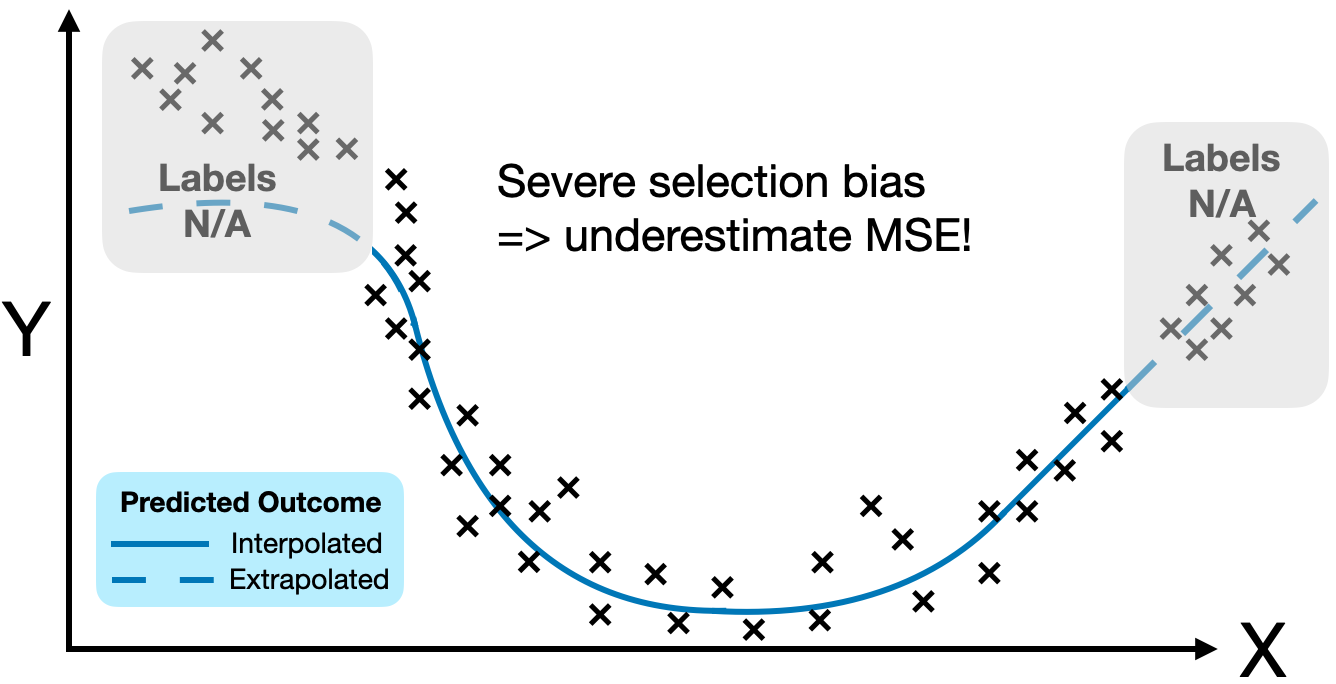}
\caption{\textbf{Adaptive labeling to reduce epistemic uncertainty over
model performance.}  Among the two clusters of unlabeled examples (left
vs. right), we must learn to prioritize labeling inputs from the left
cluster to better evaluate mean squared error.}
\label{fig:toy_illustration}
\end{figure}

\begin{example}[Estimating average treatment effects (ATE)]
A common goal in natural and social sciences is to estimate the average treatment effect of a treatment $Z=1$ 
compared to a control $Z=0$ across a population characterized by features $X$. When there is minimal overlap between the subset of the population for which observations are available and the population on which the treatment effect is to be estimated, an \emph{active} measurement effort becomes necessary to assess the ATE. The objective in such cases is to sequentially select batches from the population for experimentation.
\end{example}

In this work, we study adaptive labeling as a way to address severe distribution shifts in $X$---between screened vs. unscreend patients or treated vs. control groups. 
Our key contribution lies in the  formulation of a \emph{planning problem} 
for sequentially selecting batches of inputs/population from a large pool
(Section~\ref{sec:formulation}). 
We view rigorous empirical evaluation 
as a first step in 
designing grounded adaptive labeling policies and restrict attention to estimation problems in this paper.  The more ambitious goal of improving AI robustness or personalized treatment policies remains an open problem. 
Recognizing practical constraints such as engineering/organizational overhead or delays in receiving feedback, we consider a few-horizon setting where at
each period, a \emph{batch} of inputs is selected for labeling. 
While a fully online scenario has its conceptual appeal, we believe our few-horizon setting is of particular practical interest since fully online policies are often infeasible to implement even in large online platforms with mature engineering infrastructure~\citep{WuEtAl19, NamkoongDaBa20, AvadhanulaEtAl22}.

Effective labeling requires prioritizing measurements in regions with high
uncertainty that can be reduced with additional samples. In line with the literature, we refer to such uncertainty as  \emph{epistemic} (actionable). In contrast, some uncertainty is \emph{aleatoric} (irreducible), e.g., idiosyncratic measurement noise. 
We develop adaptive policies that
incorporate how beliefs on the estimand of interest (model performance or ATE) get sharper as more labels/outcomes are collected.   We model our current belief on the data generation process $Y|X$ (or $Y|X,Z$ in causal estimation settings) as the current state. After observing a new batch of data and it's corresponding outcomes/labels, we update
our belief on the epistemic uncertainty via a posterior update (``state
transition''). 
Our goal is to minimize the uncertainty over 
the estimand of interest
at the end of label collection, 
as measured by the variance under the
posterior after the final batch is observed. 

\begin{figure}[t]
\vspace{-1.5cm}
\centering \includegraphics[width=\textwidth, height=5.5cm]{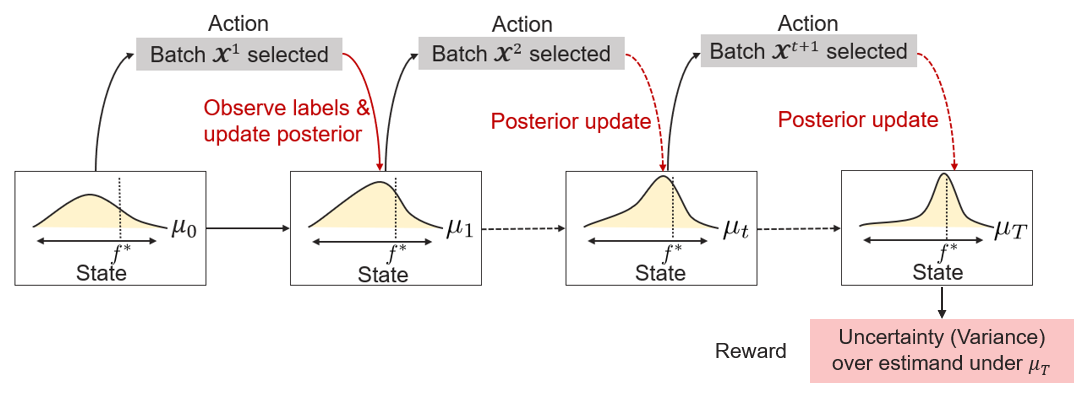}
\caption{\textbf{Overview of our adaptive sampling framework.}  At each
period, we select batch of inputs $\mathcal{X}^t$ to be labeled, and obtain a new
labeled data $\mc{D}_t$.  We view posterior beliefs $\mu_t(\cdot)$ on
$f\opt(Y|X)$ (or $f\opt(Y|X,Z)$) as the ``state'', and update it as additional labeled data is
collected. Our goal is to minimize uncertainty on the estimand of interest (performance of predictive model or ATE) at the end of $T$ periods.
}
\label{fig:overview}
\vspace{-.5cm}
\end{figure}

Modeling each batch as a time
period, we arrive at a Markov Decision Process (MDP) where states are posterior
beliefs, and actions correspond to 
selecting subsets (batches) from the pool of inputs/population (Figure~\ref{fig:overview}).
 Solving our planning problem 
is computationally challenging due to the
combinatorial action space and continuous state space. To address this, we approximately solve the problem by performing policy gradient updates over ``roll-outs" to minimize the uncertainty over the estimand of interest. Specifically, since the dynamics
of the MDP (posterior updates) are known, we execute ``roll-outs'' as follows: using the
current posterior beliefs we simulate potential outcomes that we may obtain if we label a particular batch,  and evaluate the updated uncertainty on 
the estimand
based on the imagined pseudo labels of this new batch.
We use continuously parameterized policies to select the batch of inputs~\citep{EfraimidisSp06} and then optimize it using policy gradients.

There are various uncertainty quantification (UQ) methodologies in the literature for estimating posteriors.   
Our approach is agnostic to the choice of  UQ methodology, and allows incorporating latest advances in deep 
learning-based UQ methods
(e.g.,~\citep{DuLiXuSpWuRuMa20, OsbandWenAsDwIbLuRo23}). While
Gaussian processes are effective in low-dimensional settings, quantifying epistemic uncertainty for high-dimensional inputs remains an active area of research, including
popular techniques such as Dropout~\citep{GalGh16}, Bayes by Backprop~\citep{Blundell15}, Ensemble+~\citep{LakshminarayananPrBl17,OsbandAsCa18}, and Epistemic Neural Networks~\citep{OsbandWenAsDwIbLuRo23}.
We stress that we do not require the posterior
updates to have a closed form (conjugacy),
and allow approximate posterior
inference using gradient-based optimizers; making it compatible with deep learning-based UQ models (e.g., ensembles) 
that rely on gradient-based posterior updates.

 Our framework supports the integration of various policy gradient methods. The well-known \textsf{REINFORCE} estimator  computes policy gradients using the ``score trick''~\citep{SuttonBa18, DuLiXuSpWuRuMa20}, relying solely on function evaluations.  However, it often suffers from high variance
and can be computationally expensive in practice. To address these limitations, we propose a computationally efficient alternative, $\mathsf{Smoothed\text{-}Autodiff}$. Direct computation of pathwise policy gradients is infeasible because the simulated
``roll-out'' of the adaptive labeling planning problem is not differentiable, owing to the combinatorial nature of the action space.  Differentiability is achieved only after smoothing the sample trajectories using the approximation we introduce.
Leveraging smoothly parameterized policies~\citep{XieEr19},
 we design a smooth approximation of the overall MDP, enabling direct backpropagation—--a method we call $\mathsf{Smoothed\text{-}Autodiff}$.
 We demonstrate our idea with an end-to-end differentiable simulator where we implement all the operations of our planning problem in an \emph{auto-differentiable} framework (Section~\ref{sec:diff-piepline}), including state transitions (posterior updates). Our method allows us to backpropagate the
observed reward to the policy parameters and calculate the \emph{smoothed-pathwise} policy gradient. 

We demonstrate our planning
framework over beliefs formed by Gaussian processes, and Deep Ensembles~\citep{LakshminarayananPrBl17,OsbandAsCa18}. Deep Ensembles leverage the inductive biases of neural
networks in high dimensions
and gradient descent methods to update probabilistic beliefs over
$Y|X$. 
Empirically, we observe that even a single planning step can yield
significant practical benefits (Section~\ref{sec:experiment}). 
On both real and
simulated datasets with severe selection bias, we demonstrate that one-step
lookahead policies based on our planning framework achieve
state-of-the-art performance outperforming
active learning-based heuristics. 

We further observe that $\mathsf{Smoothed\text{-}Autodiff}$
can often outperform \textsf{REINFORCE} policy gradients; 
although our original problem is nonsmooth and discrete,
auto-differentiating over the smoothed MDP can provide reliable policy optimization methods. 
Empirically, $\mathsf{Smoothed\text{-}Autodiff}$ requires 100-1000 times fewer samples to achieve the same level of estimation accuracy as the \textsf{REINFORCE} estimator. We observe the statistical efficiency gain translates to better downstream task performance: even using $\mathsf{Smoothed\text{-}Autodiff}$ over 
a \emph{single} sample trajectory leads to policies that outperform those optimized using the standard \textsf{REINFORCE} estimator.
To better understand this empirical phenomenon, we offer theoretical insights into when the $\mathsf{Smoothed\text{-}Autodiff}$ gradients might outperform \textsf{REINFORCE}-based gradients (Section \ref{sec:gradient_estimation_anlaysis}). In particular, our analysis sheds light on the steep bias-variance trade-off and illustrates when it may be favorable to take a small bias to significantly reduce variance.

Finally, our thorough empirical evaluation highlights several open
research directions   to help scale our framework (Section~\ref{sec:practical_consideration}). 
First, although recent
advances in Bayesian modeling have achieved substantial progress in one-shot
uncertainty quantification, we observe it is often difficult
to maintain accurate beliefs as more data are collected (Section~\ref{sec:eval_posterior_consis}).
Second, we detail
engineering challenges in implementing our auto-differentiation framework over
multi-period roll-outs, highlighting the need for efficient implementations of
high-order auto-differentiation (Section~\ref{sec:auto-diff-bottleneck}).


\section{Related work}
\label{sec:related-work}

\paragraph{Active Learning} Adaptive labeling is a classical topic in machine learning.
When applied to model evaluation, our framework posits a less  ambitious goal than the active learning literature~\citep{AggarwalKoGuHaPh14,Settles09} where labels are adaptively collected  to \emph{improve} model performance in classification problems. Typical active learning heuristics are greedy and select inputs based on uncertainty sampling techniques such as margin-sampling or entropy, or by leveraging disagreements among different models (e.g., query-by-committee and Bayesian Active Learning by Disagreement (BALD)~\citep{HoulsbyHuGhLe11}). 
Batching is a longstanding challenge for these approaches as it requires collecting labels over diverse inputs~\citep{Settles09}. Heuristic criteria based on diversity and density are sometimes incorporated to account for the feature space distribution~\citep{Settles09},  and recent extensions continue to rely on greedy algorithms (e.g., BatchBALD~\citep{KirschVaGa19}, BatchMIG~\citep{WangSuGr21}).

We leverage our limited scope on the efficient estimation of model performance (or ATE), rather than model improvement, 
to formalize a unified and novel computational framework that allows planning for the future.
Our formulation can flexibly handle different objectives and uncertainty quantification methodologies, and seamlessly handle regression and classification problems alike, in contrast to the traditional focus on classification in active learning.
We provide lookahead policies~\citep{BertsekasTs96,EfroniDaScMa18, EfroniGhMa20} that plan for the future, and show they dominate active learning heuristics in Section~\ref{sec:experiment}.

\paragraph{Simulation Optimization}
Bayesian optimization considers black-box functions that are computationally challenging to 
evaluate by modeling the function as a draw from a Gaussian process~\citep{Frazier18}. 
The knowledge gradient algorithm~\citep{FrazierPoDa08}
maximizes the single period expected increase in the black-box function value and can be viewed as a one-step lookahead policy; several authors propose extensions to non-myopic and cost-aware problems~\citep{AstudilloJiBaBaFr21}.
We extend these ideas to adaptive labeling  by incorporating batching and combinatorial action spaces, and accommodate more advanced uncertainty quantification methods from deep learning.

Our work is closely related to the  simulation optimization literature~\citep{AmaranSaShBu16},
 particularly to ranking and selection~\citep{ChenLiYuCh00,GlynnJu04, KimNe07, ChenChLePu15, HongNeXu15}.
Our adaptive labeling problem can be viewed as a ranking \& selection or pure-exploration bandit problem. However, in contrast to conventional formulations, our central focus on batching
introduces combinatorial action spaces.
Additionally, our formulation involves continuous covariates and a large number of alternatives,   making it  intractable for standard methods, whose performance deteriorates as the number of alternatives grow larger. 

Instead, we parameterize our discrete optimization problem through a continuous parametrization policy and use gradient-based methods to optimize it. We are able to exploit the similarity between the alternatives through this approximation, which is in contrast to the  traditional simulation optimization literature that mostly considers independent alternatives. 

\paragraph{Gradient Estimation} 

There is extensive literature on gradient estimation~\citep{Glynn87, Glasserman90, Glasserman92, FuHu12}, with applications in probabilistic modeling~\citep{KingmaWe14,JangGuPo17} and reinforcement learning~\citep{Williams92,SuttonMcSiMa99}. 
The \textsf{REINFORCE} approach uses the score-function estimator~\citep{Williams92,SuttonMcSiMa99}
$\nabla_\theta \E_{X \sim \pi_\theta}[G(X)] = \E_{X \sim \pi_\theta}[G(X) \nabla_\theta\log \pi_\theta (X)]$. While unbiased, these estimators suffer high variance~\citep{GreensmithBaBa04} and require many tricks to be effective for downstream applications~\citep{HuangDoRaKaWa22}.
Similarly, finite difference approximations~\citep{FuHi97}  and finite perturbation analysis~\citep{Cao87,HoCaCh89} also only require zero-th order function evalautions, but perform poorly in high-dimensional problems~\citep{Glynn89, Glasserman04}.

Alternatively, the reparameterization trick~\citep{MaddisonMnTe17,JangGuPo17, PaulusMaKr21} 
takes a random variable $Z$ whose distribution does not depend on the parameter of interest $\theta$ and 
exploits the identity
$\nabla_\theta \E_{X \sim \pi_\theta}[G(X)]
= \nabla_\theta \E_{Z}[G(h(Z, \theta))]$ for some function $h$.
 Gradient estimators based on the reparameterization trick 
 typically leads to much smaller
variance~\citep{MohamedRoFiMn20}, 
but can only be applied in special cases where $G(h(Z, \theta))$ is differentiable with respect to $\theta$.  
Infinitesimal Perturbation Analysis (IPA)~\citep{HoEyCh83, JohnsonJa89,Glasserman90} 
is a standard framework for constructing such pathwise gradient estimators, but conditions ensuring IPA to be valid~\citep{Cao85,HeidelbergerCaZaMiSu88,Glasserman90, Glasserman92} 
are often restrictive  and does not apply in our setting. 
In fact, in our adaptive labeling problem, the function $G(h(Z, \theta))$ of interest
is non-differentiable and only becomes differentiable in expectation.
To address this challenge, we construct a smooth approximation of the function $G(h(\cdot))$, enabling the application of pathwise gradient estimators by leveraging the known dynamics (posterior updates) of our MDP.

We are inspired  by a line of work on differentiable 
simulators~\citep{deAvilaFiSmAlTeKl18,HuangHuDuZhSuTeGa21,MoraAnHaVeCo21, XuMaNaRaMaGaMa21,SuhSiZhTe22} with applications in reinforcement learning~\citep{MoraAnHaVeCo21, XuMaNaRaMaGaMa21,MadekaToEiLuFoKa22, AlvoRuKa23}.
Theoretical analysis in stochastic optimization~\citep{GhadimiLa13, MohamedRoFiMn20} emphasizes the benefit of first-order estimators over
zeroth-order counterparts, and~\citet{SuhSiZhTe22} notes first-order estimators
perform well when the objective is sufficiently smooth and continuous. 
In our setting, although the simulation path is not differentiable, we introduce a smooth approximation of the path, enabling the application of first-order gradient estimates.
Our approach provides low-variance gradient estimates at the cost of bias, and  offers a novel way to balance bias and variance for non-differentiable functions~\citep{BengioLeCo13, JangGuPo17, TuckerMnMaLaSo17}.
Biased gradient estimates have not been studied much in the simulation optimization literature, 
aside from two notable exceptions:~\citet{EckmanHe20} show the utility of biased estimators when in the neighborhood of a local minimizer, and~\citet{CheDoNa24} study queueing network control and propose a smoothing scheme for computing biased policy gradient estimators. Focusing on adaptive data collection, we propose a tailored smoothing approach for  handling combinatorial action spaces and (approximate) posterior state transitions. 

\section{Adaptive labeling as a Markov decision process} 
\label{sec:formulation}

We illustrate our formulation for model evaluation, and extend it to the ATE estimation setting at the end of the section. 
Our goal is to evaluate the performance of a prediction model $\model: \statdomain \to \mathcal \labeldomain$ over the input distribution $P_X$ that we expect to see during deployment.  Given inputs $X  \in \mc{X}$,   labels/outcomes are generated
 from some unknown function $f\opt$: $
      Y = f\opt(X) + \varepsilon$, where $\varepsilon$ is the noise.
When ground truth outcomes are costly to obtain, previously collected labeled data $\mc{D}^0 := \{(X_i,Y_i)\}_{i \in \mc{I}}$ 
typically suffers selection bias and covers only a subset of the support of input distribution $P_X$ over which we aim to evaluate the model performance. 

Assuming we have a   pool of data $\xpool$, we design
 adaptive sampling algorithms that iteratively select
inputs in $\xpool$ to be labeled.
Since labeling inputs takes time in practice, we model
real-world instances by considering \emph{batched} settings. Our goal is to sequentially label batches of data to accurately estimate model performance over $P_X$ and therefore we assume we have access to a set of inputs $\xeval \sim P_X$. 
We use the squared loss to illustrate our framework,
where our goal is to evaluate $\E_{X \sim P_X}[ (Y - \model(X))^2]$. Under the ``likelihood" function $p(y | f, x) = p_{\varepsilon}(y - f(x))$,  let $g(f)$ be the performance of the AI model $\model(\cdot)$ under the  data generating function $f$, which we refer to as our estimand of interest.
When we consider the mean squared loss,  $g(f)$ is given by 
\begin{align}
    g(f) \defeq \E_{X \sim P_X}\left[ \E_{Y \sim p(\cdot|f,X) } \Big[ (Y - \model(X))^2 \Big] \mid f \right]. \label{eqn:l2-g-f}
\end{align}
Our framework is general and can be extended to other settings. For example, a clinically useful  metric is \texttt{Recall}, defined as the fraction of individuals that the model $\model(\cdot)$ correctly labels as positive  among all the individuals who actually have the positive label 
\begin{align*}
    g(f) \defeq  \E_{X \sim P_X}\left[ \E_{Y \sim p(\cdot|f,X) } \Big[\indic{\model(X)>0}|Y=1\Big] \mid f\right].
\end{align*}
 
  Since the true function $f\opt$ is unknown, we  model it from a Bayesian perspective by formulating a posterior given the available supervised data. We refer to uncertainty over the data generating function $f$ as \emph{epistemic} uncertainty---since we can resolve it with more data---and that over
 the measurement noise $\varepsilon$ as \emph{aleatoric} uncertainty. 
Assuming independence given features $X$, we model the  likelihood of the data via the product 
$p({Y}_{1:m}|f, {X}_{1:m}) = \prod_{i=1}^m p(Y_i|f,X_i)$.
 Our prior belief  $\mu$ over functions $f$   reflects our uncertainty about how
labels are generated given features. 
To adaptively label inputs from $\mc{X}_{\rm pool}$, we assume access to an uncertainty quantification (UQ) method that provides posterior beliefs $\mu(f \mid \mc{D})$ given
any supervised data $\mc{D}:= \{(X_i,Y_i)\}_{i \in \mc{I}}$. As we detail  in Section~\ref{sec:uq}, our framework can leverage both classical 
Bayesian models like Gaussian processes and recent advancements in deep learning-based UQ  methods.

As new batches are labeled, we update our posterior beliefs about $f$ over time, which we view as ``state transitions'' of a dynamical system.
Recalling the Markov decision process depicted in Figure~\ref{fig:overview}, we sequentially label a batch of inputs from $\mc{X}_{\rm pool}$ (actions), which lead to state transitions (posterior updates).
Specifically, our initial state is given by $\mu_0(\cdot) = \mu(\cdot \mid \mc{D}^0)$, where $\mc{D}^0$ represents the initial labeled dataset.
At each period $t$, we label a batch of $K_t$ inputs $\mc{X}^{t+1} \subset \mc{X}_{\rm pool}$ resulting in labeled data $ \mc{D}^{t+1} = (\mc{X}^{t+1}, \datay^{t+1})$. After acquiring the labels at each step $t$, we update the posterior state to $\mu_{t+1}(\cdot) = \mu_t(\cdot \mid \mc{D}^{t+1})$. Modeling practical instances, we consider a small horizon problem with limited adaptivity $T$. Formulating an MDP over posterior states has long conceptual roots, dating back to the Gittin's index for multi-armed bandits~\citep{Gittins79}.

 We denote by $\pi_t$ the adaptive labeling policy at period $t$. We account for randomized policies $\datax^{t+1} \sim \pi_t(\mu_t)$ with a flexible batch size $|\datax^{t+1}| = \batchsize_t$.   
We assume $\pi_t$ is $\mc{F}_t-$measurable for all $t < T$, where $\mc{F}_t$ is the filtration generated by the observations up to the end of step $t$.
 Observe that $\mu_{t+1}$ contains randomness in the policy $\pi_t$ as well as randomness in $\datay^{t+1} \mid (\datax^{t+1},\mu_t)$. Letting $\pi = \set{\pi_0,....,\pi_{T-1}}$,  we minimize the uncertainty over $g(f)$
 at the end of data collection
\begin{align}
H(\pi) \defeq \E_{\mc{D}^{1:T} \sim \pi} \left[G(\mu_{T}) \right]  \defeq  \E_{\mc{D}^{1:T} \sim \pi} \left[G(\mu(\cdot \mid \mc{D}^{0:T})) \right]
     = \E_{\mc{D}^{1:T} \sim \pi} \left[ \V_{f \sim \mu(\cdot \mid \mc{D}^{0:T})}  g(f)  \right],
     \label{eqn:general-obj}
\end{align}   
where $G(\mu_T) = \V_{f \sim \mu_T}  g(f)$.
In the above objective~\eqref{eqn:general-obj}, we assume
that the modeler pays a fixed and equal cost for each outcome. 
Our framework can also seamlessly accommodate variable labeling cost. Specifically, we can define a cost function $c(\cdot)$ 
 applied on the selected subsets 
 and update the objective~\eqref{eqn:general-obj} accordingly to include the term  $\lambda c(\mc{D}^{1:T})$,
 where $\lambda$
 is the penalty factor that controls the trade-off between minimizing variance and cost of acquiring samples.

Our framework can be easily extended to causal estimation problems.  Consider a feature vector ${X}$ and suppose we have two treatment arms $Z \in \{0,1\}$. Our objective is to evaluate the average treatment effect over the population distribution $P_X$.  Given feature vector $X$, and treatment $Z$,  outcomes are generated from an unknown function $f\opt$: 
$Y = f\opt(X,Z) + \varepsilon.$
The available data is denoted by $\mc{D}^0 := \{X_i,Y_i,Z_i\}_{i \in \mc{I}}$ and given a pool of candidates 
$\xpool$, we want an
 adaptive sampling algorithms that iteratively select
candidates in $\xpool$ to be assigned a random treatment so that we can estimate average treatment effect efficiently. Under the ``likelihood" function $p(y | f, x, z) = p_{\varepsilon}(y - f(x,z))$,  let $g(f)$ represent  the average treatment effect, which is our estimand of interest. Formally, this is expressed as:
\begin{align}
g(f) \defeq \E_{X \sim P_X} \left[\E_{Y_1 \sim p(\cdot|f,X,Z=1) , Y_0 \sim p(\cdot|f,X,Z=0)} \left[Y_1 -  Y_0 \right]\mid f \right]. \label{eqn:ate-g-f}
\end{align}

\begin{figure}[ht]
\centering
\begin{tikzpicture}
[
roundnode/.style={circle, draw=black!60, very thick, minimum size=10mm},
squarednode/.style={rectangle, draw=black!60, very thick, minimum size=10mm, align =center,text width = 26mm},
]
\node[roundnode]      (maintopic)                              {$\mu$};
\node[roundnode]        (circle1)       [right=20mm of maintopic] {$\mu_0$};
\node[roundnode]      (circle2)       [right=5mm of circle1] {$\mu_t$};
\node[roundnode]        (circle3)       [right=20mm of circle2] {$\mu_{t+1}$};
\node[roundnode]        (circle4)       [right=5mm of circle3] {$\mu_T$};
\node[squarednode]        (circle5)       [right=5mm of circle4] {Reward/Cost $\E \left[ \V_{\mu_T} (g(f))\right]$};

\draw[thick, ->, >=stealth] (maintopic.east) -- node[anchor=south] {$(\mathcal{X}^0,\mathcal{Y}^0,\mathcal{Z}^0)$} (circle1.west);
\draw[thick, ->, dashed] (circle1.east) --  (circle2.west);
\draw[thick, ->] (circle2.east)  -- node[above, align =center, text width = 18mm] { Query $(\mathcal{X}^t,\mathcal{Y}^t,\mathcal{Z}^t)$} (circle3.west);
\draw[thick, ->,dashed] (circle3.east) --  (circle4.west);
\draw[thick, ->] (circle4.east) --  (circle5.west);

\end{tikzpicture}
\caption{MDP framework for adaptive labeling to efficiently estimate the average treatment effect (ATE).}
\label{fig:MDP_framework_flowchart}
\end{figure}
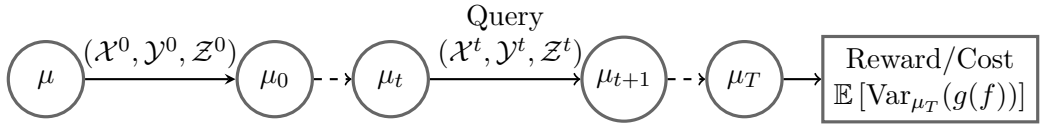

 Our prior belief  $\mu$ over functions $f$, now 
 reflects our uncertainty about how
outcomes are generated given features and treatments. 
  We sequentially observe outcome of a batch of inputs from $\mc{X}_{\rm pool}$ (actions), and treatments assigned to this batch. We assume that selected batch of inputs $\mc{X}^t$ is randomly assigned treatments $\mc{Z}^t$ with each $Z\sim p_Z$. We summarize our formulation in Figure~\ref{fig:MDP_framework_flowchart}.
Again, we assume $\mu_t$ is $\mc{F}_t-$measurable for all $t < T$, where $\mc{F}_t$ is the filtration generated by the observations up to the end of step $t$.
 Observe that $\mu_{t+1}$ contains randomness in the policy $\pi_t$, randomness in treatment assignment $\mc{Z}^{t+1}$ and randomness in $\datay^{t+1} \mid (\datax^{t+1}, \mc{Z}^{t+1},\mu_t)$. Letting $\pi = \set{\pi_0,....,\pi_{T-1}}$,  we minimize the uncertainty over $g(f)$
 at the end of data collection:
\begin{align}
\E_{\mc{D}^{1:T} \sim \pi} \left[G(\mu_{T}) \right] \defeq
\E_{\mc{D}^{1:T} \sim \pi} \left[ \V_{f \sim \mu_{T}}  g(f)  \right]
= \E_{\mc{D}^{1:T} \sim \pi} \left[ \V_{f \sim \mu(\cdot \mid \mc{D}^{0:T})}  g(f)  \right].
\label{eqn:general-ate-obj}
\end{align}

\section{Continuous Reformulation}
\label{sec:methodology}

The dynamic programs~\eqref{eqn:general-obj} and~\eqref{eqn:general-ate-obj} involve
continuous states and action spaces
that are combinatorially large in the number of data points in $\xpool$. In order to bring to bear the power of planning through policy gradients, 
we introduce continuous reformulations of the original planning problems.
First, we illustrate how common UQ approaches allow us to formulate a posterior belief $\mu(f \mid \mc{D})$ given any supervised data $\mc{D}$,
and update it when additional labels are gathered. 
In Section~\ref{sec:uq}, we describe  both classical 
Bayesian models such as 
Gaussian processes and recent advancements in deep learning-based UQ  methods that our framework can leverage.
Second, to deal with the combinatorial action space,
we propose a continuous policy parameterization $\pi_{t,\theta}$  for a single-batch policy and use $\batchsize$-subset sampling~\citep{EfraimidisSp06} to choose
$\batchsize$ samples, described in  Section\ref{sec:k-subset}.
Finally, to reliably optimize the policy $\pi_{t,\theta}$,  we simulate trajectories (``roll-out'') through which we approximate policy gradients 
(Section~\ref{sec:roll-out}). 

Though our conceptual framework is general and can leverage multi-step 
lookaheads~\citep{BertsekasTs96,EfroniDaScMa18, EfroniGhMa20}, we focus on one-step lookaheads. In subsequent sections, we  empirically demonstrate that even one-step lookaheads can  achieve significant improvement in sampling efficiency over other heuristic baselines. 

\subsection{Uncertainty Quantification (UQ) Module}
\label{sec:uq}

The UQ module maintains an estimate $\mu_t$ over posterior beliefs of $\dgm$ over time, enabling our planning framework. Our method is agnostic to the UQ module, and
we illustrate this using two instantiations: i) Gaussian Processes (GP) that are extensively used in the Bayesian Optimization literature and are known to work well in low dimensional data and regression setting, and ii) recently developed neural network-based UQ methods such as  
\ensembles/\ensembleplus~\citep{OsbandWeAsDwLuIbLaHaDoRo22}. 

\vspace{-10pt}
\paragraph{Gaussian Processes} $\mc{GP}(m(\cdot), \mc{K}(\cdot,\cdot))$ are defined by a mean function $m(\cdot)$ and kernel (or covariance function) $\mc{K}(\cdot,\cdot)$. For any set of  inputs $\mathcal{X} \equiv \{X_i\}_{i\in \mathcal{I}}$, the process $\bm{f}$ (where $f_i = {f}(X_i)$) follows a  multivariate normal distribution $\mc{N} (m(\mc{X}), \mc{K}(\mc{X},\mc{X}))$. Specifically, the mean and covariance are given by  $\E\left({f(X_i)}\right)=m(X_i)$  and $\cov(f(X_i), f(X_j)) = \mc{K}(X_i,X_j)$. 
Additionally, the observed output $\mathcal{Y}$ includes noise ${\varepsilon} \sim \mc{N}(0, \sigma^2 I)$, such that  $\mathcal{Y} = \bm{f} + \varepsilon$, where $\sigma > 0$  represents the noise level. 

Given training data $(\mathcal{X}, \mathcal{Y})$, and test points ${\mathcal{X}}'$, closed form posterior estimates for $\bm{{f}'}$ can be computed as follows:
\begin{align*}
\bm{{f}'} \mid {\mathcal{X}, \mathcal{Y}, \mathcal{X}'} & \sim \mc{N}(\bar{\bm{f'}}, \mathbb{V}), \\ 
\text{where }  \bar{\bm{f'}} &\defeq \mc{K}(\mc{X}', \mathcal{X}) [\mc{K}(\mc{X},\mc{X}) + \sigma^2 I]^{-1} \mc{Y} \\ 
\mathbb{V} & \defeq \mc{K}(\mc{X}', \mc{X}') - \mc{K}(\mc{X}', \mc{X}) [\mc{K}(\mc{X},\mc{X}) + \sigma^2 I]^{-1} \mc{K}(\mc{X},\mc{X}').
\end{align*} 


\vspace{-10pt}
\paragraph{Ensembles}\ensembles~\citep{LakshminarayananPrBl17}  learn an ensemble of neural networks from given data, with each network having independently initialized weights and bootstrap sampling. 
\ensembleplus\citep{OsbandAsCa18} extends this approach by combining ensembles of neural networks with randomized prior functions~\citep{OsbandVa15}. The prior function is added to each network in the ensemble, trained using L2 regularization. Recently,  
Epistemic neural networks (ENNs)~\citep{OsbandWenAsDwIbLuRo23} have also been shown to be an effective way of quantifying uncertainty. All these deep learning models are parametrized by some parameter $\eta$. For a given dataset $\set{(X_i,Y_i)}_{i \in \mc{I}}$, the model weights $\eta$ are update through gradient descent under a loss function $\loss(.)$, with the update rule expressed as: 
\begin{align}
    \eta_{\rm new} = \eta - 
    \sum_{i \in \mc{I}} \nabla_{\eta} \loss( X_i,Y_i,\eta). \label{eqn:ENN-update-naive}
\end{align}

\begin{algorithm}[t]
 \caption{ One-step lookahead planning}\label{alg:complete}
 \begin{algorithmic}[1] 
 \State \textbf{Inputs:} Initial labeled data $\mc{D}^0$, horizon $T$, UQ module,  pool data  $\xpool$,  batch size $\batchsize$
 
\noindent \textbf{Returns:} 
  Selected batches $(\datax^t, \datay^t)$ for $1 \le t \le T$
 and updated estimate of the objective $G(\mu_{T})$
 \State \textbf{Initialization}: Compute initial posterior state $\mu_0$ of the UQ module based 
 on initial labeled data $\mc{D}^0$.
\For{$0 \le t \le T-1$:}
\State Optimize $\pi_{t,\theta}$ using policy gradient updates  (based on  current posterior state $\mu_t$, pool  $\xpool$)
\State $\datax^{t+1} = \{X_j : X_j \in \texttt{SortDescending}(\xpool; \pi_{t,\theta}) \text{ for } 1\le j \le K \}$ 
\State Obtain labels $\datay^{t+1}$ to create $\mc{D}^{t+1}$
\State Update posterior state: $\mu_{t+1} \defeq \mu_t(\cdot\mid \mc{D}^{t+1})$
\State Estimate the objective $G(\mu_{t+1})$
\EndFor
 \end{algorithmic}
\end{algorithm}
\subsection{Sampling Policy}
\label{sec:k-subset}

Since we are using one-step lookahead policies, at each time step $t$,
we initiate a parametrized policy $\pi_{t,\theta}\defeq\pi_\theta$ that selects a  batch of $\batchsize$ samples to be queried. The samples are selected from a pool $\xpool$ of size $n$, based on weights ${\bm{w}}(\theta)$. 
 Specifically, given weights $\bm{w} \ge \bm{0}$ and a batch size $\batchsize$, the  $\batchsize$-subset sampling mechanism generates a random vector ${S} \in \{0,1\}^{n}$ such that $\sum_{i=1}^n S_i = \batchsize$, whose distribution depends on $\bm{w}$.
Let 
$\bm{e}^j$ denote the $j$-th unit vector, and consider a sequence of unit vectors $[\bm{e}^{i_1} \cdots \bm{e}^{i_K}] \defeq s$. Using a parametrization known as  weighted reservoir sampling ($\wrs$)~\citep{EfraimidisSp06}, the probability of  selecting a sequence $s$ is given by:
\begin{align}
    p_{\wrs}([\bm{e}^{i_1}, \cdots, \bm{e}^{i_K}]|\bm{w}) \defeq \frac{w_{i_1}}{\sum_{j=1}^n w_j}\frac{w_{i_2}}{\sum_{j=1}^n w_j-w_{i_1}}...\frac{w_{i_K}}{\sum_{j=1}^n w_j-\sum_{j=1}^{K-1}w_{i_j}}.
    \label{eqn:p-S-w-def}
 \end {align}

The probability of selecting a batch $S$ can then be defined as: $ p(S|\bm{w}) \defeq \sum_{s \in \Pi(S)} p_{\wrs}(s|\bm{w}),$
where $\Pi(S) = \set{s:  \sum_{j=1}^K s[j]= S}$ denote all sequences of unit vectors with sum equal to $S$.

\subsection{Policy optimization using roll-outs}
\label{sec:roll-out}

At each time step $t$, we optimize the policy to minimize the uncertainty over the one-step target estimand $G(\mu_{t+1})$ based on current posterior belief $\mu_t$.
Once we optimize the policy, our algorithm  selects a batch of inputs $\datax^{t+1}$ to be queried based on the optimized policy ($\pi_{t,\theta}$).  After selecting the batch $\datax^{t+1}$, we observe the corresponding labels $\datay^{t+1}$ and update posterior belief over the data-generating function $f$ to $\mu_{t+1}$. Algorithm~\ref{alg:complete} summarizes each of these steps of the overall procedure.   The policy optimization is done using policy gradients estimated through simulated ``roll-outs" which we explain below.

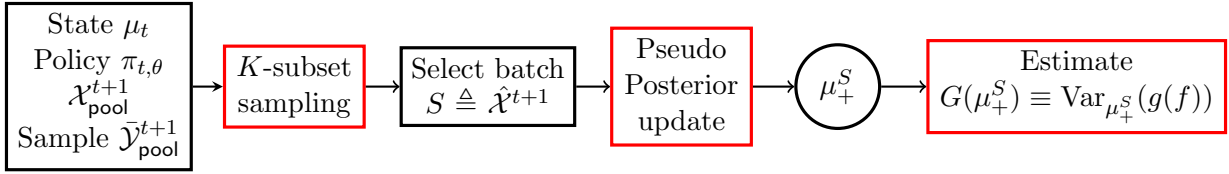
\begin{figure}[t]
\centering
\begin{tikzpicture}
[
roundnode/.style={circle, draw=black, very thick, minimum size=10mm, align=center, text width = 6mm},
roundnode2/.style={circle, draw=black, very thick, minimum size=10mm, align=center, text width = 8mm},
squarednodea/.style={rectangle, draw=black, very thick, minimum size=10mm, align =center,text width = 22mm},
squarednode1/.style={rectangle, draw=black, very thick, minimum size=10mm, align =center,text width = 20.5mm},
squarednode2/.style={rectangle, draw=red, very thick, minimum size=10mm, align =center,text width = 16mm},
squarednode3/.style={rectangle, draw=red, very thick, minimum size=10mm, align =center,text width = 37mm},
squarednode4/.style={rectangle, draw=red, very thick, minimum size=10.5mm, align =center,text width = 24mm},
squarednode4a/.style={rectangle, draw=red, very thick, minimum size=10.5mm, align =center,text width = 26mm},
squarednode5/.style={rectangle, draw=red, very thick, minimum size=10.5mm, align =center,text width = 22mm},
]
\node[squarednodea]      (maintopic)                              {State $\mu_t$ \\ Policy $\pi_{t,\theta} $ \\ $  \xpoolj{t+1}$\\ Sample $\ybpool^{t+1}$};
\node[squarednode2]        (circle1)       [right=4mm of maintopic] {$\batchsize$-subset sampling};
\node[squarednode1]      (circle2)       [right=4.5mm of circle1] {Select batch $S\defeq\hat{\mathcal{X}}^{t+1}$};
\node[squarednode2]        (circle3)       [right=4.5mm of circle2] {Pseudo Posterior update};
\node[roundnode]        (circle4)       [right=5.5mm of circle3] {${\mu}_{+}^S$};
\node[squarednode3]        (circle5)       [right=6mm of circle4] {Estimate\\ $ G(\mu_{+}^S) \equiv  \V_{{\mu}^S_{+}} (g(f))$};


\draw[thick, ->, >=stealth] (maintopic.east) --  (circle1.west);
\draw[thick, ->] (circle1.east) --  (circle2.west);
\draw[thick, ->] (circle2.east)  --  (circle3.west);
\draw[thick, ->] (circle3.east) --  (circle4.west);
\draw[thick, ->] (circle4.east) --  (circle5.west);


\end{tikzpicture}
\caption{One-step lookahead roll-out for policy gradient estimation}
\label{fig:one_step_look_ahead_general}
\end{figure}

We optimize the policy $\pi_{t,\theta}$ using policy gradients and we estimate these policy gradients by  simulating the ``roll-outs" of posterior beliefs over $\dgm$ from the current belief $\mu_t$. Specifically, we first generate pseudo labels $\ybpool$ for the $\xpool$ using current belief $\mu_t$. We then use $\pi_{t,\theta}$ ($K$-subset sampling) to select batch $S\defeq\hat{\mathcal{X}}^{t+1}$ and do a pseudo-posterior update ($\mu_{+}^S$) using selected batch $\hat{\mathcal{X}}^{t+1}$ and the corresponding labels from $\ybpool$. Using the pseudo posterior  we estimate the objective $ G(\mu_{+}^S) \equiv \V_{f\sim\mu_{+}^S} (g(f))$ using Monte-Carlo sampling. 
 Figure \ref{fig:one_step_look_ahead_general} demonstrates a typical roll-out at time step $t$. Once we do the roll-out, we then estimate the policy gradient based on this roll-out as we describe in the next section. 

\section{Policy Gradient Estimation}
 \label{sec:diff-piepline}

Under the continuous policy parameterization policy $\pi(t,\theta) \defeq \pi_\theta$ we develop in the previous section, our objective is to minimize the \textit{one-step lookahead objective}
\begin{align*}
H(\theta) \defeq \E_{\mc{D}^{t+1} \sim \pi_\theta} \left[G(\mu_{t+1}) \right]  \defeq  \E_{\mc{D}^{t+1} \sim \pi_\theta} \left[G(\mu_t(\cdot \mid \mc{D}^{t+1})) \right]\defeq \E_{\mc{D}^{t+1} \sim \pi_\theta} \left[ \V_{f \sim \mu_{t+1}}  g(f)  \right].
\end{align*}
Let the gradient be defined as $ \nabla_\theta H(\theta) \defeq \nabla_\theta \E_{A\sim\pi_\theta}[G(A)]$. Here, we slightly abuse notations by letting  $A\defeq\mc{D}^{1:T}$ and $G(A)\defeq G(\mu(\cdot \mid \mc{D}^{0}, \mc{D}^{1:T}))$,  where the distribution of $A$ depends on $\pi_\theta$.

To estimate policy gradients, a popular approach is to use the score-trick (\textsf{REINFORCE}~\citep{Williams92}):
 $\nabla_\theta \E_{A \sim \pi_{\theta}}[G(A)] 
= \E_{A \sim \pi_{\theta}}[G(A) 
\nabla_\theta \log \pi_{\theta}(A)]
$.  Using the simulated roll-out described in the previous section, the \textsf{REINFORCE} estimator  $\frac{1}{N} \sum_{i=1}^N G(A^{(i)}) 
\nabla_\theta \log \pi_{\theta}(A^{(i)})$ can be used to optimize the  policy $\pi_{t,\theta}\defeq \pi_\theta$. 
While unbiased and only requiring zeroth order access to $G$, the \textsf{REINFORCE} estimator often suffers high variance~\citep{RezendeMoWi14}, especially when $\pi_{\theta}(A)$ is small. 
Although a stream of work strives to provide variance reduction techniques for \textsf{REINFORCE}~\citep{MnihGr14,PapiniBiCaPiRe18},
they require value function estimates that are also challenging to compute in our planning problem~\eqref{eqn:general-obj}. 

To remedy this,  we now introduce a new framework through which we can approximate policy gradients using direct backpropagation.
A fundamental challenge we grapple with is that pathwise gradients do not exist in our formulation: the objective is non-differentiable since ultimately inputs are either labeled or not, causing a fundamental discretness in the dynamics. 
We thus  devise a smooth approximation of the simulated roll-out to make it pathwise differentiable, which we now detail. The policies derived using our efficient differentiable simulator exploits the system structure and can achieve significantly improved performance compared to    policies 
that do not rely on gradient information,
as we illustrate in Section~\ref{sec:experiment}.

\subsection{Smoothed-pathwise  policy gradients}

Since the dynamics of our planning problem---posterior updates---is known, instead of relying on high-variance score-based gradients
our goal is to leverage backpropagation and auto-differentiation to directly estimate approximate pathwise policy gradients~\eqref{eqn:grad-estimator}. 
A standard approach to pathwise gradient estimation is to find a random variable $Z \sim p_Z$ distributed independent of policy $\pi_{t,\theta}\defeq\pi_{\theta}$,  such that $A = {h}(Z,\theta)$ and
\begin{align}
    \nabla_\theta \E_{A \sim \pi_{\theta}}[G(A)] = \nabla_{\theta}\E_{Z\sim p_Z}[ G(h(Z, \theta))] \stackrel{(a)}{=}  \E_{Z\sim p_Z}[\nabla_{\theta} G(h(Z, \theta))]
    \label{eqn:grad-estimator-ideal}
\end{align} 
However,  in our formulations $h(Z, \theta)$ is non-differentiable w.r.t. $\theta$ because the actions are discrete; an input $X$ is either labeled or not.
As a result, the equality $(a)$ in~\eqref{eqn:grad-estimator-ideal}
no longer makes sense as $\nabla_{\theta}G(h(Z, \theta))$ does not exist. 

To address this, we  use a smoothed approximation, ${h}_{\tau}(Z,\theta)$, such that  $G({h}_{\tau}(Z,\theta))$ becomes differentiable
\begin{align}
\nabla_\theta \E_{A \sim \pi_{\theta}}[G(A)] \approx  \nabla_{\theta} \E[ G(h_{\tau}(Z, \theta))]
\approx \frac{1}{N}\sum_{i=1}^N \nabla_\theta G(h_\tau(Z_i,\theta)),
\label{eqn:grad-estimator}
\end{align}
where $\tau$ is a temperature parameter that controls the degree of smoothing.
It is important to note that sometimes even $G(\cdot)$ is non-differentiable, and in that case,
we must also consider a smooth approximation of $G(\cdot)$ 
(e.g., refer to Section~\ref{sec:details-recall-smoothing} for details on smoothing the \texttt{Recall} objective discussed earlier in Section~\ref{sec:formulation}). 

To ease notation, rewrite the one-step objective parametrized by $\theta$ 
\begin{align}
   H(\theta) \defeq \E_{{\ybpool \sim \mu; S \sim \pi_\theta }}  [G\paran{\mu_+^S}],
  \label{eqn:obj-original-unsmoothed}
\end{align} 
where $G$ was defined in~\eqref{eqn:general-obj}. Here, $\pi_\theta \equiv p(\cdot|\bm{w}(\theta))$ is the distribution over subsets governed by $\bm{w}$ as defined by~\eqref{eqn:p-S-w-def}. Further, $\mu$ is the current posterior state and $\mu_+^S$ is the updated posterior state after incorporating the the additional batch  $S$ selected from $\xpool$ and the corresponding pseudo-outcomes from $\ybpool$ (drawn based on current posterior state $\mu$).

We now describe a fully differentiable smoothed-pipeline by ensuring that each component of the ``roll-out'' pipeline: i) $\batchsize$-subset sampling, ii) posterior updates (UQ module), and iii) the objective is differentiable.

\subsection{Smoothing sampling policy: Soft $\batchsize$-subset sampling}

Recall the sampling policy introduced in Section \ref{sec:k-subset}. While the sampling policy features a continuous parametrization, it samples a discrete subset $S$ from the given pool $\mathcal{X}^{pool}$, making the operation non differentiable.  To address this, we adopt the \textit{``soft $\batchsize$-subset sampling procedure''} (Algorithm~\ref{alg:k-subset})  proposed by~\citet{XieEr19}, which enables the differentiability of sampling process. Broadly, this procedure  generates a random vector 
$\bm{a}(\theta) \in [0,1]^n$, a smoothed approximation of $S \in \{0,1\}^n$,  ensuring that $\sum_{i=1}^{n}a_i = \batchsize$. The vector  $\bm{a}({\theta})$ is approximately sampled from the sampling policy $ \pi_{\theta} \equiv p(\cdot|\bm{w}(\theta))$ defined in \eqref{eqn:p-S-w-def}, and it is differentiable with respect to $\theta$. The soft $\batchsize$-subset sampling procedure can be viewed as a generalization of the Gumbel-softmax trick, which smooths the sampling of a single item. For additional details, we refer readers to~\citep{XieEr19}.

\begin{algorithm}
\caption{Soft $\batchsize$-subset sampling   algorithm}\label{alg:k-subset}
\begin{algorithmic}[1]
\State \textbf{Inputs:} Weight vector $\bm{w}(\theta) \in \R_+^n$
\State  Sample $n$ independent standard Gumbel random variables $g_1,g_2,\ldots,g_n$
 \State Compute keys $\hat{r}_i = g_i + \log(w_i)$ for all $i$
\State
Initialize $\kappa^1_i = \hat{r}_i$ for all $i=1,\ldots,n$ 
\State  
For $j=1,..,\batchsize$, set
\begin{align*}
    a_i^j = \frac{\exp(\kappa_i^j/\tau)}{\sum_{k=1}^n \exp(\kappa_k^j/\tau)} \text{~for~all~} i=1,\ldots, n,
\end{align*}
$\kappa_i^{j+1} = \kappa_i^{j} + \log(1-a_i^j)$ for all $i=1,...,n$.
\State \textbf{Return:} The soft vector, $\bm{a} = \bm{a}^1+\bm{a}^2+...+\bm{a}^\batchsize$.  
\end{algorithmic}
\end{algorithm}

\subsection{Smoothing the Uncertainty Quantification (UQ) Module} 

Recall that our posterior updates rely on the subset $S$ of $\xpool$ selected using the policy $\pi(\theta)$. 
However, we have now introduced  soft $\batchsize$-subset samples $\bm{a}(\theta) \in [0,1]^n$, a smoothed approximation of $S\in\{0,1\}^n$, which does not represent a discrete subset of $\xpool$. This raises a question: how can posterior updates be performed using $\bm{a}(\theta)$?   
To address this, we reinterpret $\bm{a}(\theta)$ as weights for the  samples in $\xpool$ and adapt the  posterior updates of the respective uncertainty quantification methodologies accordingly. Notably, using weighted samples for posterior updates (to enable policy gradients) may hold independent value for other sequential decision-making tasks that rely on dynamic optimization.

We modify the closed-form Gaussian Process (GP) posterior updates to account for weighted inputs. Given training data $(\mathcal{X}, \mathcal{Y})$ and corresponding weights $\bm{a}$, the closed form posterior estimates for  $\bm{{f}'}$ at test points ${\mathcal{X}}'$ can be computed as follows:

\begin{align*}
\bm{{f}'} \mid {\mathcal{X}, \mathcal{Y}, \mathcal{X}'} & \sim \mc{N}(\bar{\bm{f'}}, \mathbb{V}), \\ 
\text{where }  \bar{\bm{f'}} &\defeq \mc{K}(\mc{X}', \mathcal{X}) \left [\mc{K}(\mc{X},\mc{X}) \left((\bm{1}-I)\bm{a}\bm{a}^T+I\right)+ \sigma^2 I\right ]^{-1} \mc{Y} \\ 
\mathbb{V} & \defeq \mc{K}(\mc{X}', \mc{X}') - \left(\bm{a} \odot\mc{K}(\mc{X}, \mc{X}')\right)^T \left[\mc{K}(\mc{X},\mc{X}) \left((\bm{1}-I)\bm{a}\bm{a}^T+I\right) + \sigma^2 I\right]^{-1} \left(\bm{a} \odot \mc{K}(\mc{X},\mc{X}') \right).
\end{align*}

It can be easily verified that if $\bm{a}$ is a $K-$hot vector, then the standard GP posterior update is recovered. 
Additionally, we present a weighted GP algorithm  (Algorithm~\ref{alg:weighted-GP}) adapted from Algorithm 2.1 in~\citep{Rasmussen06}, to efficiently compute the proposed posterior updates. 

\begin{algorithm}
\caption{Weighted Gaussian process regression}\label{alg:weighted-GP}
\begin{algorithmic}[1]
\State Input: $\mc{X}, \mc{Y}, \mc{X}', \sigma^2, \bm{a}$
       
       Return: $(\bar{\bm{f}}',\mathbb{V})$

       Define: 
$K \defeq \mc{K}(\mc{X},\mc{X}), K_{*} \defeq \mc{K}(\mc{X},\mc{X}')$ and $K_{*,*} \defeq \mc{K}(\mc{X}', \mc{X}')$
\State $K_{\bm{a}} = K ( (\bm{1}-I)   \bm{a}\bm{a}^\top + I)$
\State $K_{\bm{a},*} = \bm{a} \odot K_*$
\State $L = \mathsf{Cholesky}(K_{\bm{a}} + \sigma^2 I)$
\State $\alpha = L^\top \setminus (L \setminus \bm{Y})$
\State $\bar{\bm{f}}' = K_{\bm{a},*}^\top \alpha$
\State $v =  L \setminus K_{\bm{a},*}$
\State $\mathbb{V} = K_{*,*} - v^\top v$
\State Return: mean and covariance: $(\bar{\bm{f}}',\mathbb{V})$ 
\end{algorithmic}
\end{algorithm}

Similarly, to perform weighted updates for deep learning-based UQ modules (\ensembles/ \ensembleplus), we modify previous gradient updates by incorporating soft sample weights $\bm{a}(\theta)$:
\begin{align*} 
\eta_1(\bm{a}(\theta)) \defeq  \eta_0 - 
\sum_{i \in \mc{I}}  a_i(\theta)   \nabla_{\eta} 
\loss (X_i,\bar{Y}_i,\eta) \mid_{\eta = \eta_0},
\end{align*}  
where $\loss$ is the loss function to train the deep learning based-UQ methodology, 
and we evaluate its gradient    at $\eta_0$. 
We interpret $\eta_1$ as an approximation of the optimal UQ module parameter with one gradient step.
Similarly, we define
a higher-order approximation $\eta_{h+1}(\bm{a}(\theta))$  as
\begin{align}
    \eta_{h+1} (\bm{a}(\theta)) \defeq  \eta_h (\bm{a}(\theta)) - 
    \sum_{i \in \mc{I}}  a_i(\theta)   \nabla_{\eta} \ell(X_i,\bar{Y}_i,\eta)\mid_{\eta = \eta_h(\bm{a}(\theta))}. \label{eqn:higher-eta-update}
\end{align}

\begin{figure}[t]
\centering
\begin{tikzpicture}
[
roundnode/.style={circle, draw=black, very thick, minimum size=10mm, align=center, text width = 6mm},
roundnode2/.style={circle, draw=black, very thick, minimum size=10mm, align=center, text width = 8mm},
squarednodea/.style={rectangle, draw=black, very thick, minimum size=10mm, align =center,text width = 22mm},
squarednode1/.style={rectangle, draw=black, very thick, minimum size=10mm, align =center,text width = 20.5mm},
squarednode2/.style={rectangle, draw=red, very thick, minimum size=10mm, align =center,text width = 16mm},
squarednode3/.style={rectangle, draw=red, very thick, minimum size=10mm, align =center,text width = 28mm},
squarednode4/.style={rectangle, draw=red, very thick, minimum size=10.5mm, align =center,text width = 24mm},
squarednode4a/.style={rectangle, draw=red, very thick, minimum size=10.5mm, align =center,text width = 26mm},
squarednode5/.style={rectangle, draw=red, very thick, minimum size=10.5mm, align =center,text width = 22mm},
]
\node[squarednodea]      (maintopic)                              {State $\mu_t$ \\ Policy $\pi_{t,\theta} $\\  $  \xpoolj{t+1}$ \\ Sample $\ybpool^{t+1}$};
\node[squarednode2]        (circle1)       [right=4mm of maintopic] {$\batchsize$-subset sampling};
\node[squarednode1]      (circle2)       [right=4.5mm of circle1] {Select batch $S\defeq\datax^{t+1}$};
\node[squarednode2]        (circle3)       [right=4.5mm of circle2] {Pseudo Posterior update};
\node[roundnode]        (circle4)       [right=5.5mm of circle3] {${\mu}_{+}^S$};
\node[squarednode3]        (circle5)       [right=6mm of circle4] {Estimate $\E \left[ \V_{{\mu}_{+}} (g(f))\right]$};

\node[squarednode4]        (circle1b)       [below=11mm of circle1] {Soft $\batchsize$-subset sampling};
\node[roundnode]      (circle2b)       [right=5.4mm of circle1b] {\centering $a(\theta)$};
\node[squarednode4a]        (circle3b)       [right=7mm of circle2b] {Soft posterior update (differentiable)};
\node[roundnode]        (circle4b)       [right=4mm of circle3b] {${\mu}_{+}^{a(\theta)}$};
\node[squarednode5]        (circle5b)       [right=5mm of circle4b] {Differentiable estimate};

\draw[thick, ->, >=stealth] (maintopic.east) --  (circle1.west);
\draw[thick, ->] (circle1.east) --  (circle2.west);
\draw[thick, ->] (circle2.east)  --  (circle3.west);
\draw[thick, ->] (circle3.east) --  (circle4.west);
\draw[thick, ->] (circle4.east) --  (circle5.west);
\draw[thick, -, dashed] (circle1.south) --  (circle1b.north);
\draw[thick, -, dashed] (circle3.south) --  (circle3b.north);
\draw[thick, -, dashed] (circle5.south) --  (circle5b.north);

\draw[thick, ->] (maintopic.south) |-  (circle1b.west);
\draw[thick, ->] (circle1b.east) --  (circle2b.west);
\draw[thick, ->] (circle2b.east)  --  (circle3b.west);
\draw[thick, ->] (circle3b.east) --  (circle4b.west);
\draw[thick, ->] (circle4b.east) --  (circle5b.west);

\end{tikzpicture}
\caption{Differentiable one-step lookahead pipeline for efficient adaptive sampling}
\label{fig:diff_one_step_look_ahead}
\end{figure}
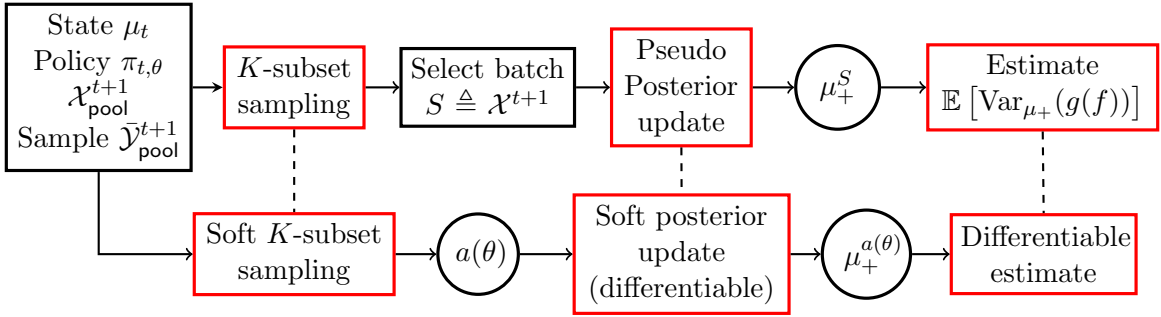

\subsection{Putting everything together}

Note that when using GPs as our UQ methodology, the gradient takes the form $\nabla_{\theta} G\left(\mu_+\left(\bm{a}(\theta)\right)\right)$. Assuming $G(\cdot)$ is differentiable, pathwise gradients can be computed since the posterior update $\mu_+$ is differentiable for GPs, and the sample $\bm{a}(\theta)$ is differentiable with respect to $\theta$.

\begin{figure}[h]
\centering
\includegraphics[width=1\textwidth, height=9cm]{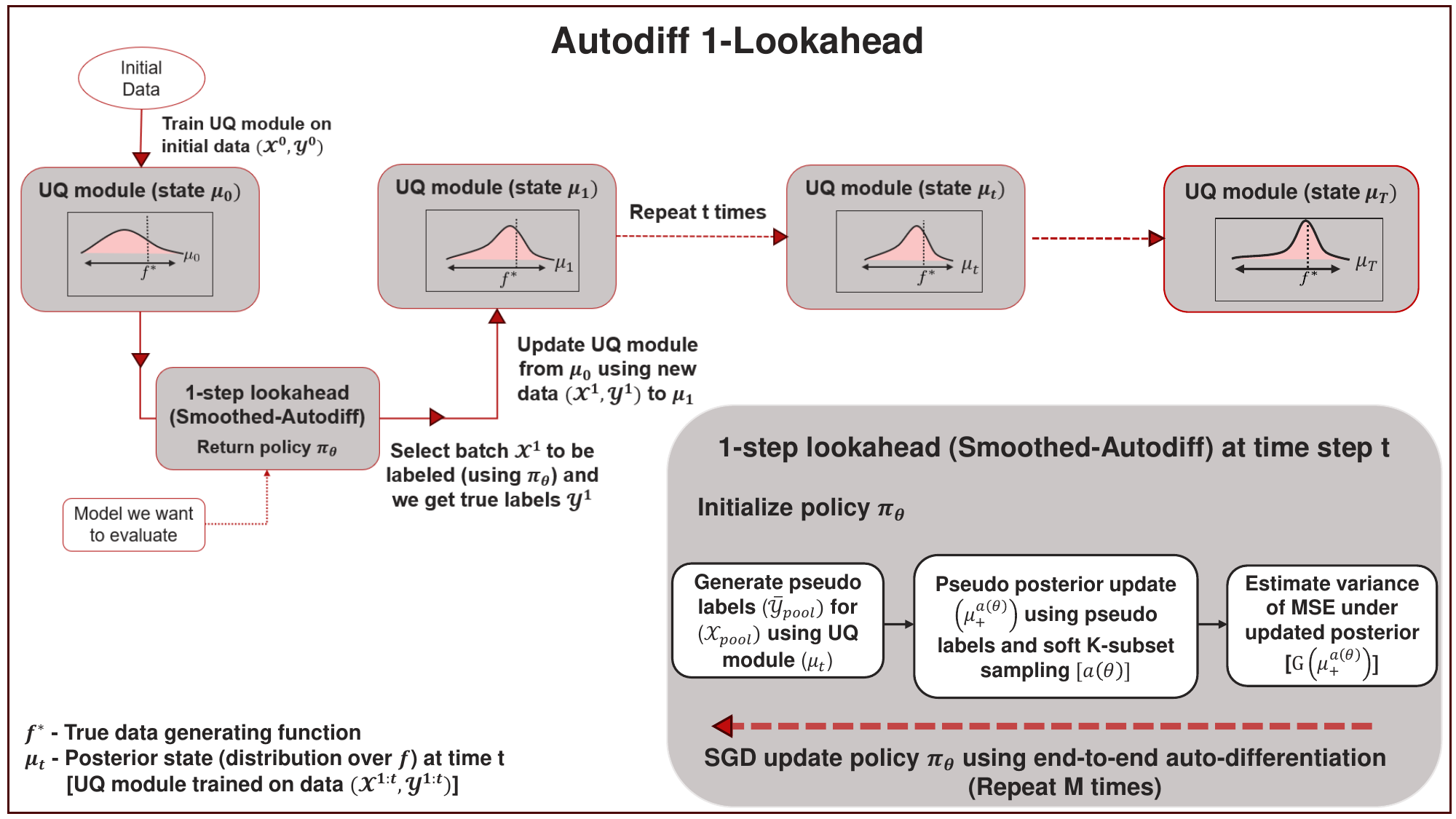}
\caption{Description of \ouralgo ~ algorithm}
\label{fig:Graphical_presentation}
\end{figure}

In the case of deep learning-based UQ methodologies, the gradient has the following form:

\[\nabla_{\theta} G\left( \argmin_{\eta} \loss \left(\eta\left(\bm{a}(\theta)\right)\right) \right) \equiv \nabla_{\theta} G\left( \eta_+\left(\bm{a}(\theta)\right) \right), \] 
where posterior beliefs are parametrized by $\eta$ (e.g., \ensembleplus weights) and $\eta_{+}$ represents the updated posterior parameters (optimized value of $\eta$). 
To compute this, we use $\mathsf{higher}$~\citep{GrefenstetteAmYaHtMoMeKiChCh19} and $\mathsf{torchopt}$~\citep{RenFeLiPaFuMaYa23}, which allows us to approximately differentiate through the argmin. In practice, there is a trade-off between the computational time and the accuracy of the gradient approximation depending on the number of training steps we do to estimate the argmin. 
The differentiable analogue of the  pipeline  is summarized in Figure~\ref{fig:diff_one_step_look_ahead}. We also explain our overall differentiable-algorithm (we call it \ouralgo) graphically in Figure~\ref{fig:Graphical_presentation}. 

\section{Experiments: Planning outperforms Heuristics}
\label{sec:experiment}

We begin our empirical demonstrations by showcasing the effectiveness of our planning framework on both synthetic and real datasets. We focus on the simplest planning algorithm, 1-step lookaheads (Algorithm~\ref{alg:complete}), and show that even basic planning can hold great promise. 
We illustrate our framework using two uncertainty quantification modules---GPs and 
\ensembles/ \ensembleplus. 

Throughout this section, we focus on evaluating the mean squared error of 
a regression model $\model$,  and develop adaptive policies that minimize uncertainty on $g(f)$ defined in~\eqref{eqn:l2-g-f}.
When GPs provide a valid model of uncertainty, 
our experiments show that our planning framework significantly outperforms other baselines. 
We further demonstrate that our conceptual framework extends to deep learning-based uncertainty quantification methods such as  \ensembleplus while highlighting computational challenges that need to be resolved in order to scale our ideas. 
For simplicity, we assume a naive predictor, i.e., $\psi(\cdot) \equiv 0$. However, we emphasize that this problem is just as complex as if we were using a sophisticated model $\psi(.)$. The performance gap between the algorithms 
primarily depends
on the level  of uncertainty in our prior beliefs.

To evaluate the performance of our algorithm, we benchmark it against several baselines. 
Our first set of baselines are from active learning~\citep{AggarwalKoGuHaPh14}:
\\ 
\textbf{(1)} 
\textsf{Uncertainty Sampling (Static):}  In this approach, we query the samples for which the model is least certain about. Specifically, we estimate the variance of the latent output $f(X)$ for each $X \in \xpool$ using the UQ module and select the top-$K$ points with the highest uncertainty. \\
\textbf{(2)} \textsf{Uncertainty Sampling (Sequential):} This is a greedy heuristic that sequentially selects the points with the highest uncertainty within a batch, while updating the posterior beliefs using pseudo labels from the current posterior state. Unlike \textsf{Uncertainty Sampling (Static)}, this method takes into account the information gained from each point within batch, and hence tries to diversify the selected points within a batch.

We also compare our approach to the  \textbf{(3)} \textsf{Random Sampling}, which selects each batch uniformly at random from the pool. Additionally, we compare solving the planning problem using  \textsf{REINFORCE}-based policy gradients with   $\mathsf{Smoothed\text{-}Autodiff}$ policy gradients.\footnote{Our code repository is available at
  \url{https://github.com/namkoong-lab/adaptive-labeling}.}


\begin{figure}[t]
\centering
\begin{minipage}[b]{0.49\textwidth}
\centering
\includegraphics[width=\textwidth, height=5cm]{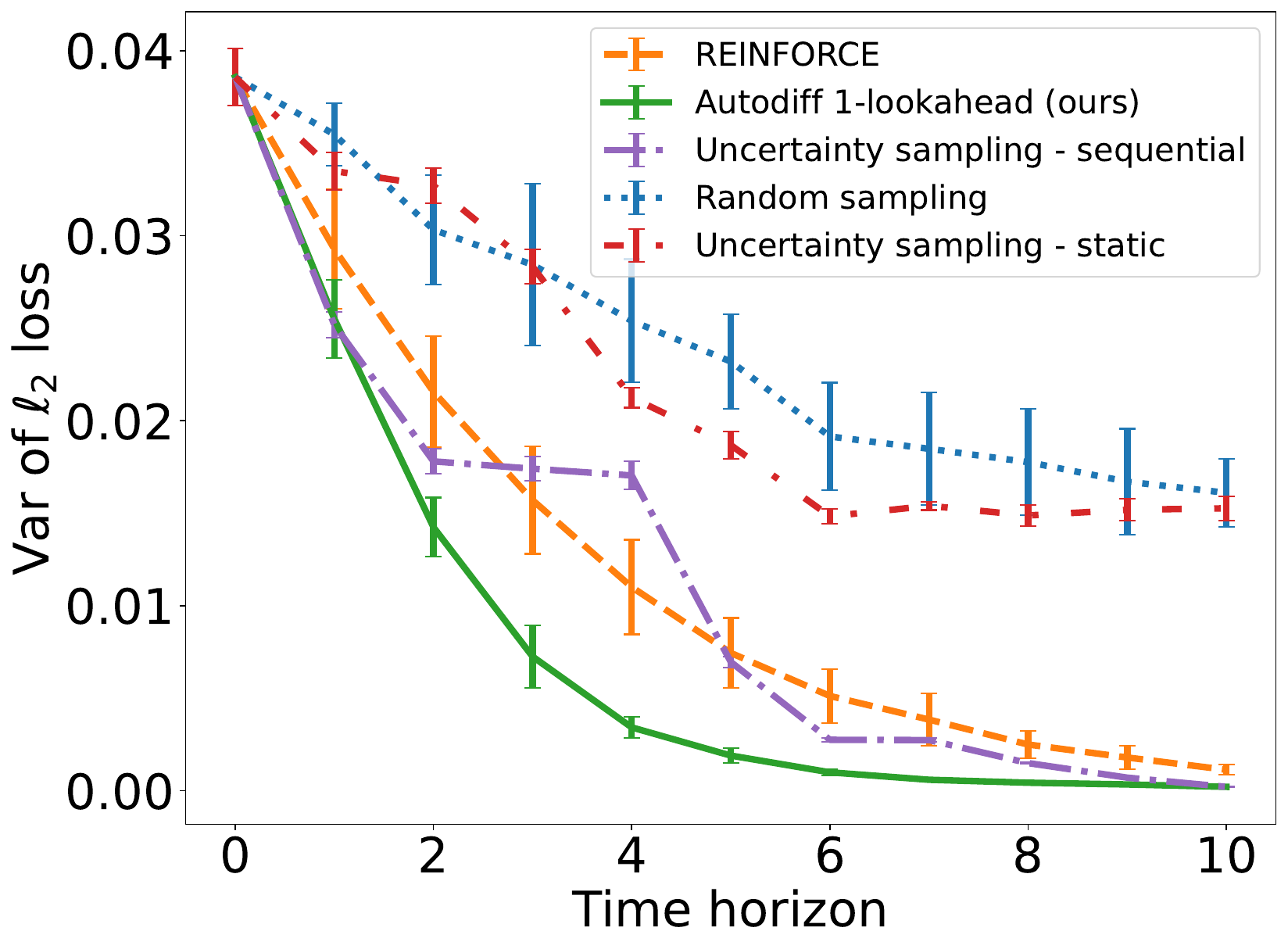}
\caption{(Synthetic data) Variance of mean squared loss evaluated through the posterior belief $\mu_t$ at each horizon $t$. This is the objective that policy gradient methods like \textsf{REINFORCE} and $\ouralgo$ optimizes. 1-step lookaheads are surprisingly effective even in long horizons.}
\label{fig:var-l2-sim}
\end{minipage}
\hfill
\begin{minipage}[b]{0.49\textwidth}
\centering \includegraphics[width=\textwidth, height=5cm]{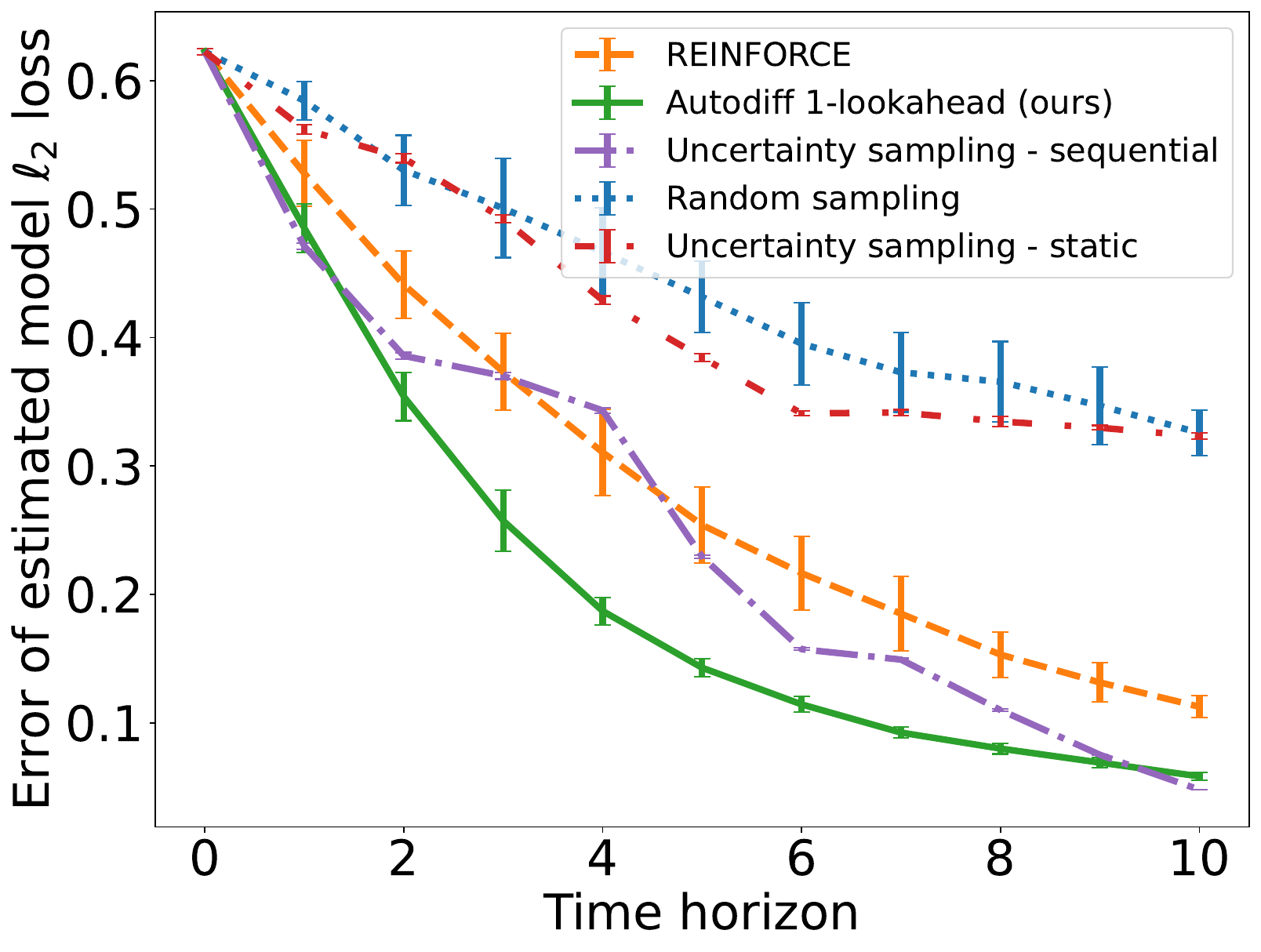}
\caption{(Synthetic data) Error between MSE calculated based on collected data $\mc{D}^{0:T}$ vs. population oracle MSE over $\mc{D}_{\rm eval} \sim P_X$. Reducing uncertainty over posteriors directly leads to better OOD evaluations. 1-step lookaheads significantly outperform active learning heuristics in small horizons.}
\label{fig:mean-l2-sim}
\end{minipage}
\end{figure}

\subsection{Planning with Gaussian processes}
\label{sec:experiment-plan-GP}
We now briefly describe the data generation process for the GP experiments,  deferring a more detailed discussion of the dataset generation to Section~\ref{sec:details-experiments}. 
We use both the synthetic data and the real data to test our methodology.
For the \emph{simulated data},  we construct a setting where the general population is distributed across \emph{51 non-overlapping clusters} while the initial labeled data $\dtrain$ just comes from one cluster. In contrast, both $\dpool \defeq (\xpool,\ypool),\deval \defeq (\xeval,\yeval)$ are generated   from all the clusters. 
We begin with a low-dimensional scenario, generating a one-dimensional regression setting using a GP. 
Although the data-generating process is not known to the algorithms,  we assume that the GP hyperparameters are known to all the algorithms
to ensure fair comparisons. This can be viewed as a setting where our prior is well-specified, allowing us to isolate the effects
of different policy optimization approaches
 without any concerns about the misspecified priors. We select $10$ batches, each of size $K=5$ across $T = 10$ time horizons.

To examine the robustness of our method against the distributional assumptions made  in the simulated case, we then move to a real dataset where the correct prior is not known. We simulate selection bias from the eICU dataset~\citep{PollardJoRaCeMaBa18}, which contains real-world patient data with in-hospital mortality outcomes. 
We conduct a $k$-means clustering to generate 51 clusters and then select data from those clusters. We view this to be a credible replication of practice, as severe distribution shifts are common due to selection bias in clinical labels.  To convert the binary mortality labels into a regression setting, we train a  random forest classifier and fit a GP on predicted scores, which serves as the UQ module for all the algorithms. As before, the task is to select 10 batches, each consisting of 5 samples, across 10 time horizons.

 In Figures~\ref{fig:var-l2-sim} and~\ref{fig:mean-l2-sim}, we present results for the simulated data. 
Figure~\ref{fig:var-l2-sim} shows the variance of $\ell_2$ loss, and Figure~\ref{fig:mean-l2-sim} presents the error in the estimated $\ell_2$ loss using $\mu_t$ (relative to true $\ell_2$ loss, that is unknown to the algorithm). 
As we can see from these plots, our method one-step lookahead  gives substantial improvements  over active learning baselines and random sampling. In addition,
compared to the one-step lookahead planning approach using \textsf{REINFORCE}-based policy gradients, 
we observe that $\mathsf{Smoothed\text{-}Autodiff}$-based policy gradients provide significantly more robust performance over all horizons.

\begin{figure}[t]
\centering
\begin{minipage}[b]{0.48\textwidth}
\centering
\includegraphics[width=\textwidth, height=5cm]{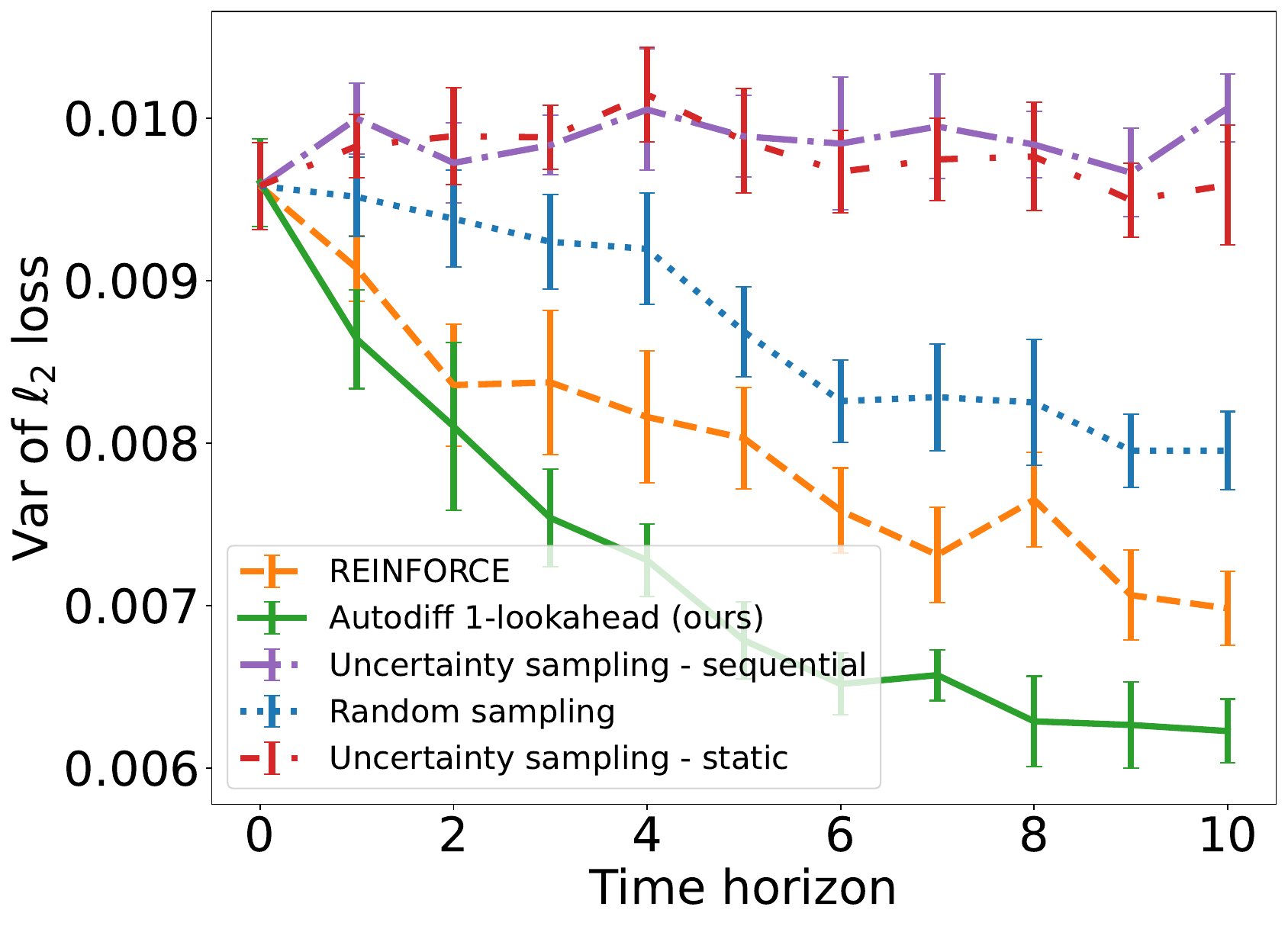}
\caption{(Real-world eICU data) Variance of mean squared loss evaluated through the posterior belief $\mu_t$ at each horizon $t$. Even 1-step lookaheads are extremely effective planners, and auto-differentiation-based pathwise policy gradients provide a reliable optimization algorithm based on low-variance gradient estimates.}
\label{fig:var-l2-real}
\end{minipage}
\hfill
\begin{minipage}[b]{0.48\textwidth}
\centering \includegraphics[width=\textwidth, height=5cm]{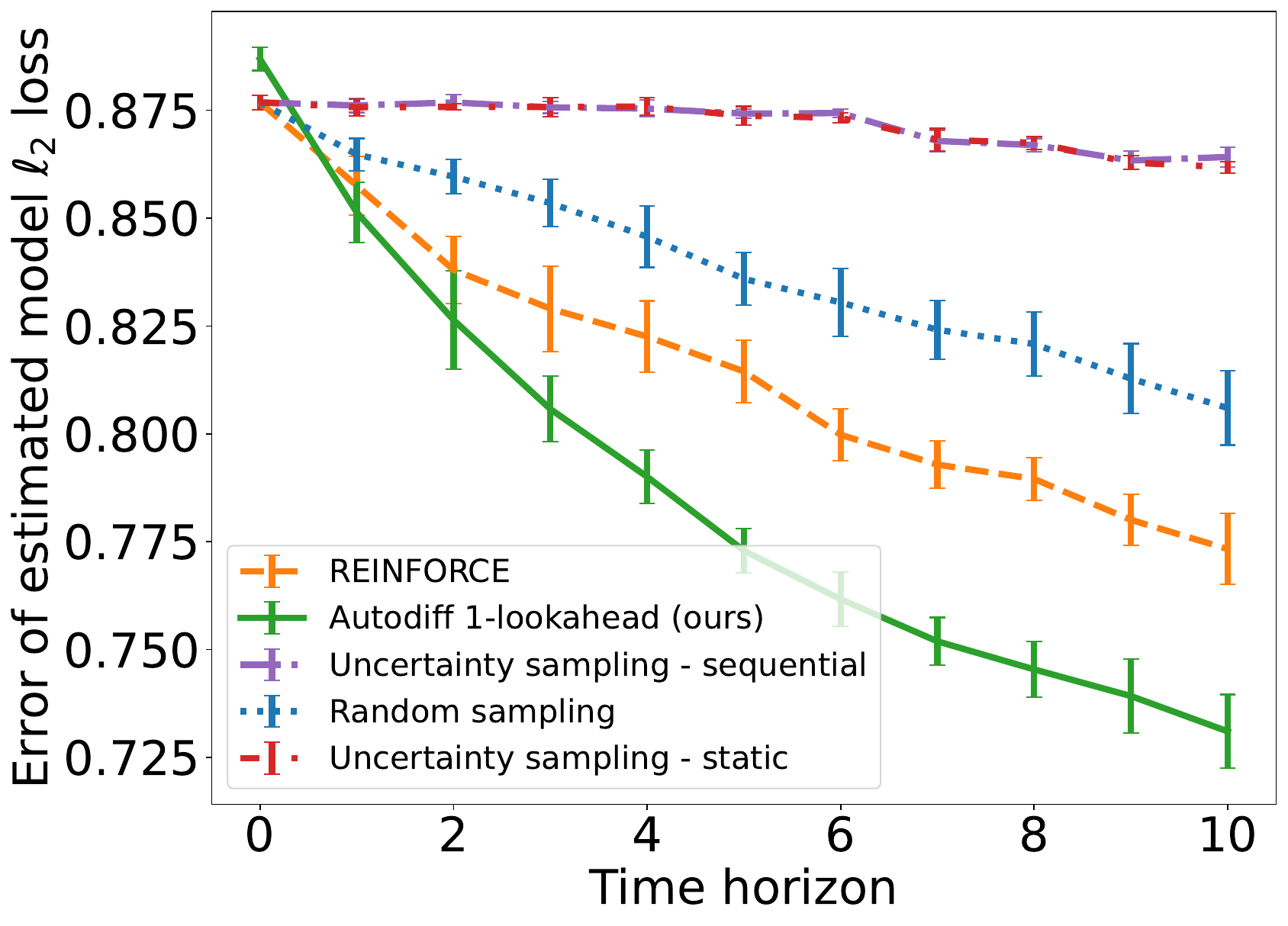}
\caption{(Real-world eICU data) Error between MSE calculated based on collected data $\mc{D}^{0:T}$ vs. population oracle MSE over $\mc{D}_{\rm eval} \sim P_X$. Reducing uncertainty over posteriors directly leads to better OOD evaluations. Our method significantly outperforms active learning-based heuristics, and random sampling.}
\label{fig:mean-l2-real}
\end{minipage}
\end{figure}

In Figures~\ref{fig:var-l2-real}~and~\ref{fig:mean-l2-real}, we observe similar findings on the eICU data. We see that planning policies (\textsf{REINFORCE} and $\mathsf{Smoothed\text{-}Autodiff}$) consistently outperform other heuristics by a large margin.  Active learning baselines perform poorly in these small-horizon batched problems and can sometimes be even worse than the random search baselines.  Overall, our results show the importance of careful planning in adaptive labeling for reliable model evaluation. 

We offer some intuition as to why one-step lookahead planning may outperform other heuristic algorithms. 
 First,  \textsf{Uncertainty sampling (Static)} while myopically selects the
 top-$K$ inputs with the highest uncertainty, it fails to consider 
the overlap in information content among the ``best” instances; see \citep{AggarwalKoGuHaPh14} for more details. 
In other words,  it might acquire points from the same region with high uncertainty while failing to induce diversity among the batch.
Although \textsf{Uncertainty Sampling (Sequential)} somewhat addresses the issue of information overlap, a significant drawback of 
this algorithm
is the disconnect between the objective we aim to optimize and the algorithm. For example, it might sample from a region with high uncertainty but very low density. 

\subsection{Planning with  neural network-based uncertainty quantification methods ($\ensembleplus$)}

We now provide a proof-of-concept that shows the generalizability of our conceptual framework  to the deep learning-based UQ modules, specifically focusing on $\ensembleplus$ due to their previously observed superior performance~\citep{OsbandWenAsDwIbLuRo23}. Recall that implementing our framework with deep learning-based UQ modules  requires us to retrain the model across multiple possible random actions $\bm{a}(\theta)$ sampled from the current policy $\pi_\theta$.
This requires significant computational resources, in sharp contrast to the GPs where the posteriors are in closed form and can be readily updated and differentiated. 

Due to the computational constraints, we test $\ensembleplus$ on a toy setting to demonstrate the generalizability of our framework. We consider a setting where the general population consists of four clusters, while the initial labeled data only comes from one cluster. Again we generate data using GPs.  The task is to select a batch of 2 points in one horizon. We detail the $\ensembleplus$ architecture in Section \ref{sec:details-experiments}, and we assume prior uncertainty to be large (depends on the scaling of the prior generating functions). 
The results are summarized in the Table~\ref{tab:UQ_ensemble}.


\begin{table}[h]
\vspace{-10pt}
\caption{Performance under \ensembleplus as the UQ module}
\centering
\begin{tabular}{|l|l|l|}
\hline
Algorithm   & Variance of $\loss_2$ loss estimate & Error of $\loss_2$ loss estimate  \\
\hline
\textsf{Random sampling} & 7129.8 $\pm$ 1027.0 & 136.2 $\pm$ 8.28 \\ \hline
\textsf{Uncertainty sampling (Static)} & 10852 $\pm$ 0.0 & 162.156 $\pm$ 0.0 \\ \hline
\textsf{Uncertainty sampling (Sequential)} & 8585.5 $\pm$ 898.9 & 144 $\pm$ 6.93 \\ \hline
\textsf{REINFORCE} & 1697.1 $\pm$ 0.0 & 45.27 $\pm$ 0.0 \\ \hline
\ouralgo & 1697.1 $\pm$ 0.0 & 45.27 $\pm$ 0.0 \\ \hline
\end{tabular}
\label{tab:UQ_ensemble}
\end{table}

\newcommand{\cosim}{\mathsf{cosim}}


\section{Analysis of Policy Gradient Estimators}
\label{sec:gradient_estimation_anlaysis}

In this section, we present a theoretical and empirical analysis of the statistical properties (bias-variance trade-off) of the \textsf{REINFORCE} and  $\mathsf{Smoothed\text{-}Autodiff}$ (smoothed-pathwise gradient) estimators and their efficacy
in downstream policy optimization task for adaptive labeling. First, in section \ref{sec:theory_gradient_estimation}, we theoretically compare the error in the $\mathsf{Smoothed\text{-}Autodiff}$ estimator and the \textsf{REINFORCE} estimator.
Next, in section \ref{sec:empirical_gradient_estimation_quality}, we empirically study the estimation quality of these two estimators within our framework, using  cosine similarity as the comparison metric. Finally, in section \ref{sec:gradient_estimation_algorithmic_performance}, we investigate the impact of gradient estimation quality on the performance of one-step lookaheads in the adaptive labeling setting.



\subsection{Analytic insights}
\label{sec:theory_gradient_estimation}

 Recall that in our optimization problem (\ref{eqn:general-obj}), we encounter a combinatorial action space.
To build intuition, in Section~\ref{sec:theory_gradient_estimation_binary},   we start with a simple two-action setting to compare the 
$\mathsf{Smoothed\text{-}Autodiff}$ estimator and the \textsf{REINFORCE} estimator. 
The more  general setting is discussed in Section~\ref{sec:theory_gradient_estimation_multi}.

 \subsubsection{Binary actions}
\label{sec:theory_gradient_estimation_binary}

To compare the two estimators, we consider a simplified setting with two possible actions 
$A \in \mathcal{A} := \{-1,1\}$, where the policy is parametrized by
$\theta \in [0,1]$, with
$\pi_\theta(-1) = \theta$ and $\pi_\theta(1) =1-\theta$. We are interested in the gradient $\nabla_\theta H(\theta)$, where $H(\theta) =\E_{A\sim \pi_\theta}G(A)$. For simplicity, we set $G(A)=A$.

\noindent Given $N$ i.i.d. samples of $A_i \sim \pi_\theta$, the \textsf{REINFORCE} estimator for $\nabla_\theta H(\theta)$ is defined as
\begin{align}
    \hat{\nabla}^{\mathsf{RF}}_N = \frac{1}{N}\sum_{i=1}^N\Big(G(A_i) \nabla_\theta \log(\pi_\theta(A_i)) \Big).
    \label{eqn:bin-reinf-def}
\end{align}
Using  the reparametrization trick, we can rewrite  $H(\theta)$ as $\E_U G(h(U,\theta))$, where 
\begin{align*}
    h(U,\theta):= 2\mc{I}(U > \theta)-1\stackrel{d}{=}A,
\end{align*}
with $U\sim \mathsf{Uni}[0,1]$. However, $h(U,\theta)$ is not differentiable with respect to $\theta$. Therefore, to enable the pathwise gradient estimation, we consider a smoothed approximation $ h_\tau(U,\theta)$ of $ h(U,\theta)$:
\begin{align*}
    h_{\tau}(U,\theta) = \frac{\exp\left(\frac{U-\theta}{\tau}\right)-1}{\exp\left(\frac{U-\theta}{\tau}\right)+1},
\end{align*}
where $\tau$ is a temperature parameter that controls the smoothness of $h_\tau(U,\theta)$.

\noindent The $\mathsf{Smoothed\text{-}Autodiff}$ estimator $\nabla_{\tau,N}^{\mathsf{grad}}$ is the $N$-sample approximation of 
$\E_{U}[\nabla_\theta G(h_{\tau}(U,\theta))]$, 
 given by
\begin{align}
    \hat{\nabla}^{\mathsf{grad}}_{\tau,N} = \frac{1}{N}\sum_{i=1}^N\Big(\nabla_\theta{G}(h_\tau(U_i,\theta))\Big).
    \label{eqn:bin-grad-def}
\end{align}
 Our result, with proof in Section~\ref{sec:proof-thm-binary},
 highlights the conditions under which 
 $\hat{\nabla}^{\mathsf{grad}}_{\tau,N}$~\eqref{eqn:bin-grad-def} achieves better gradient estimation (lower mean squared error)  compared to $\hat{\nabla}^{\mathsf{RF}}_{N}$~\eqref{eqn:bin-reinf-def}.
\begin{theorem}
For $\theta \in [0,1]$ and $N \le  \frac{1}{4\theta(1-\theta)} - 1$,
there exists $\Tilde{\tau}$ depending on $(N, \theta)$ such that 
\begin{align*}
\mse(\hat{\nabla}^{\mathsf{grad}}_{\Tilde{\tau},N})
< 
\mse(\hat{\nabla}^{\mathsf{RF}}_{N}).
\end{align*}
Additionally, for any $N$, $\theta = \frac{1}{k}$, and $k \to \infty$, we have that
$\mse(\hat{\nabla}^{\mathsf{RF}}_{N})  = \Omega(k) 
$, $\mse(\hat{\nabla}^{\mathsf{grad}}_{\tilde{\tau},N}) < 4$. 
The same statement holds for $\theta =  1-\frac{1}{k}$ as well. 
This implies that the MSE of $\hat{\nabla}^{\mathsf{RF}}_{N}$ is unbounded, while the MSE of gradient estimator is bounded.
\label{thm:binary}
\end{theorem}
\noindent In particular, this example illustrates how $\mathsf{Smoothed\text{-}Autodiff}$ gradients shine when 
the sample size $N$ is small and the policy takes on extreme values.

\subsubsection{General case}
\label{sec:theory_gradient_estimation_multi}

We now extend our analysis from Section~\ref{sec:theory_gradient_estimation_binary} to the setting with more than two actions.
Consider an  action space  $\mathcal{A}:= \{a_1,a_2,...,a_m\} \subset \mathbb{R}$ with $m \ge 2$ actions. 
Actions $A\in \mathcal{A}$ are drawn according to a policy $\pi_\theta$, such that $A=a_i$ with probability $\pi_\theta(a_i) = \frac{\theta_i}{\sum_{j=1}^m \theta_j}$. 
As in Section~\ref{sec:theory_gradient_estimation_binary}, we are interested in the gradient $\nabla_\theta H(\theta)$, where $H(\theta) =\E_{A\sim \pi_\theta}G(A)$. We assume that 
$G(\cdot)$ is differentiable and  sufficiently smooth such that
$G'(a) \le \bar{G}$ for all $a \in \mathsf{conv}(a_1,...,a_m)$. 










\noindent As in~\eqref{eqn:bin-reinf-def}, given $N$ i.i.d. samples of $A_j \sim \pi_\theta$,
the \textsf{REINFORCE} estimator 
is defined as
\begin{align}
\hat{\nabla}^{\mathsf{RF}}_{N,\theta_i}
\defeq 
\frac{1}{N}\sum_{j=1}^N\Big(G(A_j) \nabla_{\theta_i} \log(\pi_\theta(A_j)) \Big).
\label{eqn:eq-def-reinforce}
\end{align}



\noindent We now generalize the $\mathsf{Smoothed\text{-}Autodiff}$ estimator in ~\eqref{eqn:bin-grad-def}. Recall that $H(\theta) = \E_{A\sim \pi_\theta} {G}(A)$,  which can be rewritten as $$H(\theta) =  \E_{A\sim \pi_\theta}\left[ G \left(\sum_{i=1}^ma_i (\mc{I}(A)=a_i) \right) \right].$$
 Using the Gumbel-Max reparametrization trick  
 from~\citep{JangGuPo17}, we have
\begin{align*}
\{\mc{I}(A=a_i)\}_{i=1}^m 
\deq 
\tilde{h}(Z,\theta) \defeq
\mathsf{one\_hot}\left[ \argmax_i\set{\log(\theta_i) + Z_i} \right]
\end{align*}
 with  $Z_i$ being i.i.d samples drawn from Gumbel (0, 1), ensuring that 
  $A \sim \pi_\theta$.

 \noindent Therefore, we can now express   $H(\theta) = \E_Z[G\left(h(Z,\theta) \right)]$, where $h(Z,\theta) = \tilde{h}(Z,\theta) \odot (a_i)_{i=1}^m $, the element-wise product  of $\tilde{h}(Z,\theta)$ and  the vector $(a_i)_{i=1}^m$. 
Since $h(Z,\theta)$ is not differentiable with respect to $\theta$, to enable pathwise gradient estimation, we consider a smoothed approximation $h_\tau(Z,\theta)$ of $ h(Z,\theta)$, as described in~\citep{JangGuPo17}:
\begin{align*}
h_\tau(Z,\theta)  
=
\left( \frac{\exp((\log (\theta_i)+Z_i)/\tau)}{\sum_{j=1}^{m}\exp((\log (\theta_j)+Z_j)/\tau)}\right)_{i=1}^m \odot (a_i)_{i=1}^m,
\end{align*} 
where $\tau$ is a temperature parameter controlling the smoothness of $h_\tau(Z,\theta)$. Thus, $$H(\theta)  \approx 
\frac{1}{N}\sum_{j=1}^N G(h_\tau(Z^{(j)},\theta)).$$
We now define the $\mathsf{Smoothed\text{-}Autodiff}$ estimator as
\begin{align}
  \hat{\nabla}^{\mathsf{grad}}_{\tau, N,\theta_i}
  \defeq \frac{1}{N}\sum_{j=1}^N\nabla_{\theta_i}G(h_\tau(Z^{(j)},\theta)).
\label{eqn:eq-def-autodiff}
\end{align}

We present our main result, Theorem~\ref{thm:multi}, for the $\mathsf{Smoothed\text{-}Autodiff}$ estimator defined in~\eqref{eqn:eq-def-autodiff} and \textsf{REINFORCE} estimator defined in~\eqref{eqn:eq-def-reinforce}.  Similar to Theorem~\ref{thm:binary},
the result below highlights 
that the \textsf{REINFORCE}  estimator will be worse
as compared to the $\mathsf{Smoothed\text{-}Autodiff}$ estimator
if the number of samples $N$ is not   large enough.
The proof is provided in Section~\ref{sec:proof-thm-multi}:



\begin{theorem}
Assuming
$G(\cdot)$ is differentiable and  sufficiently smooth such that
$|G'(a)| \le \bar{G}$ for all $a \in \mathsf{conv}(a_1,...,a_m)$. 
Fix $i \in [m]$.
For $\theta \in [0,1]^m$, any $\epsilon>0$ and 
\begin{align*}
N \le 
\frac{1}{\paran{\nabla_{\theta_i} \E_{A\sim \pi_\theta}(G(A))+\epsilon}^2}
\paran{
\frac{\sum_{j=1}^m\theta_j(G(a_j))^2}{(\sum_{j=1}^m\theta_j)^3} + \frac{(G(a_i))^2}{\theta_i\sum_{j=1}^m \theta_j} - 2 \frac{(G(a_i))^2}{(\sum_{j=1}^m \theta_j)^2}  - \left(\nabla_{\theta_i} \E_{A\sim \pi_\theta}(G(A))\right)^2},
\end{align*}
there exists $\Tilde{\tau}$ depending on $ (N, \theta)$ such that 
\begin{align*}
\mse(\hat{\nabla}^{\mathsf{grad}}_{\Tilde{\tau},N,\theta_i})
\le 
\left(\nabla_{\theta_i}\E_{A\sim \pi_\theta}(G(A))\right)^2 + {\epsilon}
\le 
\mse(\hat{\nabla}^{\mathsf{RF}}_{N,\theta_i}).
\end{align*}

Additionally, for any $N$, $\theta_i = \frac{1}{k}$, and $k \to \infty$, we have the following
\begin{align*}
\mse(\hat{\nabla}^{\mathsf{RF}}_{N,\theta_i}) & = \Omega(k),
 \mse(\hat{\nabla}^{\mathsf{grad}}_{\tilde{\tau},N,\theta_i})
 \le 
\left(\nabla_{\theta_i} \E_{A\sim \pi_\theta}(G(A))\right)^2.
\end{align*}

The same statement holds for $\theta_i =  1-\frac{1}{k}$ as well. This implies that the mse of $\hat{\nabla}^{\mathsf{RF}}_{N}$ is unbounded, while the mse of gradient estimator is bounded.
\label{thm:multi}
\end{theorem}



\subsection{Empirical comparison of gradient estimators}

\label{sec:empirical_gradient_estimation_quality}

In  the preceding theoretical analysis, we demonstrate  that $\mathsf{Smoothed\text{-}Autodiff}$ gradients can outperform \textsf{REINFORCE} gradients and there exists a temperature for which the mean squared error (MSE) of the $\mathsf{Smoothed\text{-}Autodiff}$ estimator is lower than that of the \textsf{REINFORCE} estimator.  While it is theoretically challenging to precisely characterize the difference between the $\mathsf{Smoothed\text{-}Autodiff}$ and \textsf{REINFORCE} gradients for varying $N$ and temperature $\tau$, we provide an empirical analysis of the gradient estimation within our adaptive labeling framework to further assess this comparison.

As discussed, we denote the true gradient by $\nabla_{\theta}H(\theta)$.  For any gradient estimator $\hat{\nabla}_N $ (either $\hat{\nabla}_{\tau,N}^{\mathsf{grad}}$ or $\hat{\nabla}_N^{\mathsf{RF}}$), we measure the following performance metric, the cosine similarity, with the true gradient, denoted by $\cosim(\hat{\nabla}_N)$:
\begin{align*}
\cosim(\hat{\nabla}_N) = \E[\cos(\hat{\nabla}_N, \nabla_{\theta}H(\theta))],
\end{align*}
where expectation is taken over the randomness of the $\hat{\nabla}_N$.
Cosine similarity measures the alignment of direction between $\hat{\nabla}_N$ and $\nabla_{\theta}H(\theta)$, capturing both the bias and variance of the gradient estimators. 
The \textsf{REINFORCE} estimator is unbiased but suffers from high variance. As a result 
$\E [\cos(\hat{\nabla}_N^{\mathsf{RF}}, \nabla_{\theta}H(\theta))]$ might be small, even though $\E[\hat{\nabla}_N^{\mathsf{RF}}]= \nabla_{\theta} H(\theta)$.  
For the  $\mathsf{Smoothed\text{-}Autodiff}$  gradient, the $\cosim$ metric $\E [\cos(\hat{\nabla}_{N,\tau}^{\mathsf{grad}}, \nabla_{\theta}H(\theta))]$ is influenced by two factors. First, the  bias in the $\mathsf{Smoothed\text{-}Autodiff}$ gradient,  $\E (\hat{\nabla}_{N,\tau}^{\mathsf{grad}}) - \E (\nabla_{\theta}H(\theta))$, increases as $\tau$ increases. 
Second, the variance of the $\mathsf{Smoothed\text{-}Autodiff}$ gradient, $\V(\hat{\nabla}_{N,\tau}^{\mathsf{grad}})$,  decreases as $\tau$ increases. Our objective is to determine whether a favorable trade-off between bias and variance can be achieved, resulting in higher $cosim$ values and improved gradient alignment for the $\mathsf{Smoothed\text{-}Autodiff}$ gradient.

\begin{figure}
\centering
\begin{minipage}[b]{0.32\textwidth}
\centering
\includegraphics[height=4cm, width=5cm]
{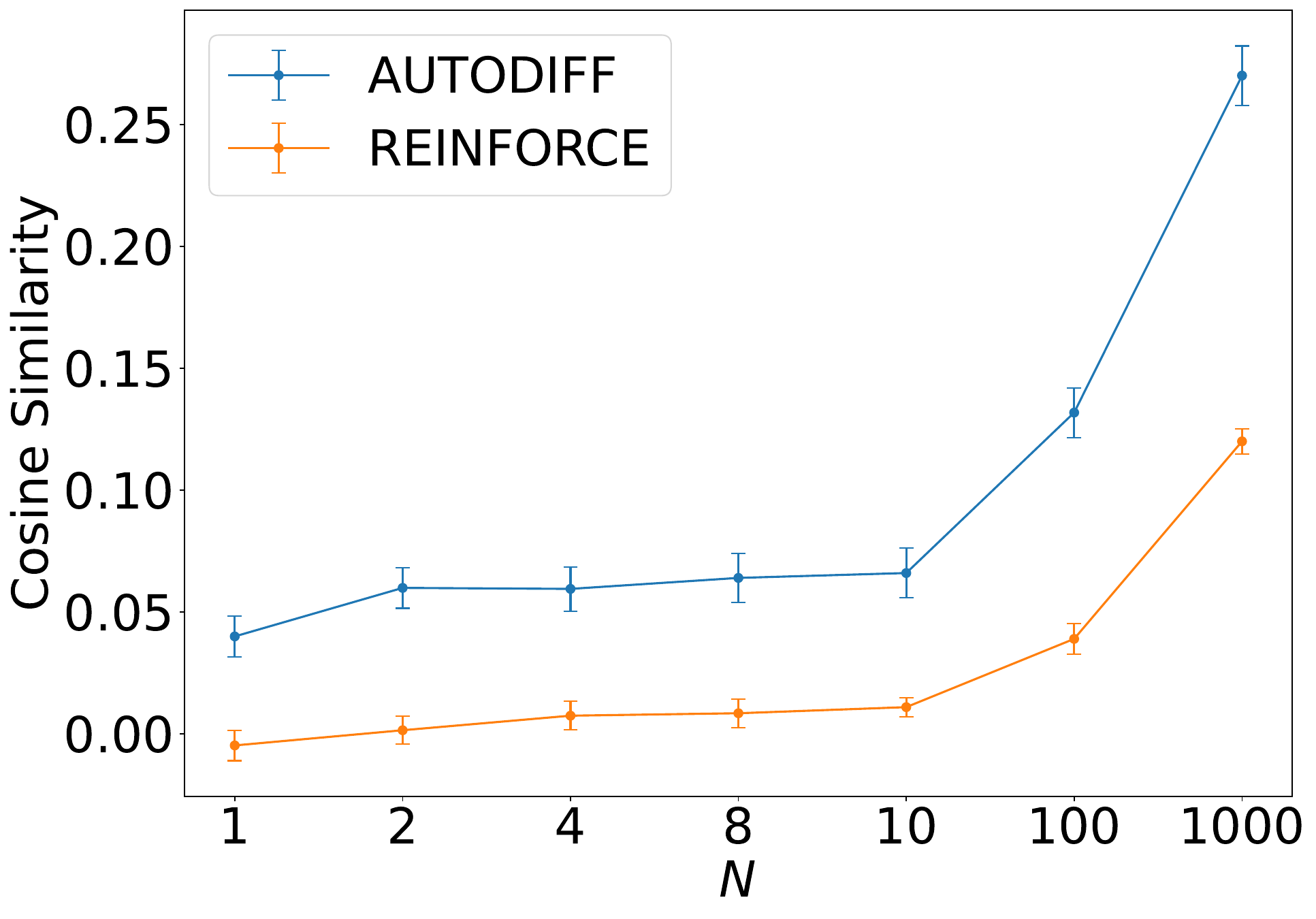}
\centering{\small{{$\tau=0.1$ }}}
\end{minipage}
\hfill
\begin{minipage}[b]{0.32\textwidth}
\centering
\includegraphics[height=4cm, width=5cm]{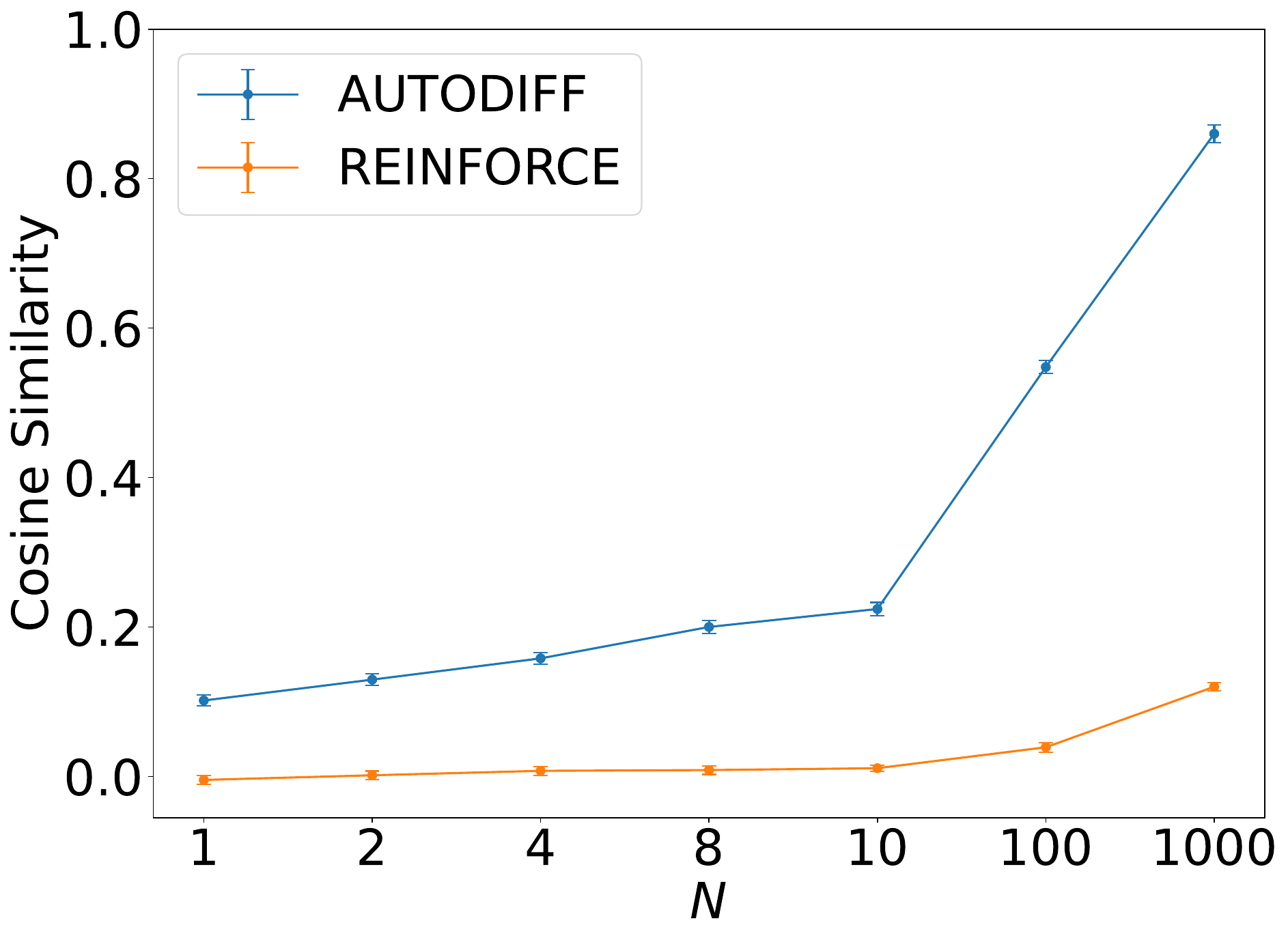}
\centering{\small{ $\tau=0.5$}}
\end{minipage}
\hfill
\begin{minipage}[b]{0.32\textwidth}
\centering \includegraphics[height=4cm, width=5cm]{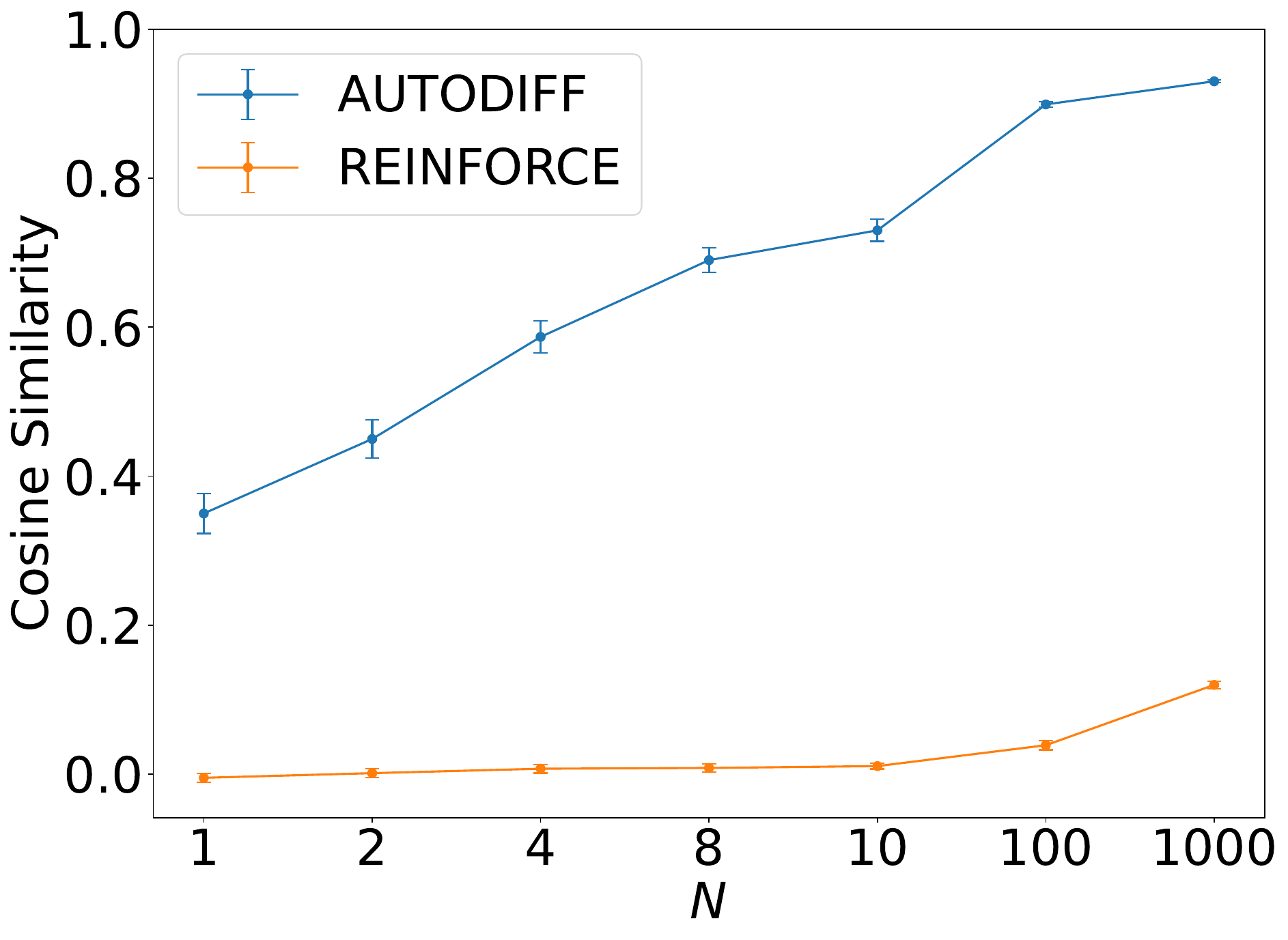}
\centering{\small{$\tau=10$}}
\end{minipage}
\caption{Comparison of $\mathsf{Smoothed\text{-}Autodiff}$ and \textsf{REINFORCE} estimators under different temperature ($\tau$) hyperparameter for the $\mathsf{Smoothed\text{-}Autodiff}$ gradient - metric used is cosine similarity}
\label{fig:compare-grad-cos}
\end{figure}

In the below mentioned results, we compare the $\mathsf{Smoothed\text{-}Autodiff}$ gradients and  \textsf{REINFORCE} gradients in the setting discussed in 
Figures~\ref{fig:var-l2-sim} and~\ref{fig:mean-l2-sim}.  
To approximate the ground truth gradient $\nabla_{\theta}H(\theta)$, we use the unbiased \textsf{REINFORCE} gradient estimated over many samples (in our case, $10^6$ samples). We evaluate both  $\mathsf{Smoothed\text{-}Autodiff}$ and \textsf{REINFORCE} gradients for different values of $N$. 
To estimate the expected values of cosine similarity, each gradient estimator was computed $200$ times. Figure~\ref{fig:compare-grad-cos} compares the $\mathsf{Smoothed\text{-}Autodiff}$ estimator and the \textsf{REINFORCE} estimator for different $N$.
We observe that for different $N$, pathwise estimator outperforms the \textsf{REINFORCE} estimator and has higher cosine similarity.
Moreover, \textsf{REINFORCE} estimator takes almost 100-1000 times more samples to reach the same level of cosine similarity.





\subsection{Impact of gradient estimation on algorithmic performance}
\label{sec:gradient_estimation_algorithmic_performance}

\begin{figure}[t]
\centering
\begin{minipage}[b]{0.49\textwidth}
\centering
\includegraphics[width=\textwidth, height=6cm]{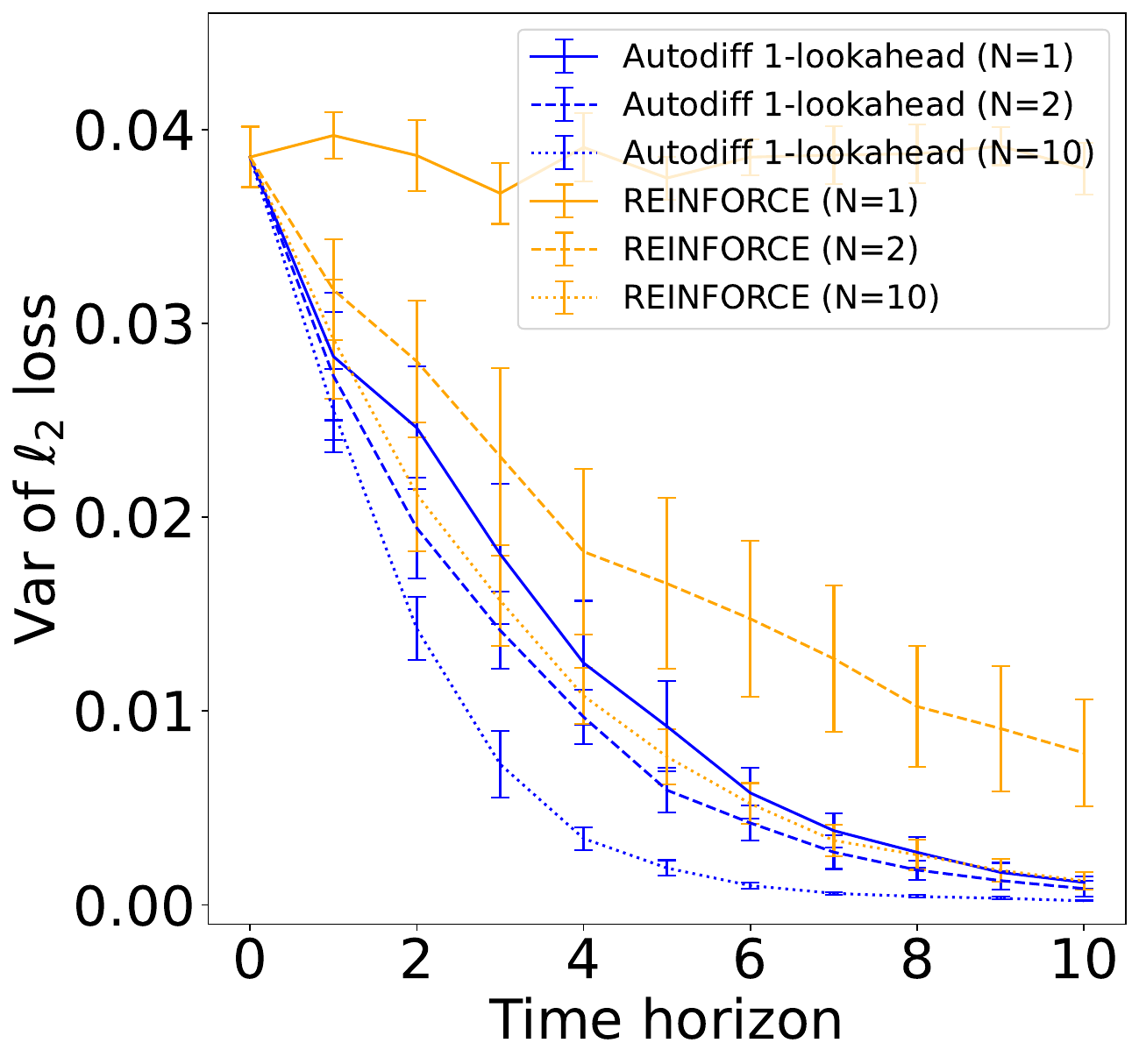}
\caption{Comparing the impact of $\mathsf{Smoothed\text{-}Autodiff}$ and \textsf{REINFORCE} gradient estimators (with different values of $N$) on the performance of  \ouralgo. (Synthetic data) Variance of mean squared loss evaluated through the posterior belief $\mu_t$ at each horizon $t$.}
\label{fig:var-l2-sim-in-N}
\end{minipage}
\hfill
\begin{minipage}[b]{0.49\textwidth}
\centering \includegraphics[width=\textwidth, height=6cm]{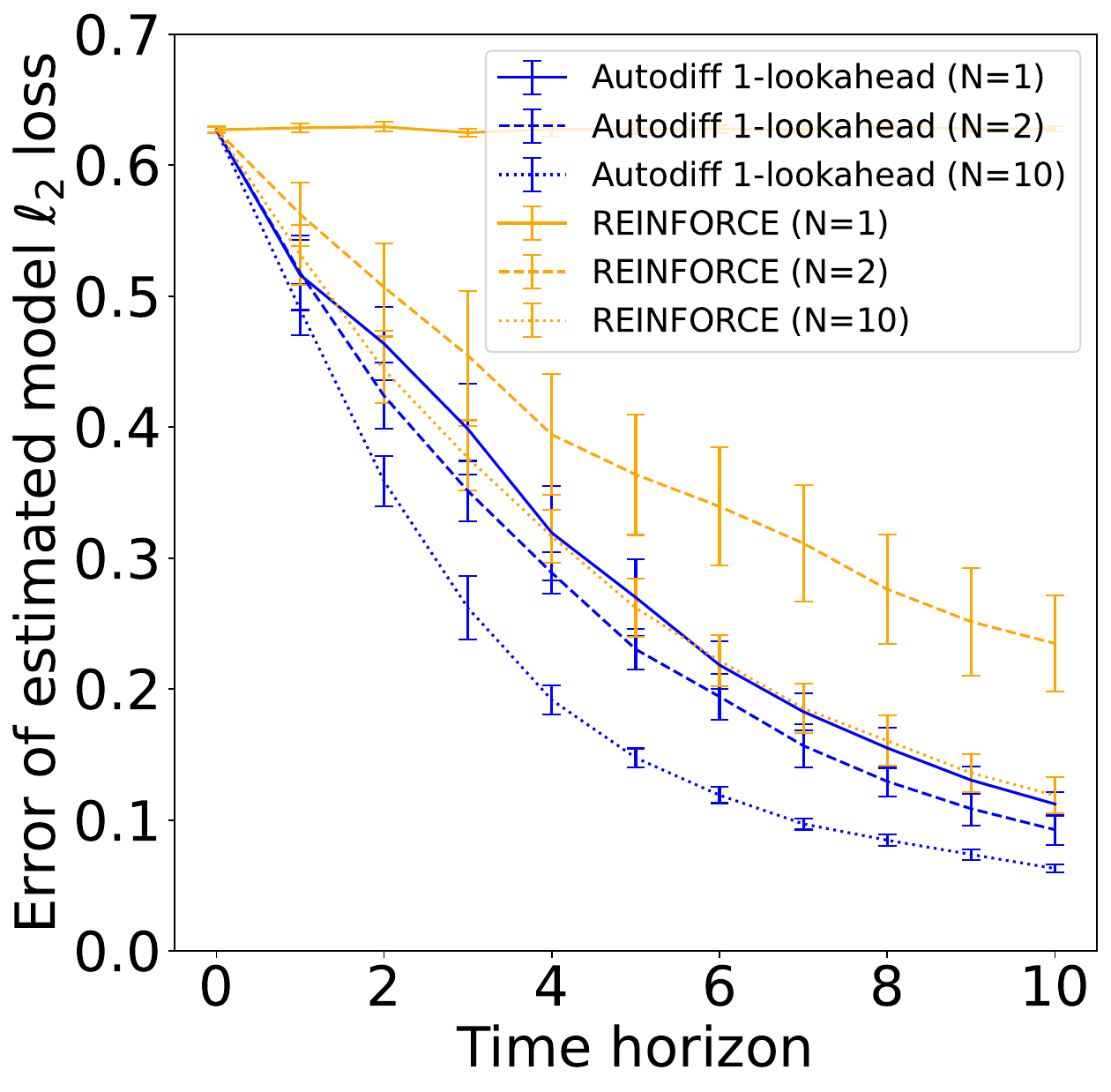}
\caption{Comparing the impact of $\mathsf{Smoothed\text{-}Autodiff}$ and \textsf{REINFORCE} gradient estimators (with different values of $N$) on the performance of  \ouralgo. (Synthetic data) Error between MSE calculated based on collected data $\mc{D}^{0:T}$ vs. population oracle MSE over $\mc{D}_{\rm eval} \sim P_X$.}
\label{fig:mean-l2-sim-N}
\end{minipage}
\end{figure}

We now compare the performance of $\mathsf{Smoothed\text{-}Autodiff}$  and \textsf{REINFORCE} gradients in the downstream optimization task of adaptive labeling, examining the effect of varying the number of samples $N$ used in the gradient estimator. Our results show that gradient quality directly impacts downstream task performance. In Figures~\ref{fig:var-l2-sim-in-N} and~\ref{fig:mean-l2-sim-N}, we   replicate the setting from 
Figure~\ref{fig:var-l2-sim} and~\ref{fig:mean-l2-sim}  in Section~\ref{sec:experiment-plan-GP}. As illustrated, $\mathsf{Smoothed\text{-}Autodiff}$ gradients perform exceptionally well even with a single sample in the estimator, with only marginal gains as $N$ increases due to limited room of improvement. In contrast, \textsf{REINFORCE} performs poorly with one sample but gradually improves as the sample size increases.


\section{Practical considerations in implementing our framework}
\label{sec:practical_consideration}

As highlighted in our previous discussion, the success of our framework relies on  two critical components.  
First, the uncertainty quantification methodology must provide reliable and efficiently computable posterior updates. 
Second, the computational efficiency of the autodifferentiation software framework is crucial, 
as it enables end-to-end gradient estimation through the entire pipeline. 
This efficiency will be particularly important in multi-step lookahead scenarios. 
In this section, we highlight practical considerations associated with these two components, 
and discuss  remaining challenges and corresponding promising future research directions. 
We begin by evaluating the reliability of current UQ methodologies in Section~\ref{sec:eval_posterior_consis}, followed by a discussion on the efficiency of the autodiff framework in Section~\ref{sec:auto-diff-bottleneck}.

\subsection{Evaluating Posterior Consistency of Uncertainty Quantification Methods}
\label{sec:eval_posterior_consis}
The reliability of a UQ methodology depends on two key properties i) reliable UQ performance on out-of-distribution (OOD) inputs, and ii) the ability to sharpen beliefs reliably as more data is gathered. We refer to these two properties collectively as  ``posterior consistency''.


Most existing literature on UQ evaluation  focuses on static single-shot settings~\citep{OsbandWenAsDwIbLuRo23}, with some literature on evaluating UQ performance on OOD data~\citep{OvadiaFeReNa19,Nadoal21}. To our knowledge, none of these literature evaluates UQ in dynamic settings, which is critical for applications in active learning and reinforcement learning settings. 
In this section,  we focus on assessing whether a UQ model, 
even  
when retrained perfectly, performs posterior updates consistently or not—--specifically, whether the posteriors are sharp on observed data and provide reliable UQ on unobserved data. 
Since UQ is foundational to trustworthy AI and active exploration in reinforcement learning, our empirical insights may be of independent interest to researchers in these domains.  

Our experiments reveal several challenges that current UQ methodologies suffer from. Due to the variation in UQ quality across hyperparameters, we observe a clear trade-off between in-distribution (ID) and out-of-distribution (OOD) performance. 
The performance disparity between ID and OOD performance is intricately tied to canonical model training practices that extend beyond the conceptual UQ paradigms. In particular, we observe that implementation details such as weight-decay, prior scales (where applicable), and stopping times have an outsized influence on the performance of UQ methodologies. Moreover, we highlight
the unreliability of typical approximate posterior update heuristics,
which rely on gradient updates to the underlying network(s). 
We observe large variation in the quality of the posterior across similar implementation details and even random seeds,
underscoring the importance of further methodological development along these dimensions. 
These challenges magnify in the dynamic setting where posteriors are updated with new observations, highlighting the need for reliable posterior update methods---an often  overlooked dimension in UQ algorithmic development predominantly focused on static evaluations.  

We demonstrate the above discussion using the eICU dataset. Additional results from other datasets, along with further details on the evaluated UQ methodologies, the selection biases introduced, and the metrics used to assess UQ performance, are provided in
Section~\ref{sec:practical_consideration_experiment_details}.
 
\begin{figure}
\centering
\begin{minipage}[b]{0.32\textwidth}
\centering
\captionsetup{labelformat=empty,labelsep=none}
\includegraphics[height=4.5cm]
{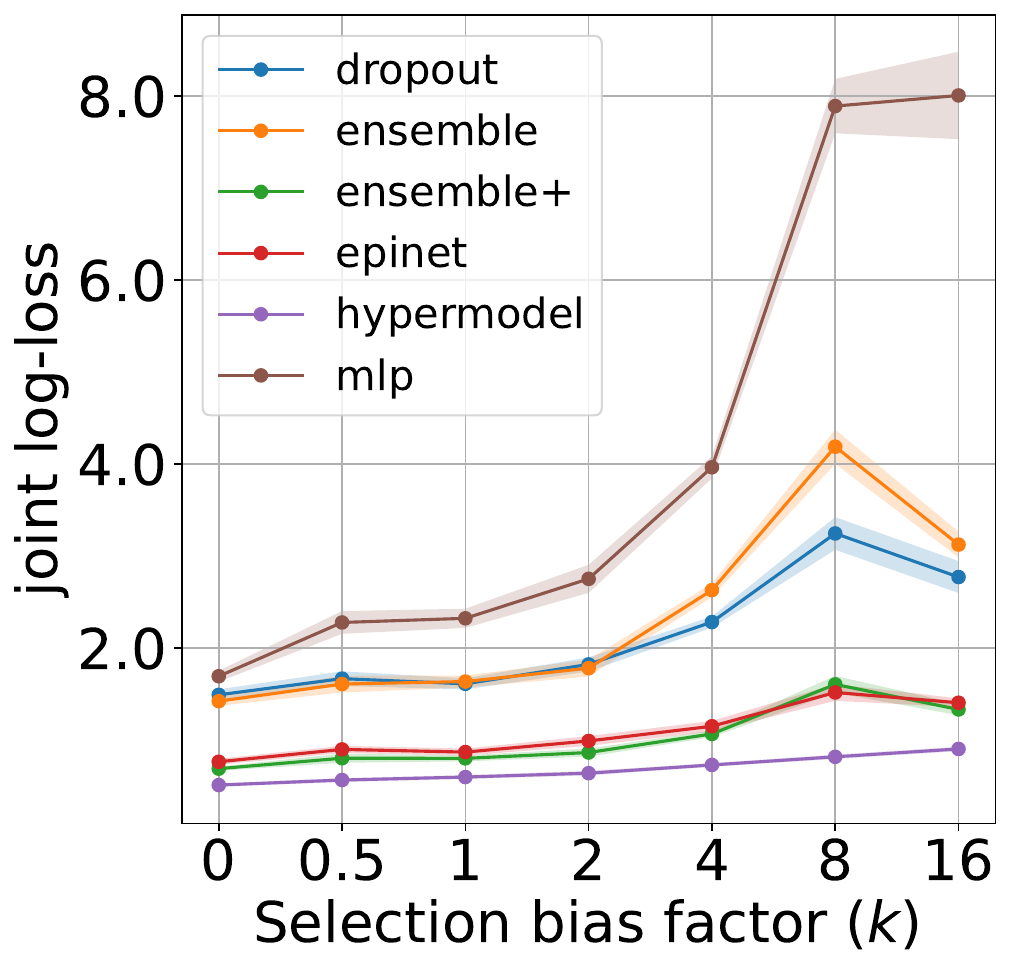}
{\small{\centering{eICU (Linear bias)}}}
\end{minipage}
\hfill
\begin{minipage}[b]{0.32\textwidth}
\centering
\includegraphics[height=4.6cm]{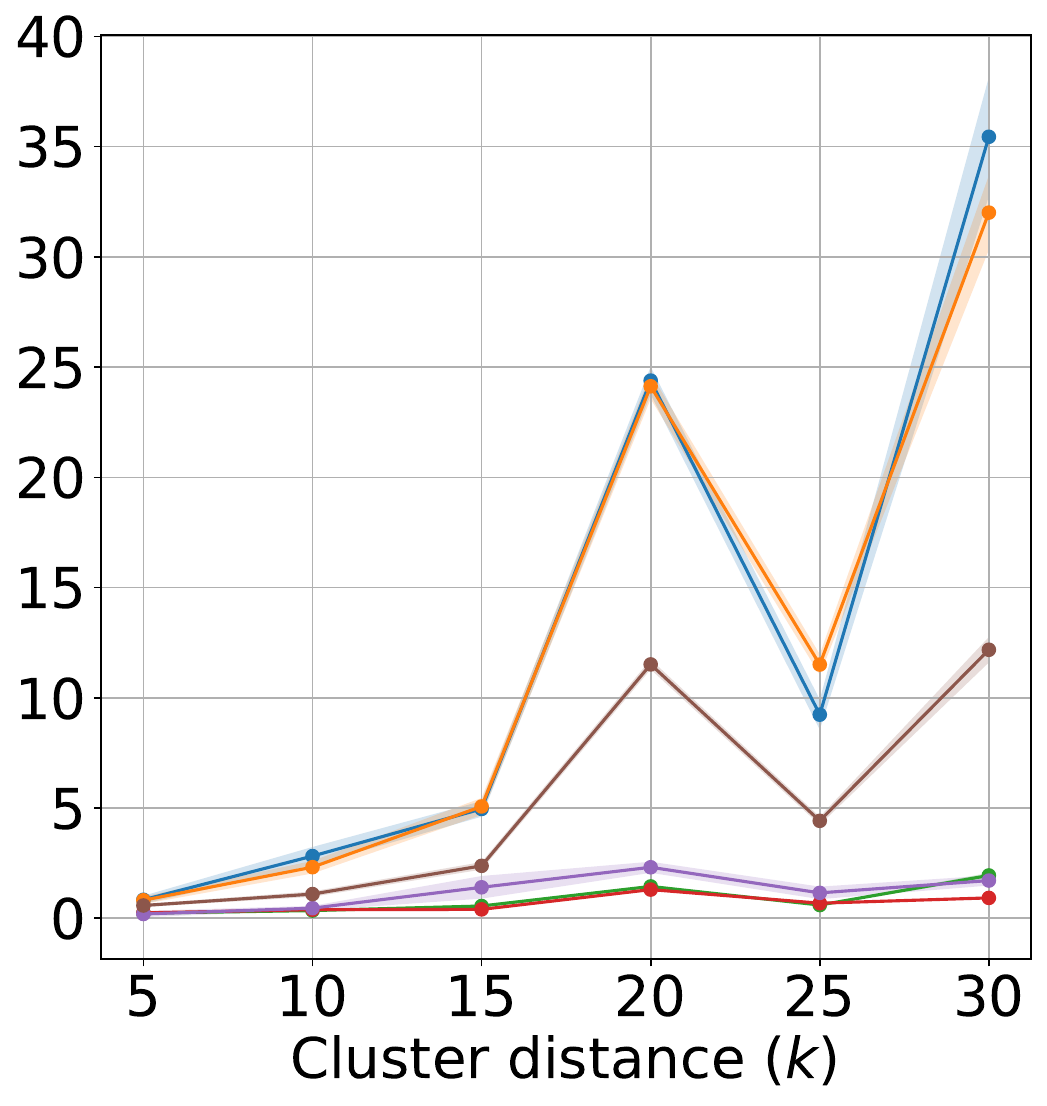}
{\small{ eICU (Clustering bias)}}
\end{minipage}
\hfill
\begin{minipage}[b]{0.32\textwidth}
\centering \includegraphics[height=4.5cm]{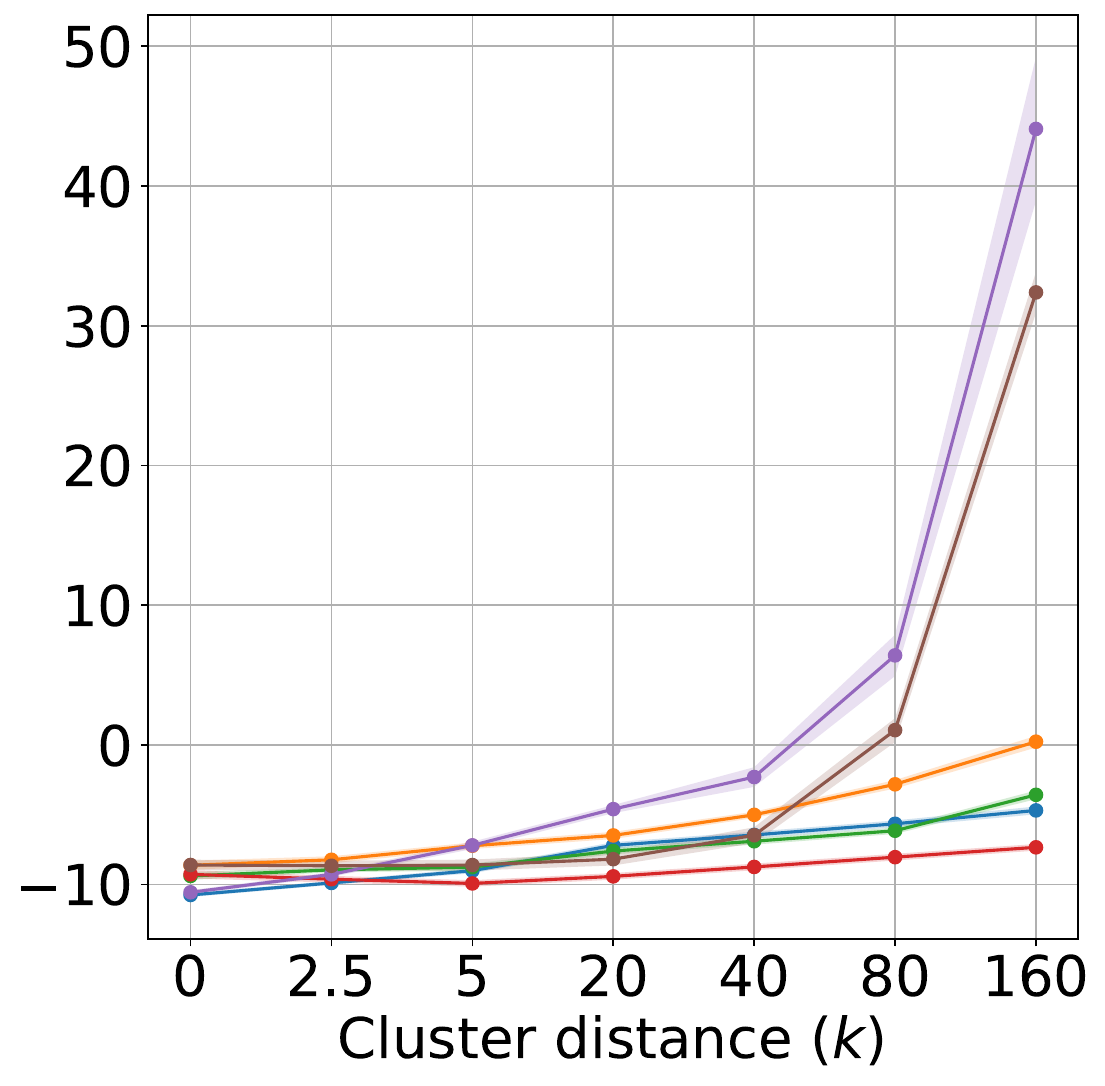}
{\small{ Synthetic (Clustering bias)}}
\end{minipage}
\caption{Joint log-loss on OOD data with increasing selection bias. 
Though some agents such as ensemble and dropout perform well under no selection bias, their performance deteriorates significantly with a large selection bias (larger $k$).}
\label{fig:task-1-joint-ood-all-data}
\end{figure}

\paragraph{Performance under distribution shifts} In Figure~\ref{fig:task-1-joint-ood-all-data}, we compare the performance of different UQ methodologies on  out-of-distribution data under increasing levels of  \emph{linear} and \emph{out-of-support} selection bias. The methodology to introduce selection bias is detailed later in Section~\ref{sec:selection-bias}.
 Figure~\ref{fig:task-1-joint-ood-all-data} shows that most UQ methodologies are sensitive to selection bias, with performance often degrading  significantly on OOD data, while in-distribution performance remains stable.  These results emphasize the importance of evaluating UQ models beyond the data it has seen so far.

\begin{figure}[h]
\centering
\begin{minipage}[b]{0.32\textwidth}
\centering
\includegraphics[height=4.5cm]{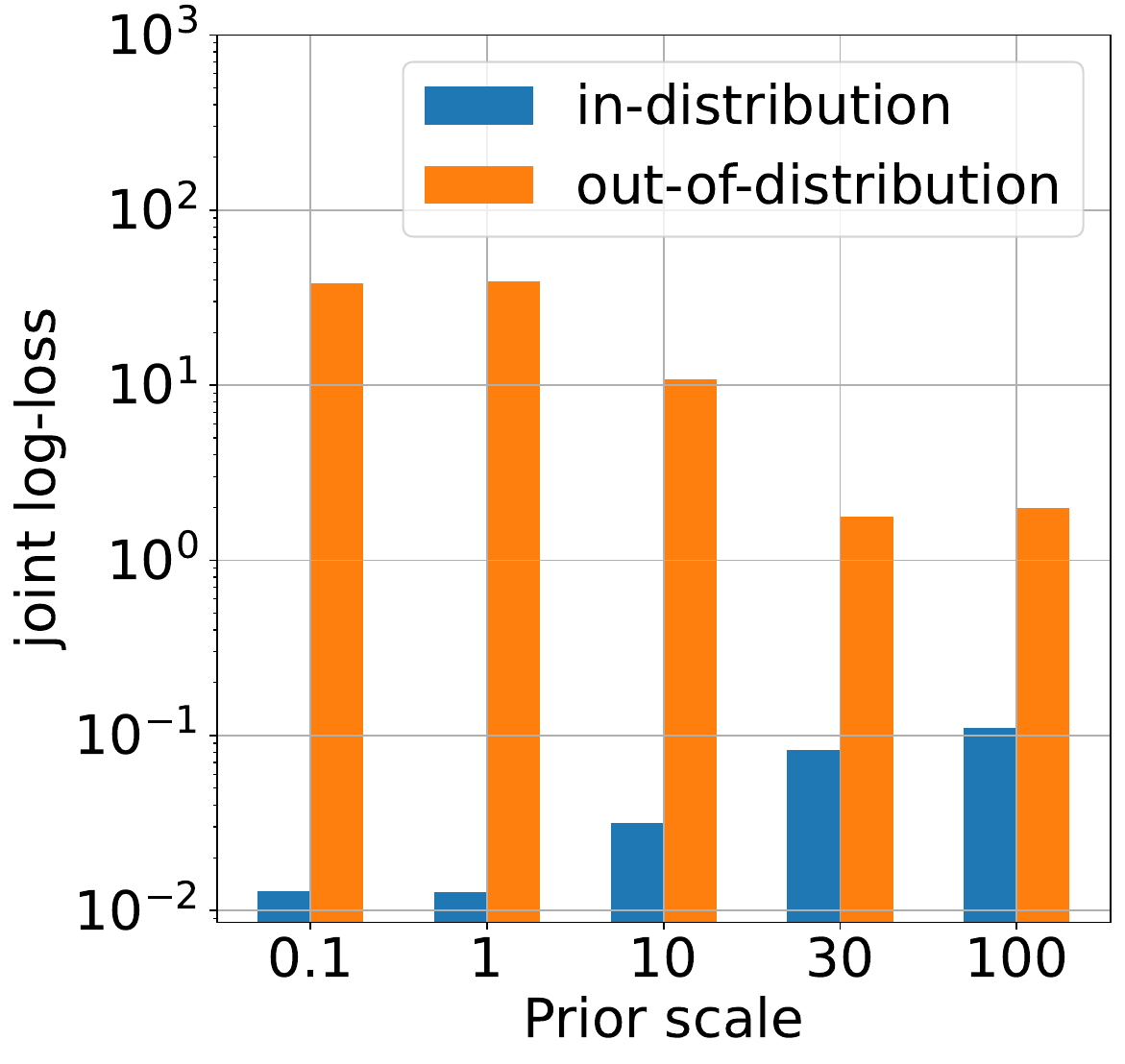}
{\small{\centering Ensemble$+$}}
\end{minipage}
\hfill
\begin{minipage}[b]{0.32\textwidth}
\centering \includegraphics[height=4.5cm]{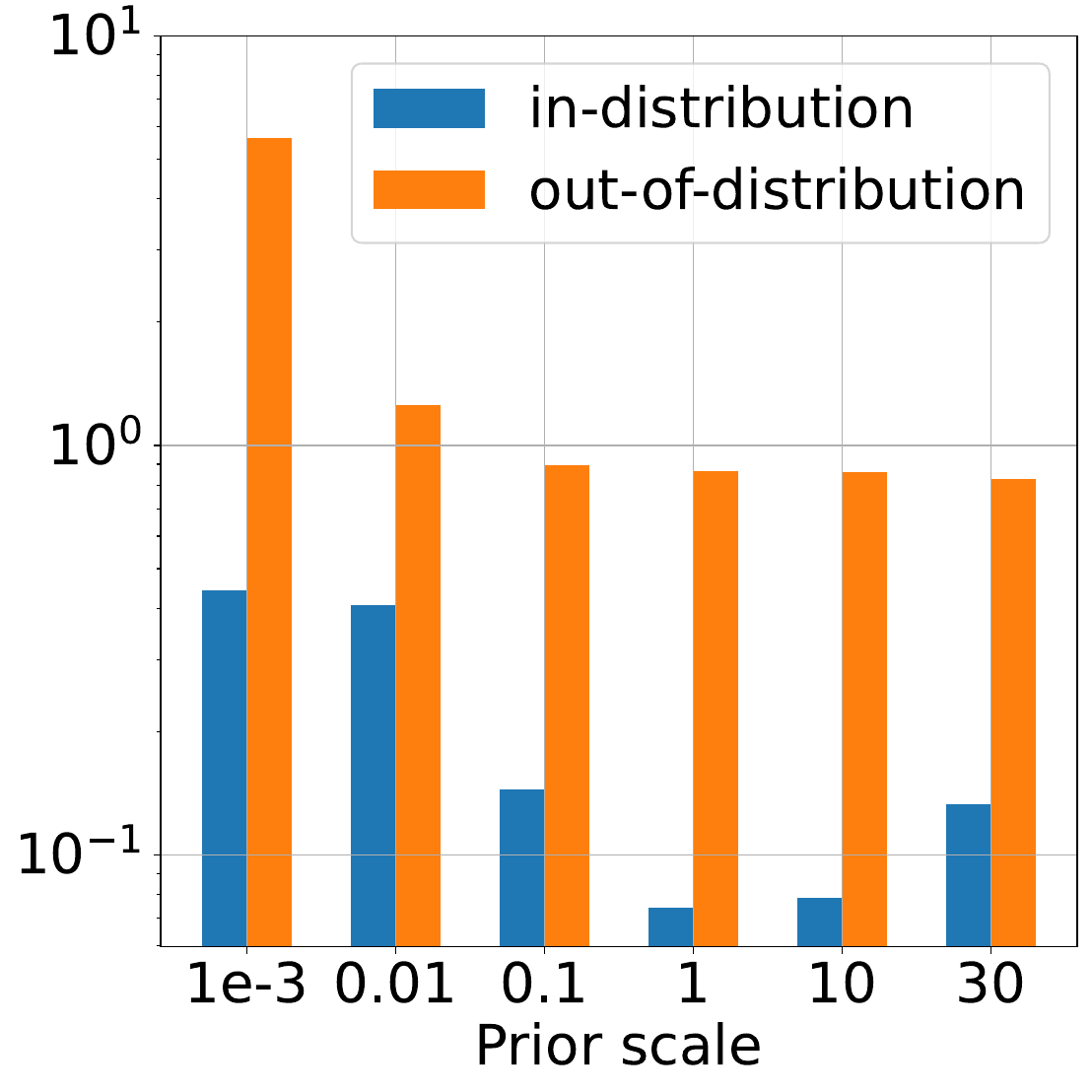}
{\small{\centering Epinet}}
\end{minipage}
\hfill
\begin{minipage}[b]{0.32\textwidth}
\centering \includegraphics[height=4.5cm]{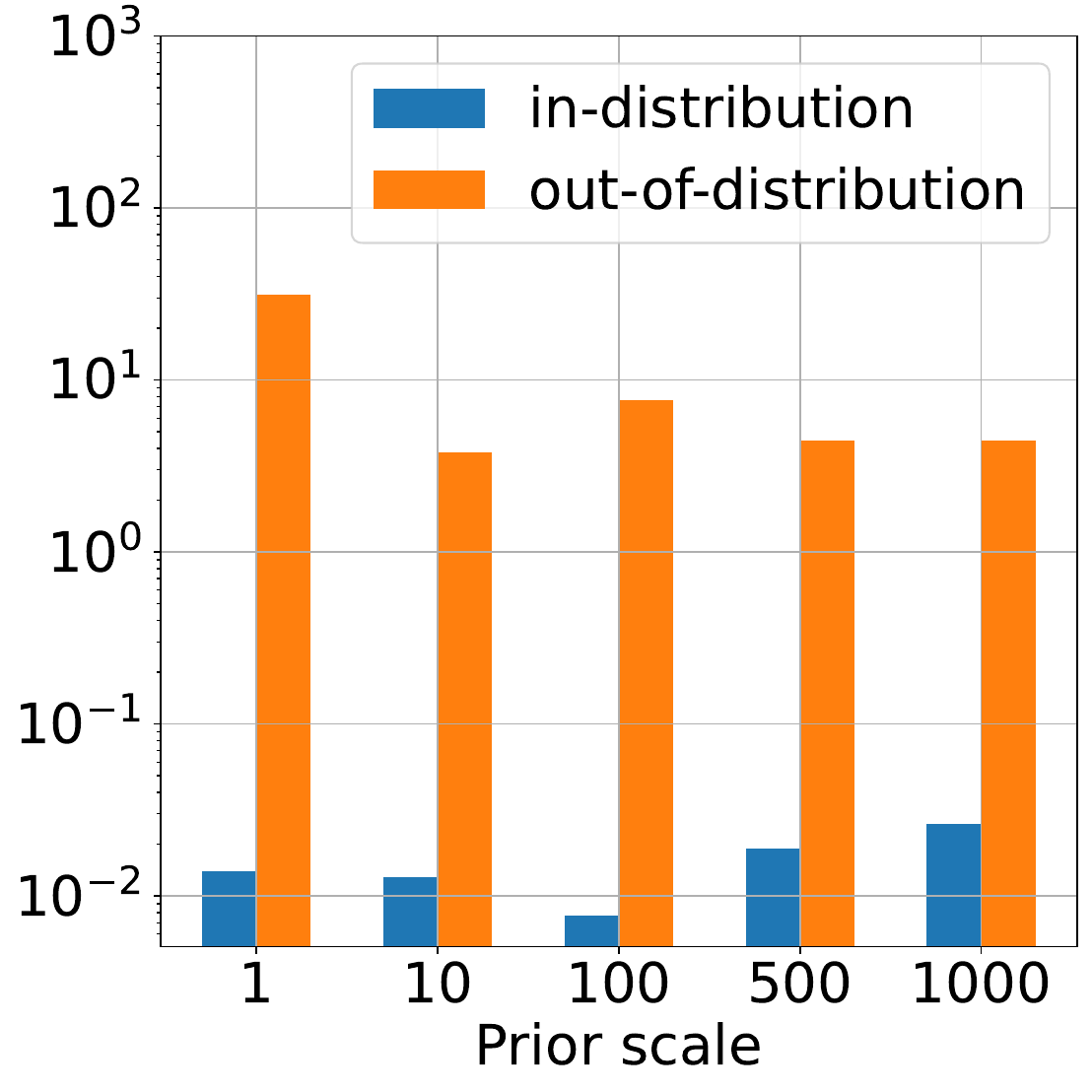}
{\small{\centering Hypermodel}}
\end{minipage}
\caption{Trade-off between  in-distribution and out-of-distribution joint log-loss with prior scale as the hyperparameter (eICU, Clustering bias).} \label{fig:difficult_to_choose_prior_scale}
\end{figure}

\paragraph{Hyperparameter tuning breaks under distribution shifts} 
UQ methodologies such as \ensembles/ \ensembleplus and Epinet require extensive hyperparameter tuning to perform well. Canonically, these hyperparameters are tuned by splitting the available dataset, $\mc{D}_0$, into a training and validation sets, and choosing the hyperparameters that yield the best results on the validation set.
However, we demonstrate that this approach has a significant limitation when the data experiences distribution shifts: the best hyperparameters for in-distribution (ID) data are not necessarily the best for out-of-distribution (OOD) test data. 
Moreover, in many cases, there is a unexpected trade-off between ID and OOD performance. In Figure~\ref{fig:difficult_to_choose_prior_scale}, we illustrate this phenomenon for  \ensembleplus, Epinets, and Hypermodels 
when tuning the Prior Scale hyperparameter (we provide a detailed explanation of each methodology in Section \ref{sec:UQ_bench_benchmark-uq-metric}). We observe that reducing prior scale for \ensembleplus improves ID  performance while the OOD performance deteriorates. Similar trends were noted for other hyperparameters, such as weight decay and stopping time, as shown in Figures~\ref{fig:difficult_to_choose_weight_decay} and~\ref{fig:difficult_to_choose_stopping_time} in Section \ref{sec:practical_consideration_experiment_details}.

\begin{figure}[h]
\centering
\begin{minipage}[b]{0.24\textwidth}
\centering
\includegraphics[width = \textwidth]{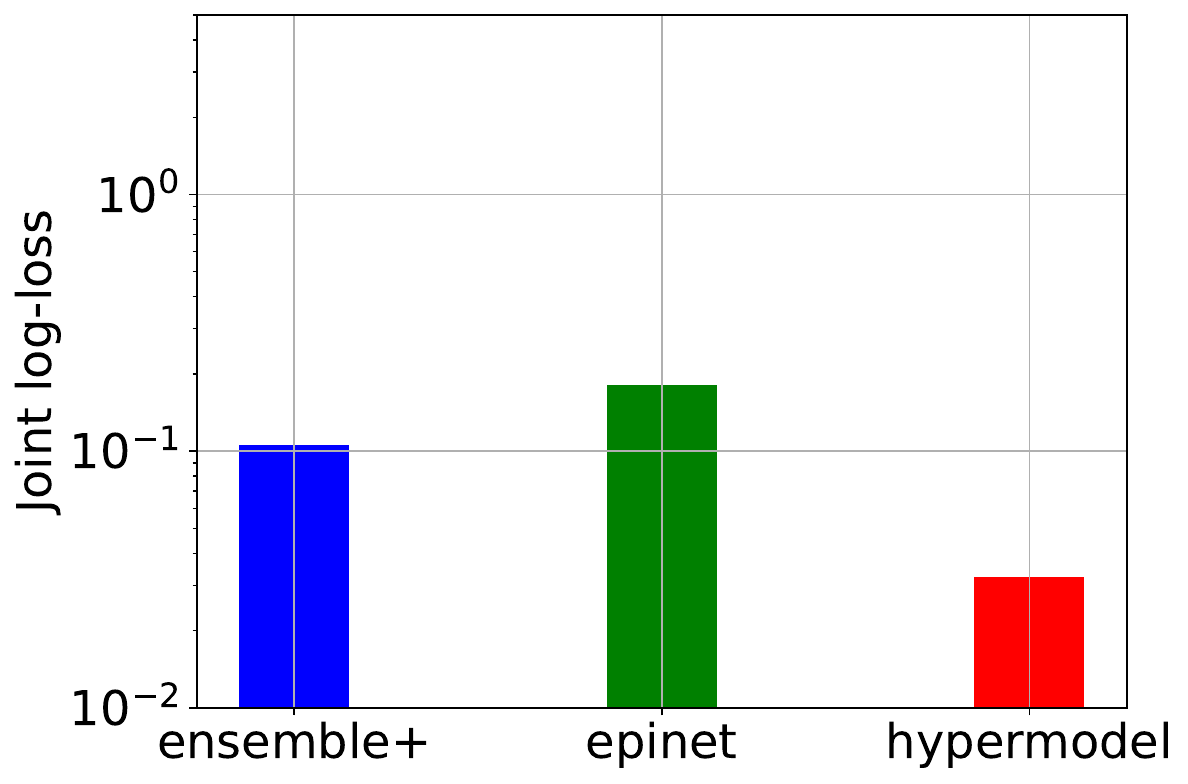}

{\small{{(a)} ID performance ($T=0$) }}
\end{minipage}
\hfill
\begin{minipage}[b]{0.24\textwidth}
\centering \includegraphics[width = \textwidth]{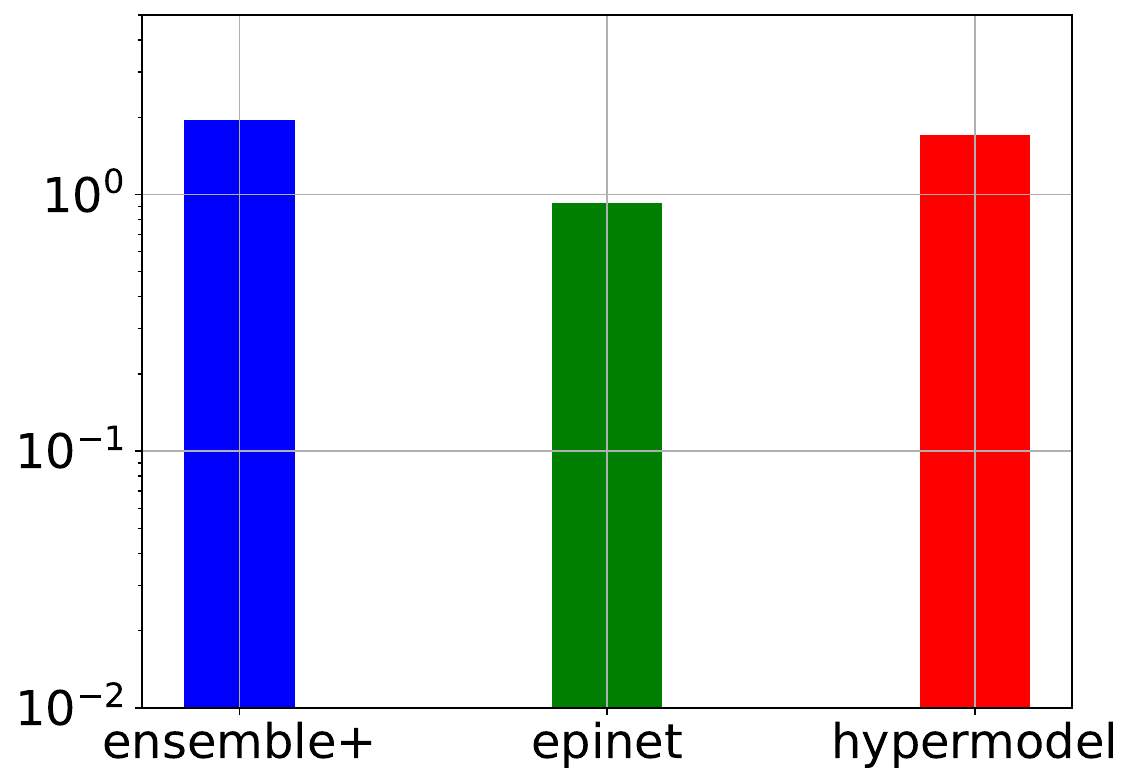}
{\small{{(b)} OOD performance, ($T=0$) }}
\end{minipage}
\hfill
\begin{minipage}[b]{0.24\textwidth}
\centering \includegraphics[width = \textwidth]{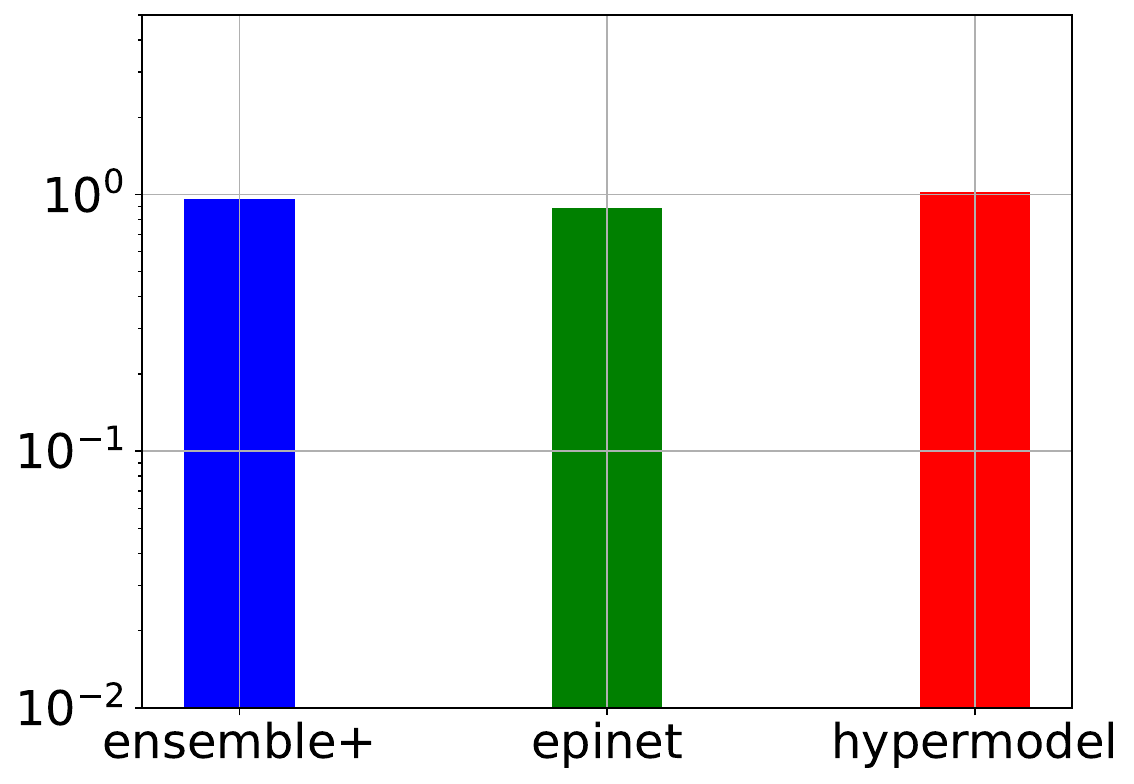}
{\small{{(c)}~ OOD performance, ($T=1$)}}
\label{fig:eicu_clustering_improvement-k=30}
\end{minipage}
\hfill
\begin{minipage}[b]{0.24\textwidth}
\centering \includegraphics[width = \textwidth]{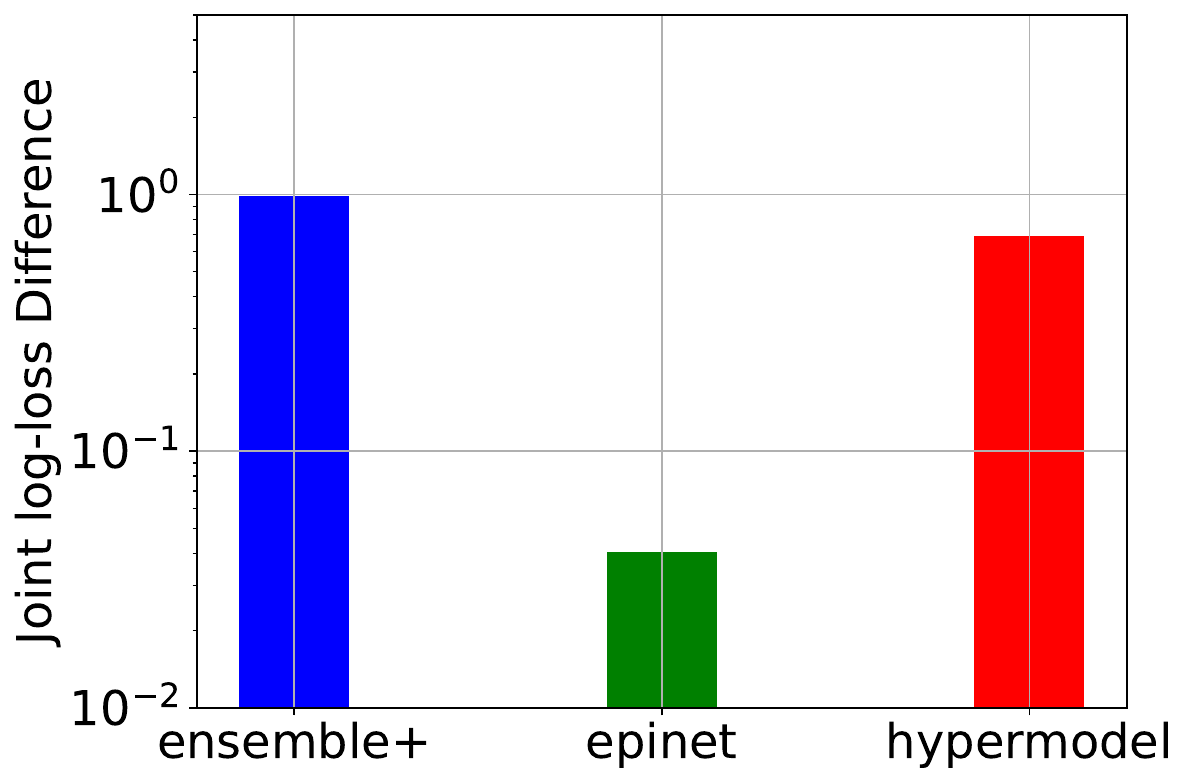}
{\small{{(d)} ~ OOD improvement ($T=0 \to 1$) }}
\end{minipage}
\caption{Performance of different UQ modules in a dynamic setting. \textbf{Experimental Setup:} Population is distributed across 2 clusters.  At $T=0$, agent  observes data $\mc{D}_0$ sampled from Cluster 1. 
Figure (a) and Figure (b) show how different UQ models perform on test data from Cluster 1 (in-distribution) and Cluster 2 (out-of-distribution). At $T=1$, UQ agent observes some data $\mc{D}_1$  sampled from Cluster 2.  Figure (c) performance on test data from Cluster 2 at $T=1$. Figure (d)  showcase performance improvement on test data from Cluster 2 at $T=1$. \textbf{Observations:} Comparison between Figure (a) and Figure (b) signifies the trade-off between  the  in-distribution and out-of-distribution performance discussed previously. Figure (d) shows Epinets, in this case, are much worse at adapting to new data compared to the other two agents.}
\label{fig:dynamic_setting_k_30}
\end{figure}

\paragraph{Performance under dynamic settings} We observe that even static OOD performance does not comprehensively capture the reliability of a UQ method. 
It is crucial that a UQ method  sharpens its beliefs as new data is observed while  still maintaining uncertainty over the unobserved space.  We capture this by assessing UQ performance in  a  dynamic setting. 
Figure~\ref{fig:dynamic_setting_k_30} summarizes  the results of this experiment. 
Figure~\ref{fig:dynamic_setting_k_30} (a) and (b)  shows that at $T=0$, models with better performance on in-distribution data are worse on 
out-of-distribution data, highlighting the previously discussed trade-off. For instance, Epinets perform better on out-of-support data, their ID performance is worse compared to \ensembleplus and Hypermodels. Figure~\ref{fig:dynamic_setting_k_30} (c) and (d)  shows performance improvement after acquiring new data, allowing us to assess the sharpness of posterior updates for each model. 
Although Epinets initially excel on OOD data, their performance improvement (posterior update) on OOD data is not as sharp as that of Hypermodels and \ensembleplus. Additionally, hyperparameter tuning remains challenging in dynamic settings (see Section \ref{sec:practical_consideration_experiment_details} for further evaluations). While one might consider re-tuning hyperparameters with each new data acquisition, no consistent procedure currently exists to ensure reliable performance. Apart from being computationally intensive, even re-tuned hyperparameters cannot be reliably trusted in dynamic OOD scenarios, as previously discussed. These experiments reveal a significant limitation of current UQ methodologies: SOTA UQ methods suffer from a trade-off between achieving sharper posterior updates on the observed data and maintaining performance on OOD data. Furthermore, there is no reliable method to control this trade-off. This inability of UQ modules to effectively balance these competing demands underscores the need for new methodological advancements in UQ to support complex downstream tasks.



\vspace{-10pt}
\paragraph{Sensitivity of posterior estimates to implementation details} Apart from assessing the fundamental trade-offs  on OOD data and under dynamic setting, we also examine performance degradations under practical computational constraints.
Our experiments show that the current UQ methodologies are sensitive to early stopping and random seeds, we defer the experimental details to Section \ref{sec:UQ_BENCH_additional_experiment}. This sensitivity may limit the practical effectiveness of these UQ methodologies.
Our findings highlight the outsized impact of these factors on UQ performance, underscoring the importance of evaluating UQ methods within realistic computational constraints rather than in idealized settings.

\subsection{Auto-differentiability bottleneck} 
\label{sec:auto-diff-bottleneck}
Our smoothed-pathwise gradient methodology relies on the automatic differentiation (Autodiff) functionality of Pytorch/Jax. As we saw in Section \ref{sec:diff-piepline}, for \ouralgo, we use Autodiff to differentiate through the whole pipeline including soft $\batchsize$-subset sampling, posterior update and the objective evaluation.

Our implementation focused on one-step lookaheads, and extending it to longer horizons  involves efficiently differentiating through hierarchical (multi-level) optimization and accurately estimating Hessians, tasks for which the current autodifferentiation frameworks lack sufficient efficiency. Hessian estimation via autodifferentiation remains an active area of research~\citep{BaydinPeRaSi17,ShabanChHaBo18,Margossian19,LorraineViDu20,Amos22,BlondelBeCuFrHoLlPeVe22, ScieurGiBePe22,Alesiani23,KotaryDiFi23}, and exploring and adapting these advancements or developing new engineering solutions combined with innovative parallelization techniques, represents a promising avenue for future reseach.

Similarly, while the posterior update in GPs is differentiable, 
  deep learning-based UQ modules rely on a finite number of SGD steps for posterior updates.
Consequently, the Autodiff gradient through the posterior update  also requires differentiation through hierarchical (multi-level) optimization. Moreover, there is a trade-off between computational time
and the accuracy of the gradient approximation, depending on the number of training steps used in the posterior update. 
If the number of steps becomes large, estimation errors may increase due to round-off and vanishing gradient issues. 
This limitation, along with previously mentioned challenges such as performance on OOD datasets and in dynamic settings, 
restricts the applicability of 
current UQ methodologies.
Recent advancements, such as Bayesian 
Transformer~\citep{NguyenGr22, MullerHoArGrHu22}, 
offer
promising solutions by providing differentiable posterior updates 
without requiring gradient steps. 
Exploring this approach could be an interesting direction for further improvement.

\section{Conclusion and Future Work}\label{sec:conclusion}

Datasets often suffer from severe selection bias when labels are expensive, 
making it difficult to reliably evaluate AI models or estimate causal effects. 
We propose a novel framework for adaptive labeling 
to improve the reliability of measurement.
Following a Bayesian framework, we formulate an MDP over posterior beliefs on the estimand of interest (model performance or ATE)
and demonstrate that 
planning problems can be efficiently solved using pathwise policy gradients, 
computed through a carefully designed auto-differentiable pipeline. 
We show that even one-step lookaheads can yield substantial improvements over heuristic algorithms inspired by active learning. 
Additionally, our results suggest that $\mathsf{Smoothed\text{-}Autodiff}$ gradient estimation can outperform \textsf{REINFORCE}-based methods, 
with potential implications for policy gradients in model-based reinforcement learning more broadly.  

Our experiments highlight several important future research directions. UQ methodologies should possess certain key properties.
Foremost,
posterior updates  should be consistent. 
Additionally, these updates should be readily available and easily differentiable.  Recent advances in  Bayesian 
Transformers~\citep{NguyenGr22, MullerHoArGrHu22} present an intriguing dimension to explore in this regard. 
Another interesting direction involves enabling multi-step lookaheads by addressing the autodiff bottleneck.








\newpage
\begin{appendix}

\section{Smoothing the recall objective} \label{sec:details-recall-smoothing}
 
Performance metric, such as the \texttt{Recall} of a model $\model(.)$,  is not differentiable. Recall that $  H(\theta) \defeq \E_{{\ybpool \sim \mu; S \sim \pi_\theta } }  [G\paran{\mu_+^S}]$,  where $G(\mu_+^S) = \V_{f \sim \mu_+^S}  g(f)$, and $g(f) \defeq  \E_{X \sim P_X}\left[ \E_{Y \sim p(\cdot|f,X) } \Big[\indic{\model(X)>0}|Y=1\Big]\mid f \right]$,  with $\mu_T$ depending on $\pi_\theta$. To estimate $\nabla_\theta H(\theta)$  in a differentiable manner, in addition to approximately sampling 
$\bm{a}(\theta)$ using Algorithm \ref{alg:k-subset},  we also need to apply a smooth approximation to $g(f)$. This can be  achieved using the softmax trick~\citep{JangGuPo17}, where we draw $2n$ i.i.d. Gumbel$(0,1)$ samples $R_{i}^{l}$ with $l \in \set{1,2}$. 
Define:
\begin{align*}
g_{\tau}(f; \bm{R}) & \defeq \frac{\sum_{i=1}^n  Y^{\tau}_i(f; \bm{R})  \model(X_i)}{\sum_{i=1}^n Y^{\tau}_i(f; \bm{R})}, \\
&\text{ where }
Y^{\tau}_i(f;\bm{R}) = \frac{\Exp{(\log f_i) + R_i^1) / \tau}}{
\Exp{(\log f_i) + R_i^1) / \tau}
+ \Exp{(\log (1-f_i)) + R_i^2) / \tau}
},
\end{align*} here  $\tau > 0$ is the temperature hyperparameter that controls the smoothing of $g_{\tau}(f; \bm{R})$. 
It is easy to show that $\E_{\bm{R}}[Y^{\tau}_i(f;\bm{R})] \approx f_i$,  making $g_{\tau}(f; \bm{R})$  a natural approximation of the previously defined \texttt{Recall} $g(f)$.


\section{Experimental details (Section \ref{sec:experiment})} \label{sec:details-experiments}

In this section, we provide detailed information about the experiments discussed in Section~\ref{sec:experiment}. 

\subsection{Planning with Gaussian Processes - synthetic data experiments}

As mentioned earlier, we generate our data using a  Gaussian Process ({GP}). Specifically, we use a {GP} with an RBF kernel: $f_i \sim \mathcal{GP}(m,\mc{K}) $, where $m(X)=0$ and $\mc{K}(X,X') = \sigma_f^2 \exp\left(-\frac{||X-X'||_2^2}{2\loss^2}\right)$, further Gaussian   noise $N(0,\sigma^2)$ is added to the outputs. We set $\loss = 1$, $\sigma_f^2 = 0.69$, and $\sigma^2 = 0.01$.
Recall that the marginal distribution $P_X$ consists of $51$ \textit{non-overlapping} clusters. To achieve this, we create a polyadic sampler, which first samples $51$ anchor points spaced at linearly increasing distances from the center to avoid overlap. These anchor points serve as our cluster centers. We sample the points around these anchor points by adding a small Gaussian noise $N(0,0.25)$. The training points are drawn from a single cluster, while the pool and evaluation points are drawn from all the $51$ clusters. The setup includes $100$ initial labeled data points, $500$ pool points and $285$ evaluation points used to estimate the objective. 

\noindent\textbf{\ouralgo~ training and hyperparameters:} As mentioned earlier, the underlying uncertainty quantification (UQ) module for these experiments is the {GP}. Although our algorithm does not know the true data generating function, but it has access to the {GP} hyperparameters. Therefore we use a $\mathcal {GP}$ with an RBF kernel: $f_i \sim \mathcal{GP}(m,\mc{K}) $, where $m(X)=0$ and $\mc{K}(X,X') = \sigma_f^2 \exp\left(-\frac{||X-X'||_2^2}{2\loss^2} \right)$, with 
Gaussian   noise $N(0,\sigma^2)$ added to the outputs. We set $\loss = 1$, $\sigma_f^2 = 0.69$ and $\sigma^2 = 0.01$. For soft $K$-subset sampling  (see Algorithm~\ref{alg:k-subset}),  we set $\tau$ to $0.1$. Further for evaluating the objective function $\V (g(f))$ - we take $100$ samples of $f(\xeval)$ from the posterior state $\mu_+^{a(\theta)}$. 
For policy optimization in each horizon, we use the Adam optimizer with a learning rate of $0.1$ to perform policy gradient steps over $100$ epochs. The results presented are averaged over $10$ different training seeds.

\noindent\textbf{Policy gradient (\rein) training and hyperparameters:} As mentioned earlier, all the algorithms have access to the true to {GP} hyperparameters. For evaluating the objective $\V (g(f))$ under a given action,  we take $100$ samples of $f(\xeval)$.  To optimize the policy in each horizon, we use the Adam optimizer with a learning rate of $0.1$ to perform policy gradient updates over $100$ epochs. The results presented are averaged over $10$ different training seeds.

\subsection{Planning with Gaussian Processes - real data (eICU) experiments}

The eICU  dataset is a healthcare dataset that contains data from various critical care units across the United States. To create a supervised learning setup from this dataset, we first extracted the $10$ most important features using a \texttt{Random Forest} classifier (number of trees $=$ $100$, criterion $=$ ``gini'').  The outcome variable was in-hospital mortality. Some examples of the extracted features include: Hospital length of stay, Number of days on the ventilator and the Last recorded temperature on Day 1 of ICU admission. We transformed the classification outcome variable (in-hospital mortality) into a regression task using the probability of in-hospital mortality predicted by the classifier.
To adapt the extracted data to our setting, we introduced selection bias in the data. We generated 51 clusters through the standard $k-$means algorithm. Our initial labeled data comes from only $1$ cluster, while the pool and evaluation data comes from all the $51$ clusters. Our dataset consists $100$ initial labeled data points, $500$ pool points, and $285$ evaluation points used to estimate the objective.

We then fitted a  {GP}  with an RBF kernel to above data using  {GP}-regression. The resulting {GP} hyperparameters were: $m(x)=0.255$, $\loss = 0.50, \sigma_f^2 = 1.0, \sigma^2 =0.0001$. This $\mathcal{GP}$ model serves as our uncertainty quantification (UQ) module.

\noindent\textbf{\ouralgo  ~ training and hyperparameters:} For soft $K$-subset sampling  (see Algorithm~\ref{alg:k-subset}),  we set $\tau$ to $0.1$. Further to evaluate the objective function $\V (g(f))$, we take 100 samples of $f(\xeval)$ from the posterior state $\mu_+^{a(\theta)}$.
For policy optimization in each horizon, we use the Adam optimizer with a learning rate of $0.1$ to perform policy gradient steps over $1000$ epochs. The results presented are averaged over $10$ different training seeds.

\noindent\textbf{Policy gradient (\rein) training and hyperparameters:} To evaluate the objective $\V (g(f))$ under an action, we take $100$ samples of $f(\xeval)$.  For optimizing the policy in each horizon, we use an Adam optimizer with learning rate of $0.1$,  performing policy gradient updates over $1000$ epochs. The results presented are averaged over $10$ different training seeds.

\subsection{Planning with \ensembleplus experiments}

  Once again, we use {GP}s as our data generating process. Specifically, we use a GP with an RBF kernel: $f_i \sim \mathcal{GP}(m,\mc{K}) $, where $m(X)=0$ and $\mc{K}(X,X') = \sigma_f^2 \exp\left(-\frac{||X-X'||_2^2}{2\loss^2} \right)$. Additionally, Gaussian noise   $N(0,\sigma^2)$ is added to the outputs. We set $\loss = 1$, $\sigma_f^2 = 0.69$, and $\sigma^2 = 0.01$.
The marginal distribution $P_X$ consists of $4$ \textit{non-overlapping} clusters. We follow the same process as in the Gaussian process experiments to form the cluster centers and sample points around them. While the training points are drawn from a single cluster, both the pool and evaluation points are drawn from all the $4$ clusters. Our setup includes $20$ initial labeled data points, $10$ pool points, and $252$ evaluation points used to evaluate the objective. 


\noindent\textbf{\ensembleplus Architecture:} We use an ensemble of 10 models. Each model being a 2-hidden layer MLP with 50 units per hidden layer. In addition, each models includes an additive prior. The additive prior function for each model is fixed to be a 2-hidden layer MLP with 50 units in each hidden layer.  
 Specifically, the $m$-th model for $1\leq m \le 10$ of the \ensembleplus takes the form
\begin{align*}
    f_{\eta_m}(X) = g_{\eta_m}(X) + \alpha p_m(X),
\end{align*}
where $g_{\eta_m}$  is the trainable part of the network while $\alpha p_m(\cdot)$ is the additive prior function. Here $\alpha$ controls our prior belief about the uncertainty ---higher the $\alpha$, the greater the uncertainty. The prior functions $p_m(\cdot)$ differ across models $m$ due to different initializations. In our setup, we set $\alpha=100$ to reflect a high level of uncertainty in the prior beliefs.

\noindent\textbf{\ensembleplus training:} For a given dataset ${\mathcal D} =(X_i,Y_i)_{i=1}^n$, we train the $m$-th model so as to minimize the following loss function 
\begin{align*}
    \loss(\eta_m, \mathcal D) =  \frac{1}{n}\sum_{i=1}^n (g_{\eta_m}(X_i) + \alpha p_m(X_i)-Y_i)^2 + \lambda \norm{\eta_m}_2^2.
\end{align*}
We tune $\lambda$ to $0.1$ and use the Adam optimizer with a tuned learning rate of $0.1$. Each model is trained for 50 iterations.

\noindent\textbf{\ouralgo  ~ training and hyperparameters:} Recall that in soft $K$-subset sampling there is a hyperparameter $\tau$  (see Algorithm~\ref{alg:k-subset}), which we set to $0.1$. To differentiate through the argmin operation, we employ the differentiable optimizer from the $\mathsf{torchopt}$ package, specifically the MetAdam Optimizer with a learning rate of $0.1$. For optimizing the sampling policy, we use the Adam optimizer with a learning rate of $0.05$, performing policy gradient updates over $500$ epochs.









\section{Proof of Theorem~\ref{thm:binary} and Theorem~\ref{thm:multi}}

\subsection{Proof of Theorem~\ref{thm:binary}}
\label{sec:proof-thm-binary}

 We begin by analyzing the $\rein$ based gradient estimator.
With $N$ samples, we have: 
\begin{align*}
\mse(\hat{\nabla}^{\mathsf{RF}}_{N})
= \V(\hat{\nabla}^{\mathsf{RF}}_{N})
=\frac{1}{N} \V_{A \sim \pi_\theta} \bigg(A \nabla_\theta \log(\pi_\theta(A)) \bigg).
\end{align*}

Using the definition of $A$, we derive the following:
\begin{align*}
\V_{A \sim \pi_\theta} \left(A \nabla_\theta \log(\pi_\theta(A)) \right) = \frac{1}{(1-\theta) \theta} - 4.
\end{align*}
As a result, we have:
\begin{align}
\mse(\hat{\nabla}^{\mathsf{RF}}_{N}) =  \frac{1}{N
}\left(\frac{1}{(1-\theta) \theta} - 4\right). \label{eqn:mse-rf-bin}
\end{align}

Next, we analyze the pathwise gradient estimator $\hat{\nabla}^{\mathsf{grad}}_{\tilde{\tau},N}$.
Let $h_{\tau}(U,\theta) = \frac{\exp\left(\frac{U-\theta}{\tau}\right)-1}{\exp\left(\frac{U-\theta}{\tau}\right)+1}$ 
be a random variable,
where $U \sim \mathsf{Uni}[0,1]$, so that:
\begin{align}
\mse(\hat{\nabla}^{\mathsf{grad}}_{\tau,N}) = \left[
\bias(\nabla_\theta h_{\tau}(U,\theta))\right]^2 + \frac{1}{N} \V(\nabla_\theta h_{\tau}(U,\theta)).
\label{eqn:mse-decompo-bin}
\end{align}

To compute the bias term in~\eqref{eqn:mse-decompo-bin}, we note that
\begin{align*}
\E(h_{\tau}(U,\theta)) & = \E \left(\frac{1}{1+\exp\left( -\frac{U-\theta}{\tau}\right)}\right) - \E\left(\frac{1}{\exp\left(\frac{U-\theta}{\tau}\right)+1} \right) =2 \tau \log \left( \frac{\exp (\frac{1-\theta}{\tau})+1}{\exp (\frac{-\theta}{\tau})+1}\right)-1, \end{align*}
\begin{align*}
\nabla_\theta \E(h_{\tau}(U,\theta)) & = 2 \left(\frac{1}{1+\exp\left(\frac{1-\theta}{\tau} \right)} - \frac{1}{1+\exp\left(\frac{-\theta}{\tau} \right)}\right).
\end{align*} 
Therefore,
\begin{align}
\left[\bias(\nabla_\theta h_{\tau}(U,\theta))\right]^2 = 4 \left(1+\frac{1}{1+\exp\left(\frac{1-\theta}{\tau} \right)} - \frac{1}{1+\exp\left(\frac{-\theta}{\tau} \right)}\right)^2.
\label{eqn:mse-bias-bin}
\end{align}
Note that the above term goes to 4 as $\tau \to \infty$. The other part contributing to the mse~\eqref{eqn:mse-decompo-bin} is  the variance term and it satisfies
\begin{align*}
\V(\nabla_\theta h_{\tau}(U,\theta)) \le 
\E\left[\nabla_\theta (h_{\tau}(U,\theta))\right]^2 =
\frac{4}{\tau} \left[ \frac{3 \exp\left(\frac{-\theta}{\tau}\right) +1}{6 \left(\exp\left(\frac{-\theta}{\tau}\right)+1\right)^3 } - \frac{3 \exp\left(\frac{1-\theta}{\tau}\right) +1}{6 \left(\exp\left(\frac{1-\theta}{\tau}\right)+1\right)^3 }  \right].
\end{align*}
Combining the above equation with~\eqref{eqn:mse-bias-bin},
we arrive at $\mse(\hat{\nabla}^{\mathsf{grad}}_{\tilde{\tau},N}) 
\le 
v_N(\tau)$, where
\begin{align*}
v_N(\tau) \defeq 
4
\left(1+\frac{1}{1+\exp\left(\frac{1-\theta}{\tau} \right)} - \frac{1}{1+\exp\left(\frac{-\theta}{\tau} \right)}\right)^2 + \frac{1}{N}
\frac{4}{\tau} \left[ \frac{3 \exp\left(\frac{-\theta}{\tau}\right) +1}{6 \left(\exp\left(\frac{-\theta}{\tau}\right)+1\right)^3 } - \frac{3 \exp\left(\frac{1-\theta}{\tau}\right) +1}{6 \left(\exp\left(\frac{1-\theta}{\tau}\right)+1\right)^3 }  \right]
\end{align*}

For all $N \ge 1$, 
we can verify that $\lim_{\tau \to \infty} v_N(\tau) = 4$
and for large enough $\tau$, $v_N(\tau)$ is increasing in $\tau$. Thus, there exists some 
$\Tilde{\tau} > 0$ such that $\mse(\hat{\nabla}^{\mathsf{grad}}_{\tilde{\tau},N}) < 4$.
Comparing this with~\eqref{eqn:mse-rf-bin} for $\mse(\hat{\nabla}^{\mathsf{RF}}_{N}) $  completes the proof.

\subsection{Proof of Theorem~\ref{thm:multi}}
\label{sec:proof-thm-multi}

 For every $j$,
$G(A) \nabla_{\theta_i} \log(\pi_\theta(A)) = \frac{G(a_j)}{\pi_\theta(a_j)}\nabla_{\theta_i} (\pi_\theta(a_j))$
with probability $\frac{\theta_j}{\sum_{j=1}^m \theta_j}$
and
\begin{align*}
\frac{G(a_j)}{\pi_\theta(a_j)}\nabla_{\theta_i} (\pi_\theta(a_j))
&= \begin{cases}
\frac{-G(a_j)}{\sum_{j=1}^m\theta_j}, & j \ne i \\ 
\frac{G(a_i)}{\theta_i} - \frac{G(a_i)}{\sum_{j=1}^m\theta_j},  & j =i.
\end{cases}
\end{align*}

Therefore,
\begin{align}
\mse(\hat{\nabla}^{\mathsf{RF}}_{N,\theta_i})
 & =   \V_{A \sim \pi_\theta} \bigg(G(A) \nabla_{\theta_i} \log(\pi_\theta(A)) \bigg) \nonumber \\
  & = \sum_{j\neq i} \frac{\theta_j}{\sum_{j=1}^m \theta_j} \frac{(G(a_j))^2}{(\sum_{j=1}^m\theta_j)^2} +  \frac{\theta_i}{\sum_{j=1}^m \theta_j} \left( \frac{G(a_i)}{\theta_i} - \frac{G(a_i)}{\sum_{j=1}^m\theta_j} \right)^2 - \left(\nabla_{\theta_i} \E_{A\sim \pi_\theta}(G(A))\right)^2  \nonumber  
\\ &= \frac{\sum_{j=1}^m\theta_j(G(a_j))^2}{(\sum_{j=1}^m\theta_j)^3} + \frac{(G(a_i))^2}{\theta_i\sum_{j=1}^m \theta_j} - 2 \frac{(G(a_i))^2}{(\sum_{j=1}^m \theta_j)^2}  - \left(\nabla_{\theta_i} \E_{A\sim \pi_\theta}(G(A))\right)^2.
\label{eqn:mse-reinforce}
\end{align}


We also note that
\begin{align*}
    \nabla_{\theta_k}{G\left(\sum_{i=1}^ma_i (\tilde{h}_\tau(Z,\theta))_i\right)} &= {G'\left(\sum_{i=1}^ma_i (\tilde{h}_\tau(Z,\theta))_i\right)}\left(\frac{\exp\left(\frac{Z_k}{\tau}\right) {\theta_k}^{\frac{1}{\tau}}}{\tau \theta_k}\right) \\& \left( \frac{a_k}{\sum_{j=1}^m{\exp\left(\frac{Z_j}{\tau}\right)}{\theta_j}^{\frac{1}{\tau}}} - \sum_{i=1}^m\frac{a_i\exp\left(\frac{Z_i}{\tau}\right){\theta_i}^{\frac{1}{\tau}}}{(\sum_{j=1}^m{\exp\left(\frac{Z_j}{\tau}\right)}{\theta_j}^{\frac{1}{\tau}})^2}\right),
\end{align*}
and therefore, 
$\lim_{\tau \to \infty} \nabla_{\theta_k}{G\left(\sum_{i=1}^ma_i (\tilde{h}_\tau(Z,\theta))_i\right)} = 0 $ for all $Z$.

Noting that $|G'(\cdot)| \le \bar{G}$, we have 
\begin{align*}
    \lim_{\tau \to \infty} \nabla_{\theta_k}
    G\left(\sum_{i=1}^m a_i (\tilde{h}_\tau(Z,\theta))_i \right) = 0.
\end{align*}

As a result,  
\begin{align*}
    \lim_{\tau \to \infty}\V_Z \left(\nabla_{\theta_k}
    G
    \left(\sum_{i=1}^m a_i (\tilde{h}_\tau(Z,\theta))_i \right)
    \right) =0,
\end{align*} and
\begin{align*}
    \lim_{\tau \to \infty} \left[\bias(\hat{\nabla}^{\mathsf{grad}}_{\tau,N, \theta_i})\right]^2 = \left(\nabla_{\theta_i} \E_{A\sim \pi_\theta}(G(A))\right)^2.
\end{align*} 
Theorem~\ref{thm:multi} follows directly from the above.

\section{Experimental details (Section \ref{sec:practical_consideration})}
\label{sec:practical_consideration_experiment_details}

\subsection{Evaluation metrics}
\label{sec:UQ_eval_metrics}

 Recall we denoted by $ \mc{D}^{0} = (\mc{X}^{0}, \datay^{0}) := (X_{1:m-1},Y_{1:m-1})$ the initial training data and let $X_i \simiid Q_X$. Further, let $P_X$ be the distribution we expect to see during deployment and  $\mc{X}^{0} $ only represents a part of this distribution (with having selection bias). 
 Also recall, given a true data generating function  $f\opt$, outcomes $Y$ are generated as following:
\begin{equation*}
    Y = f\opt(X) + \varepsilon.
\end{equation*}
for some random noise $\varepsilon$,
We view UQ models as outputting a probability distribution over a number of data points given contexts, which allows us to evaluate UQ models based on their ability to explain held-out observed sequences. 
This view is well-founded in  prequential statistics~\citep{Roberts65,Dawid84}, sequence modeling view of Bayesian models~\citep{FortiniPe23}, and information theory~\citep{BarronRiYu98}.
Most recently,~\citet{OsbandWeAsSeDwLuVa22} propose this approach to evaluating UQ models.
Suppose we wish to predict $\tau$ outcomes 
$\ypred \defeq \ypredsetl$ for features $\xpred \defeq \xpredsetl$ where $\xpredsetl \simiid P_X$. The corresponding true outcomes are $y_{m+1:m+\tau}$.  Further, let $\mu(f|\mc{D}^0)$ be the posterior from some UQ module after seeing data $\mc{D}^0$. We  set our evaluation metric to be

\begin{align}
- \log {\left[\int \left(\prod_{i=m}^{m+\tau} p_\epsilon (y_i -f(X_i))\right) d\mu(f|\mc{D}^0) \right]}.
\label{joint_log_loss}
\end{align}

\citet{OsbandWeAsSeDwLuVa22}~referred to the above metric as ``\emph{marginal} log-loss'' when 
$\tau=1$ and as ``\emph{joint} log-loss'' when $\tau > 1$. Marginal log-loss represents the expected negative log-likelihood of predicting a single test example, based on the UQ module’s predictive distribution for that specific input. On the other hand, joint log-loss is defined as the expected negative log-likelihood for a batch of test examples, calculated under the model’s joint predictive distribution across various label combinations for the entire batch of inputs. This effectively provides a unified measure to compare the performance of UQ methods. 

\paragraph{Joint log-loss with dyadic sampling} 
To evaluate the quality of joint predictions as discussed, batches of inputs $(X_1, \cdots, X_\tau)$ need to be sampled multiple times, and their log-loss is assessed against corresponding labels $(y_1, \cdots, y_\tau)$. 
Typically, batch size $\tau$ required to effectively distinguish joint predictions that reflect label interdependencies becomes large as dimensionality of data increases, and it becomes impractical as dimension increases. 
Dyadic sampling~\citep{OsbandWeAsDwLuIbLaHaDoRo22} offers a practical heuristic to estimate joint likelihood.
The basic dyadic sampling method starts by independently sampling two anchor points, $X_1$ and $X_2$, from the input distribution. Then, $\tau$ points are sampled independently and with equal probability from these two anchor points $\set{X_1, X_2}$ that leads to a random sample $X_{(1)}, \cdots, X_{(\tau)}$.
An agent's joint prediction of labels is evaluated on this batch of size $\tau$.  In our experiments, we use $\tau = 10$ and it seems to be an effective estimator of the joint log-loss.

\paragraph{In-distribution and out-of-distribution evaluations} If training distribution and prediction distribution are same, that is $Q_X =P_X$, then we call it in-distribution performance. If $Q_X \neq P_X$  then it is the case of distribution shifts (potentially out-of-support).




\paragraph{Dynamic setting} As specified in Section \ref{sec:eval_posterior_consis}, we  also evaluate posterior updates of different UQ methodologies in dynamic setting. We restrict attention to single-step updates as there is little algorithmic difference in single-step vs. multi-step updates.
Consider a setting where  $t=0$ we observe data $\mc{D}^0$ and our UQ methodology has posterior $\mu(f|\mc{D}^0)$.
At time $t = 1$, we see additional data $\mc{D}^1$ and
the posterior is updated to  $\mu(f|\mc{D}^0\cup \mc{D}^1)$. Performance under the updated posterior is:
\begin{align*}
    - \log {\left[\int \left(\prod_{i=m}^{m+\tau} p_\epsilon (y_i -f(X_i))\right) d\mu(f|\mc{D}^0 \cup \mc{D}^1) \right]}.
\end{align*}

\paragraph{Sensitivity to implementation details} Canonical engineering practices involve several hyperparameters (e.g., weight-decay, prior scale) and exogeneous randomness. In applications that require frequent posterior updates (e.g., active exploration in RL), posterior inference $\mu(\cdot | \mc{D}^0, \mc{D}^1)$ incurs a huge computational cost and heuristics like early stopping are used as an approximation. Additionally, the integral in (\ref{joint_log_loss}) is often challenging to compute and approximations are made; typically, the final estimator depends on the UQ methodology, as well as implementation details such as initialization and training randomness.  We also  evaluate the sensitivity of UQ methodologies against these implementation choices and observe that
they have an outsized influence in OOD performance, particularly in dynamic settings.


\subsection{Datasets}  \label{sec:UQ_BENCH_datasets}
As we specify in Section \ref{sec:eval_posterior_consis}, we study both synthetic and real-world datasets
to allow comparison against oracle methods that know the true model class (synthetic), and represent realistic application areas (real). 

\noindent{\bf{Synthetic data.}} We simulate a regression task using Gaussian Processes (GPs).  
 As before, to introduce out-of-support bias, we sample features $X$ from multiple clusters distributed over $\R^n$, where $n$ is the feature dimension.  Specifically,  we create two anchor points, one at the origin and other at a linearly increasing distance from the origin parameterized by $k$. 
We then sample the $X$ data around these
anchor points by adding a small zero mean Gaussian noise to the anchor points. 
 Outcomes $Y$ are generated using the GP 
with an RBF kernel: $f_i \sim \mathcal{GP}(m,k) $, with $m(X)$ being the mean and $k(X,X') = \sigma_f^2 \exp\left(-\frac{(X-X')^2}{2\loss^2}\right)$ being the covariance kernel, further we add Gaussian noise $N(0,\sigma^2)$ to $f(X)$ to get $Y$.  We set $\loss = 2$, $\sigma_f^2 = 0.69$ and $\sigma^2 = 0.01$.
We select 100 points near the first anchor point as the training data and 
200 points randomly sampled from the two clusters form the test data, which we divide into in-distribution and out-of-distribution data depending on which cluster the data sample comes from.


\noindent{\bf{eICU dataset.}} The eICU Collaborative Database  is a large multi-center critical care electronic health record (EHR) database consisting of patient data collected as part of critical care~\citep{PollardJoRaCeMaBa18}. We assess UQ performance for in-hospital mortality prediction using physiologic and laboratory measurements. Some examples of the features
include the  length of stay at the hospital, number of days on a ventilator, and  recorded temperature on Day 1 of ICU admission.

\noindent{\bf{Fraud dataset.}} 
The Fraud Dataset Benchmark (FDB)~\citep{GroverXuTiCheLiZaLiZh22} collects publicly available data  covering a broad range of fraud detection tasks, including content moderation. We demonstrate UQ performance on four fraud datasets: \texttt{IEEE-CIS Fraud Detection, Credit Card fraud, Fraud Ecommerce, Vehicle Loan Default Prediction}. 

\noindent{\bf{ACS dataset.}} For economic data, we use \texttt{ACS Employment, Income}  in New York datasets from the US-wide \texttt{ACS PUMS} data~\citep{DingHaMoSc21}, where
the outcome is employment, and whether an individual’s income exceeds $50k$.   

\begin{table}[h!]
\centering
\begin{tabular}{|c|c|c|c|}
\hline
\textbf{Dataset name} & \textbf{\# Train} & \textbf{\# Test} & \textbf{\# Features} \\ \hline
eICU & 38,534 & 16,515 & 48 \\ \hline
$\ieeecis$ & 561,013 & 28,527 & 67 \\ \hline
$\ccfraud$ & 227,845 & 56,962 & 28 \\ \hline
$\fraudecom$& 120,889 & 30,223 & 6 \\ \hline
$\vehicleloan$& 186,523 & 46,631 & 38 \\ \hline
$\acsemployment (\mathtt{NY})$ & 137,876 & 59,091 & 98 \\ \hline
$\acsincome (\mathtt{NY})$& 72,114 & 30,907 & 811 \\ \hline
\end{tabular}
\caption{All datasets used in this section. 
For all datasets, we use standard one-hot encoding for categorical features.}
\label{table:data-details}
\end{table}

\subsection{Simulating out-of-distribution data} \label{sec:selection-bias} 

We introduce mild distribution shifts using a logit-based likelihood ratio---(log) ``linear selection bias''---and out-of-support shifts using a \emph{clustering-based} selection bias procedure, leveraging a ``polyadic sampler''. 

\noindent{\bf{Linear section bias}} Given a data $\mc{D}$, we standardize the data $x_i \in \R^d$, and then sample each point $i$ with probability 
$\mathsf{sigmoid}\paran{k \cdot \beta^\top x_i/d}$,
 where $\beta$ is a random vector in $[-1,1]^d$ and $k \ge 0$ is a parameter governing the intensity of selection bias to create a random subset of data $\dtrain \subseteq \mc{D}$.

\noindent{\bf{Clustering-based selection bias}} 
 We cluster the data using  $k$-means clustering and select the training data just from one cluster while the test data comes from all the clusters to create out-of-support shifts. Further, we vary the distance between the cluster centers to model the severity of the distribution shift. Clusters spaced far apart serve as out-of-support shifts from the training data.  
To enable out-of-support shifts in {\bf{dynamic}} settings, we restrict initial training data $\mc{D}^0$ to come  from one of the clusters, following which, the agent observes data sequentially, in batches, from other clusters. 

Specifically for the real datasets,  
we construct 
50 clusters using the $k$-means algorithm for each of the above dataset.
To make the clustering efficient,
we randomly sample 50,000 rows from all datasets except eICU, and we only apply clustering to this sampled data.
We define the largest cluster as the training cluster which we label as Cluster 1.  
We then relabel each cluster based on their distance  to the training cluster, the closest  one being labeled as Cluster 2 and so on. 
Next, we  select 250 points from Cluster 1 as the training data, 250 points from Cluster $k \in \set{5,10,15,20,25,30}$ as the out-of-distribution test data
and 250 points from Cluster 1 as in-distribution test data.
We also select 50 points from Cluster $k$ as the ``pool data'' that the modeler can acquire in the dynamic setting.
For linear selection bias,   we 
first draw $\beta \sim \mathsf{Uni}[-1,1]^d$ where $d$ is the dimension of the feature, we fix $\beta$ and  for each $k \in \set{0,0.5,1,2,4,8,16}$, 
we sample each point $x_i$ with probability 
$\mathsf{sigmoid}\paran{k \cdot \beta^\top x_i/d}$.



\subsection{UQ methodologies}
\label{sec:UQ_bench_benchmark-uq-metric}

We compare Dropout, Ensemble, \ensembleplus, Epinet, Hypermodels with varying  hyperparameters for each model (see Section \ref{sec:UQ_BENCH_hyperparameter-sweep}). 
  For synthetic datasets we also compare to GPs as one of the UQ methods, which serves as an oracle baseline.  As specified in  Section~\ref{sec:eval_posterior_consis}, we explain different UQ methodologies here. We also specify the configurations of different methodologies that we use, mostly inspired from~\citep{OsbandWeAsDwLuIbLaHaDoRo22, OsbandWenAsDwIbLuRo23}. In the end of this section, we also discuss some aspects of varying these configurations of different methodologies.

\subsubsection{MLP}

We use Multi-layer perceptron (MLP) as a baseline to compare the benefits of the other UQ methodologies over a simple MLP.  We train the MLP on mean-squared error loss for regression, while for the classification we train it on cross-entropy loss, both with $L_2$ weight decay. In our experiments, we use  MLP with two hidden layers, with each layer having 50 hidden units with ReLU activation.

  \subsubsection{Ensemble}
Deep Ensemble is a basic approach introduced in~\citep{LakshminarayananPrBl17} for posterior approximation. 
Broadly, we learn an ensemble of networks (e.g., MLPs) on the given training data,  with each networks differing from each other through  different random initialization of weights. Specifically, the ensemble consists of $M$ models.  The $m$-th model of the Ensemble takes the form  $f_{\eta_m}(X)$ and is parameterized by $\eta_m$,
where $f_{\eta_m}(\cdot)$  is   trained on the given data and differs across $m$ through random weight initializations. In regression setting, the model is trained using the mean-squared error loss, while in classification setting ,it is trained using the cross-entropy loss. In both settings, we also use $L_2$ weight regularization. In our experiments, we use an ensemble of 100 MLPs where each MLP has two hidden layers with 50 hidden units each and we use the ReLU activation.

 \subsubsection{\ensembleplus}
 
As mentioned earlier, \ensembleplus was introduced in~\citep{OsbandAsCa18} building on deep ensembles. 
 \ensembleplus consists of ensemble of MLPs along with randomized prior function for each MLP and bootstrap sampling of the training data. Randomized prior functions and bootstrap sampling help in maintaining diversity among the models on the unseen data and is crucial for performance improvement over Ensembles. 
 Specifically, assume that Ensemble $+$ consists of $m$ models. 
 Then the $m$-th model ($1\leq m \le M$) of the Ensemble $+$ takes the form
\begin{align*}
    f_{\eta_m}(X) = g_{\eta_m}(X) + \alpha p_m(X),
\end{align*}
where $g_{\eta_m}$  is the trainable part of the network (parameterized by $\eta_m$) and $\alpha p_m(\cdot)$ is the additive prior function, where $\alpha$ (prior scale) controls our prior belief about the uncertainty -- a higher   $\alpha$ means a higher  uncertainty. 
Here $g_{\eta_m}(\cdot)$ and $p_m(\cdot)$ differs across $m$ through random weight initializations.  
 As in ensembles, we train the model on mean-squared error loss (in regression setting) or on the cross-entropy loss (in classification setting) along with $L_2$ weight regularization. In our experiments, we use an ensemble of 100 MLPs where each MLP has two hidden layers, each with 50 hidden units and ReLU activation. We use an identical MLP (with different weight initialization) as the prior generating function for each particle of the ensemble.

\subsubsection{Hypermodels}

A Hypermodel agent~\citep{DwaracherlaLuIbOsWeVa20} consists of a base model parameterized by $\theta \in \Theta$. Given 
$\theta$ and an input $X$, base model gives the output $ f_\theta(X)$. A hypermodel $g_\nu: {\mathcal Z} \to \Theta$ is parameterized by $\nu$ and $z \in \zdist$ is an index determining a specific instance of the base model. 
Along with this, a reference distribution $p_z$ is also specified for the index $z$ to determine the distribution over the base models. 
Therefore, overall the output corresponding to the index $z$ and input $X$ is $f_{g_\nu (z)}(X)$. 
In addition to this, we can have additive prior for the hypermodel similar to   Ensemble $+$. In our experiments, we use an MLP with two hidden layers, each with 50 hidden units and ReLU activation as our base model. We also use an additive prior with one hidden layer and 10 hidden units. 



 \subsubsection{Epistemic Neural Networks (ENNs - Epinet)}

Epistemic Neural Networks were introduced in~\citep{OsbandWenAsDwIbLuRo23}. 
We focus on Epinets   $f_\theta(X,z)$) that can be parameterized by $\theta$. 
It consists of a base model $\mu_\zeta(X)$ parameterized by 
$\zeta$ and an epinet $\sigma_\eta(\mathrm{sg}[\phi_\zeta(X),z])$ parameterized by $\eta$. 
Here $\phi_\zeta(X)$ is a subset of inputs and outputs of each layer of the base model and ``sg'' is the stop gradient 
operation that was shown to stabilize the training~\citep{OsbandWenAsDwIbLuRo23}.
In addition, $z$ is an epistemic index drawn from a Gaussian distribution. 
Therefore, ENN has parameters $\theta=(\zeta,\eta)$. Moreover, the epinet has a learnable part 
$\sigma_\eta^L(\mathrm{sg}[\phi_\zeta(X),z])$ and a prior net $\sigma^P_\eta(\mathrm{sg}[\phi_\zeta(X),z])$.
Overall,
\begin{align*}
    f_\theta(X,z) = \mu_\zeta(X) + \sigma_\eta^L(\mathrm{sg}[\phi_\zeta(X),z]) + \sigma^P_\eta(\mathrm{sg}[\phi_\zeta(X),z]).
\end{align*}
We use Algorithm 1 in~\citep{OsbandWenAsDwIbLuRo23} for training epinets, and for loss function we  again use  mean squared error (for regression) or cross-entropy loss (for classification) with $L_2$ weight decay. For our experiments, we set a two-layer MLP as our basenet. This MLP has two hidden layers with 50 hidden units in each layer and ReLU activation. For learning part of the epinet,  we take $\sigma_\eta^L([\phi_\zeta(X),z]) = g_\eta([\phi_\zeta(X),z])^Tz$  where $g_\eta(.)$ is an MLP with two hidden layers with 15 hidden units and ReLU activation. The prior function of the epinet 
takes the form $\sigma^P_\eta([x,z]) = \alpha \sum_{i=1}^{D_z}p^i(x)z_i$, where $D_z$ is the dimension of the index $z$, $\alpha$ is the prior scale controlling the prior belief of the uncertainty and each $p_i$ is an MLP with two hidden layers, each with 5 hidden units and ReLU activation.

\subsubsection{Dropout}

Monte Carlo dropout was introduced in ~\citep{GalGh16} for posterior approximation. Broadly, the agent applies  dropout on each layer of a fully connected MLP with ReLU activation at both and training and test time, making the predictions random to approximate the posterior. For regression, we train the model on mean-squared error loss while for the classification, we train it on cross-entropy loss, both with $L_2$ weight decay. In our experiments, we apply dropout to the  MLP with two hidden layers, with each layer having 50 hidden units each with ReLU activation.

  \subsubsection{Gaussian Processes (GP)}
GPs~\citep{Rasmussen03} are extensively used in the Bayesian optimization literature and are known to work well for low dimensional data. 
As specified earlier, we use GPs as one our baselines for the synthetic regression experiments. Specifically, we use $\mathcal{GP}$ with an RBF kernel: $f_i \sim \mathcal{GP}(m,\mc{K}) $, where $m(x)$ is the mean and $\mc{K}(x,x') = \sigma_f^2 \exp\left(-\frac{(x-x')^2}{2\loss^2}\right)$,  we further add Gaussian   noise $N(0,\sigma^2)$ to the outputs. We use the same values for $m(x), \loss, \sigma_f, \sigma$ through which synthetic data is generated to have a well specified prior setting and an oracle baseline. 


\paragraph{Discussion - configurations of different methodologies} Note that the configurations we use for different UQ methodologies such as Ensemble, \ensembleplus, Epinets and Hypermodels have different numbers of parameters. 
For instance, ensemble $+$ with 100 particles has many more parameters than Epinets. And as we saw in Section \ref{sec:eval_posterior_consis}, Epinets perform comparable to   \ensembleplus on in-distribution and out-of-distribution with these configuration. In a way, we can say Epinets perform as well as \ensembleplus with much fewer number of parameters. As mentioned in~\citep{OsbandWeAsDwLuIbLaHaDoRo22, OsbandWenAsDwIbLuRo23}, if we reduce the number of particles in \ensembleplus, its performance deteriorates. 
In our experiments, we do not concern ourselves with comparing different UQ methodologies with the same number of parameters. 
Rather, the main aim for our experiments is to showcase that the models (methodologies + configurations) performing similarly on in-distribution (ID) data  differ drastically in their performance on OOD performance and under dynamic settings. 

\paragraph{Evaluation} In all of our experiments, we set $\tau = 10$ and employ the dyadic sampling heuristic from ~\citet{OsbandWenAsDwIbLuRo23} to evaluate the joint log-loss~\eqref{joint_log_loss} (see Section~\ref{sec:UQ_eval_metrics} for details).

\subsection{Hyperparameter  sweep}
\label{sec:UQ_BENCH_hyperparameter-sweep}

In Table \ref{tab:hyperparameter_sweeps} we summarize the hyperparmeters we tune for each of the methodologies.

\begin{table}[h!]
\centering
\begin{tabular}{|c|c|}
\hline
\textbf{Agent} & \textbf{Hyperparameters} \\ \hline
\textbf{mlp} & learning rate, L2 weight decay \\ \hline
\textbf{dropout} & learning rate, length scale, dropout rate \\ \hline
\textbf{ensemble} & learning rate,  L2 weight decay \\ \hline
\textbf{ensemble+} & learning rate,  L2 weight decay, prior scale \\ \hline
\textbf{hypermodel} & learning rate, L2 weight decay, prior scale \\ \hline
\textbf{epinet} & learning rate,  L2 weight decay, prior scale \\ \hline
\end{tabular}
\caption{Hyperparameter sweeps for each agent}
\label{tab:hyperparameter_sweeps}
\end{table}

For learning rates, we swept over $\set{1e-4, 0.001, 0.01,0.1 }$ for each agent and found that learning rate of $0.001$ works well for all agents. We fix this learning rate for the rest of the experiments. For other hyperparameters, we sweep over the following sets for each agent:
\begin{itemize}
    \item For \textbf{mlp}, we sweep over L2 weight decay in $\set{0.01, 0.1, 1, 10, 100}$.
    \item For \textbf{dropout}, we sweep over  length scale in $\set{0.0001, 0.001, 0.01, 0.1, 1, 10, 100}$, and dropout rate in $\set{0.0, 0.1, 0.2, 0.3, 0.4, 0.5, 0.6}$.
    \item For \textbf{ensemble}, we sweep over L2 weight decay in $\set{0.01, 0.1, 1, 10, 100, 1000}$.
    \item For \textbf{ensemble+}, we sweep over  L2 weight decay in $\set{0.01, 0.1, 1, 10, 100}$, and prior scale in $\set{1, 3, 10, 30, 100}$.
    \item For \textbf{hypermodel}, we sweep over  L2 weight decay in $\set{0.01, 0.1, 1, 10, 100}$, and prior scale in $\set{1, 3, 10, 30, 100}$.
    \item For \textbf{epinet}, we sweep over L2 weight decay in $\set{0.001, 0.01, 0.1, 1, 10, 100}$, and prior scale in $\set{0.01, 0.1, 1, 10, 100}$.
\end{itemize}

\subsection{Other experimental details}
For Figures~\ref{fig:task-1-joint-ood-all-data} 
and~\ref{fig:task-1-joint-ood-all-data-with-oracle},  we average the results over 10 different inference seeds  with each seed evaluating 1000 batches for the joint log-loss (with dyadic sampling) and randomizing over Monte Carlo approximation of the posterior.  
For the rest of the experiments in Section \ref{sec:UQ_BENCH_additional_experiment}, we fix the seed unless we specify it otherwise. 



\subsection{Additional Experiments}
\label{sec:UQ_BENCH_additional_experiment}
Here we present additional experiments from Section \ref{sec:eval_posterior_consis}.

\subsubsection{Comparison with a Gaussian oracle}
\begin{figure}
\centering 
\centering \includegraphics[height=4.5cm]{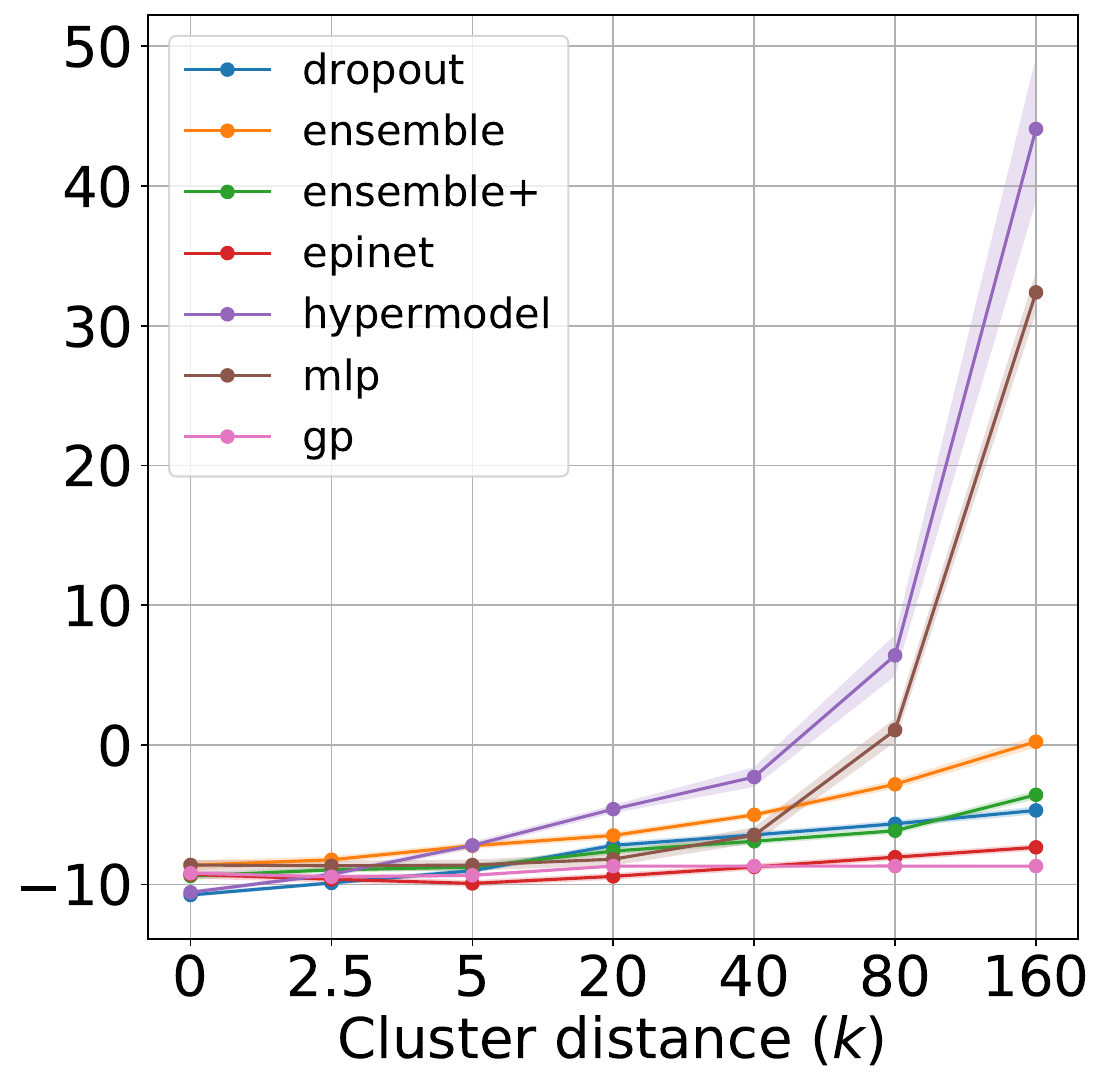}

\centering{\small{ Synthetic Data (Clustering bias)}}
\caption{Joint log-loss on OOD data with increasing selection bias with a GP oracle.}
\label{fig:task-1-joint-ood-all-data-with-oracle}
\end{figure}

In synthetic data generation setting we compare different UQ methodologies to the  Gaussian process oracle baseline with access to the parameters of the the data generation process (which is also an GP).  Figure~\ref{fig:task-1-joint-ood-all-data-with-oracle} compares  different UQ methodolgies and the GP oracle (the setting is same to that in Figure~\ref{fig:task-1-joint-ood-all-data}).
From Figure~\ref{fig:task-1-joint-ood-all-data-with-oracle}, we observe that UQ methodology ``GP'', which is the oracle, performs better for a large $k$, and for small $k$ values, many other UQ methodlogies are quite similar in performance. 

\subsubsection{Hyperparameter tuning breaks under distribution shifts}

\label{sec:hyperparameter_tuning_breaks_under_dis_shifts_experiment_simple}

\begin{figure}[h]
\centering
\begin{minipage}[b]{0.32\textwidth}
\centering
\includegraphics[height=4.5cm]{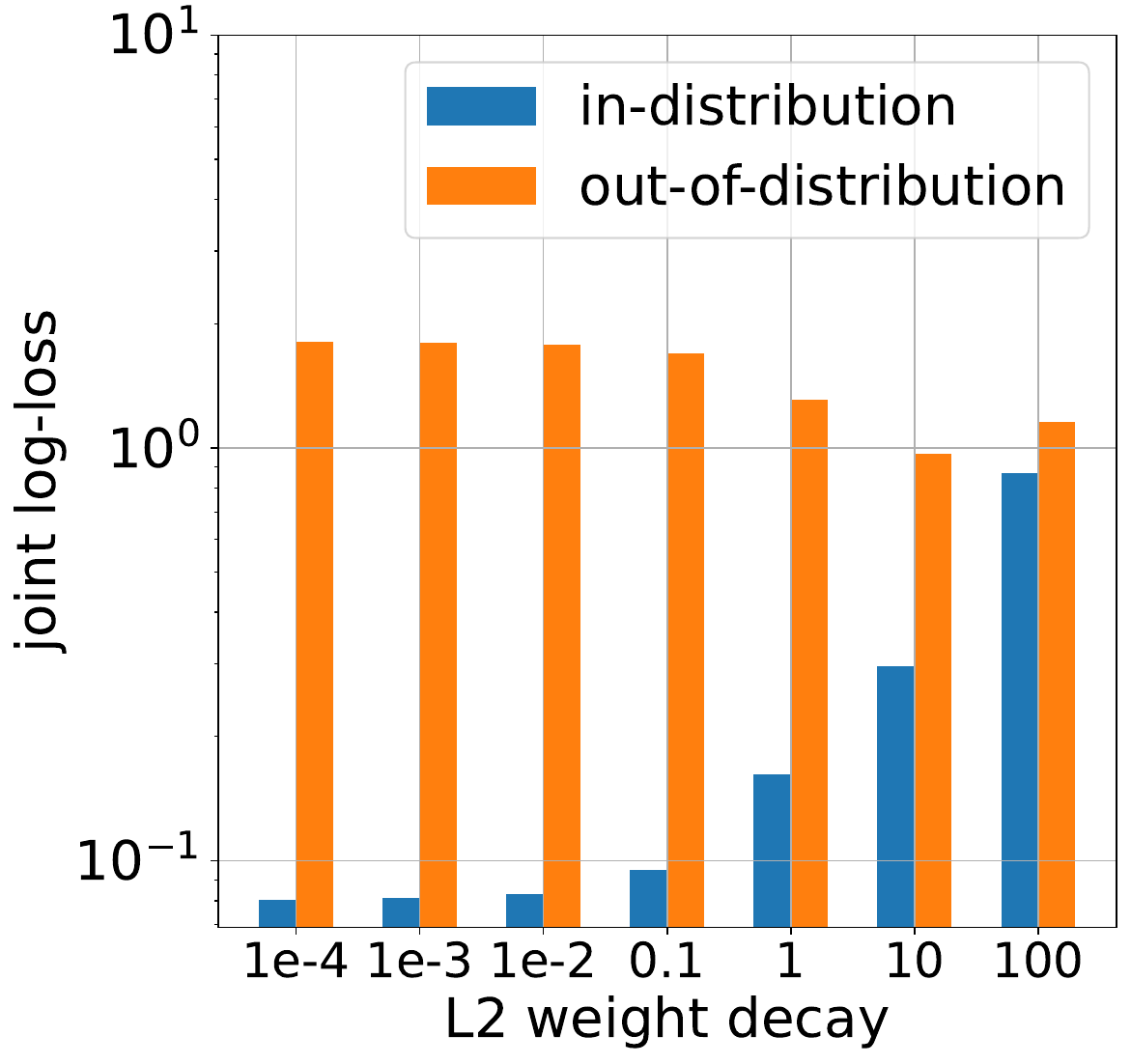}

\centering {\small{\ensembleplus}}
\end{minipage}
\hfill
\begin{minipage}[b]{0.32\textwidth}
\centering \includegraphics[height=4.5cm]{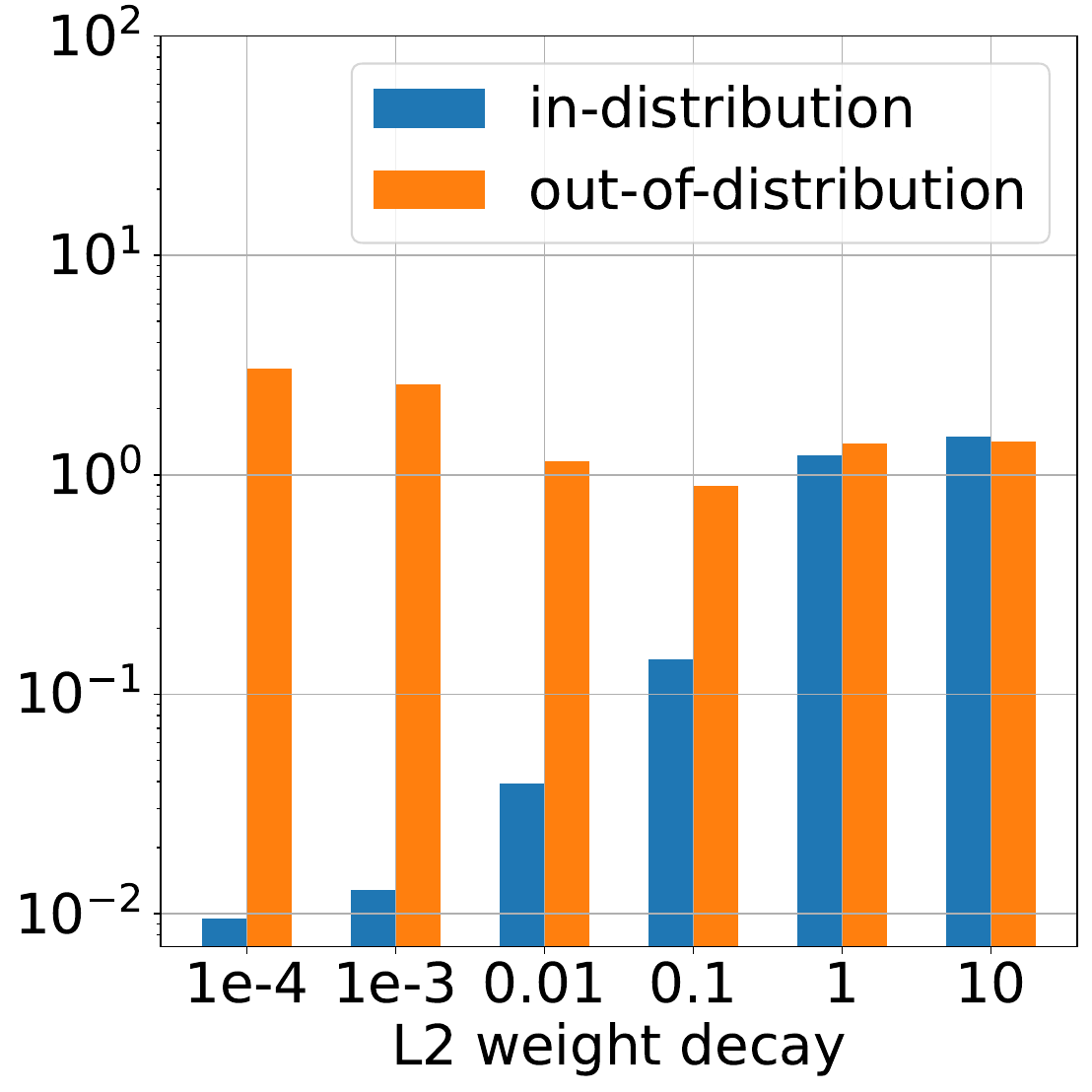}
\centering {\small{Epinet}}
\end{minipage}
\hfill
\begin{minipage}[b]{0.32\textwidth}
\centering \includegraphics[height=4.5cm]{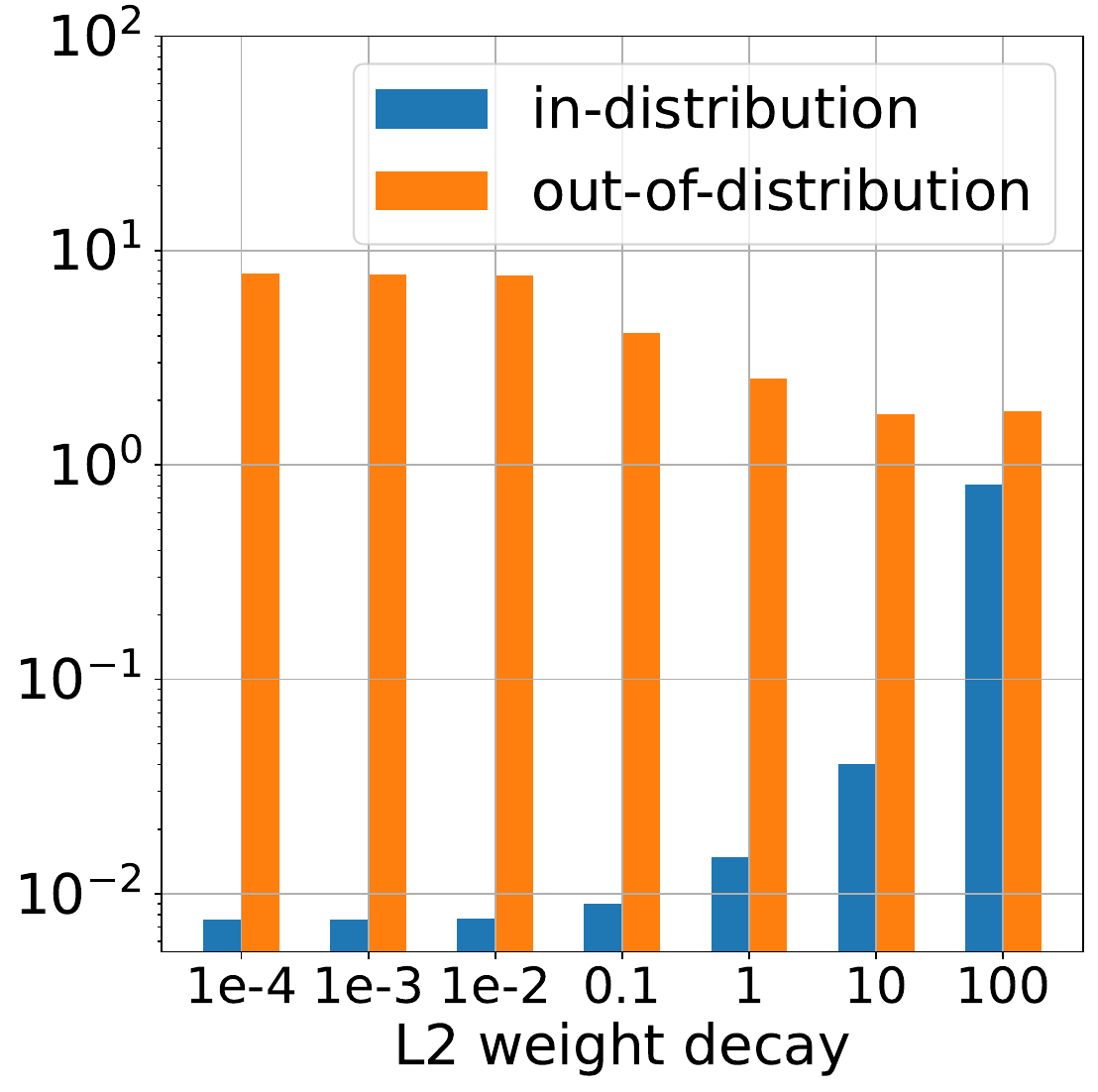}
\centering {\small{Hypermodel}}
\end{minipage}
\caption{Trade-off between  in-distribution (ID) and out-of-distribution (OOD) performance with weight decay as the hyperparameter (eICU, Clustering bias). }
\label{fig:difficult_to_choose_weight_decay}
\end{figure}

\begin{figure}[h]
\centering
\begin{minipage}[b]{0.31\textwidth}
\centering
\includegraphics[height=4cm, width=\textwidth]{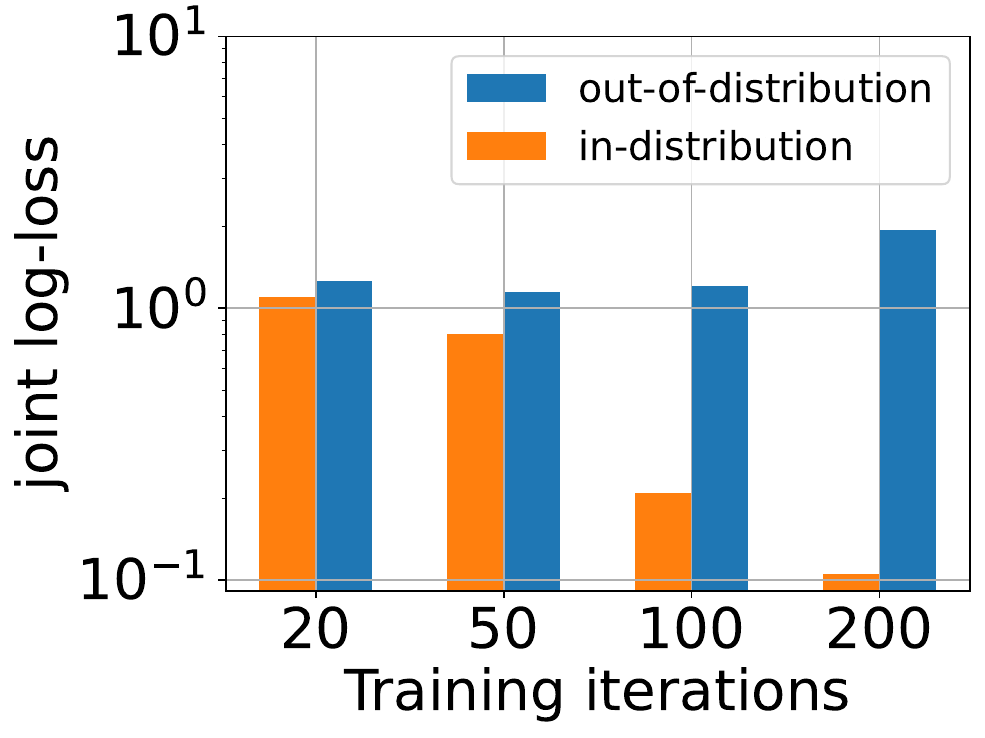}
\centering {\small{\ensembleplus}}
\label{fig:ensemble+_difficult_to_choose_stopping_time}
\end{minipage}
\hfill
\begin{minipage}[b]{0.31\textwidth}
\centering \includegraphics[height=4cm, width=\textwidth]{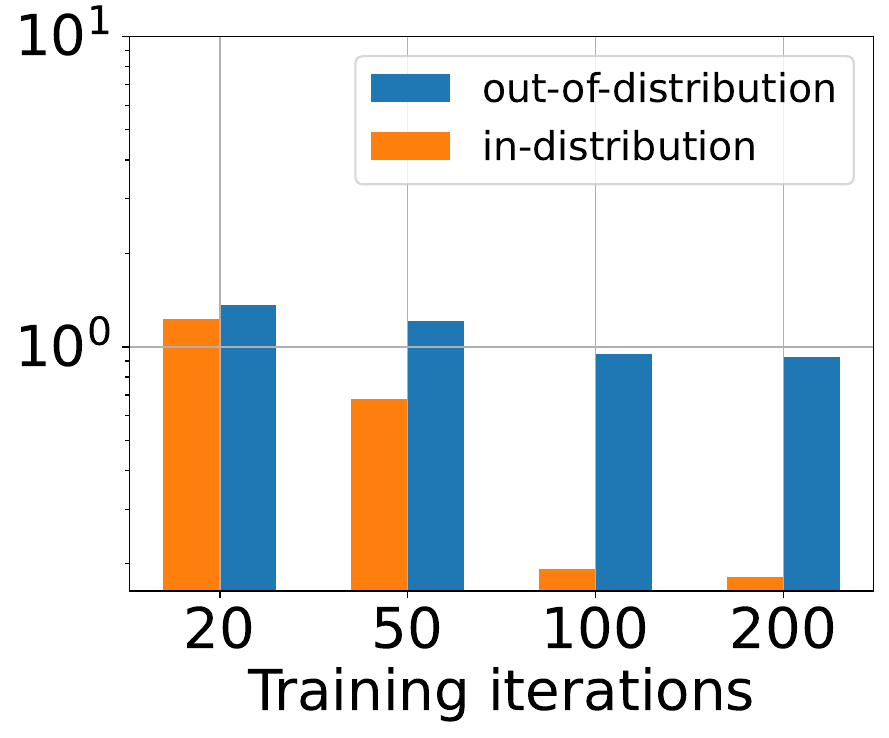}
\centering {\small{Epinet}}
\label{fig:epinet_difficult_to_choose_stopping_time}
\end{minipage}
\hfill
\begin{minipage}[b]{0.31\textwidth}
\centering \includegraphics[height=4cm, width=\textwidth]{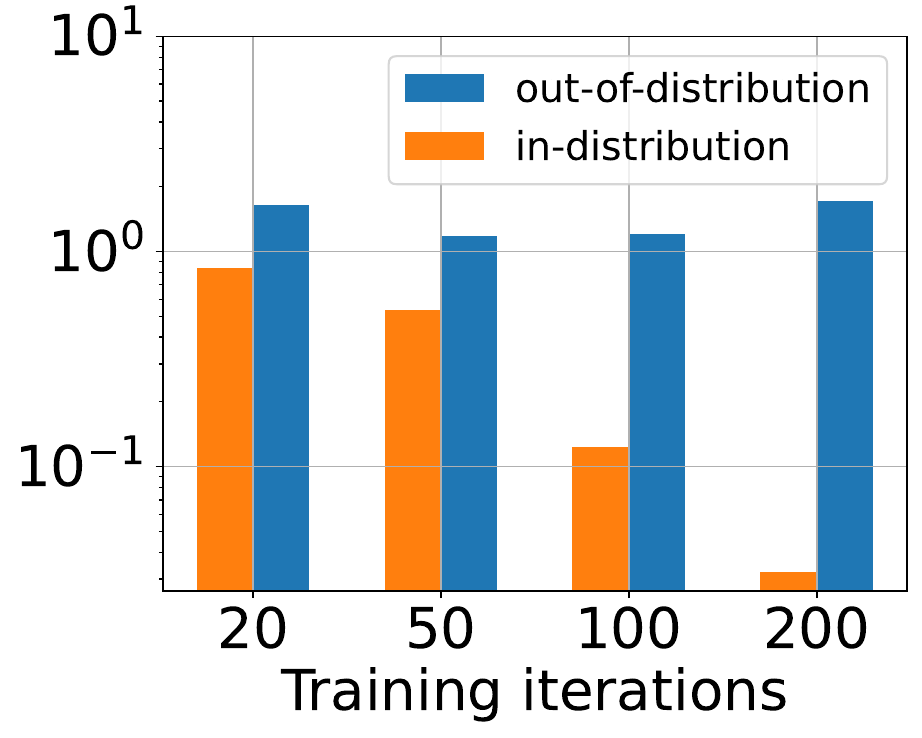}
\centering {\small{Hypermodel}}
\label{fig:hypermodel_difficult_to_choose_stopping_time}
\end{minipage}
\caption{Trade-off between  in-distribution (ID) performance and out-of-distribution (OOD) performance with stopping time as the hyperparameter (eICU, clustering).}
\label{fig:difficult_to_choose_stopping_time}
\end{figure}

Figure~\ref{fig:difficult_to_choose_weight_decay} demonstrates that increasing the weight decay deteriorates ID performance while first improving OOD performance and subsequently deteriorating.  
Similarly, Figure~\ref{fig:difficult_to_choose_stopping_time} shows that as number of iterations increases, ID performance improves on the test set.  On the other hand, test OOD performance first improves with increasing training iterations but later deteriorates for \ensembleplus and Hypermodels.

\subsubsection{Dynamic settings for   other datasets}

\begin{figure}[h]
\centering
\begin{minipage}[b]{0.24\textwidth}
\centering
\includegraphics[width = \textwidth, height=2.5cm]{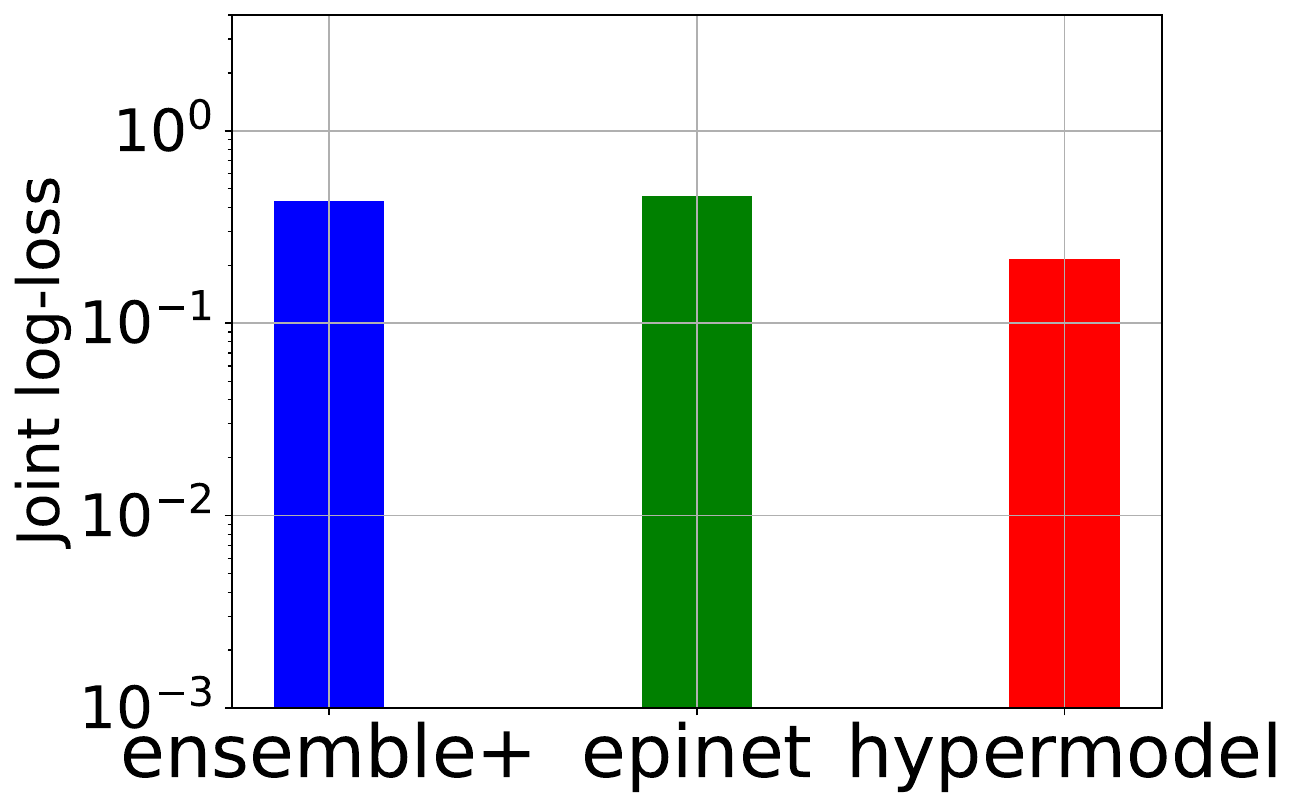}
{\small{{(a)} ID performance ($T=0$) }}
\end{minipage}
\hfill
\begin{minipage}[b]{0.24\textwidth}
\centering \includegraphics[width = \textwidth, height=2.5cm]{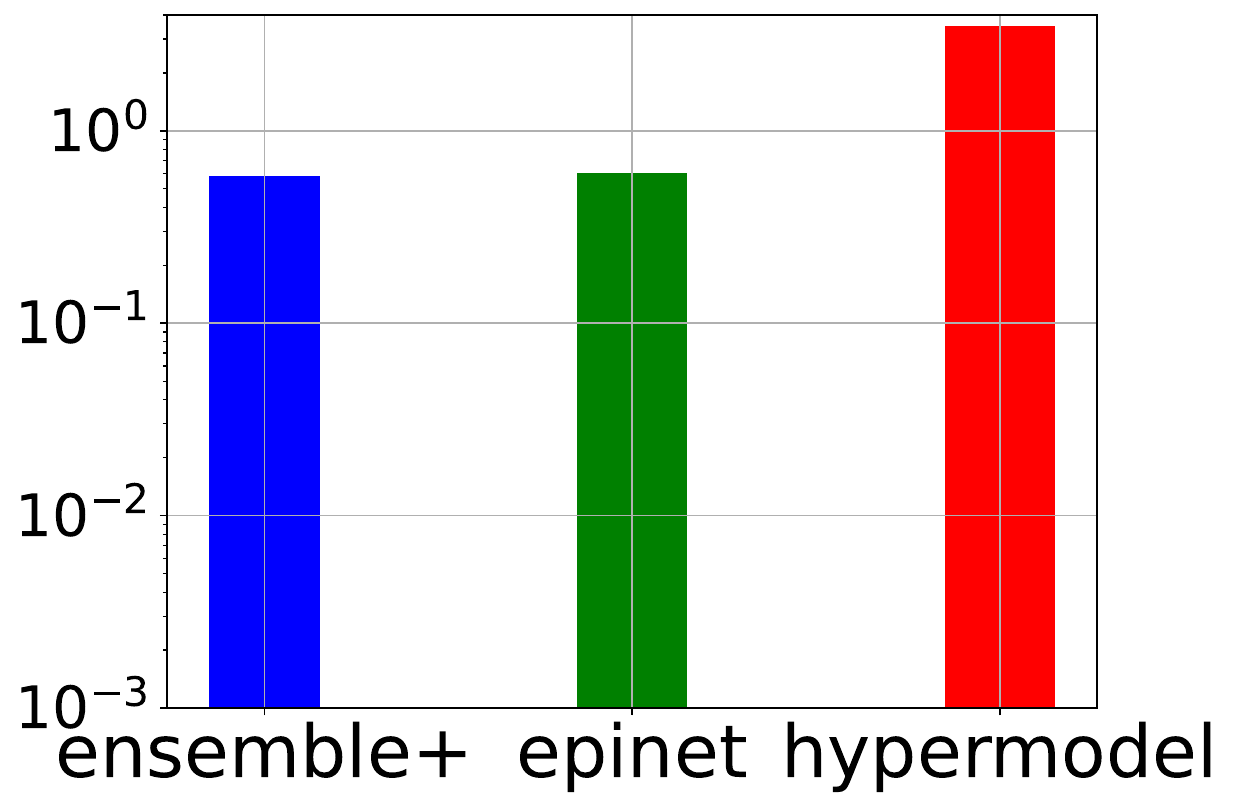}
{\small{{(b)} OOD performance ($T=0$) }}
\end{minipage}
\hfill
\begin{minipage}[b]{0.24\textwidth}
\centering \includegraphics[width = \textwidth, height=2.5cm]{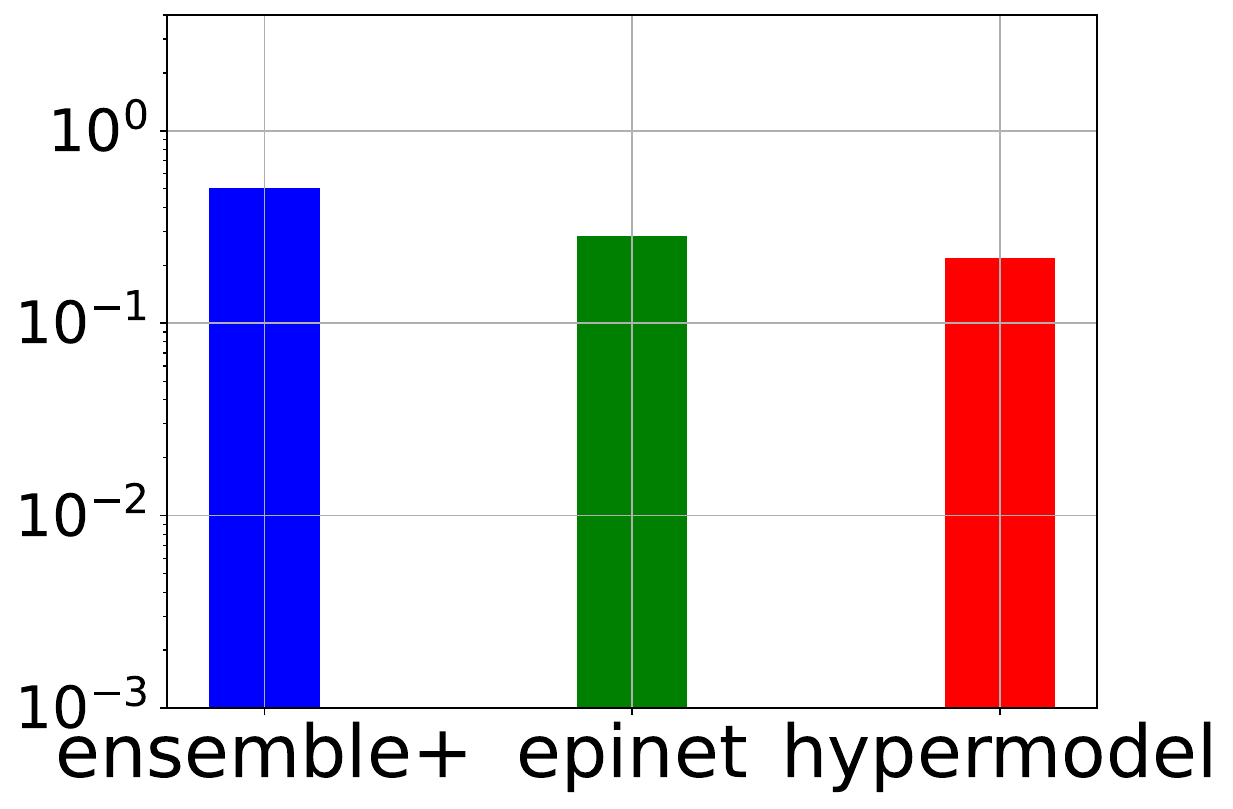}
{\small{{(c)} OOD performance ($T=1$) }}
\end{minipage}
\hfill
\begin{minipage}[b]{0.24\textwidth}
\centering \includegraphics[width = \textwidth, height=2.5cm]{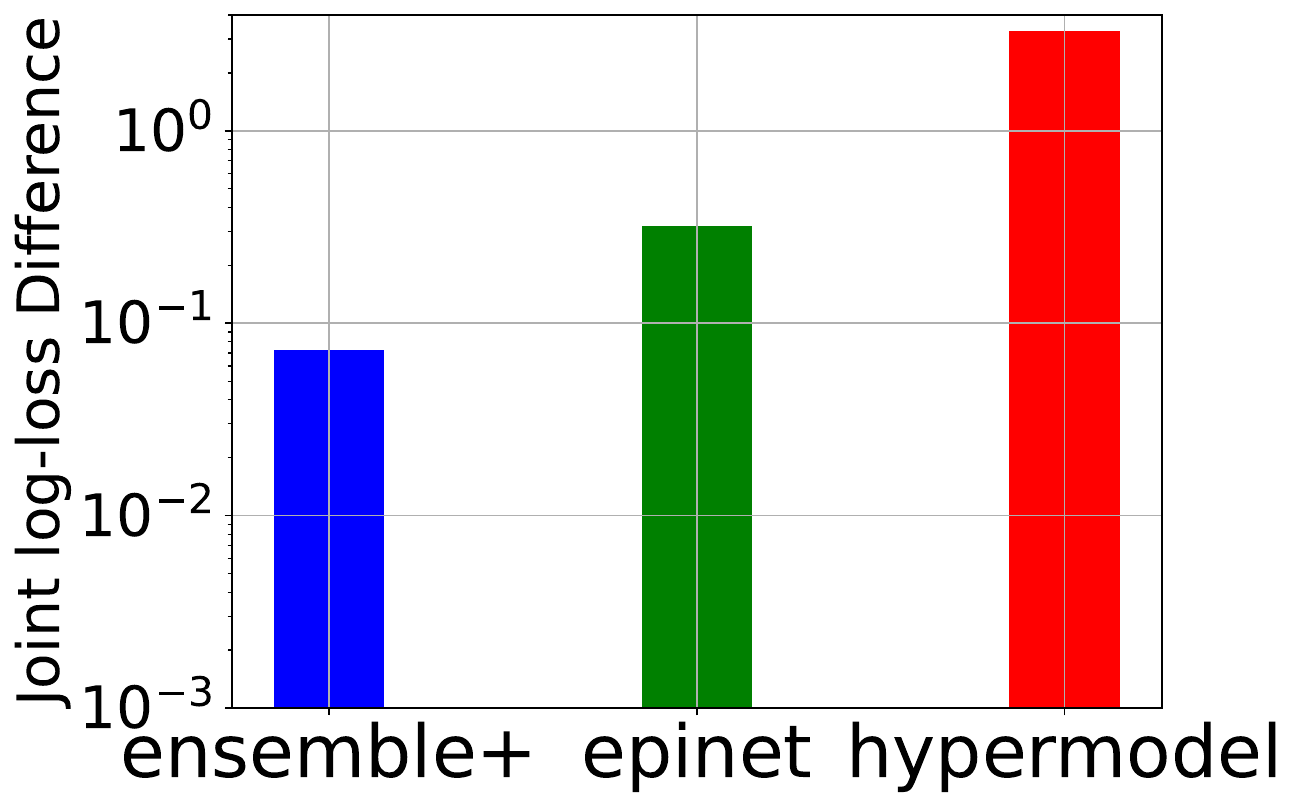}
{\small{{(d)} OOD improvement ($T=0 \to 1$) }} \end{minipage}
\caption{Performance of different UQ modules in a dynamic setting for \ieeecis.
Although Hypermodels perform the best in the ID setting (a), they perform the worst in the OOD setting at $T=0$. However, we see that Hypermodels are good at adapting to new data for this dataset and quickly improve the performance as shown in plot (d).
}
\label{fig:dynamic_setting_IEEECIS}
\end{figure}

\begin{figure}[h]
\centering
\begin{minipage}[b]{0.24\textwidth}
\centering
\includegraphics[width = \textwidth, height=2.5cm]{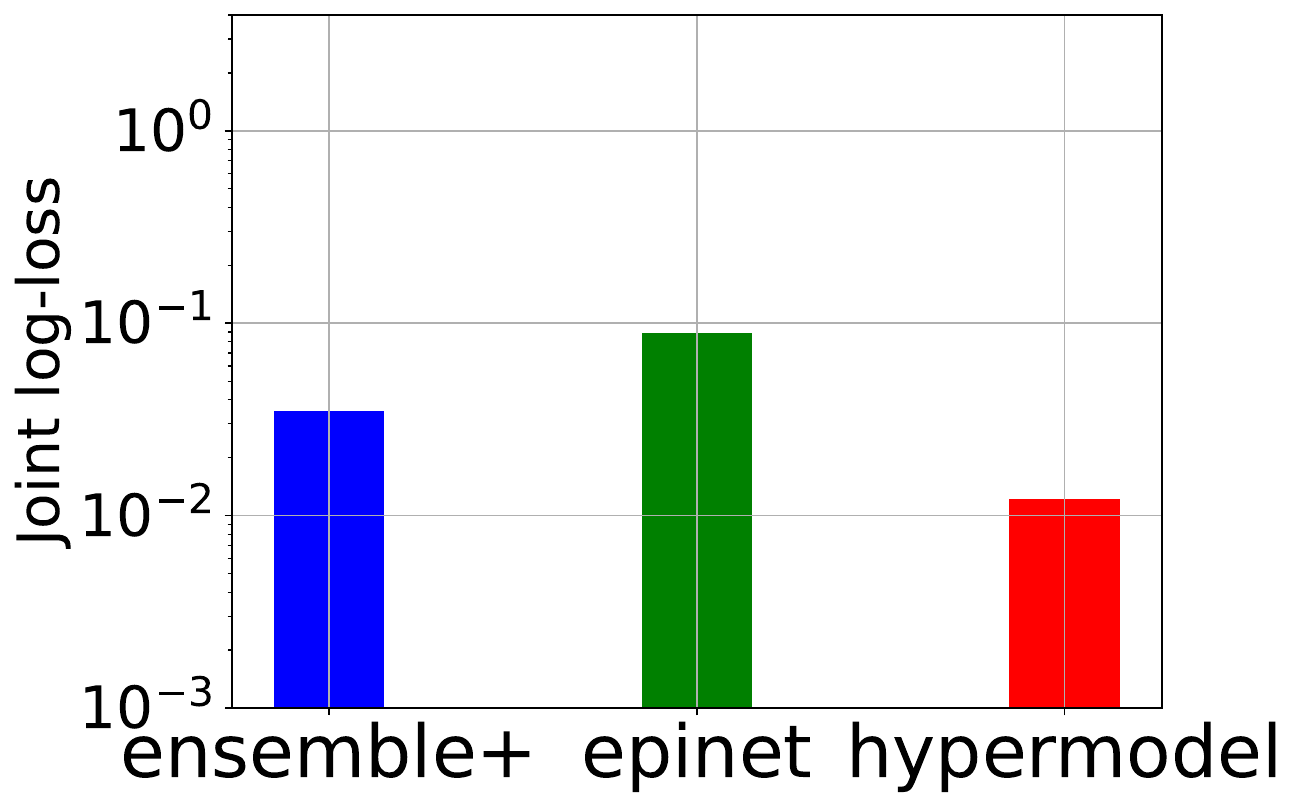}
{\small{{(a)} ID performance ($T=0$) }}  
\end{minipage}
\hfill
\begin{minipage}[b]{0.24\textwidth}
\centering \includegraphics[width = \textwidth, height=2.5cm]{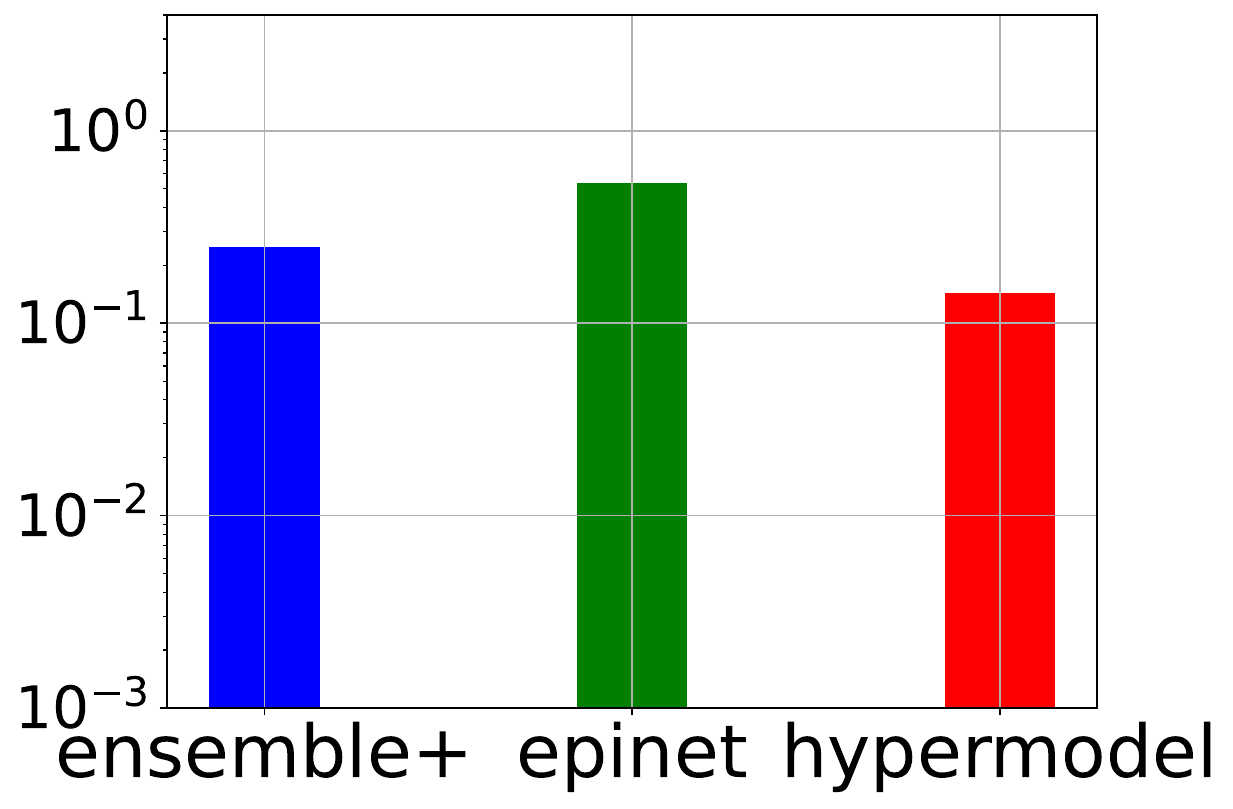}
{\small{{(b)} OOD performance ($T=0$) }} 
\end{minipage}
\hfill
\begin{minipage}[b]{0.24\textwidth}
\centering \includegraphics[width = \textwidth, height=2.5cm]{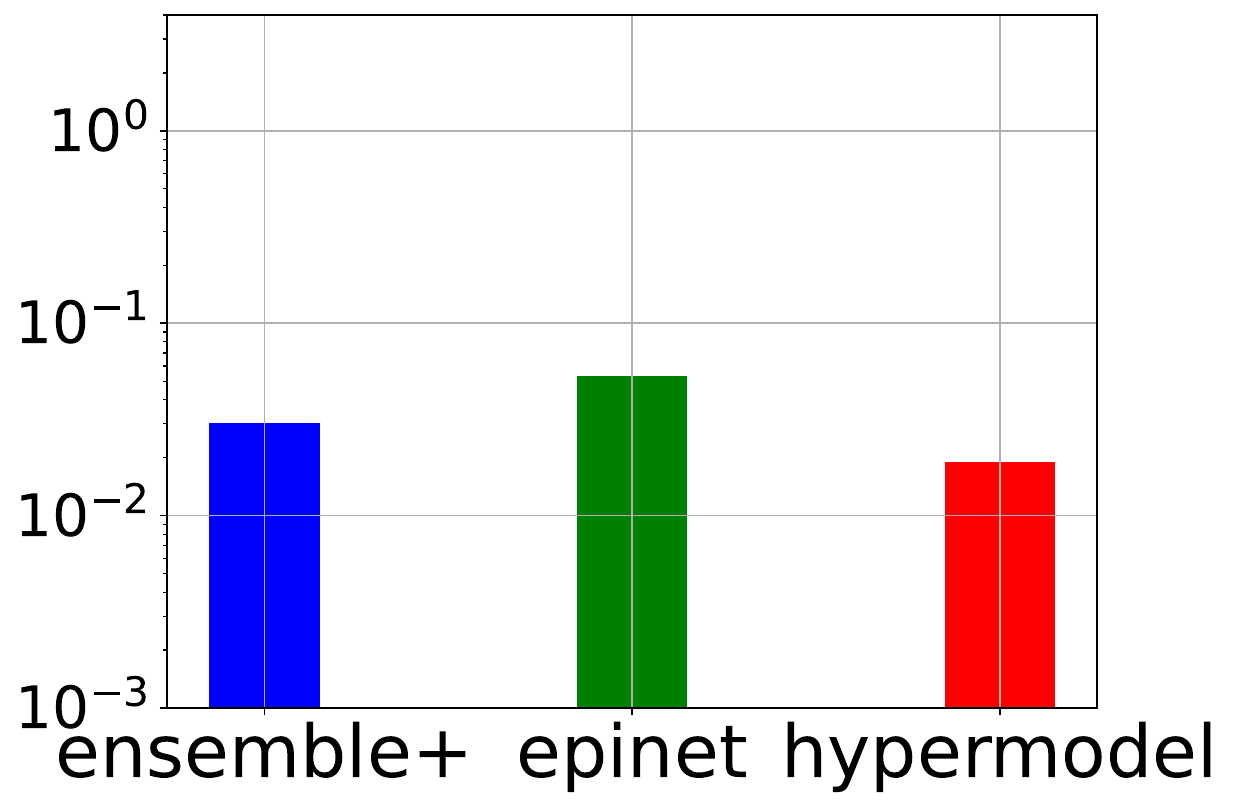}
{\small{{(c)} OOD performance ($T= 1$) }} 
\end{minipage}
\hfill
\begin{minipage}[b]{0.24\textwidth}
\centering \includegraphics[width = \textwidth, height=2.5cm]{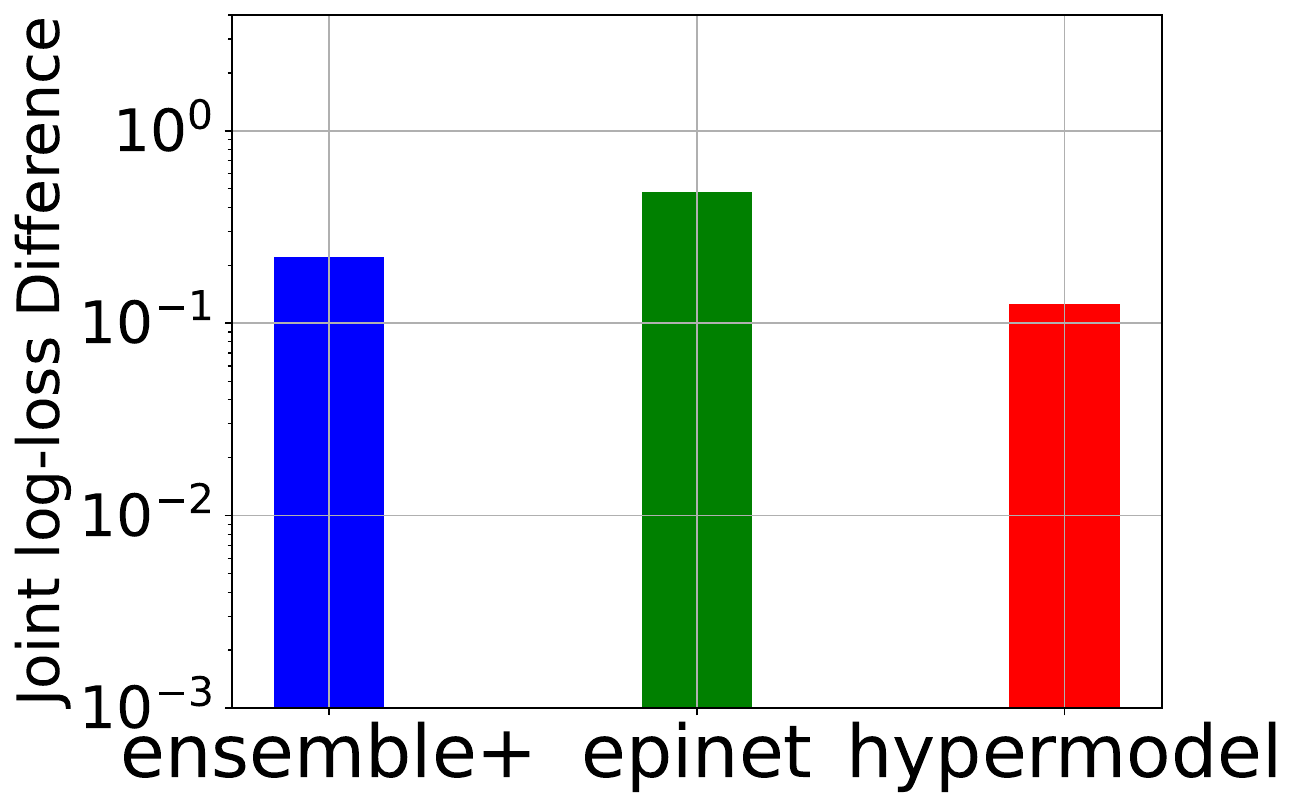}
{\small{{(d)} OOD improvement ($T=0 \to 1$) }} 
\end{minipage}
\caption{Performance of different UQ modules in a dynamic setting for \ccfraud. 
In this case hypermodels perform the best for both the ID setting (a) and the OOD setting (b) at $T=0$. However, we can see that the OOD performance improvement (d) is the best for the epinets, which again showcases the trade-off between having sharper posteriors and the performance on the  OOD data.}
\label{fig:dynamic_setting_ccfraud}
\end{figure}

\begin{figure}[h]
\centering
\begin{minipage}[b]{0.24\textwidth}
\centering
\includegraphics[width = \textwidth, height=2.5cm]{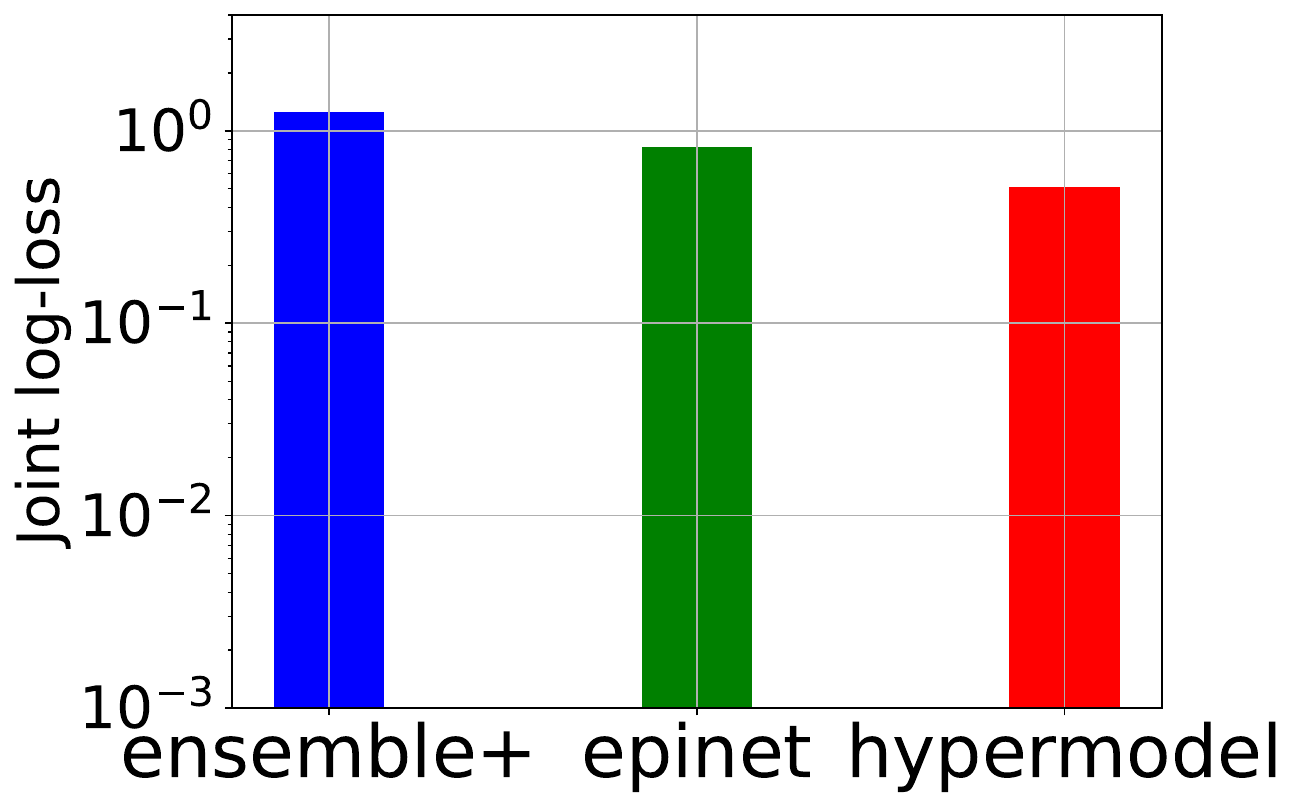}
{\small{{(a)} ID performance ($T=0$) }} 
\end{minipage}
\hfill
\begin{minipage}[b]{0.24\textwidth}
\centering \includegraphics[width = \textwidth, height=2.5cm]{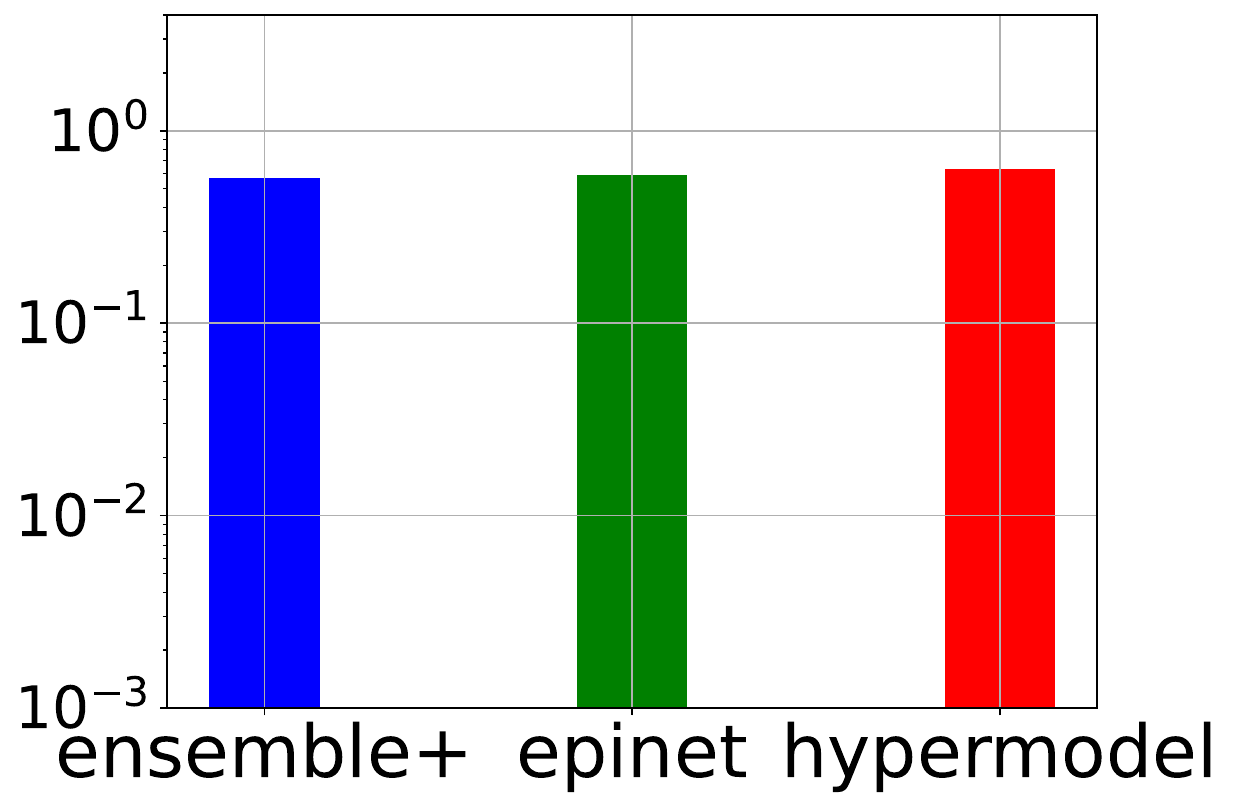}
{\small{{(b)} OOD performance ($T=0$) }} 
\end{minipage}
\hfill
\begin{minipage}[b]{0.24\textwidth}
\centering \includegraphics[width = \textwidth, height=2.5cm]{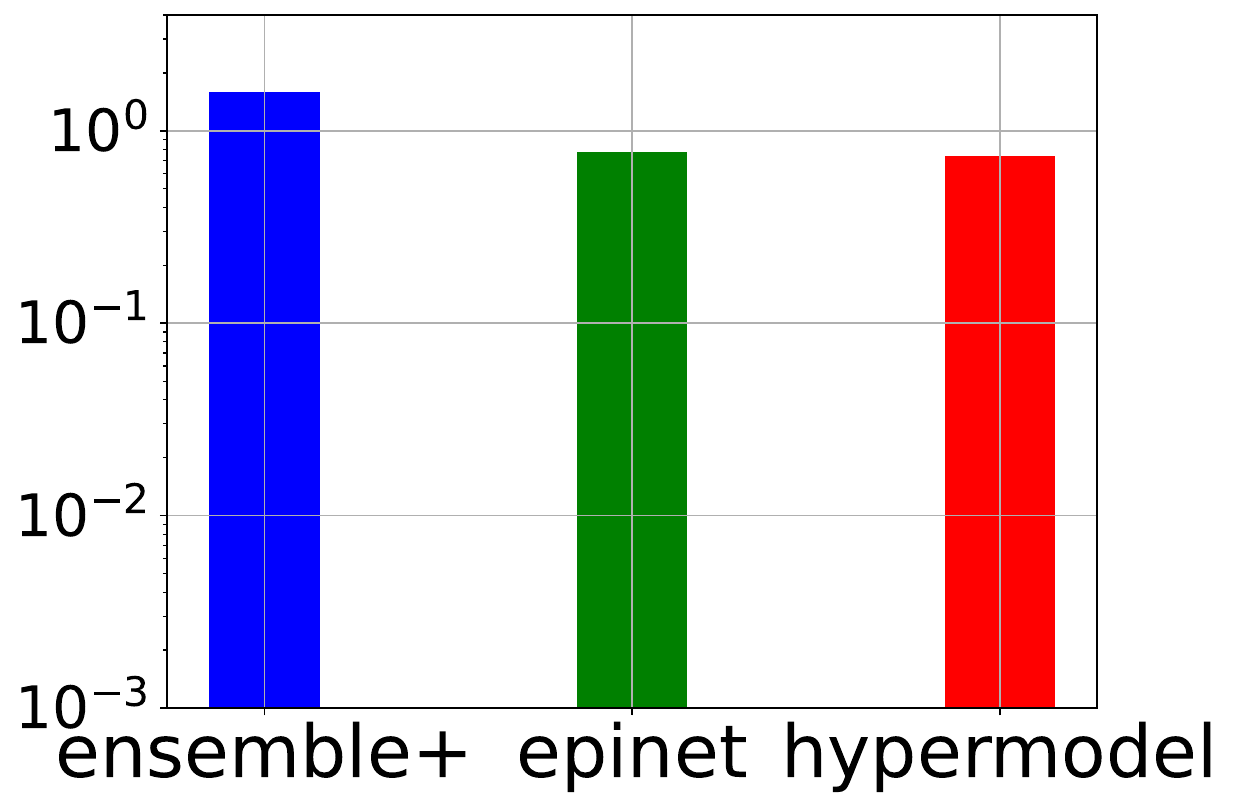}
{\small{{(c)} OOD performance ($T=1$) }} 
\end{minipage}
\hfill
\begin{minipage}[b]{0.24\textwidth}
\centering \includegraphics[width = \textwidth, height=2.5cm]{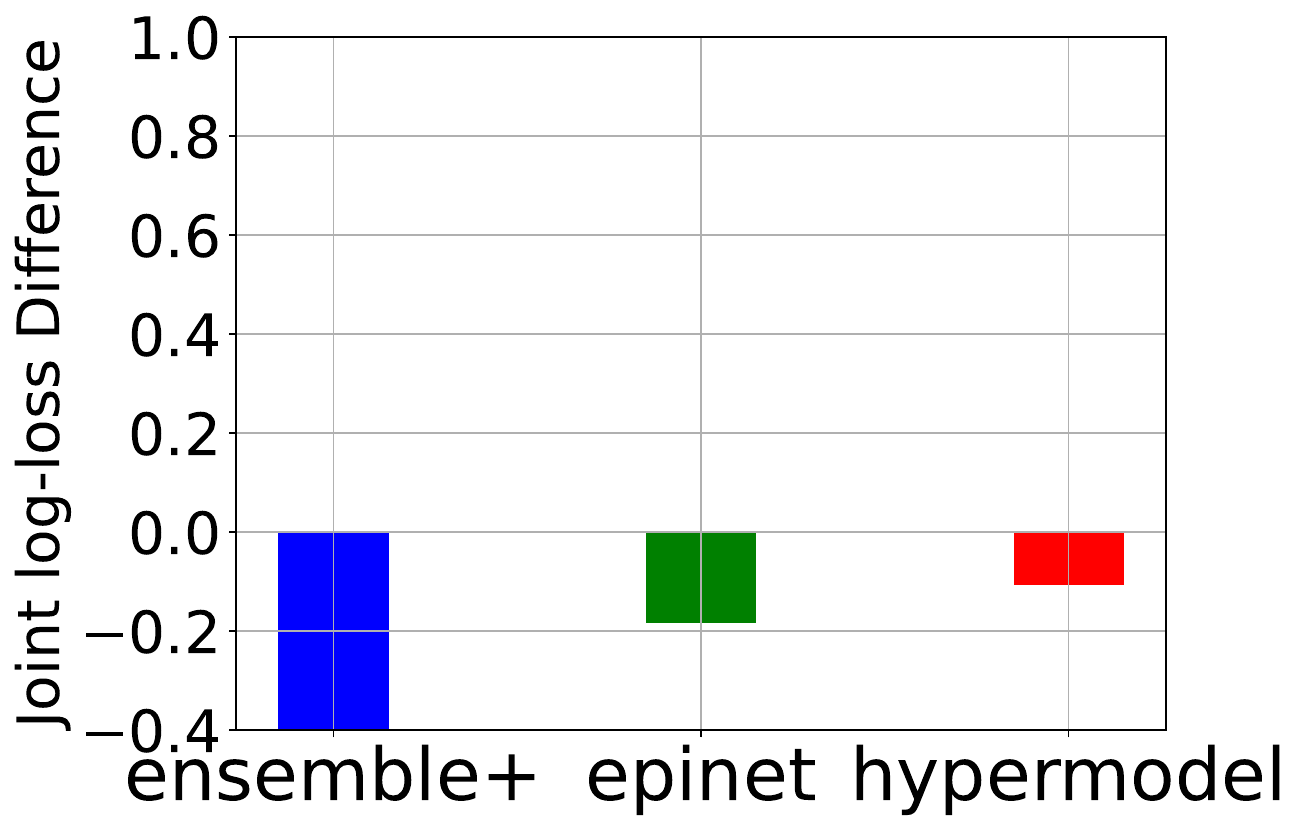}
{\small{{(d)} OOD improvement ($T=0 \to 1$) }} 
\end{minipage}
\caption{Performance of different agents in a dynamic setting for \fraudecom. Surprisingly, we observe that all three models have a larger OOD joint log-loss after seeing new data. The reason for such a behaviour might be that the models start from a prior belief with lower level of uncertainty and as they see additional data they become more uncertain. }
\label{fig:dynamic_setting_fraudecom}
\end{figure}

\begin{figure}[h]
\centering
\begin{minipage}[b]{0.24\textwidth}
\centering
\includegraphics[width = \textwidth]{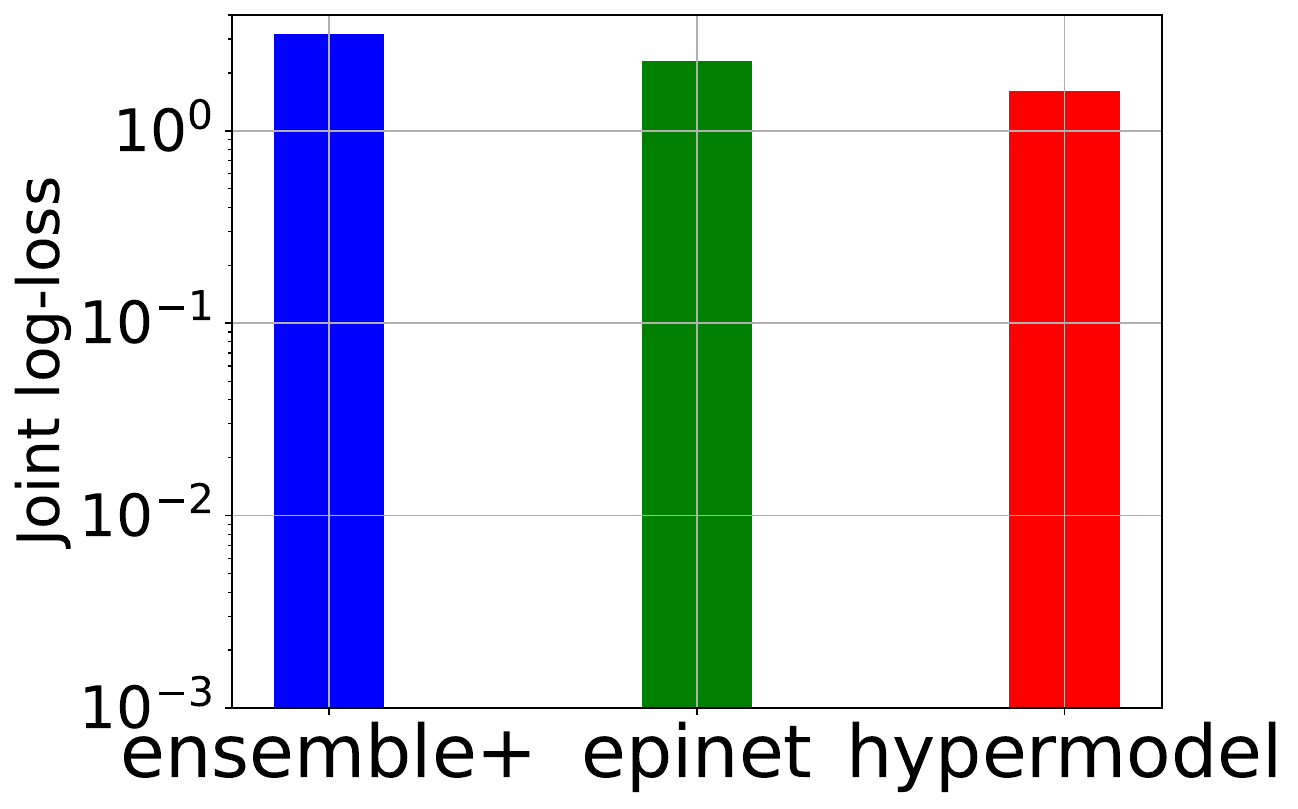}
{\small{{(a)} ID performance ($T=0$) }} 
\end{minipage}
\hfill
\begin{minipage}[b]{0.24\textwidth}
\centering \includegraphics[width = \textwidth, height=2.5cm]{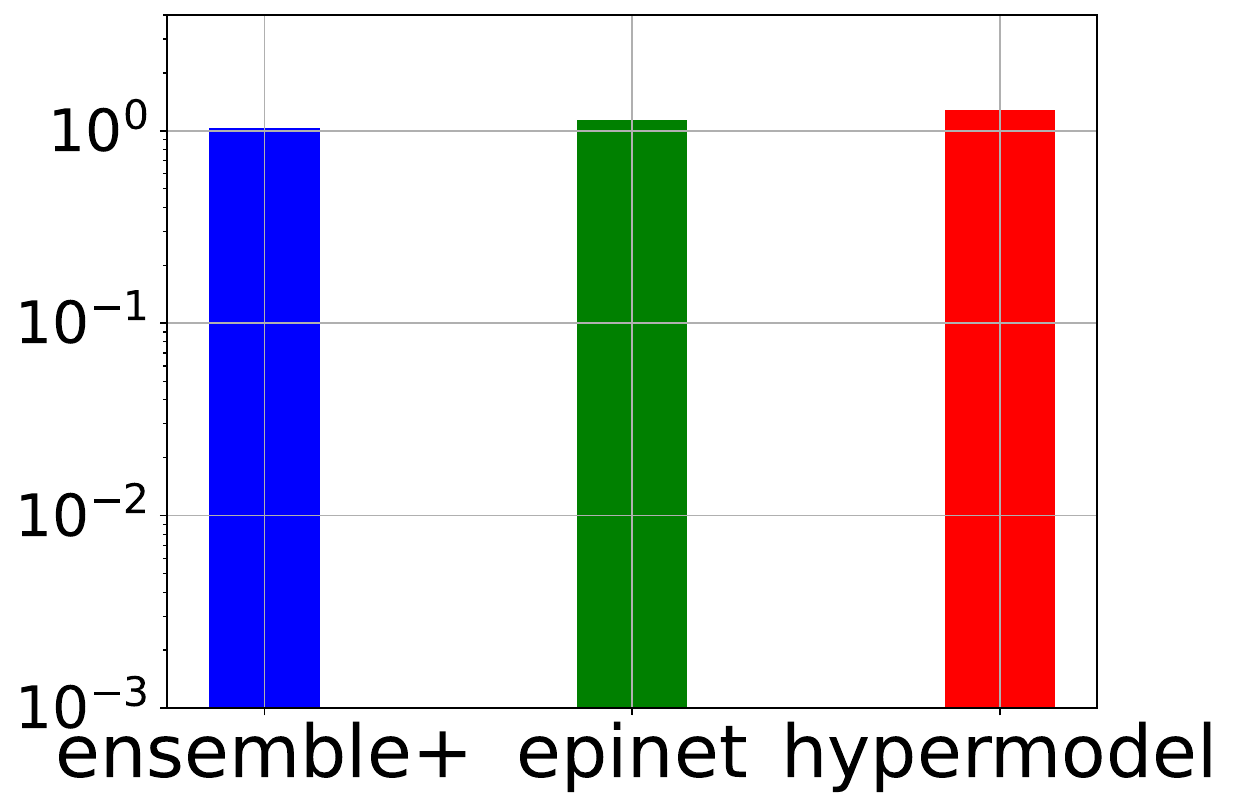}
{\small{{(b)} OOD performance ($T=0$) }} 
\end{minipage}
\hfill
\begin{minipage}[b]{0.24\textwidth}
\centering \includegraphics[width = \textwidth]{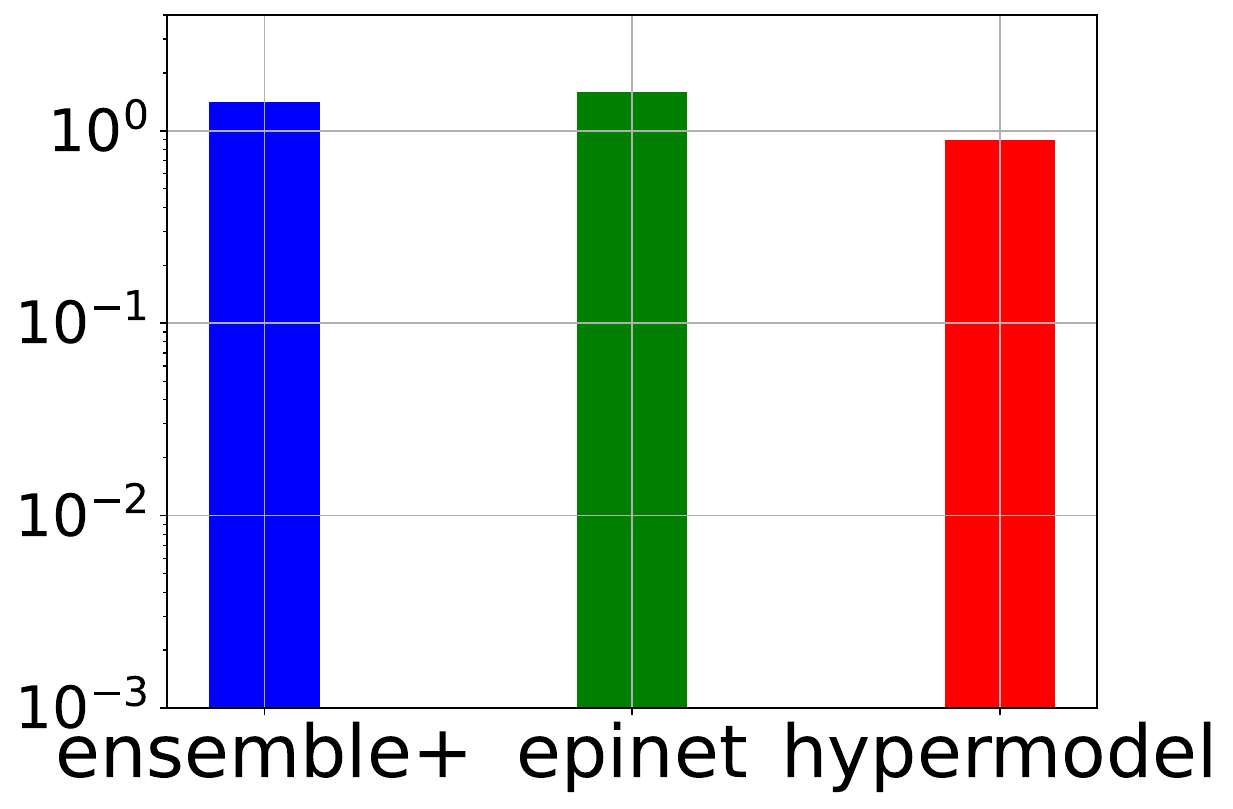}
{\small{{(c)} OOD performance ($T=1$) }}
\end{minipage}
\hfill
\begin{minipage}[b]{0.24\textwidth}
\centering \includegraphics[width = \textwidth, height=2.5cm]{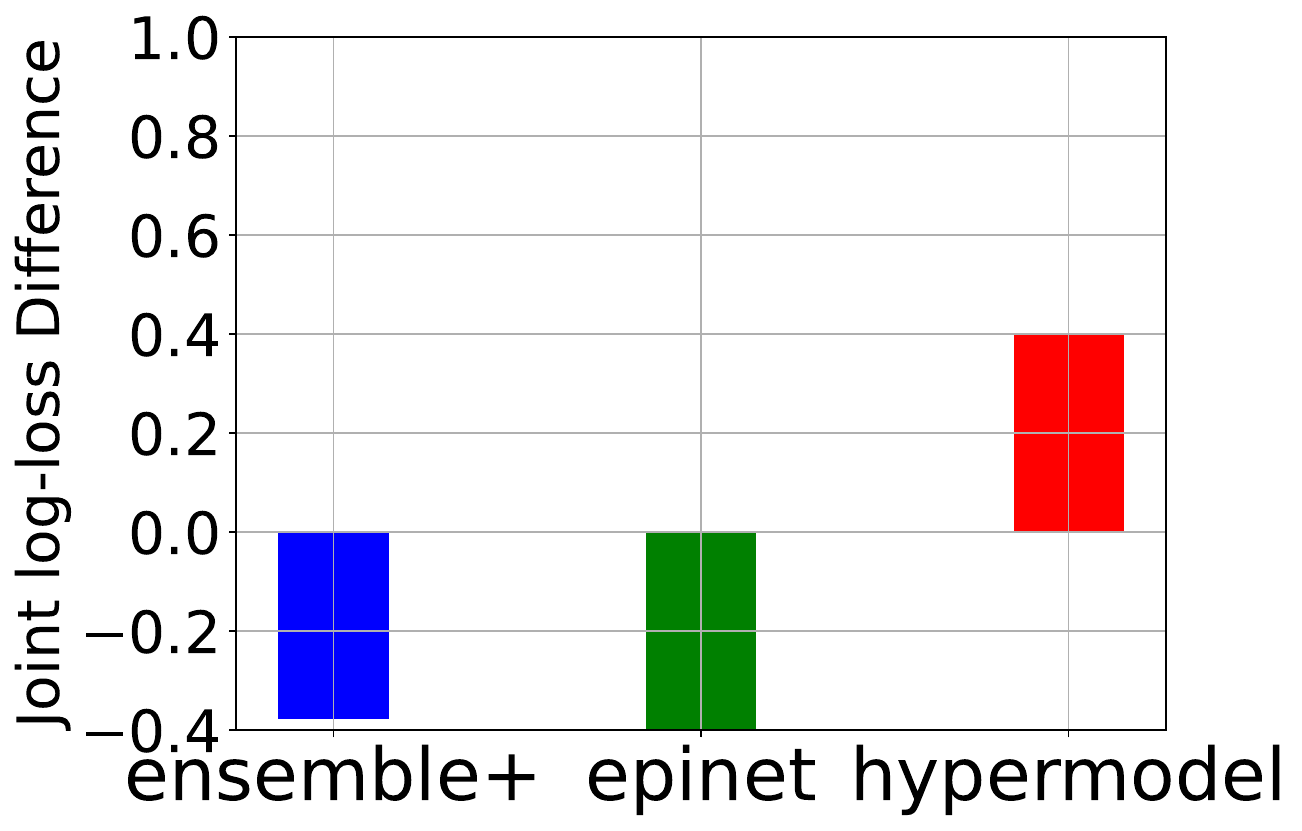}
{\small{{(d)} OOD improvement ($T=0 \to 1$) }}
\end{minipage}
\caption{Performance of different agents in a dynamic setting for \vehicleloan.
Although hypermodels perform the best in the ID setting (a), they perform the worst in the OOD setting at $T=0$. However, we see that hypermodels are good at adapting to new data for this dataset and quickly improve the performance as shown in plot (d). In addition, hypermodels are the only model that have a lower joint log-loss after seeing new data.
}
\label{fig:dynamic_setting_vehicleloan}
\end{figure}

\begin{figure}[h]
\centering
\begin{minipage}[b]{0.24\textwidth}
\centering
\includegraphics[width = \textwidth, height=2.5cm]{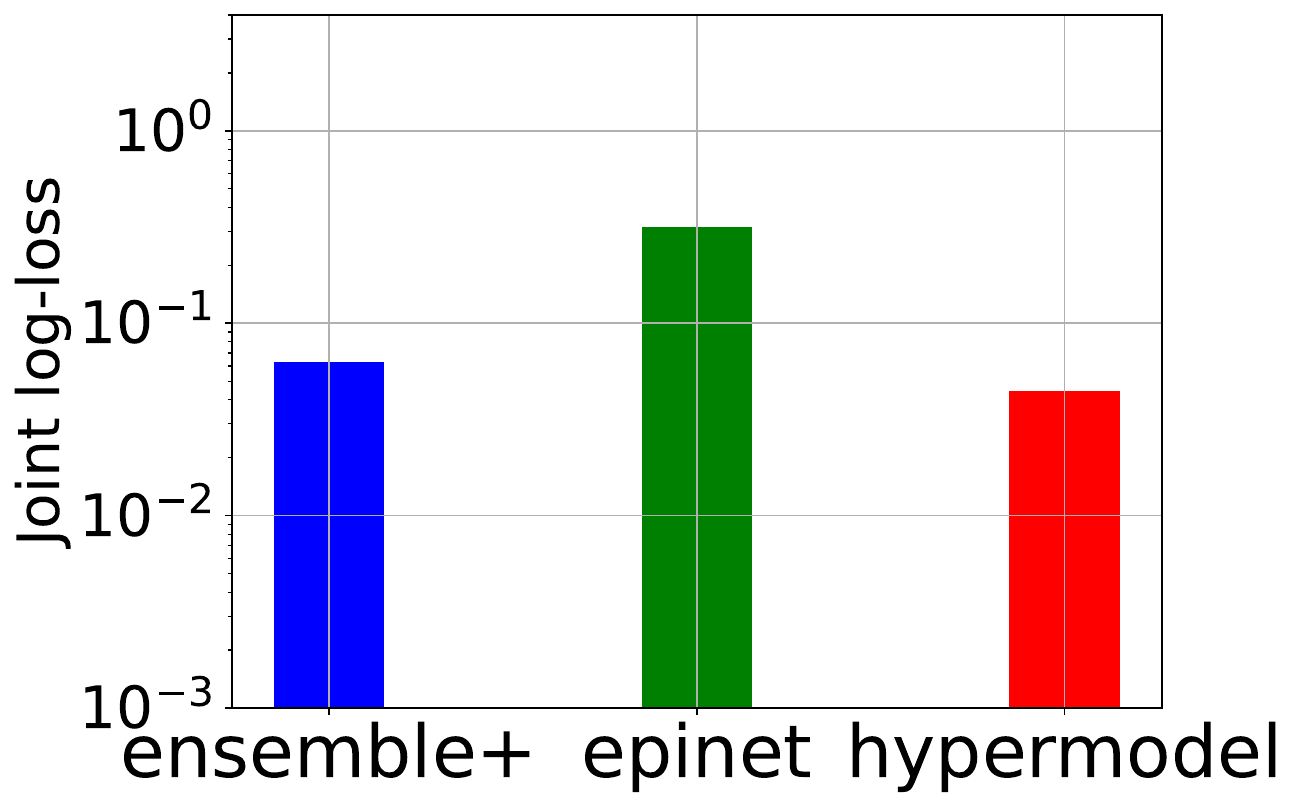}
{\small{{(a)} ID performance ($T=0$) }}
\end{minipage}
\hfill
\begin{minipage}[b]{0.24\textwidth}
\centering \includegraphics[width = \textwidth, height=2.5cm]{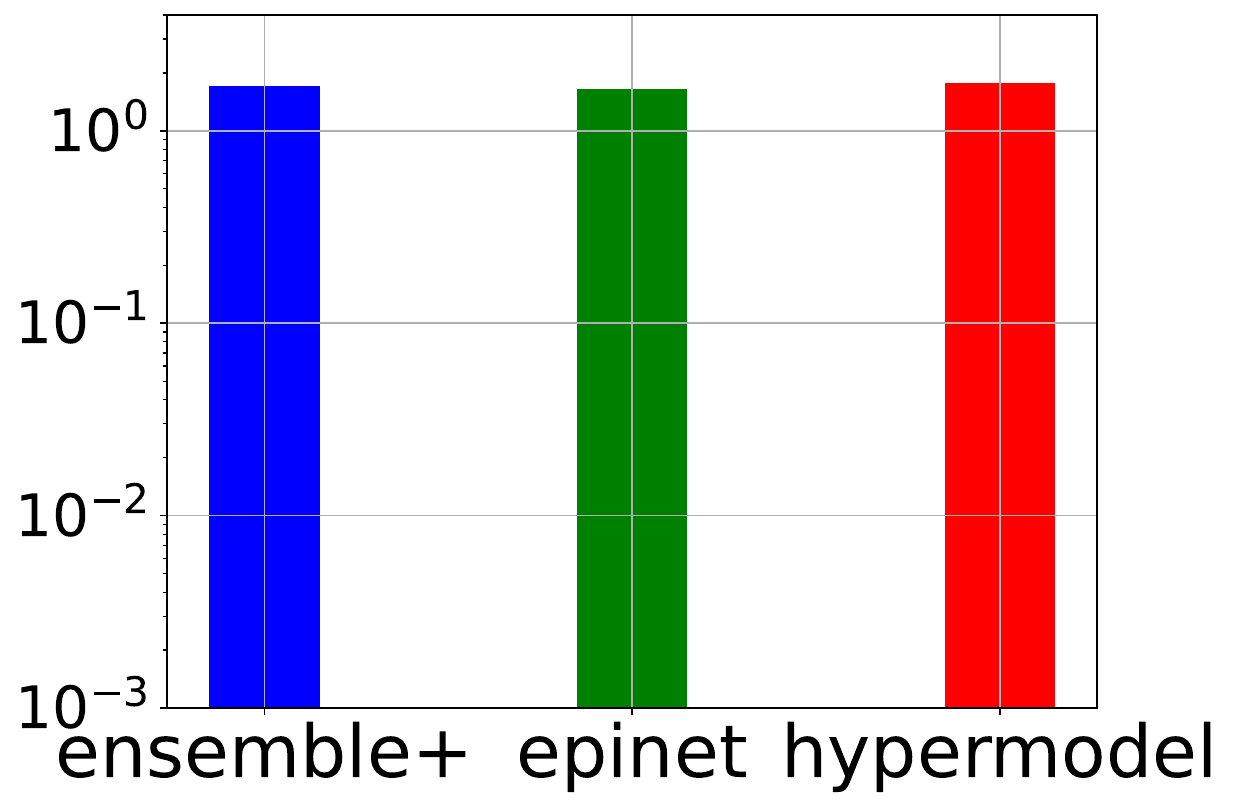}
{\small{{(b)} OOD performance ($T=0$) }}
\end{minipage}
\hfill
\begin{minipage}[b]{0.24\textwidth}
\centering \includegraphics[width = \textwidth, height=2.5cm]{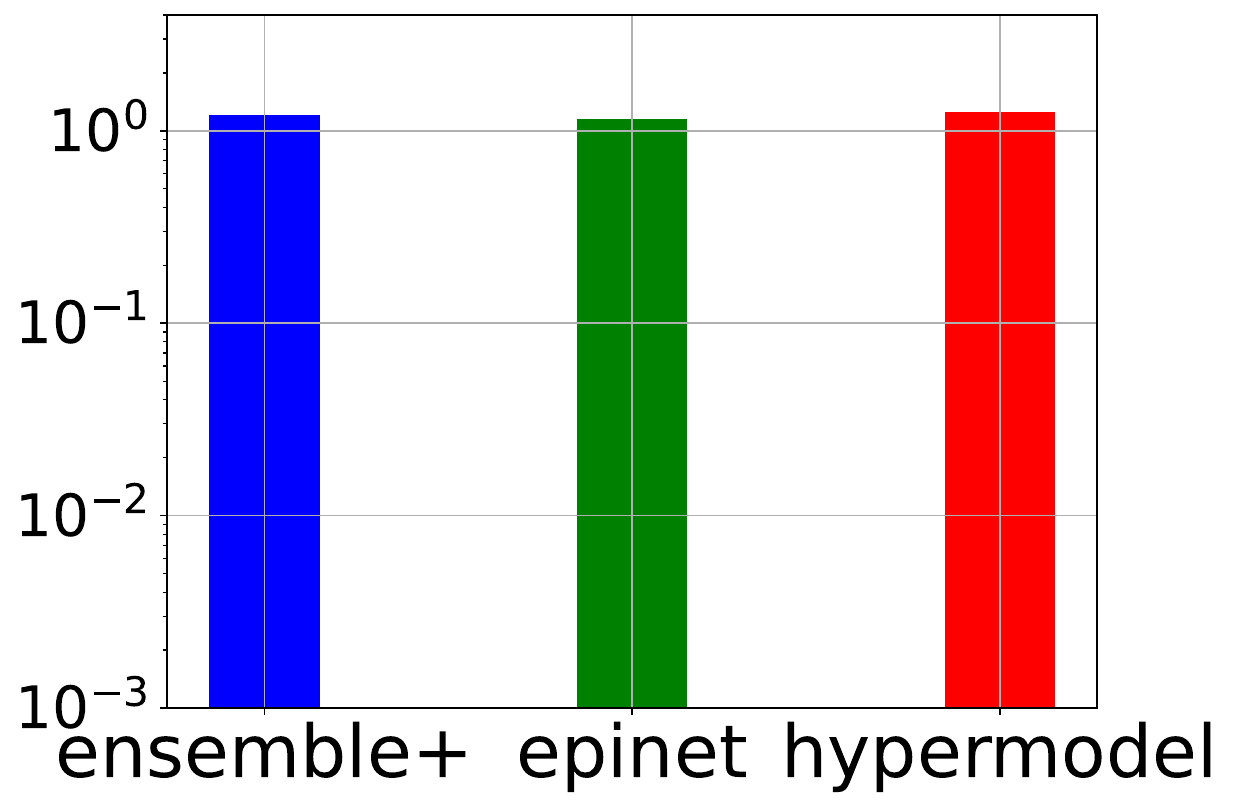}
{\small{{(c)} OOD performance ($T= 1$) }}
\end{minipage}
\hfill
\begin{minipage}[b]{0.24\textwidth}
\centering \includegraphics[width = \textwidth, height=2.5cm]{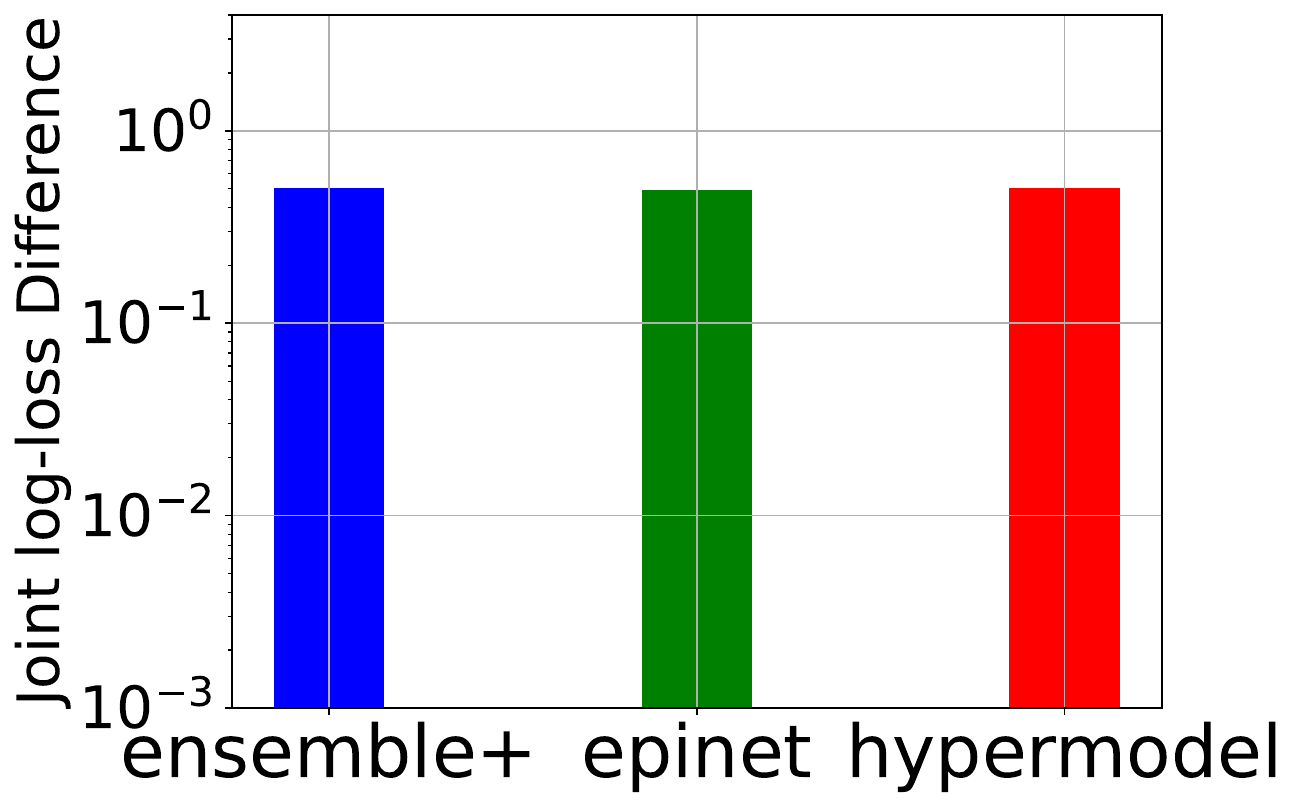}
{\small{{(d)} OOD improvement ($T=0 \to 1$) }}
\end{minipage}
\caption{Performance of different agents in a dynamic setting for \acsemployment. In this case hypermodels perform the best in both the ID setting (a) but in the OOD setting [(b),(c), and (d)], all the models perform similarly at $T=0$ and $T=1$.}
\label{fig:dynamic_setting_ACS_employment}
\end{figure}

\begin{figure}[h]
\centering
\begin{minipage}[b]{0.24\textwidth}
\centering
\includegraphics[width = \textwidth, height=2.5cm]{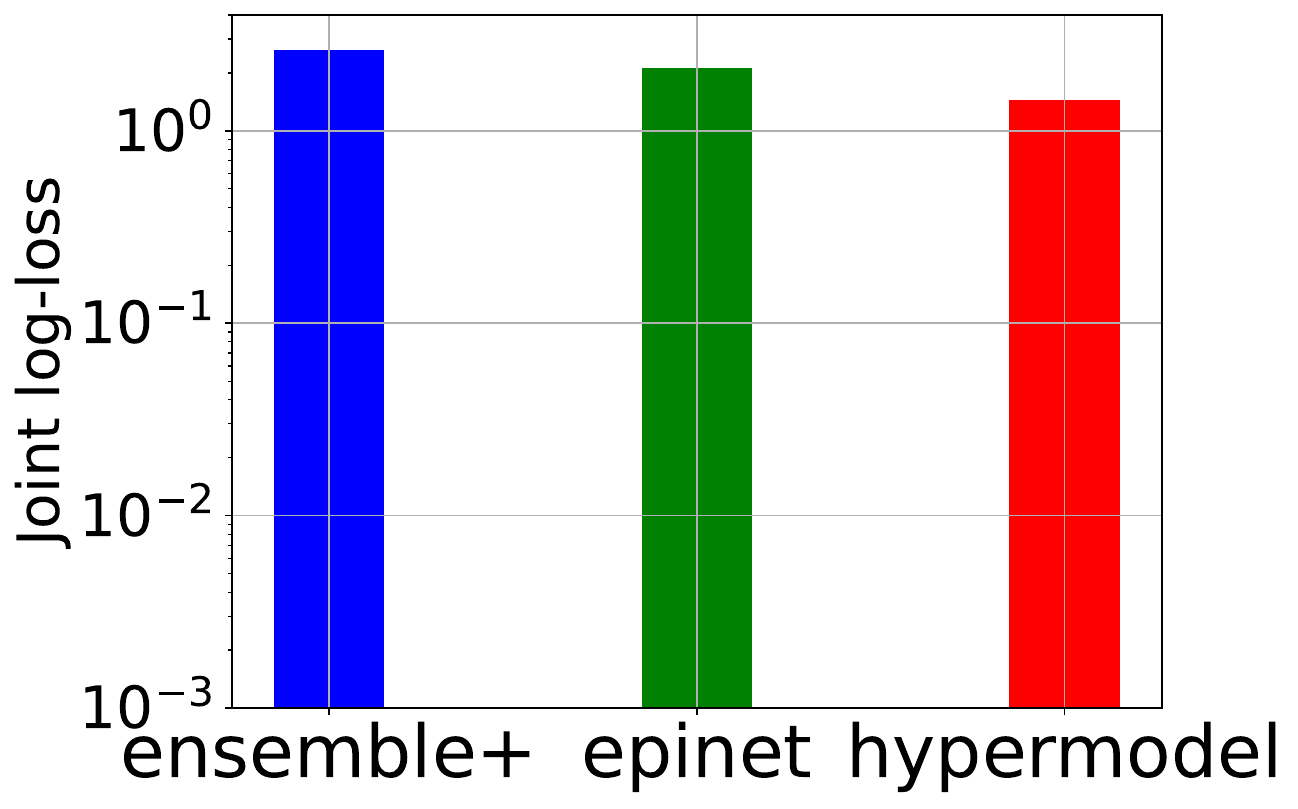}
{\small{{(a)} ID performance ($T=0$) }}
\end{minipage}
\hfill
\begin{minipage}[b]{0.24\textwidth}
\centering \includegraphics[width = \textwidth, height=2.5cm]{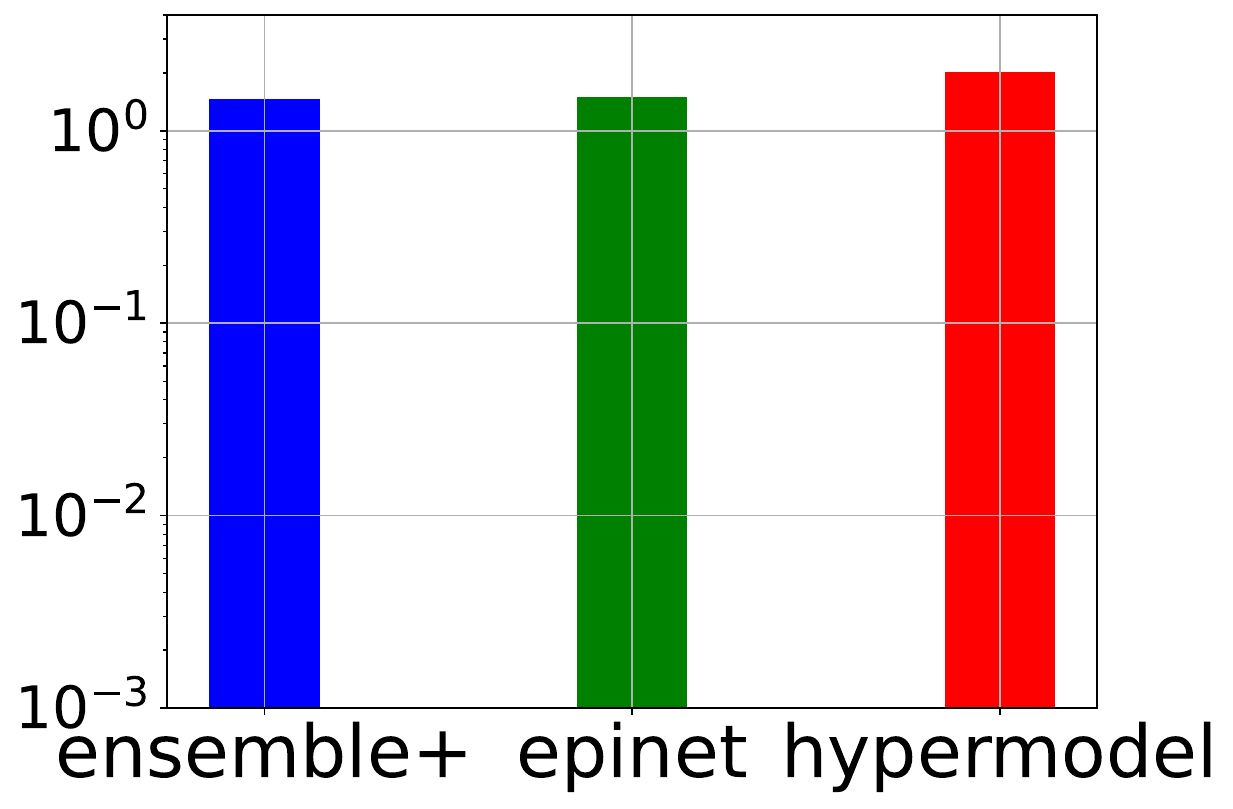}
{\small{{(b)} OOD performance ($T=0$) }}
\end{minipage}
\hfill
\begin{minipage}[b]{0.24\textwidth}
\centering \includegraphics[width = \textwidth, height=2.5cm]{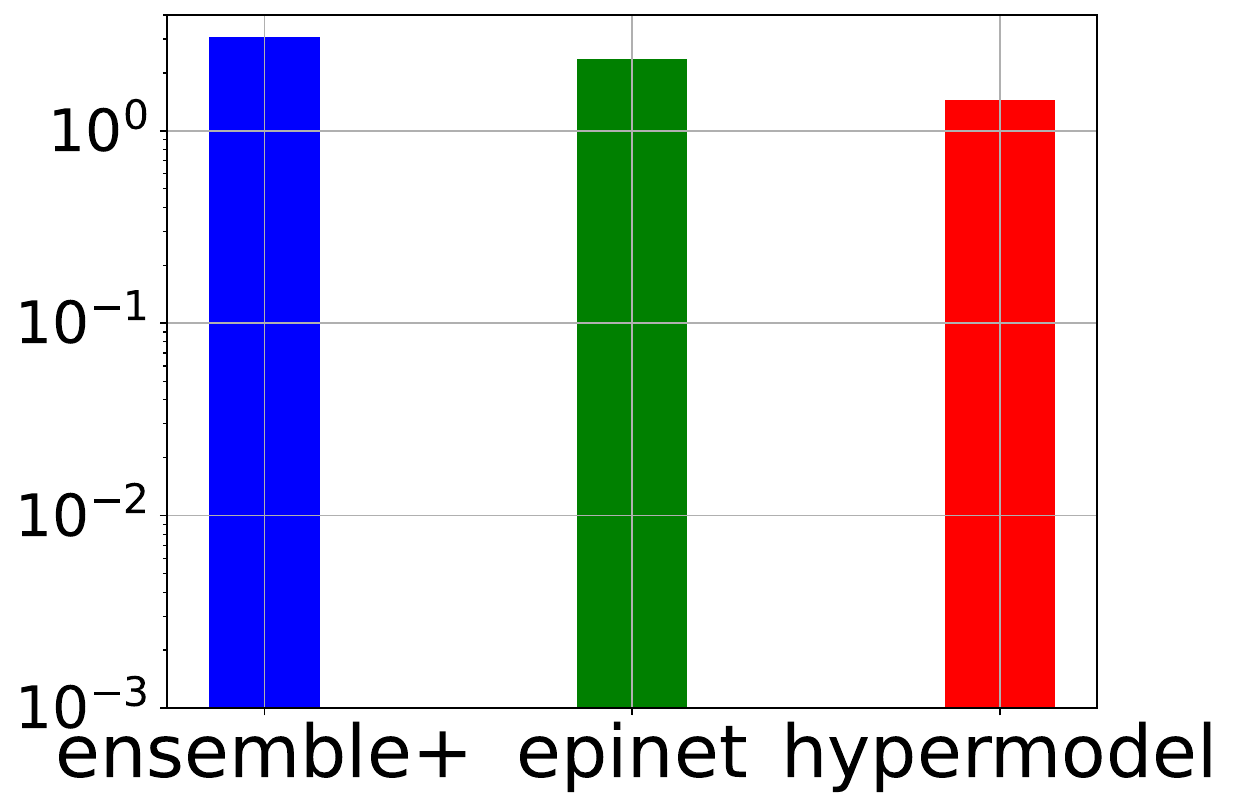}
{\small{{(c)} OOD performance ($T=1$) }}
\end{minipage}
\hfill
\begin{minipage}[b]{0.24\textwidth}
\centering \includegraphics[width = \textwidth, height=2.5cm]{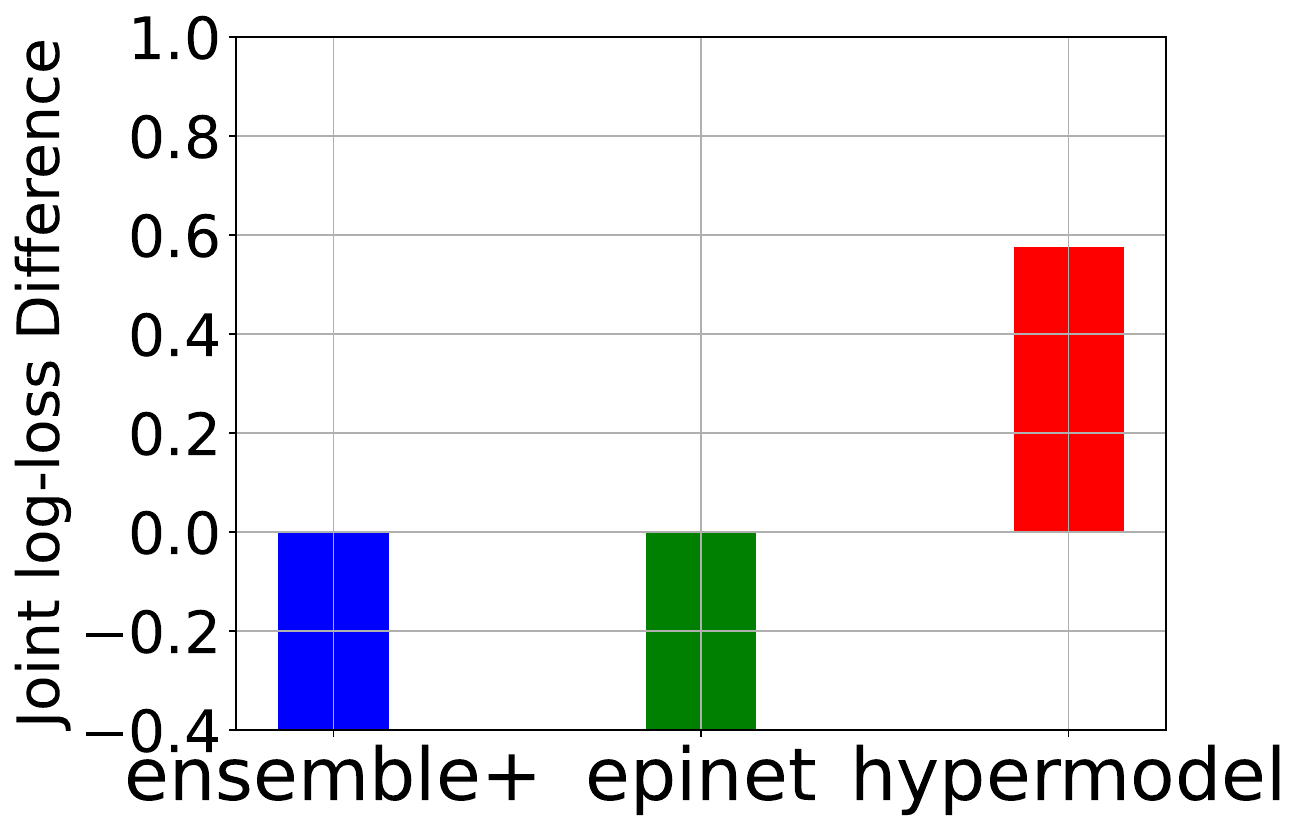}
{\small{{(d)} OOD improvement ($T=0 \to 1$) }}
\end{minipage}
\caption{Performance of different agents in a dynamic setting for \acsincome.
Although hypermodels perform the best in the ID setting (a), they perform the worst in the OOD setting at $T=0$. However, we see that hypermodels are good at adapting to new data for this dataset and quickly improve the performance as shown in plot (d). In addition, hypermodels are the only model that have a lower joint log-loss after seeing new data.
}
\label{fig:dynamic_setting_ACSincome_NY_}
\end{figure}

To further examine our findings in Figure~\ref{fig:dynamic_setting_k_30}, we conduct similar experiments using clustering bias as described in Section~\ref{sec:selection-bias} for all datasets in Section~\ref{sec:UQ_BENCH_datasets}.
We summarize our results in Figure~\ref{fig:dynamic_setting_IEEECIS}
for \ieeecis, Figure~\ref{fig:dynamic_setting_ccfraud}
for \ccfraud, Figure~\ref{fig:dynamic_setting_fraudecom}
for \fraudecom,  Figure~\ref{fig:dynamic_setting_vehicleloan}
for \vehicleloan,  Figure~\ref{fig:dynamic_setting_ACS_employment}
for \acsemployment, and 
Figure~\ref{fig:dynamic_setting_ACSincome_NY_}
for \acsincome. 
In summary, our results are consistent with the results from Figure~\ref{fig:dynamic_setting_k_30}. 
Interestingly, we observe some UQ agents can actually have a larger joint log-loss after adapting to new data.

\subsubsection{Hyperaprameter tuning suffers in dynamic settings}
\begin{figure}[h]
\centering
\begin{minipage}[b]{0.24\textwidth}
\centering
\includegraphics[width = \textwidth, height=3.25cm]{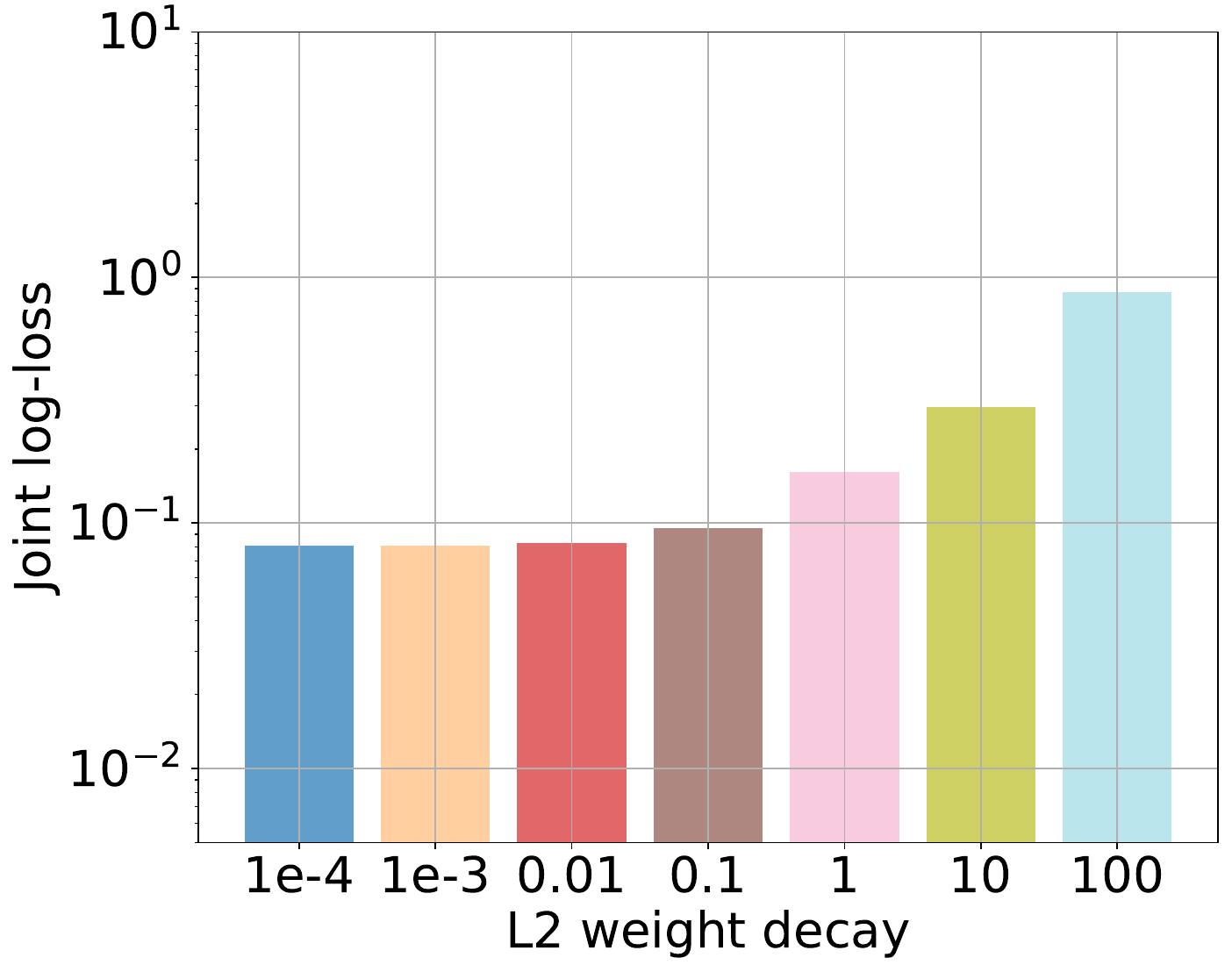}
{\small{{(a)} ID performance ($T=0 $) }}
\end{minipage}
\hfill
\begin{minipage}[b]{0.24\textwidth}
\centering \includegraphics[width = \textwidth, height=3.25cm]{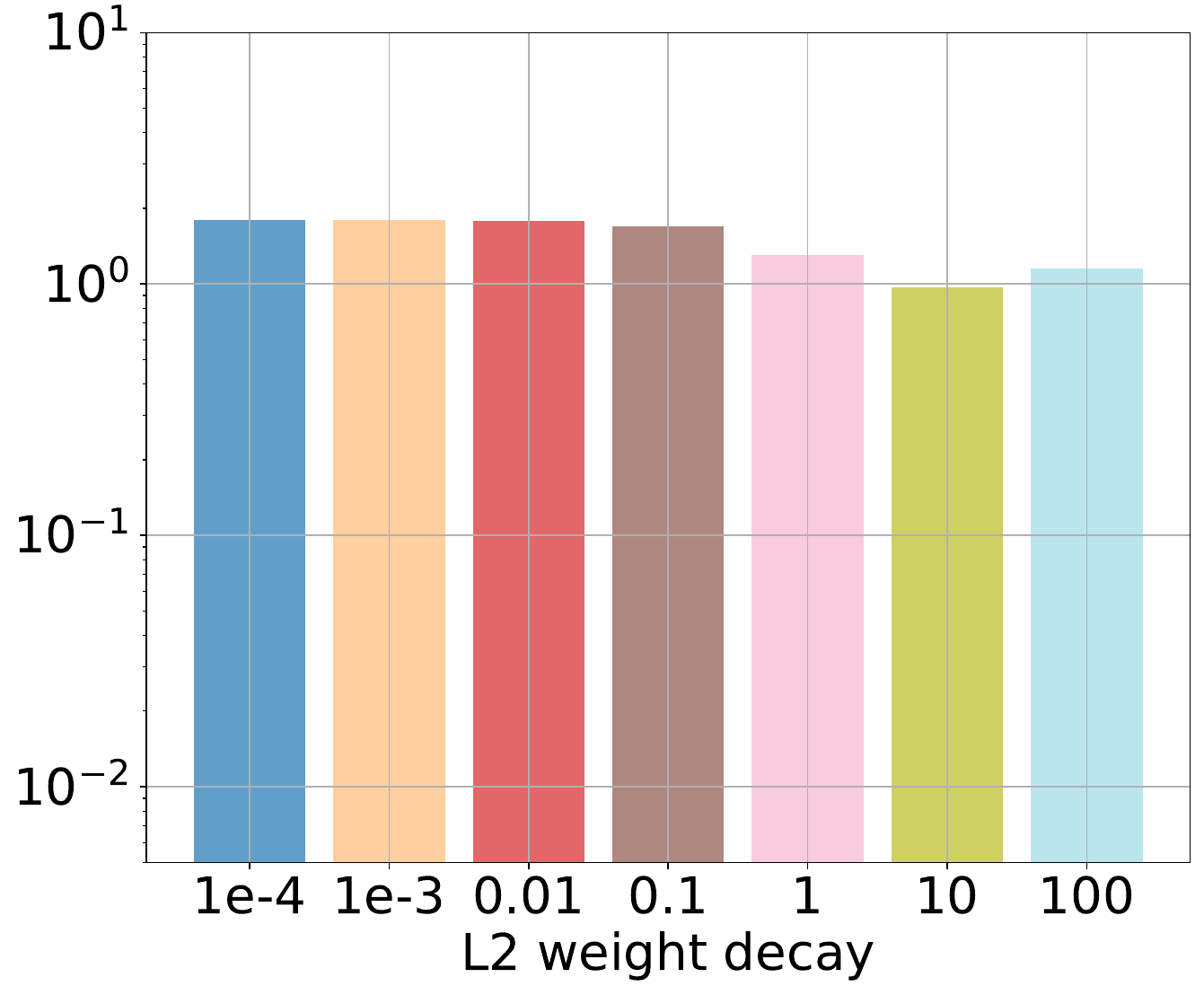}
{\small{{(b)} OOD performance ($T=0$) }}
\end{minipage}
\hfill
\begin{minipage}[b]{0.24\textwidth}
\centering \includegraphics[width = \textwidth, height=3.25cm]{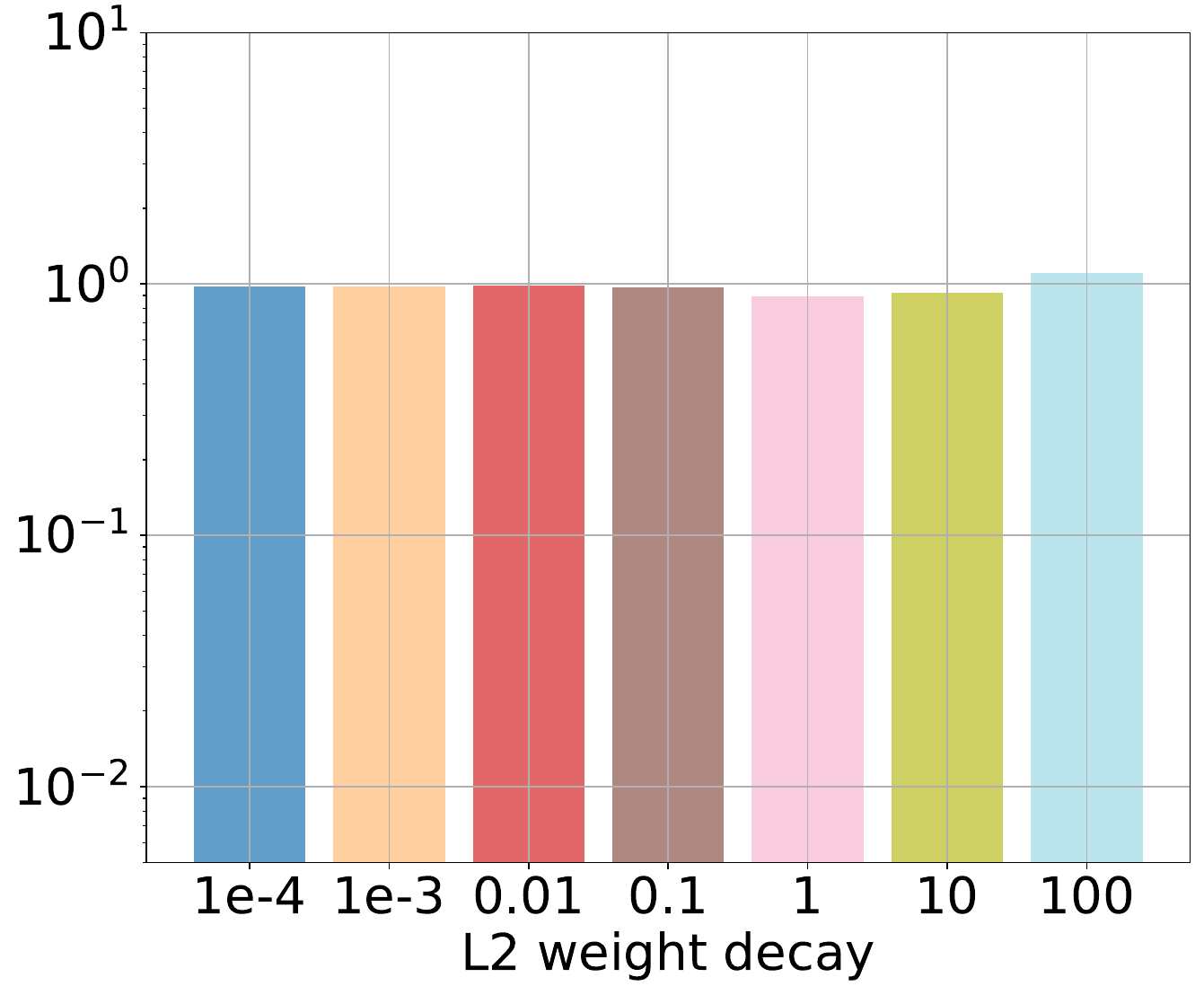}
{\small{{(c)} OOD performance ($T=1$) }}
\end{minipage}
\hfill
\begin{minipage}[b]{0.24\textwidth}
\centering \includegraphics[width = \textwidth, height=3.25cm]{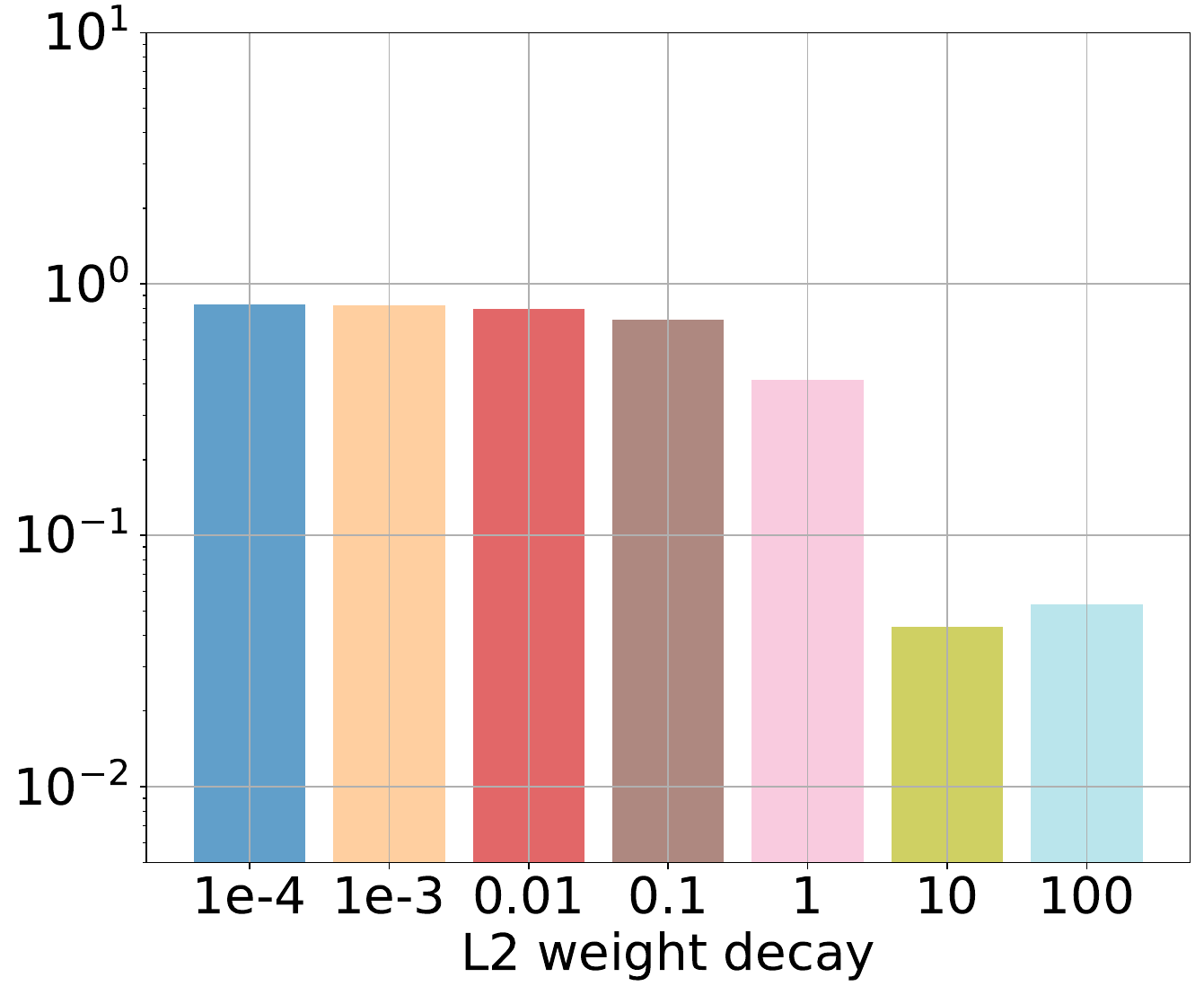}
{\small{{(d)} OOD improvement ($T=0 \to 1$) }}
\end{minipage}
\caption{Performance of ensemble $+$ in a dynamic setting with varying weight decays [eICU data with clustering bias].
We can see the weight decay with values from $1e-4$ to $1e-2$ is the  best for the in-distribution data performance [plot (a)], while weight decay of $10$ is the best for the OOD data performance [plot (b)] at $T=0$. 
This again underlines the trade-off between the in-distribution performance and out-of-distribution performance. In plot (d), we can see that the performance improvement is maximum for the weight decay from $1e-4$ to $1e-2$ while it is the least for the weight decay of $10$, showcasing the trade-off between sharper posteriors and OOD performance. 
}
\label{fig:Ensemble+_weight_decay}
\end{figure}

\begin{figure}[h]
\centering
\begin{minipage}[b]{0.24\textwidth}
\centering
\includegraphics[width = \textwidth, height=3.25cm]{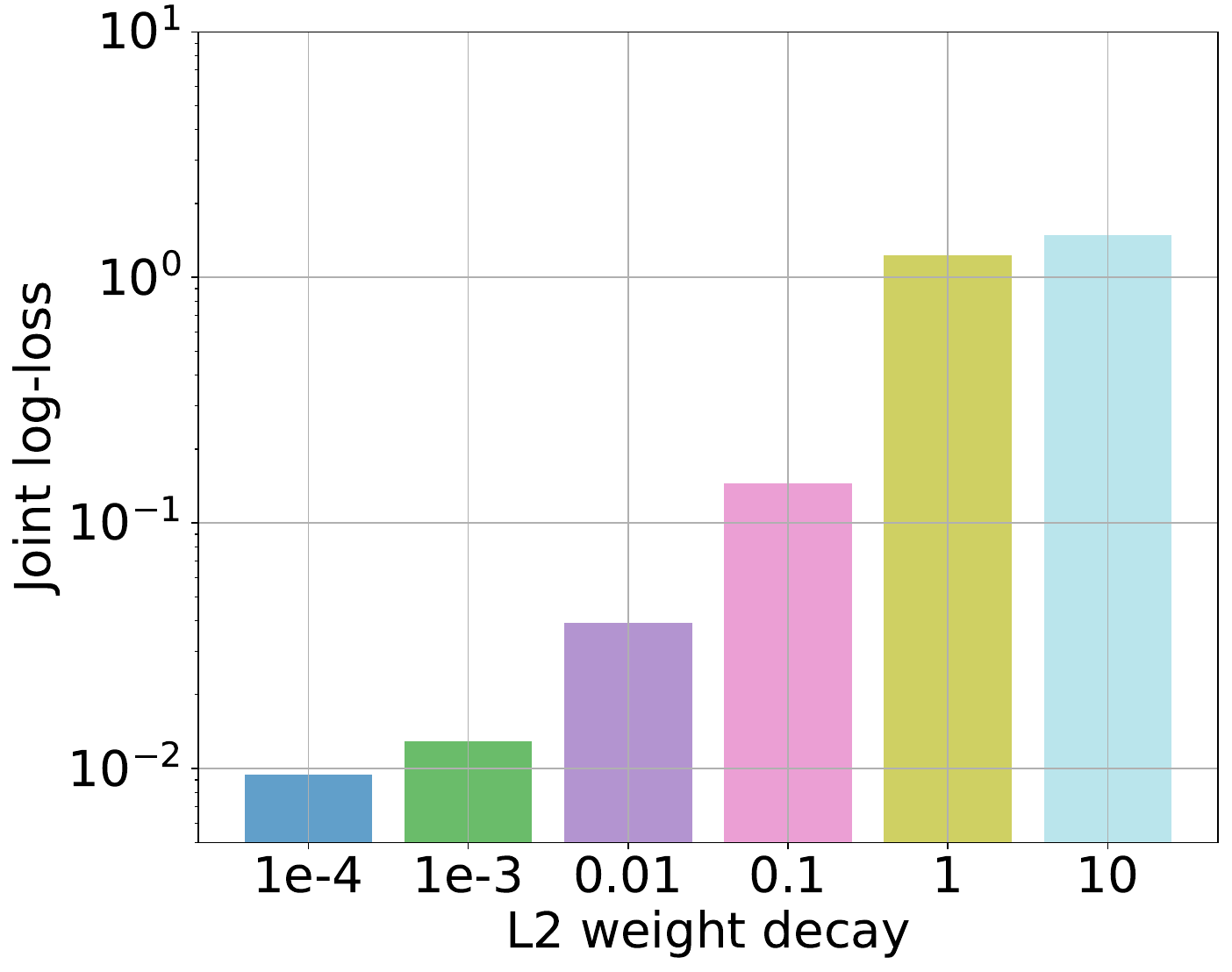}
{\small{{(a)} ID performance ($T=0$) }}
\end{minipage}
\hfill
\begin{minipage}[b]{0.24\textwidth}
\centering \includegraphics[width = \textwidth, height=3.25cm]{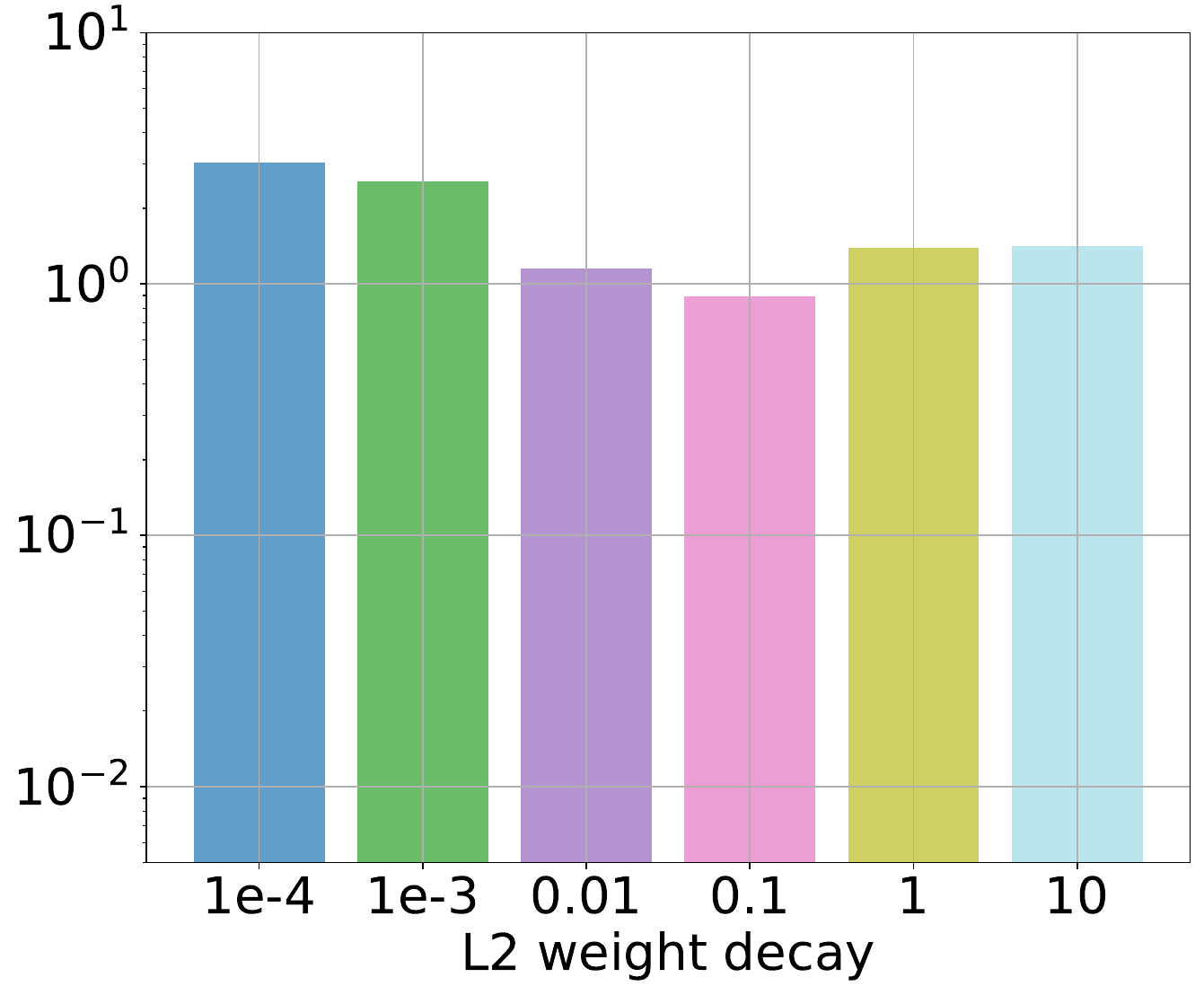}
{\small{{(b)} OOD performance ($T=0$) }}
\end{minipage}
\hfill
\begin{minipage}[b]{0.24\textwidth}
\centering \includegraphics[width = \textwidth, height=3.25cm]{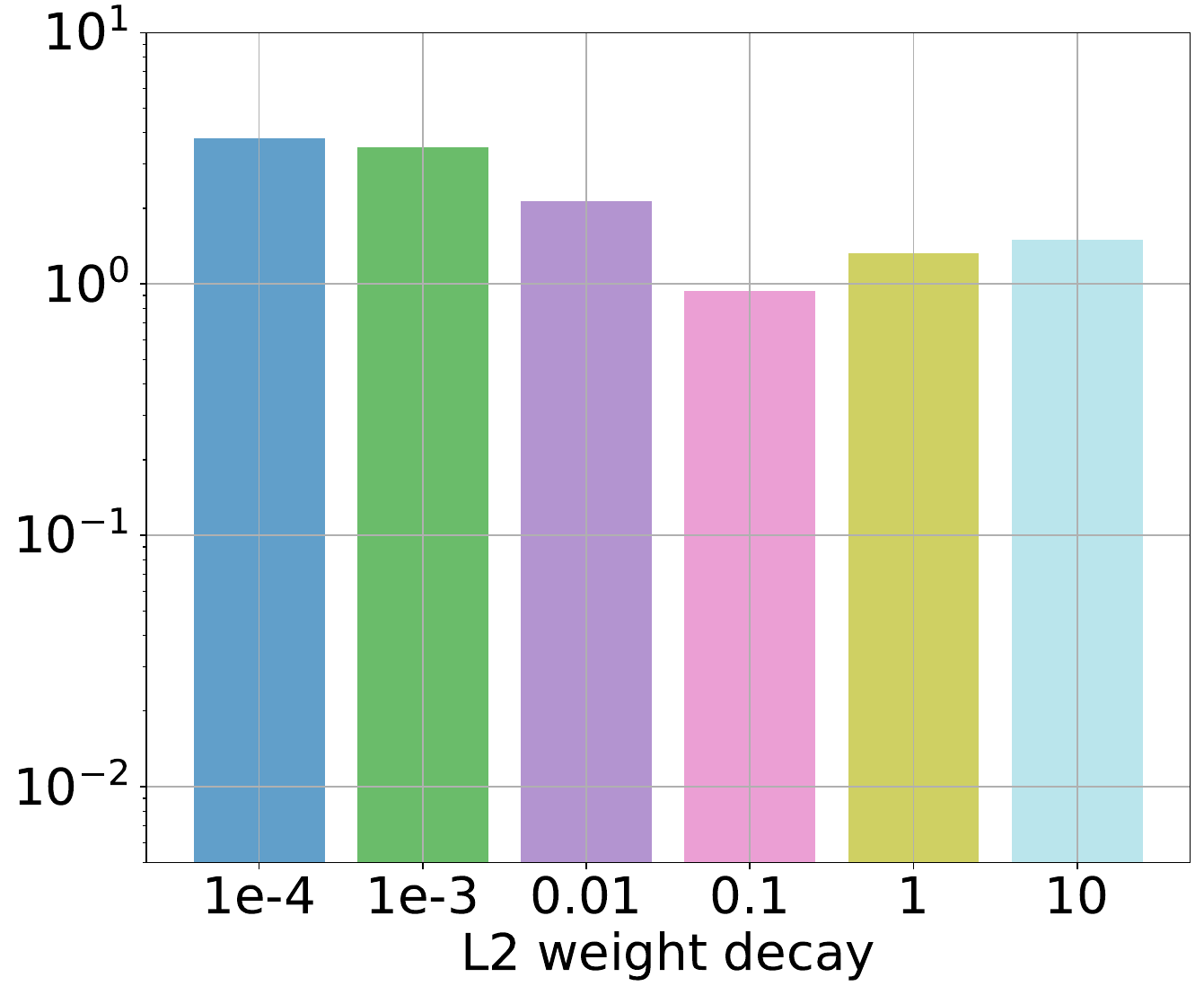}
{\small{{(c)} OOD performance ($T=1$) }}
\end{minipage}
\hfill
\begin{minipage}[b]{0.24\textwidth}
\centering \includegraphics[width = \textwidth, height=3.25cm]{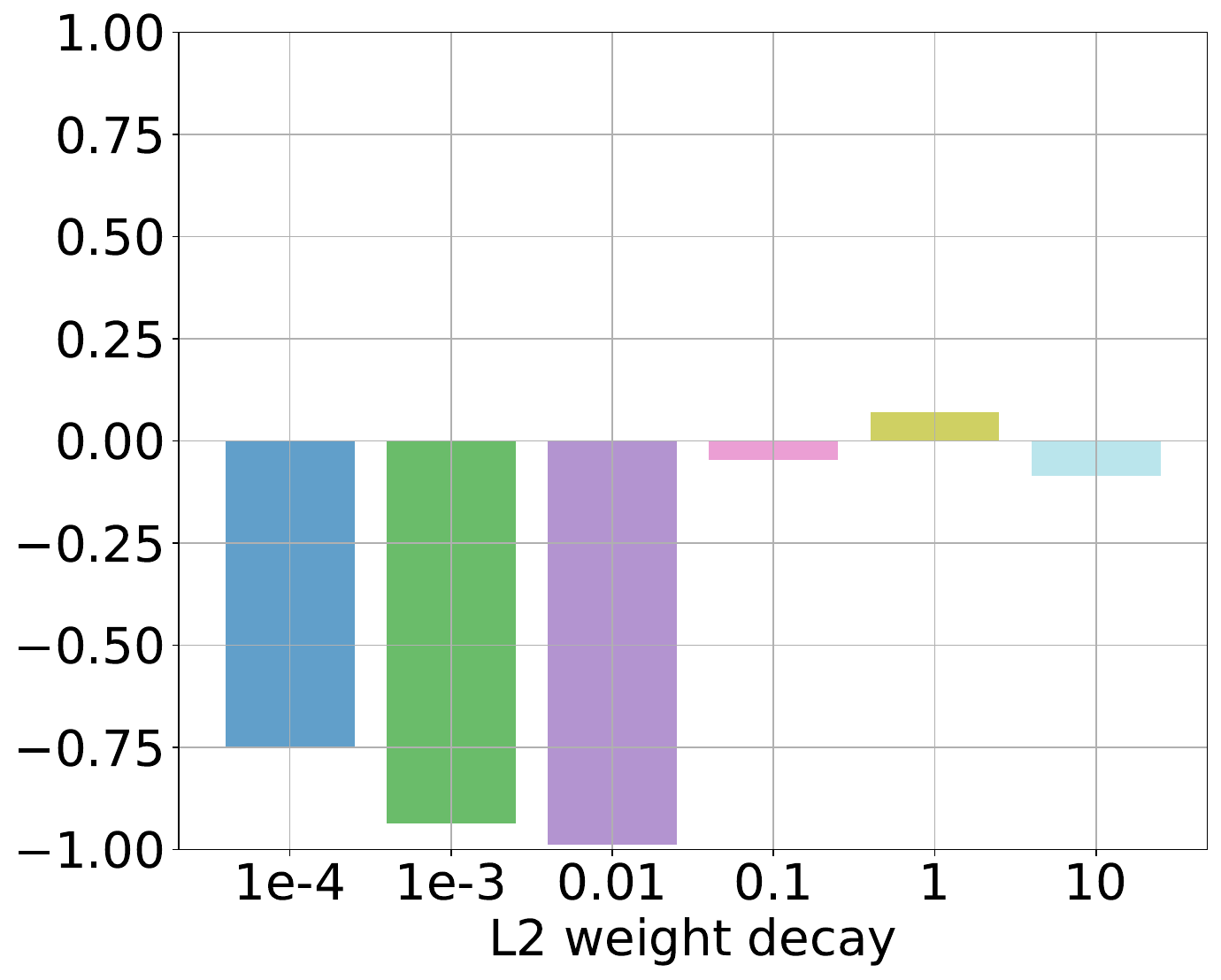}
{\small{{(d)} OOD improvement ($T=0 \to 1$) }}
\end{minipage}
\caption{Performance of epinets in a dynamic setting with varying weight decays [eICU data with clustering bias].
We can see the weight decay of $1e-4$ is the best for the in-distribution data performance [plot (a)], while  weight decay of $0.1$ is the best for the OOD data performance [plot (b)] at $T=0$.  
In plot (d), we can see that the performance improves only for the weight decay of $1.0$, while for other values it deteriorates even after acquiring the new data. 
}
\label{fig:Epinet_weight_decay}
\end{figure}

\begin{figure}[h]
\centering
\begin{minipage}[b]{0.24\textwidth}
\centering
\includegraphics[width = \textwidth, height=3.25cm]{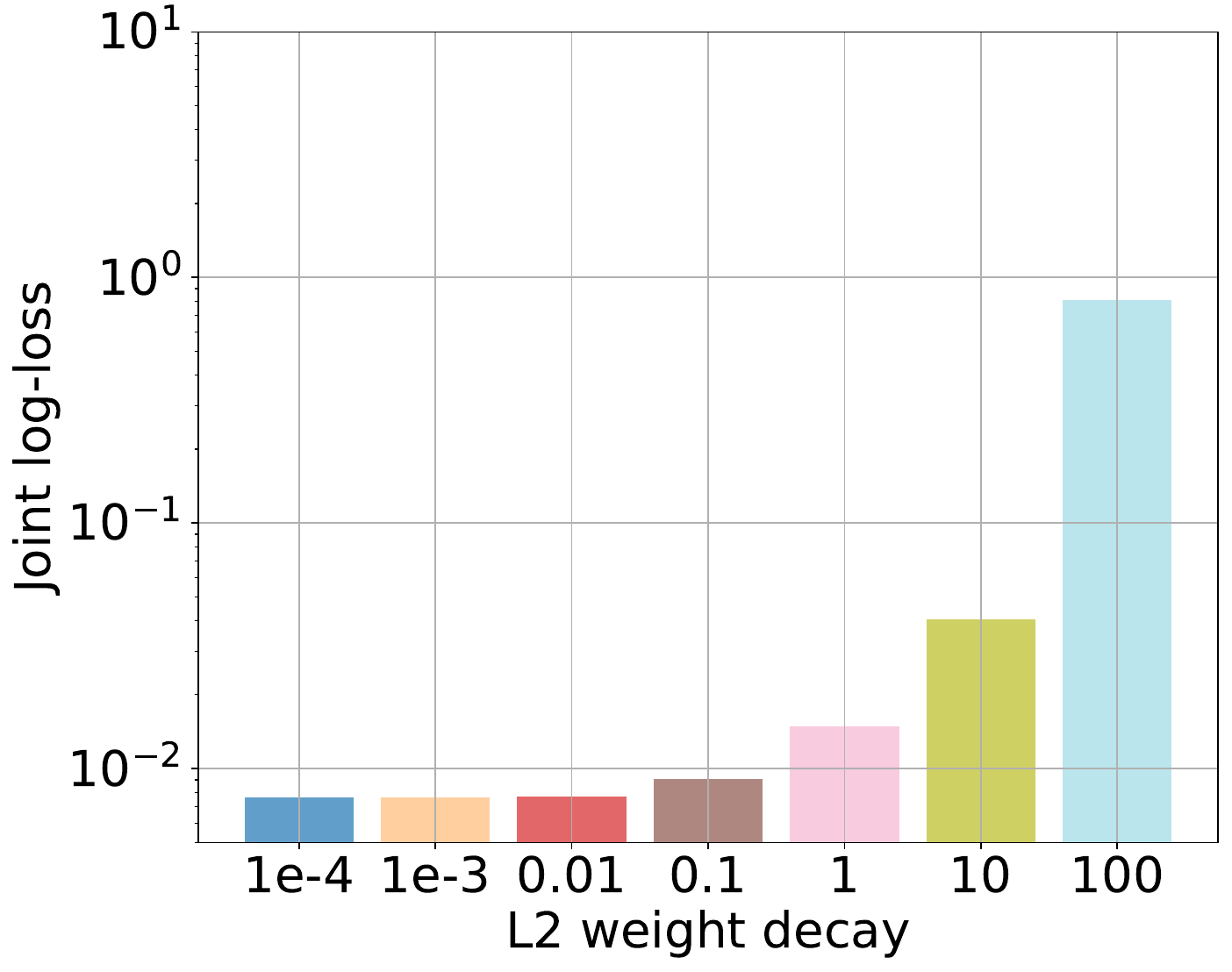}
{\small{{(a)} ID performance ($T=0$) }}
\end{minipage}
\hfill
\begin{minipage}[b]{0.24\textwidth}
\centering \includegraphics[width = \textwidth, height=3.25cm]{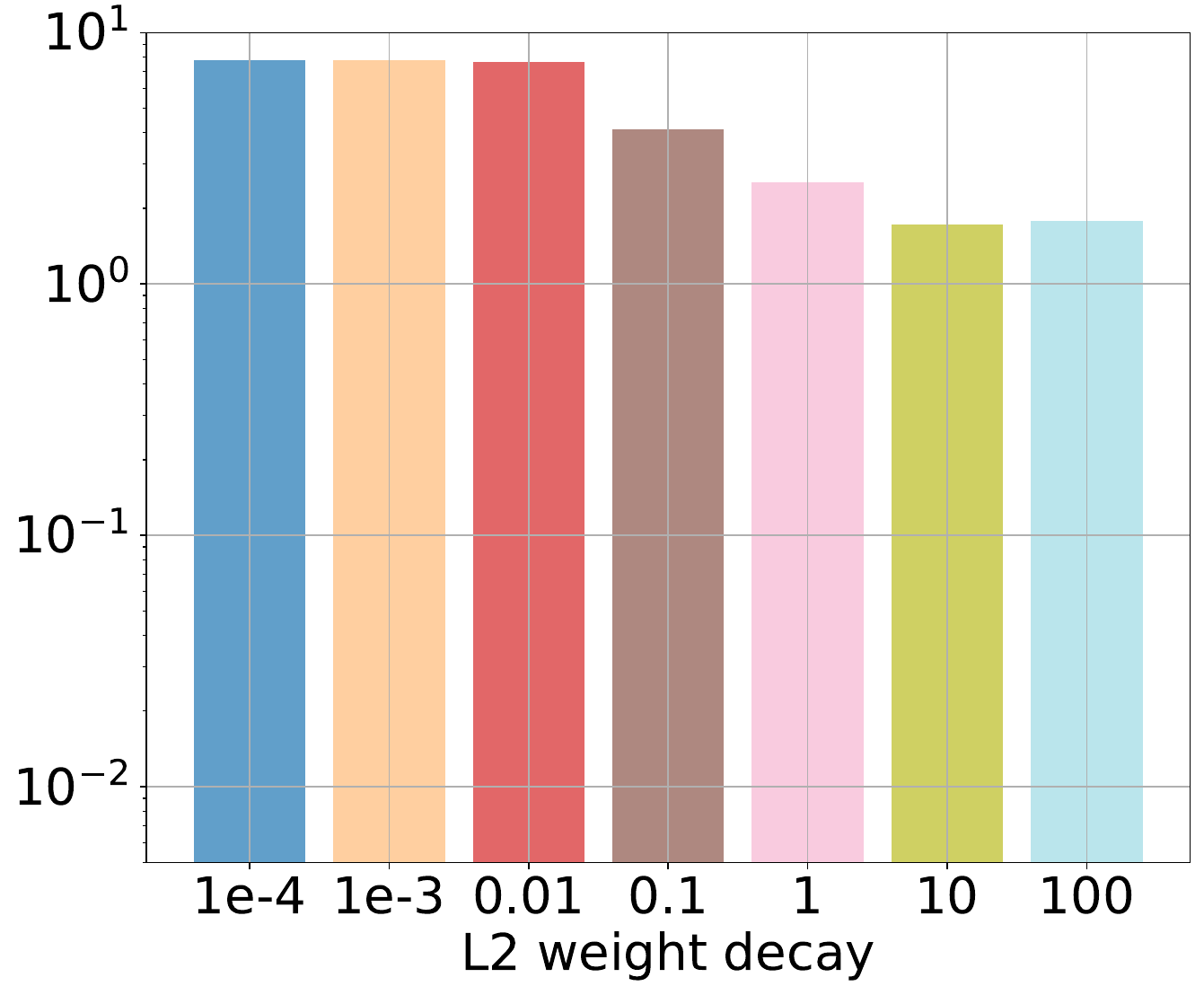}
{\small{{(b)} OOD performance ($T=0$) }}
\end{minipage}
\hfill
\begin{minipage}[b]{0.24\textwidth}
\centering \includegraphics[width = \textwidth, height=3.25cm]{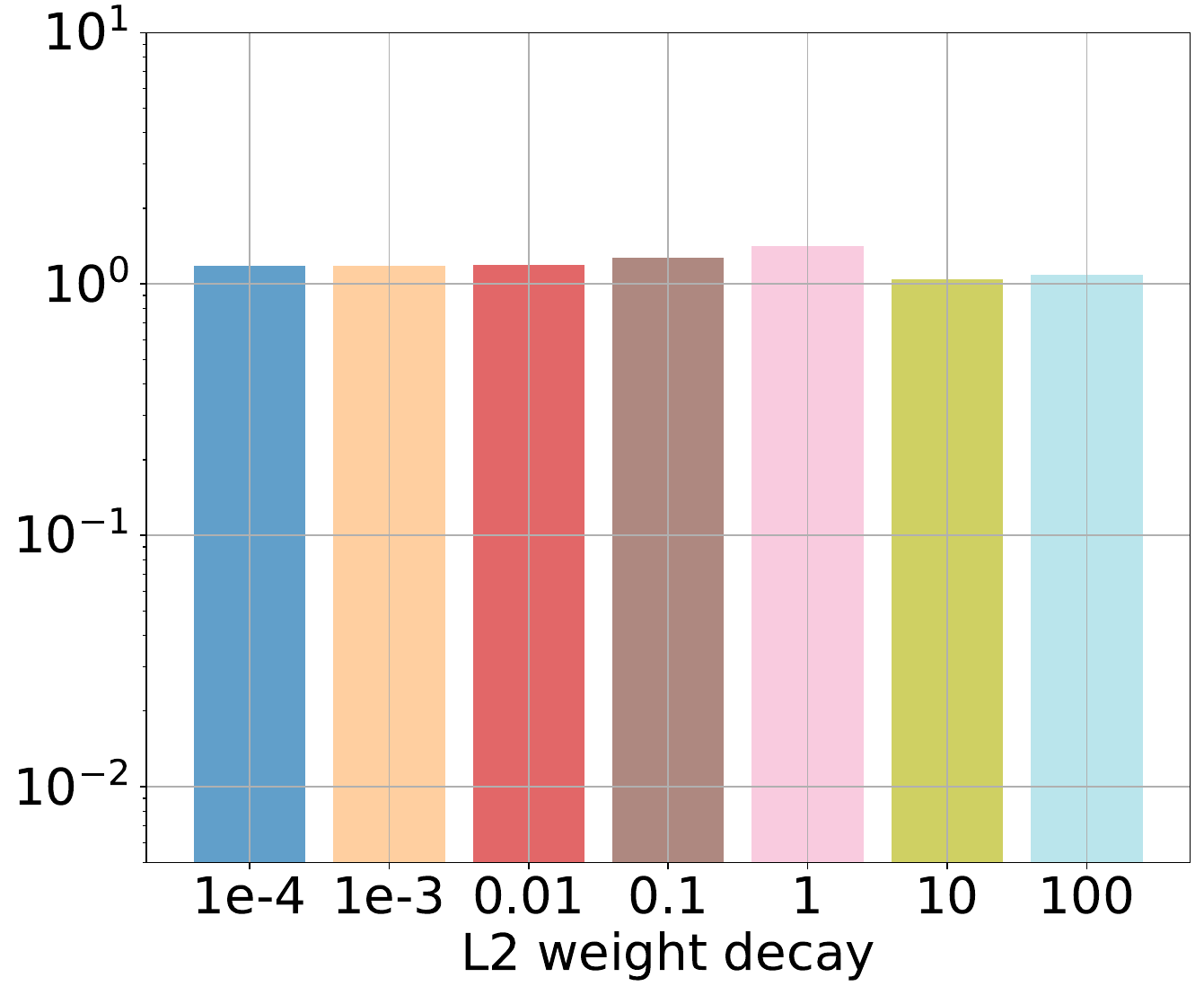}
{\small{{(c)} OOD performance ($T= 1$) }}
\end{minipage}
\hfill
\begin{minipage}[b]{0.24\textwidth}
\centering \includegraphics[width = \textwidth, height=3.25cm]{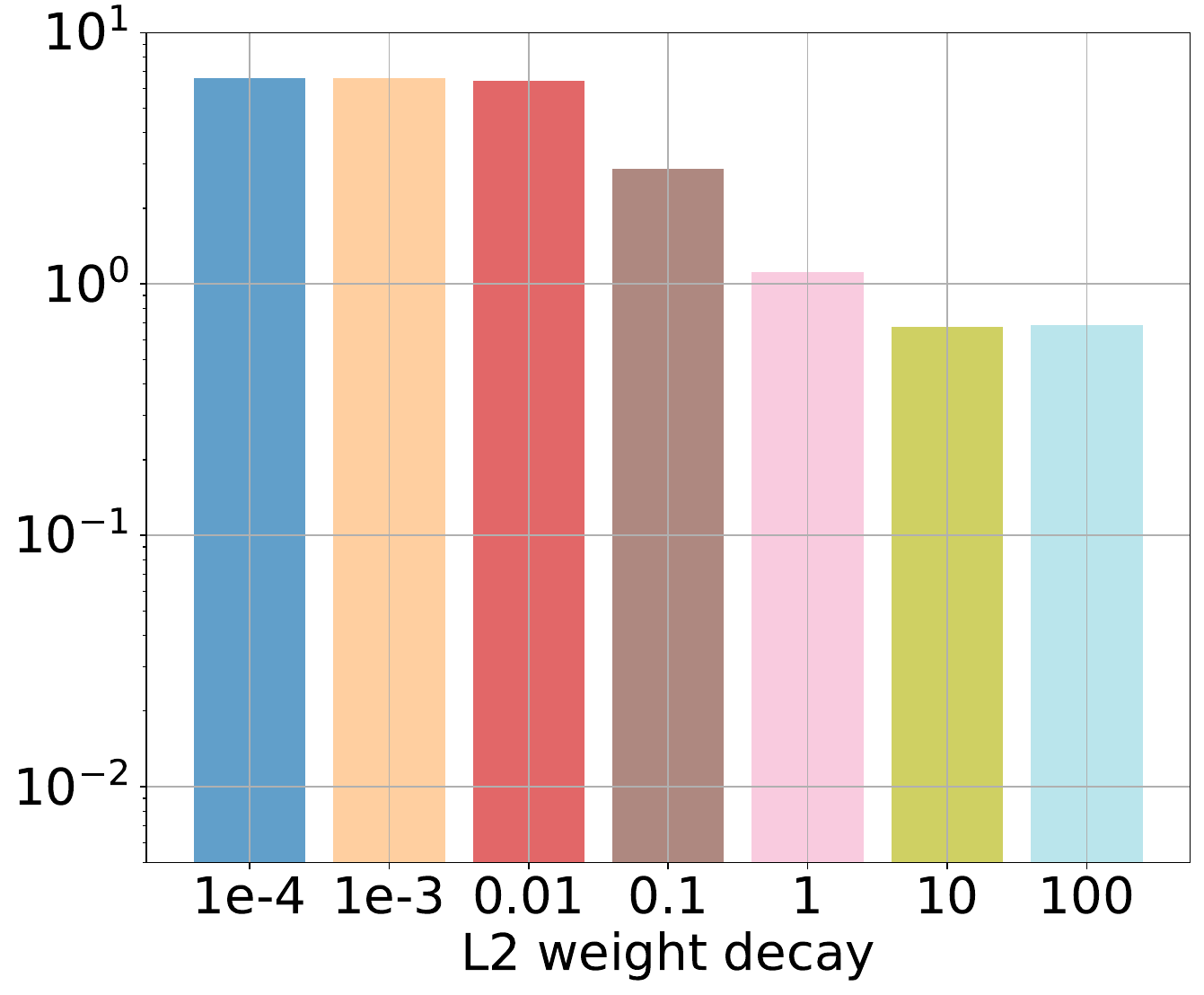}
{\small{{(d)} OOD improvement ($T=0 \to 1$) }}
\end{minipage}
\caption{Performance of hypermodels in a dynamic setting with varying weight decay [eICU data with clustering bias].
We can see the weight decay with values from $1e-4$ to $1e-2$ is the best for the in-distribution data performance [plot (a)], while weight decay of $10$ to $100$ is best for the OOD data performance [plot (b)] at $T=0$, underlying the trade-off between the in-distribution performance and out-of-distribution performance. In plot (d), we can see that the performance improvement is maximum for the weight decays taking values from $1e-4$ to $1e-2$, while it is the least for the weight decay with values from $10$ to $100$,  showcasing the trade-off between sharper posteriors and OOD performance. 
}
\label{fig:Hypermodel_weight_decay}
\end{figure}

\begin{figure}[h]
\centering
\begin{minipage}[b]{0.24\textwidth}
\centering
\includegraphics[width = \textwidth, height=4.0cm]{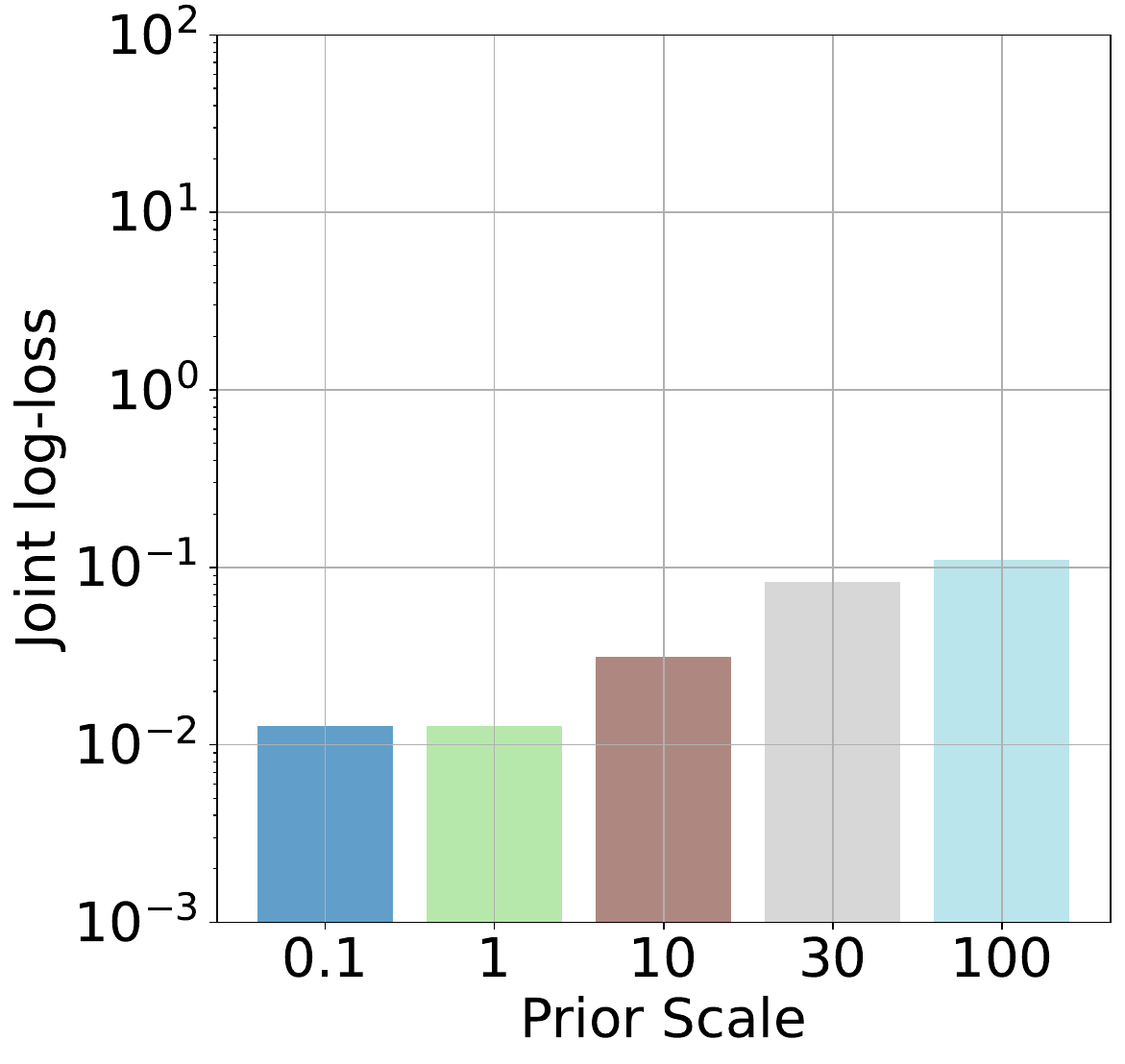}
{\small{{(a)} ID performance ($T=0$) }}
\end{minipage}
\hfill
\begin{minipage}[b]{0.24\textwidth}
\centering \includegraphics[width = \textwidth, height=4.0cm]{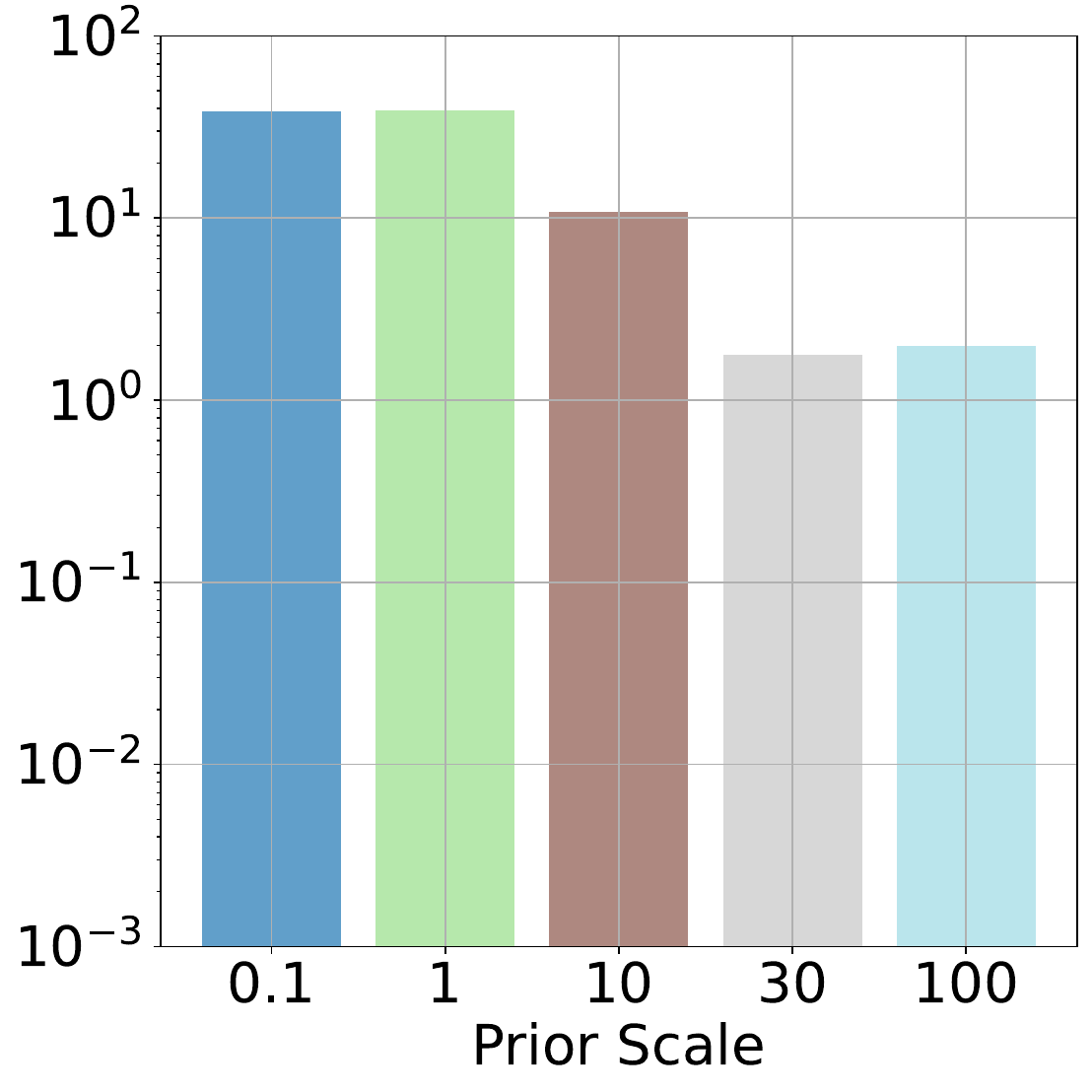}
{\small{{(b)} OOD performance ($T=0$) }}
\end{minipage}
\hfill
\begin{minipage}[b]{0.24\textwidth}
\centering \includegraphics[width = \textwidth, height=4.0cm]{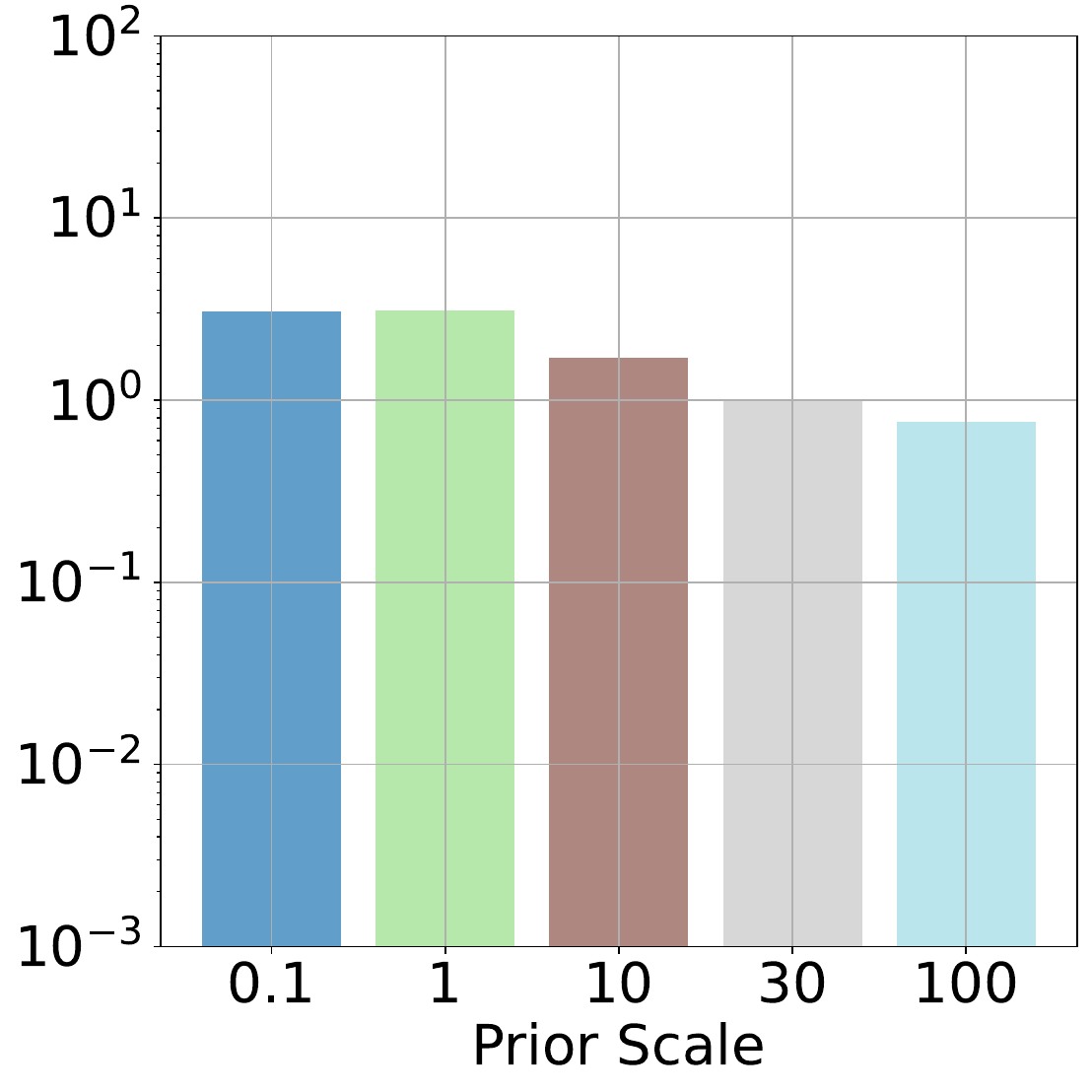}
{\small{{(c)} OOD performance ($T= 1$) }}\end{minipage}
\hfill
\begin{minipage}[b]{0.24\textwidth}
\centering \includegraphics[width = \textwidth, height=4.0cm]{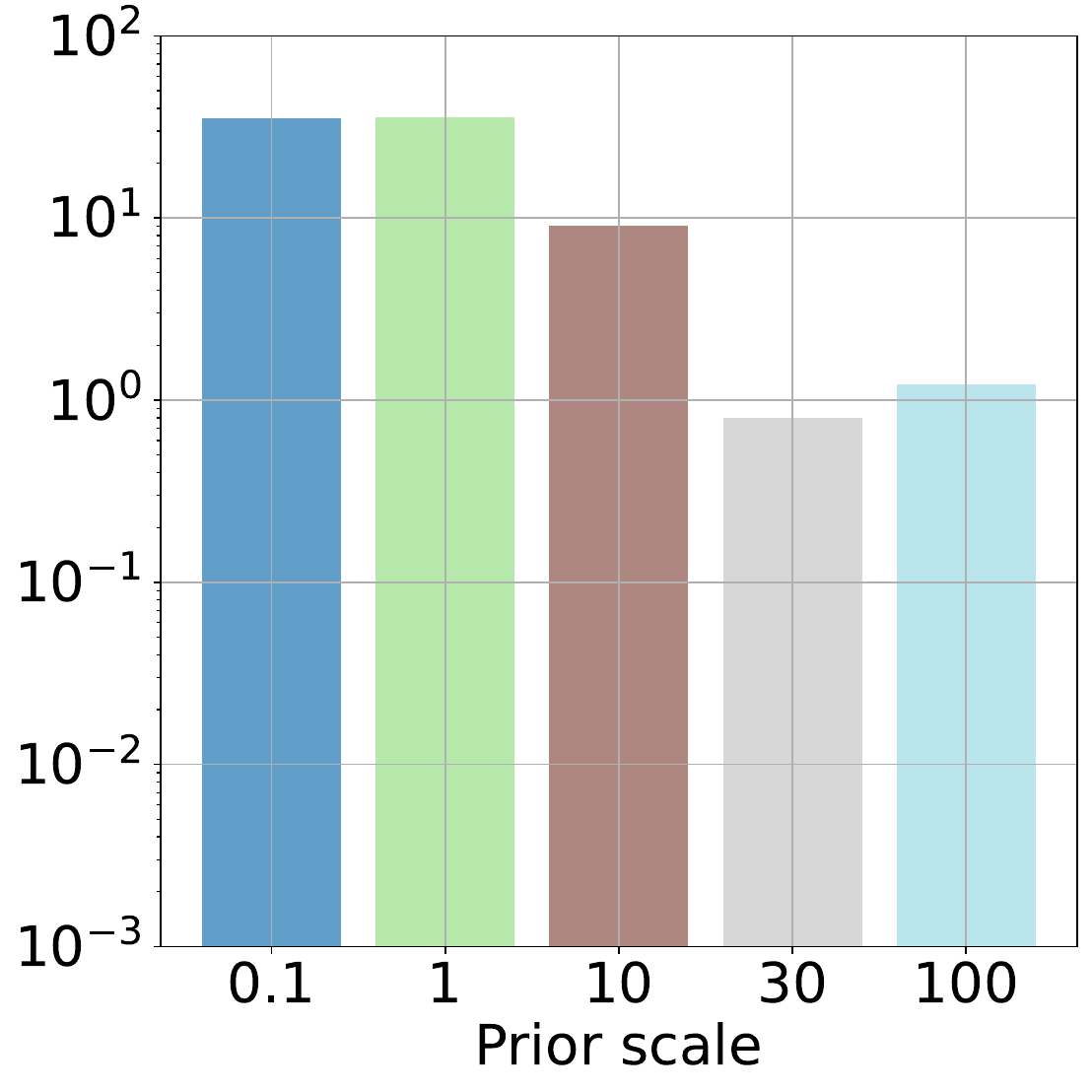}
{\small{{(d)} OOD improvement ($T=0 \to 1$) }}
\end{minipage}
\caption{Performance of ensemble $+$ in a dynamic setting with varying prior scale [eICU data with clustering bias]. We can see that the prior scale $0.1$ to $1$ is  the best for the in-distribution data performance [plot (a)]. 
However, prior scale of $10$ is the best for the OOD data performance [plot (b)] at $T=0$, underlying the trade-off between the in-distribution performance and out-of-distribution performance. In plot (d), we can see that the performance improvement is maximum for the prior scales  of values from $0.1$ to $1$, while it the least for the prior scale of $10$, showcasing the trade-off between sharper posteriors and OOD performance. 
}
\label{fig:Ensemble+_prior_scale}
\end{figure}

\begin{figure}[h]
\centering
\begin{minipage}[b]{0.24\textwidth}
\centering
\includegraphics[width = \textwidth, height = 4.0cm]{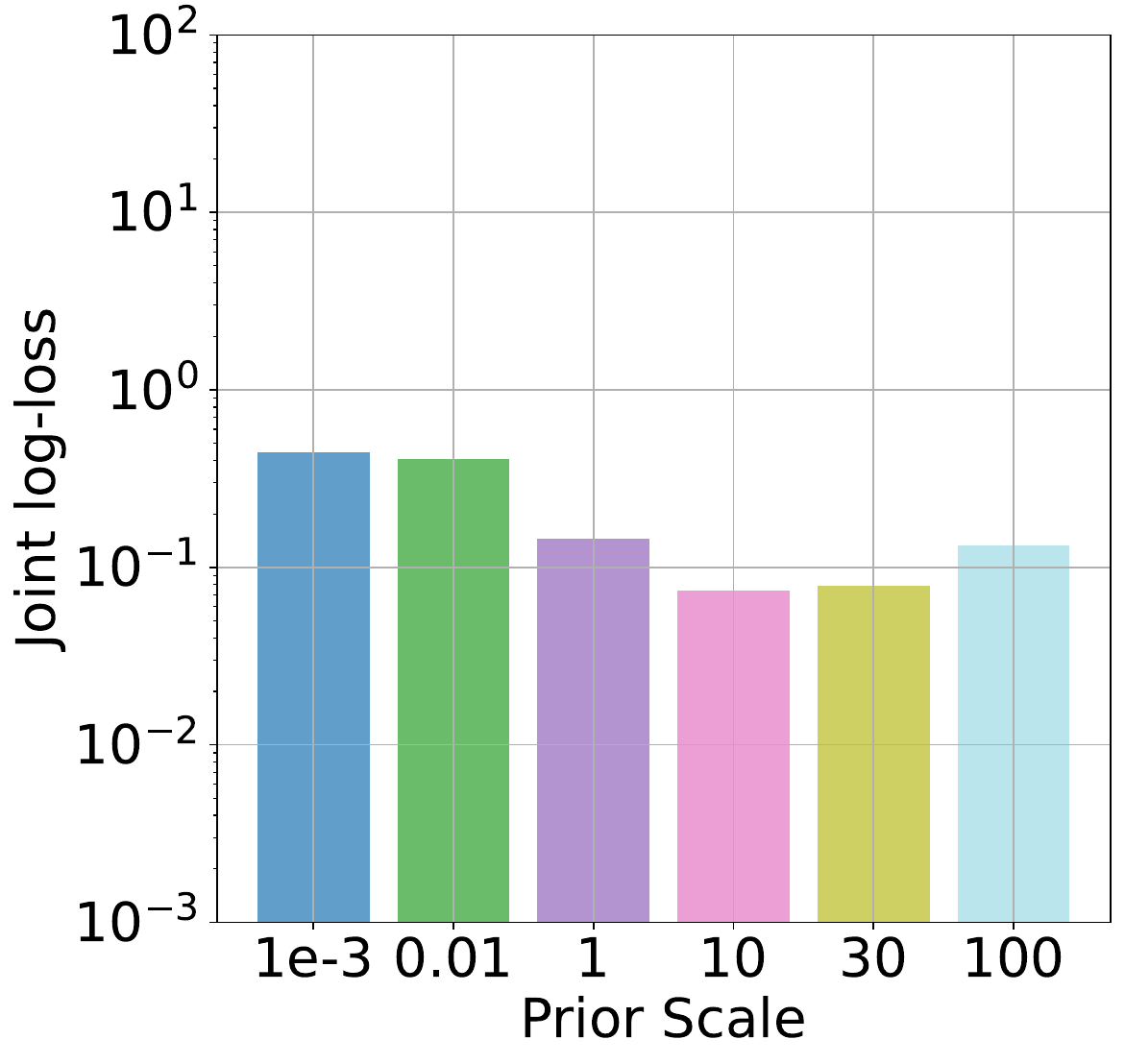}
{\small{{(a)} ID performance ($T=0$) }}
\end{minipage}
\hfill
\begin{minipage}[b]{0.24\textwidth}
\centering \includegraphics[width = \textwidth, height = 4.0cm]{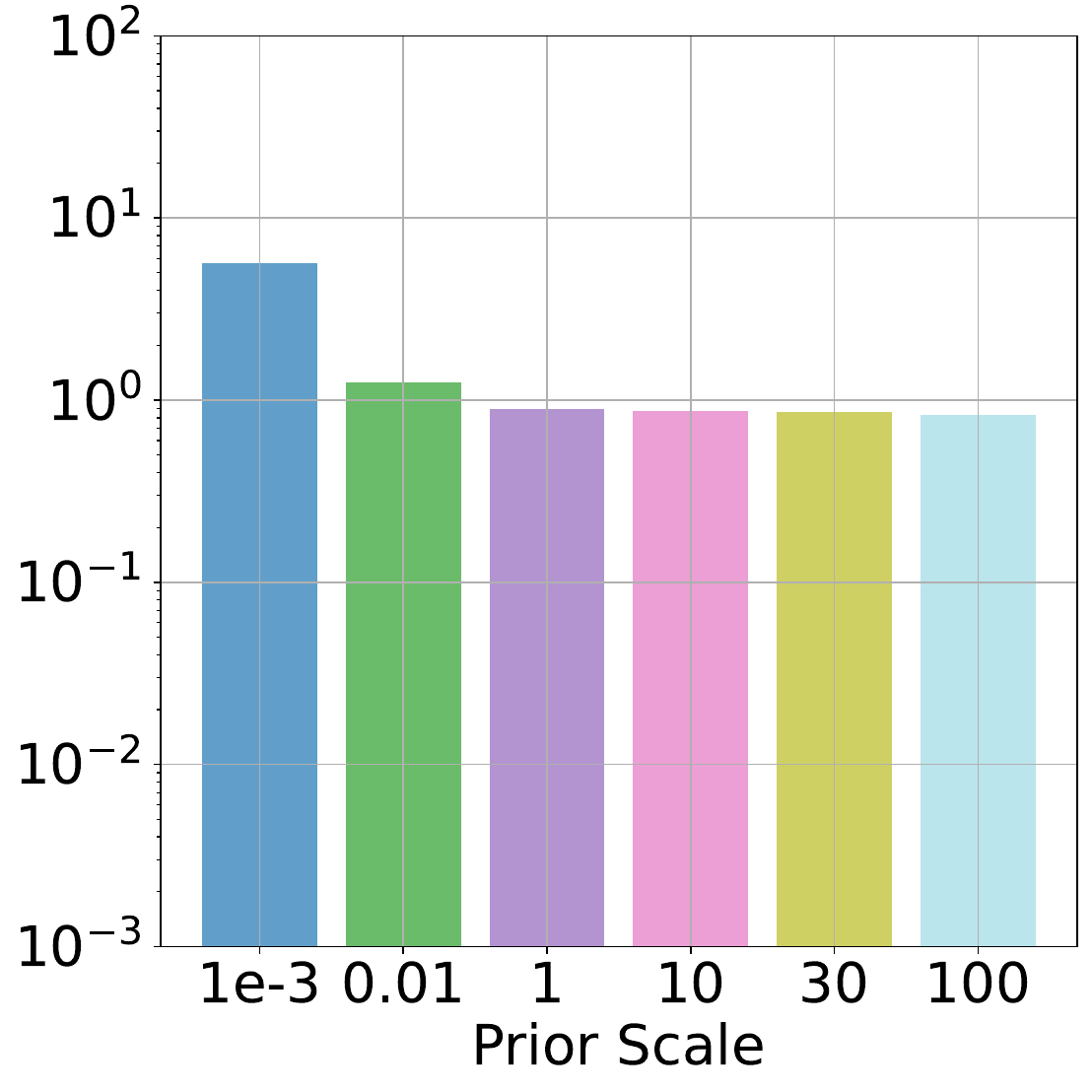}
{\small{{(b)} OOD performance ($T=0$) }}
\end{minipage}
\hfill
\begin{minipage}[b]{0.24\textwidth}
\centering \includegraphics[width = \textwidth, height = 4.0cm]{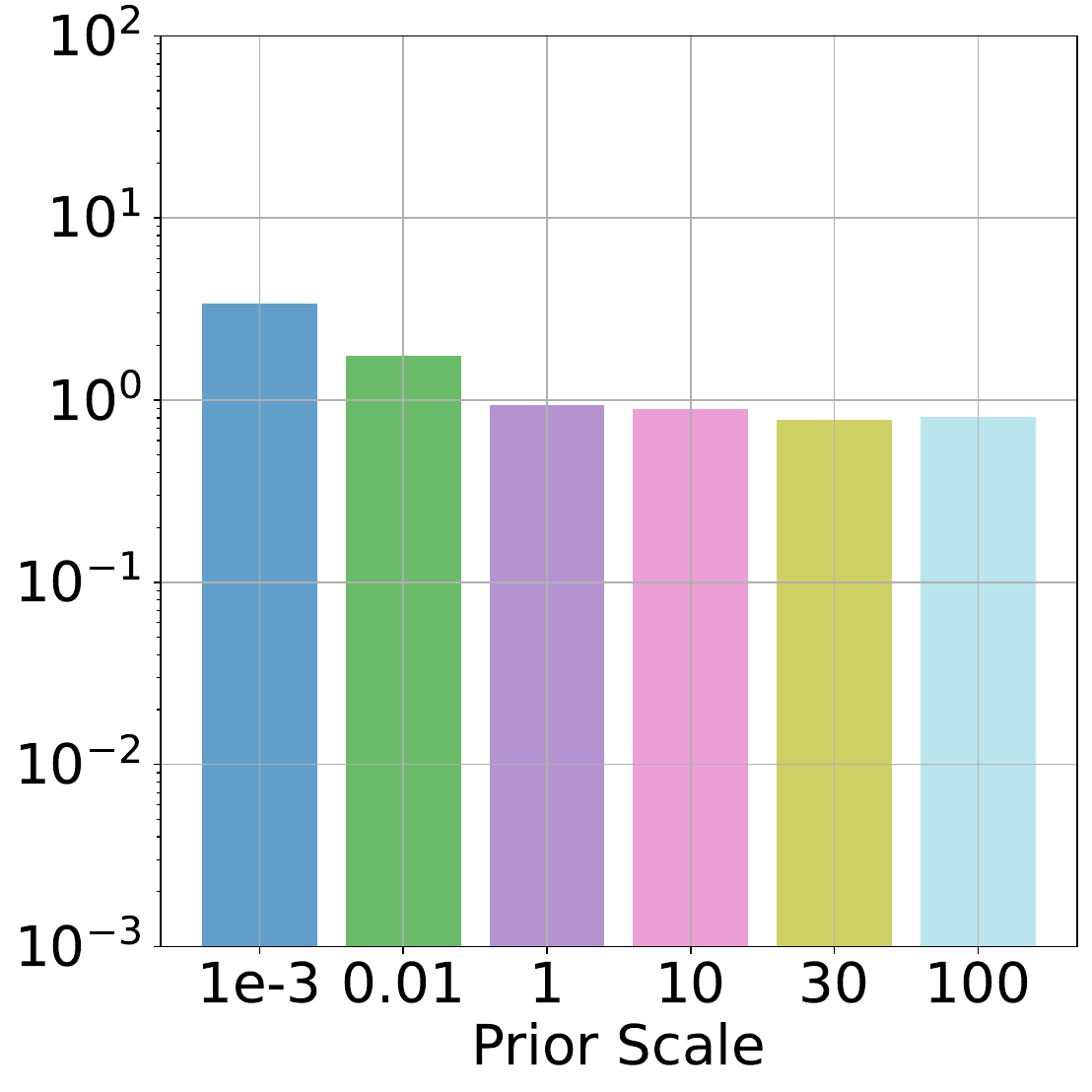}
{\small{{(c)} OOD performance ($T=1$) }}
\end{minipage}
\hfill
\begin{minipage}[b]{0.24\textwidth}
\centering \includegraphics[width = \textwidth, height = 4.0cm]{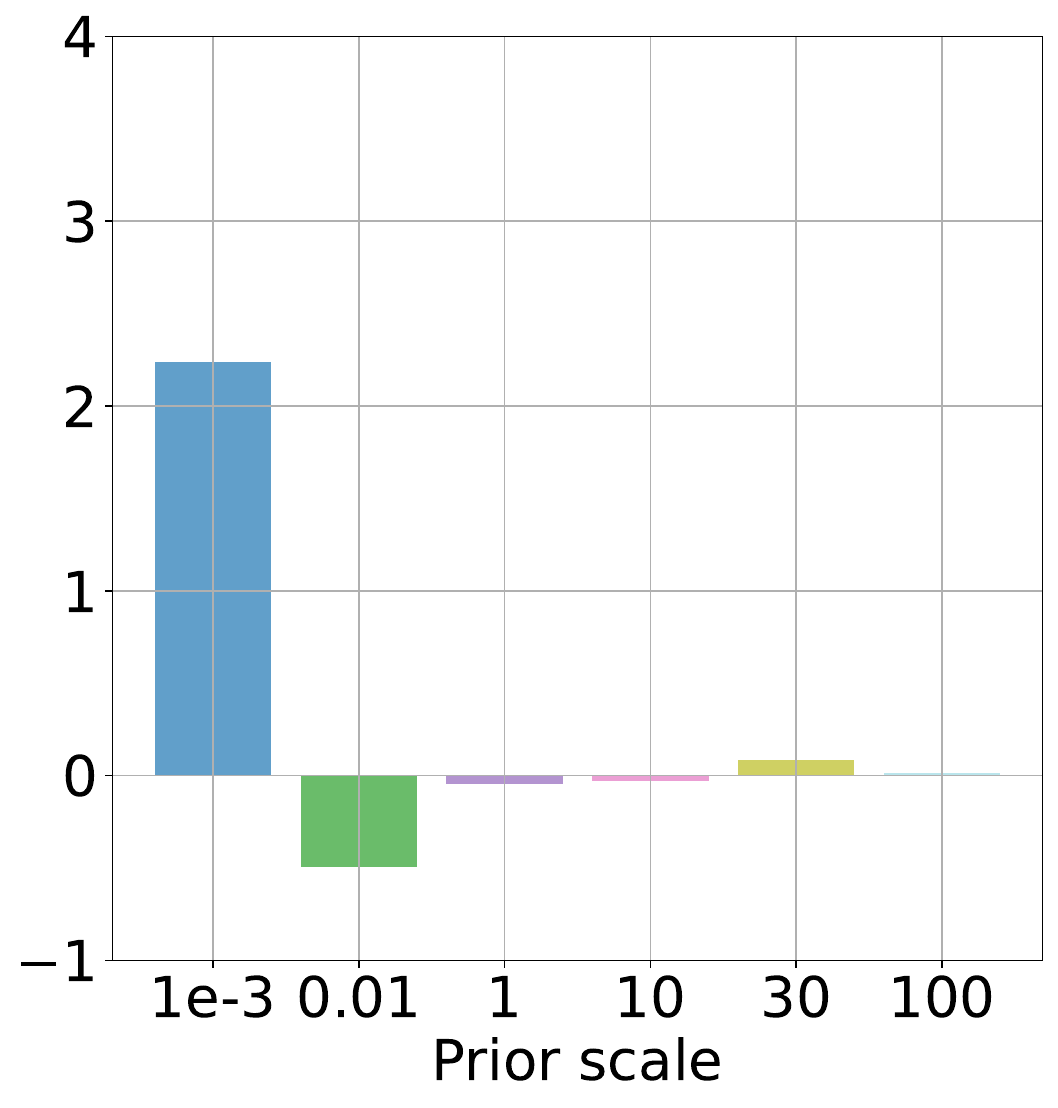}
{\small{{(d)} OOD improvement ($T=0 \to 1$) }}
\end{minipage}
\caption{Performance of epinets in a dynamic setting with varying prior scale [eICU data with clustering bias]. We can see the prior scale of $10$ to $30$ works best for both the in-distribution data performance [plot (a)]  the OOD data performance [plot (b)] at $T=0$. However, in plot (d), we can see that the OOD performance improvement is maximum for the prior scale of $1e-3$. 
On the other hand, for other prior scales improvement is marginal or sometimes even deteriorates.
}
\label{fig:Epinet_prior_scale}
\end{figure}

\begin{figure}[h]
\centering
\begin{minipage}[b]{0.24\textwidth}
\centering
\includegraphics[width = \textwidth, height = 4.0cm]{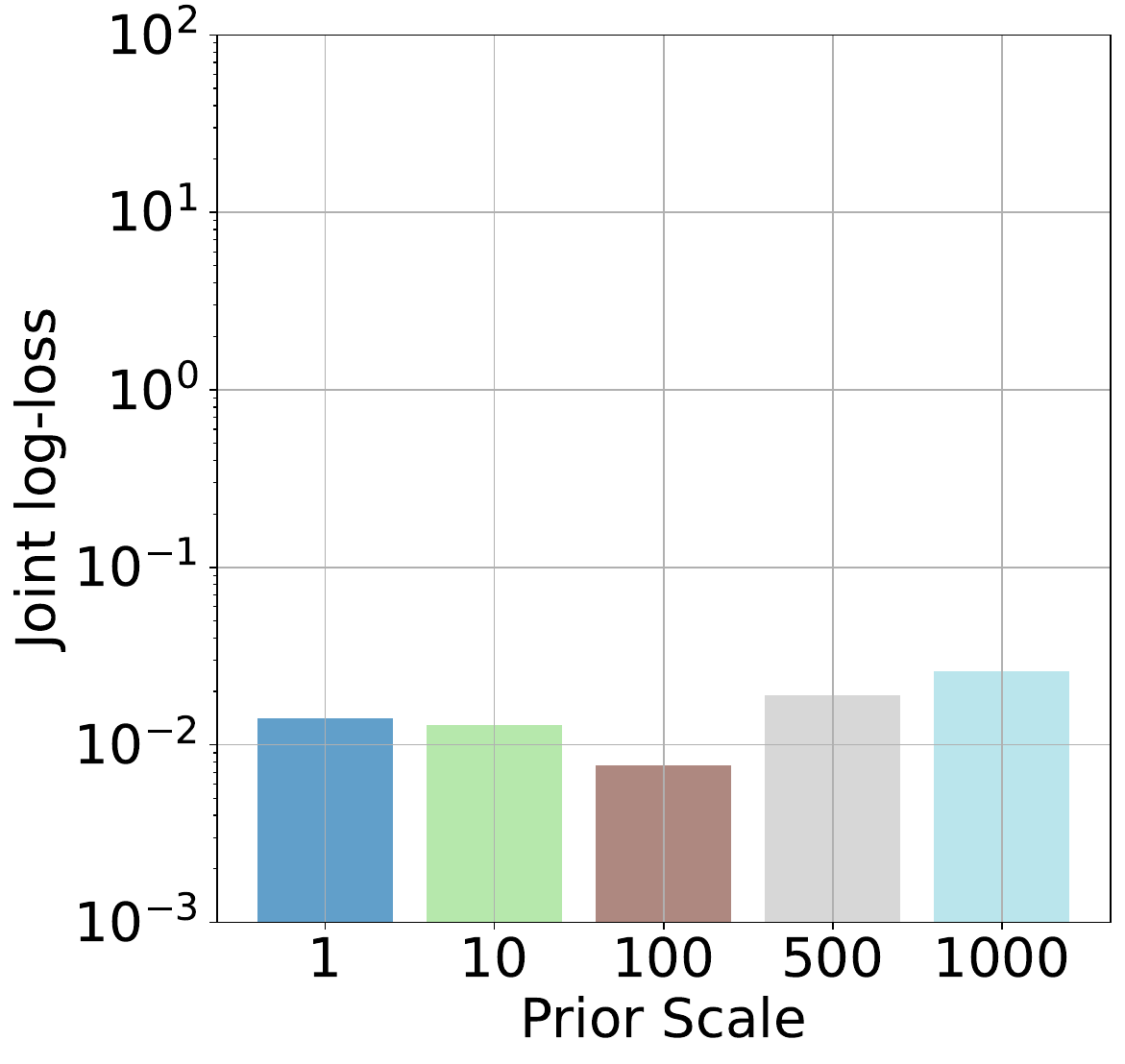}
{\small{{(a)} ID performance ($T=0$) }}
\end{minipage}
\hfill
\begin{minipage}[b]{0.24\textwidth}
\centering \includegraphics[width = \textwidth, height = 4.0cm]{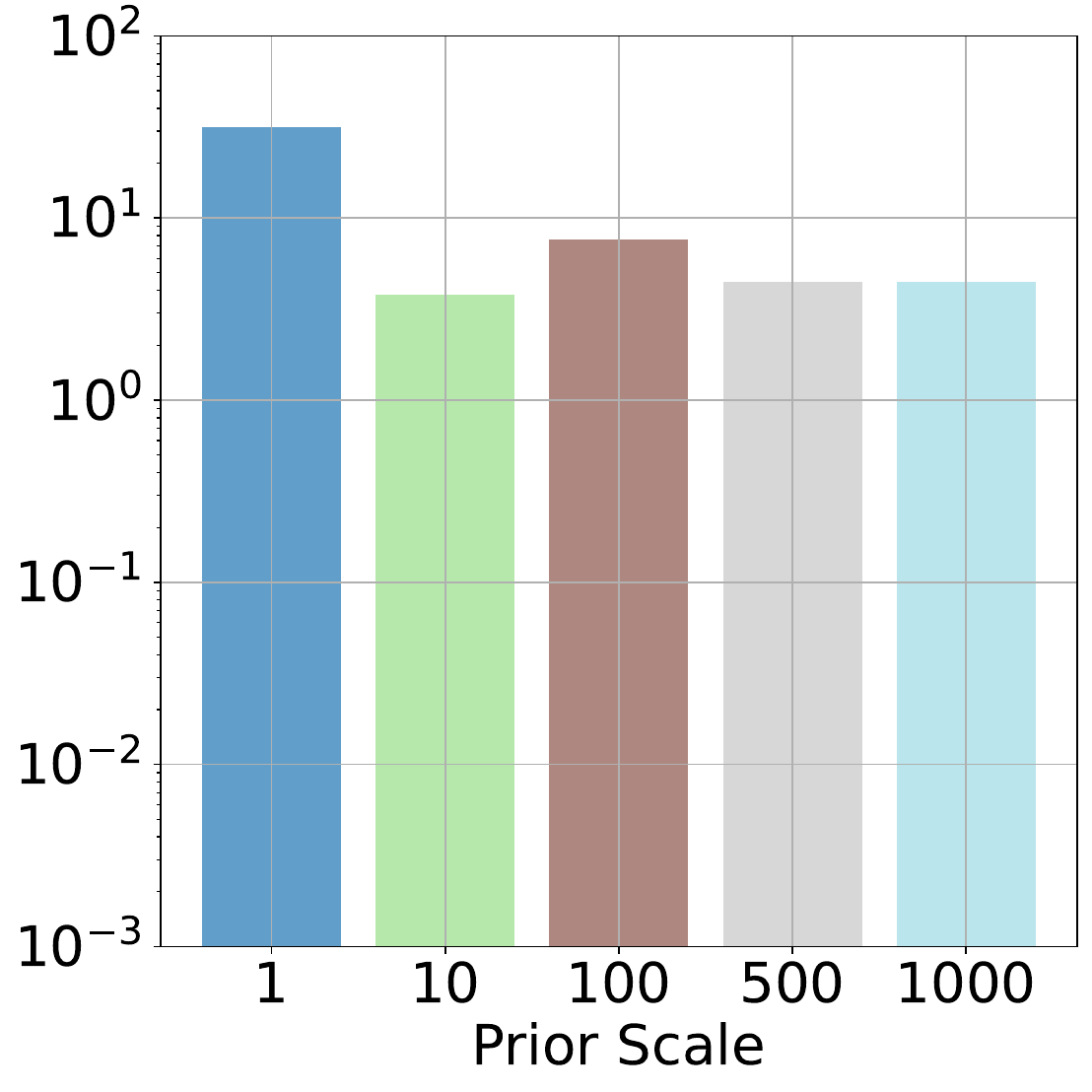}
{\small{{(b)} OOD performance ($T=0$) } }
\end{minipage}
\hfill
\begin{minipage}[b]{0.24\textwidth}
\centering \includegraphics[width = \textwidth, height = 4.0cm]{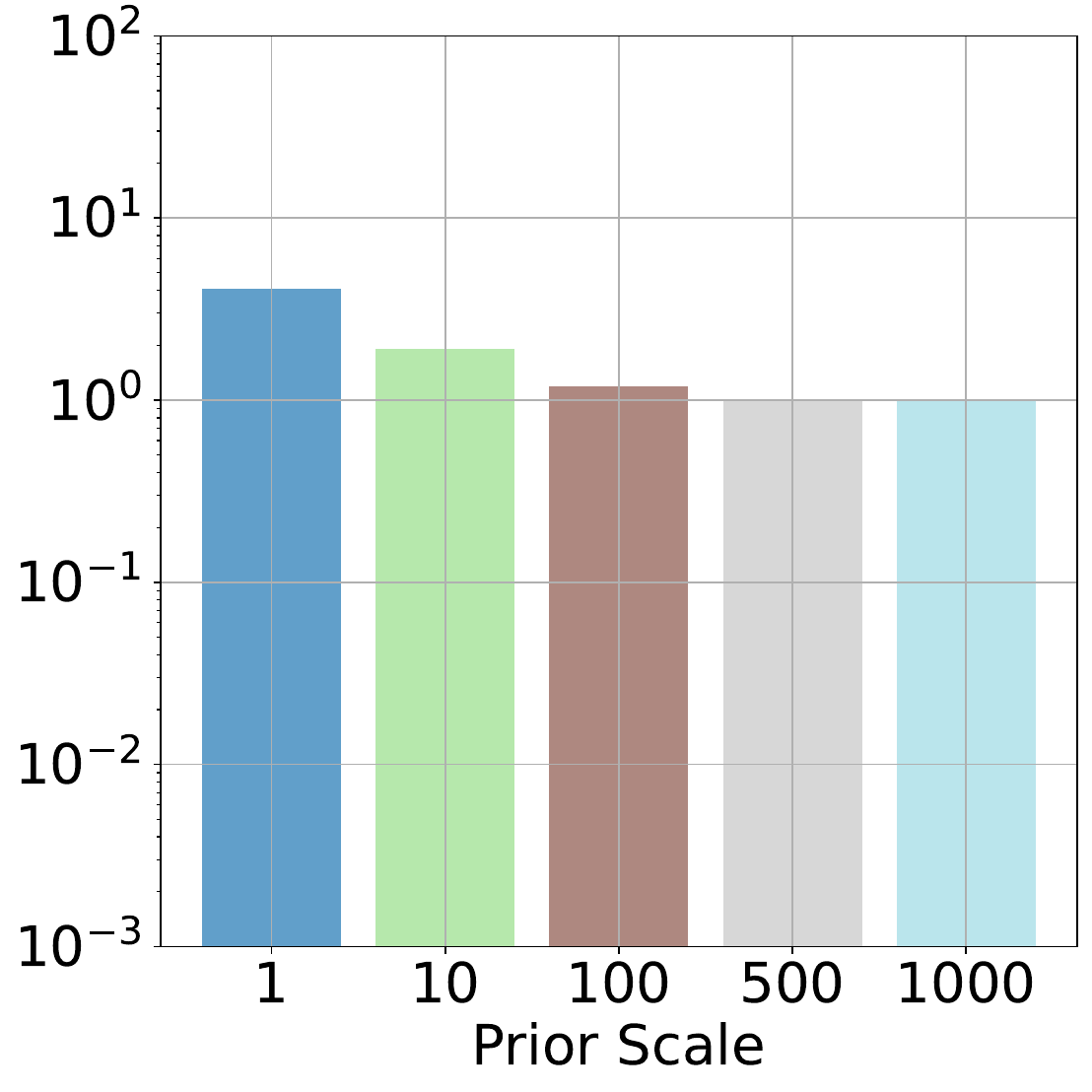}
{\small{{(c)} OOD performance ($T=1$)}}
\end{minipage}
\hfill
\begin{minipage}[b]{0.24\textwidth}
\centering \includegraphics[width = \textwidth, height = 4.0cm]{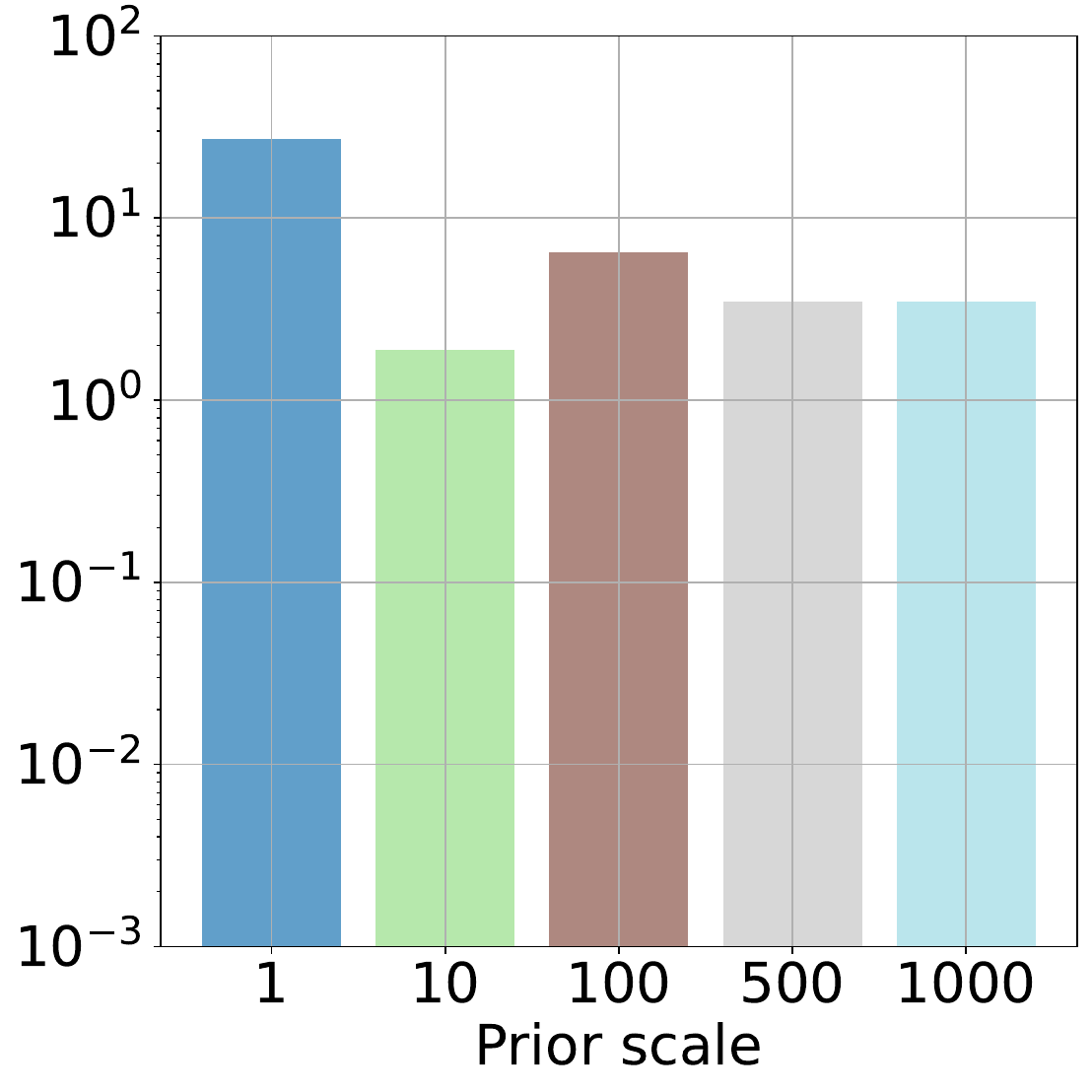}
{\small{{(d)} OOD improvement ($T=0 \to 1$) }}
\end{minipage}
\caption{Performance of Hypermodels in a dynamic setting with varying prior scale [eICU data with clustering bias]. We can see the prior scale $100$ is best for the in-distribution data performance [plot (a)] while, prior scale of $10$ is the best for the OOD data performance [plot (b)] at $T=0$. In plot (d), we can see that the performance improvement is maximum for the prior scale of $1$, while it is least for prior scale of $10$, showcasing the trade-off between sharper posteriors and OOD performance. 
}
\label{fig:Hypermodel_prior_scale}
\end{figure}

As referred to in Section~\ref{sec:eval_posterior_consis}, hyperparameter tuning suffers in dynamic settings as well. 
In  Figures~\ref{fig:Ensemble+_weight_decay} 
to~\ref{fig:Hypermodel_prior_scale}, we demonstrate this for \ensembleplus, Epinets, and Hypermodels with hyperparameters weight decay and prior scale. We consider the eICU dataset with clustering bias in a dynamic setting for these experiments.
 Figure \ref{fig:Ensemble+_weight_decay}  shows that at $T=0$, weight decay that works the best for the \ensembleplus for the in-distribution data becomes worse related to other values for the 
out-of-distribution data, which  signifies the trade-off between in-distribution performance and out-of-distribution performance. 
Figure~\ref{fig:Ensemble+_weight_decay} (d) also shows  the performance improvement after acquiring new data.
As we can see, the weight decay for which the OOD performance was the best has the least performance improvement.
This shows the trade-off between having sharp posteriors and the OOD performance when we optimize for the hyperparameter weight decay. 
Similarly, we can see for the other models (Hypermodels and Epinets) this trade-off 
for the weight decay and prior scale hyperparameter.
We can also see that in some settings, performance even gets deteriorated (see Figures~\ref{fig:Epinet_weight_decay} 
and~\ref{fig:Epinet_prior_scale}) as we acquire more data points. 
The performance deterioration might be due to the following reason: UQ methodologies starts from a prior belief with lower level of the uncertainty and as the agents acquire more data at ($T=1$),
they become more uncertain about the outcomes as compared to the previous status ($T=0$). 
 

\subsubsection{Sensitivity to inference seeds and early stopping time}

\begin{figure}[h]
\centering
\begin{minipage}[b]{0.40\textwidth}
\centering
\includegraphics[width=\textwidth, height=5cm]{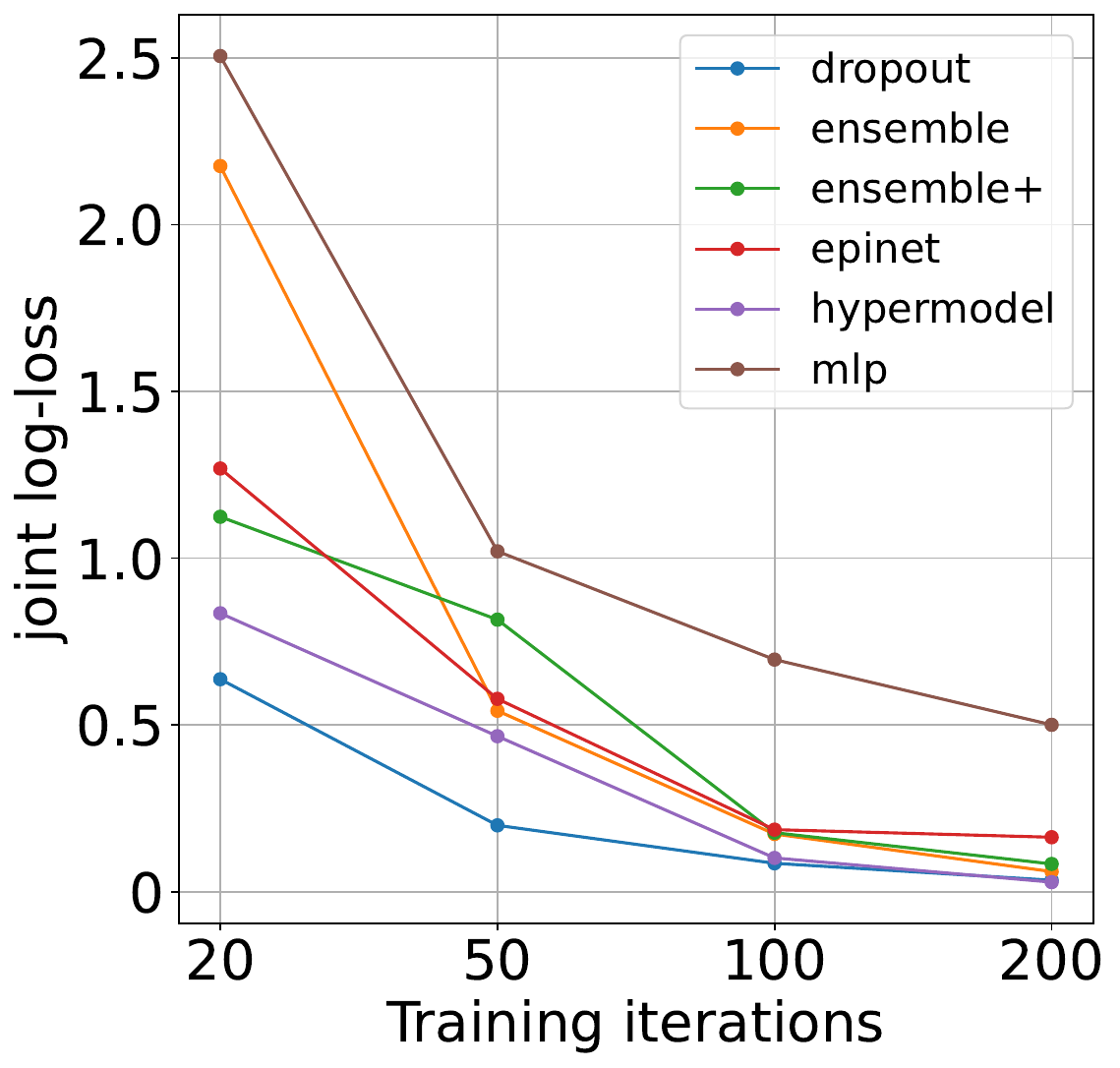}
\caption{Performance of UQ methodologies deteriorates with early stopping - Joint log-loss for in-distribution (ID) data - (eICU data with clustering bias)}
\label{fig:eicu-clustering-task-4-k-0}
\end{minipage}
\hfill
\begin{minipage}[b]{0.40\textwidth}
\centering
\includegraphics[width=\textwidth, height=5cm]{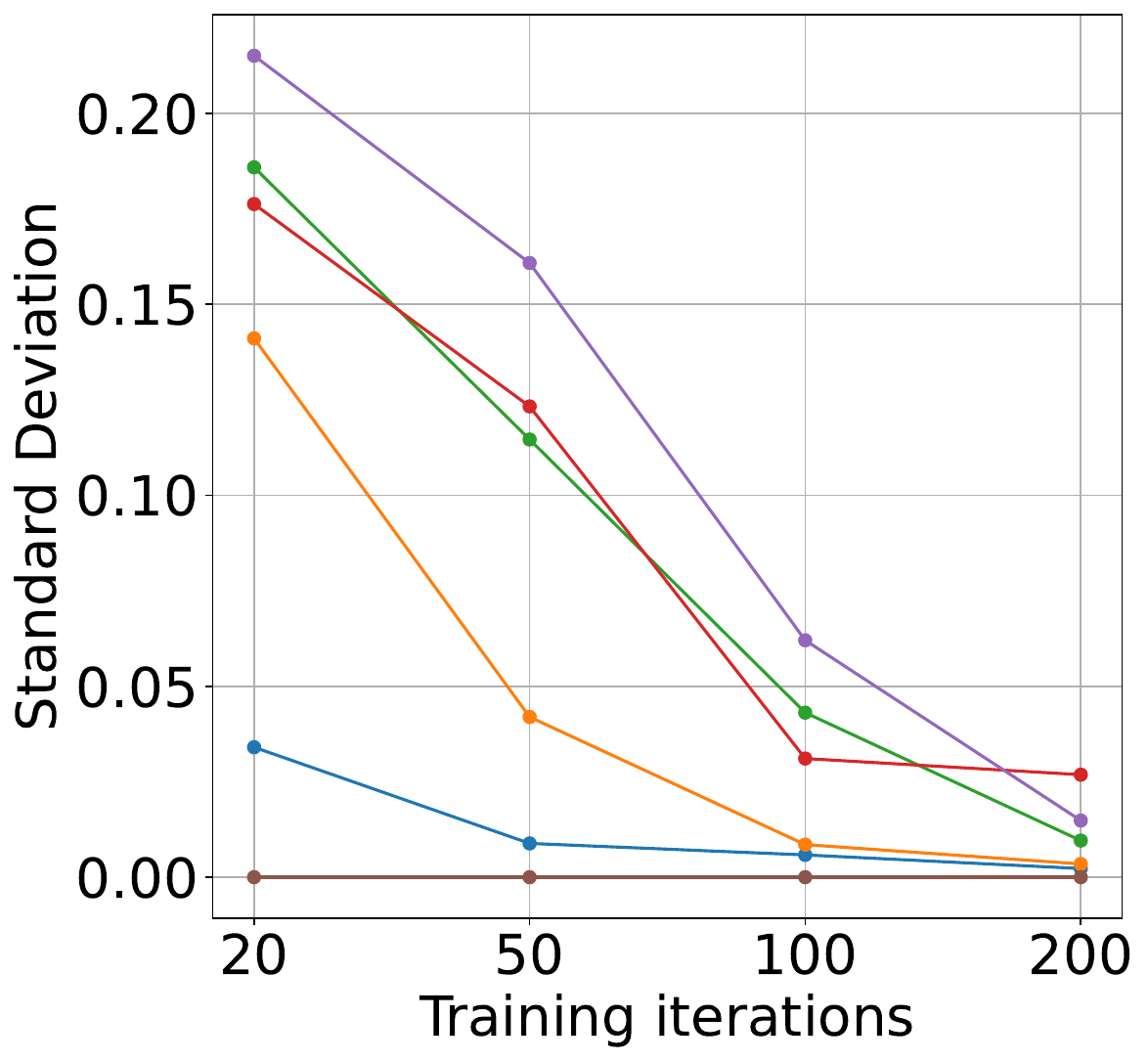}
\caption{Sensitivity of UQ methodologies to random inference seeds - Standard deviation of joint log-loss for in-distribution (ID) data - (eICU data with clustering bias)}
\label{fig:eicu-clustering-task-2-tau10-id}
\end{minipage}
\end{figure}

\begin{figure}[h]
\centering
\begin{minipage}[b]{0.40\textwidth}
\centering
\includegraphics[width=\textwidth, height=5cm]{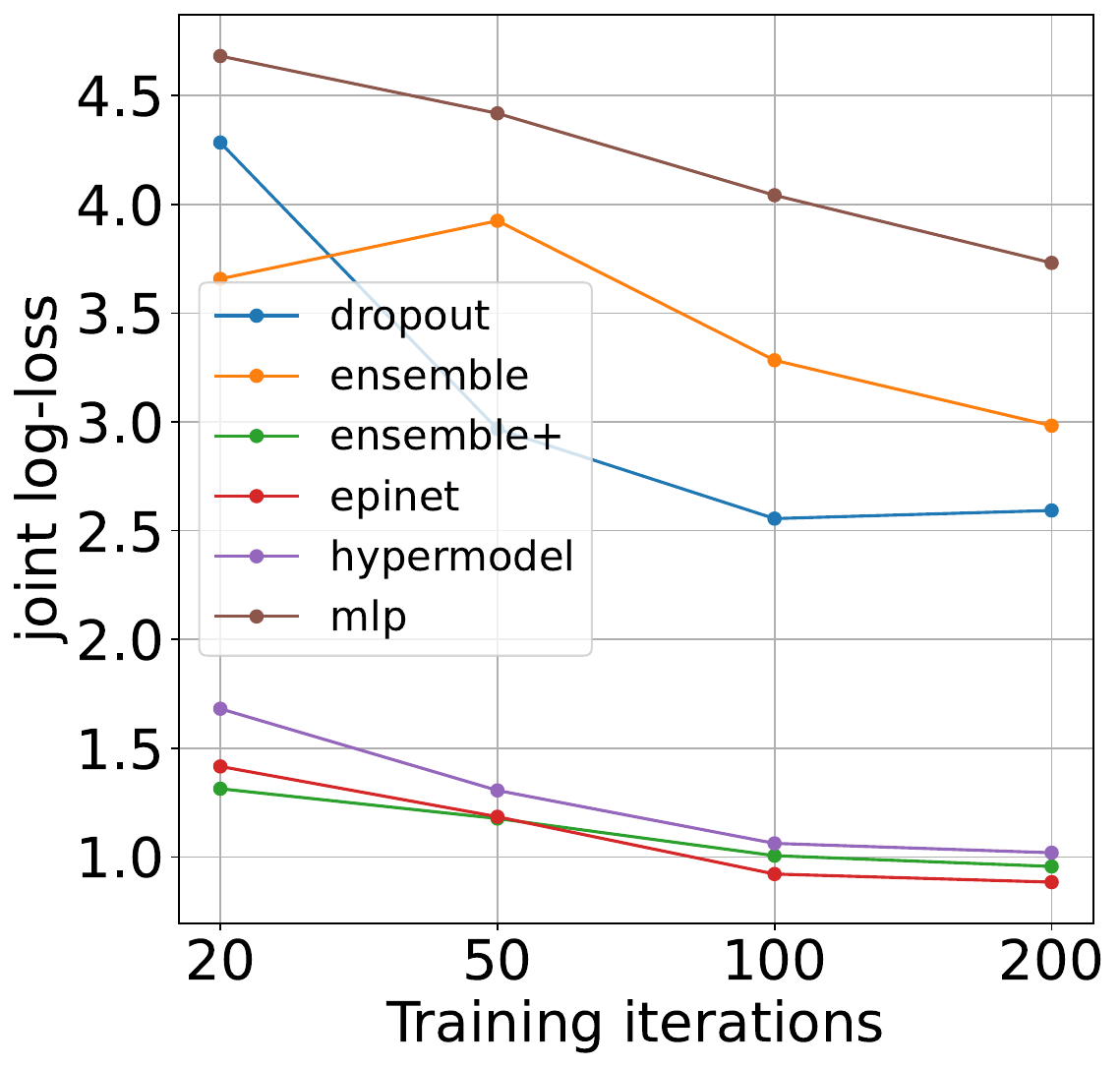}
\caption{Performance of UQ methodologies deteriorates with early stopping - Joint log-loss for out-of-distribution (OOD) data  - (eICU data with clustering bias)}
\label{fig:eicu-clustering-task-4-k-0-ood}
\end{minipage}
\hfill
\begin{minipage}[b]{0.40\textwidth}
\centering \includegraphics[width=\textwidth, height=5cm]{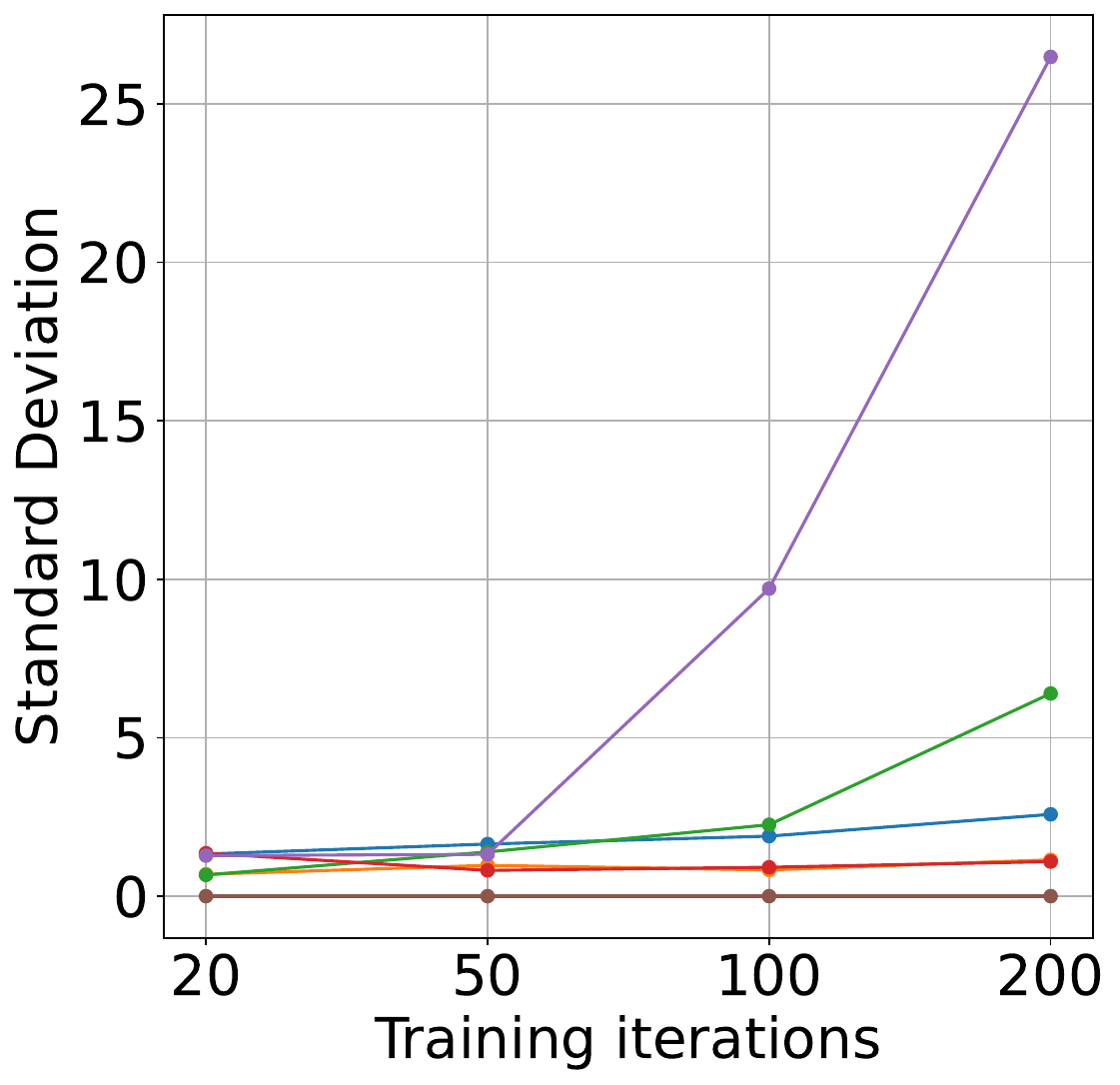}
\caption{Sensitivity of UQ methodologies to random inference seeds - standard deviation of Joint log-loss for out-of-distribution (OOD) data - (eICU data with clustering bias)}
\label{fig:eicu-clustering-task-2-tau10-ood}
\end{minipage}
\end{figure}

In Figures~\ref{fig:eicu-clustering-task-4-k-0} to~\ref{fig:eicu-clustering-task-2-tau10-ood}, we demonstrate the effect of early stopping and random seeds on the posterior inference of different 
 UQ methodologies. 
 We conduct these experiments for the eICU dataset with clustering bias. 
 In Figures~\ref{fig:eicu-clustering-task-4-k-0} to~\ref{fig:eicu-clustering-task-2-tau10-ood}, 
 we   early stop the training of different UQ methodologies and see its impact on the performance on in-distribution data and out-of-distribution data. Figure~\ref{fig:eicu-clustering-task-4-k-0} demonstrates the effect of early stopping on the in-distribution performance. 
 We can see that the performance deteriorates. Similar deterioration can be seen on the out-of-distribution performance with early stopping (Figure~\ref{fig:eicu-clustering-task-4-k-0-ood}).
To measure the sensitivity of different methodologies to the random inference seeds, we take the variance of inference (joint log-loss) across 10 different seeds - within each seed we sample 20 instances of the posterior models and evaluate it on 100 test samples.
 In Figure~\ref{fig:eicu-clustering-task-2-tau10-id} we showcase the sensitivity of the inference on in-distribution data due to different random seeds for different methodologies. 
 We can see that as we stop earlier, the sensitivity (standard deviation) of the inference (joint log-loss) increases for the in-distribution. 
 Moreover, the sensitivity is larger for \ensembleplus, Epinets and Hypermodels as compared to other methodologies. 
 An important thing to note here is that for Ensemble and \ensembleplus models, we had $100$ particles and in principle we can derive the posterior exactly without any Monte Carlo approximation and then the standard deviation will reduce to 0 (same as MLP) for these models. However, in current experiments, as specified earlier, we sample $20$ instances of the posterior from these $100$ particles. For agents such as Hypermodels and Epinets, we have to do   Monte Carlo approximation as the reference distributions for index $z$ follows a Gaussian distribution.  
 Figure~\ref{fig:eicu-clustering-task-2-tau10-ood} reports the sensitivity of the inference on out-of-distribution data due to different random seeds for different methodologies. 
 In this case, as we train more, sensitivity increases for some models such as hypermodels and ensemble $+$.
 This seems to be an effect of the performance deterioration on the OOD data as we saw earlier in Figure~\ref{fig:difficult_to_choose_stopping_time} in Section~\ref{sec:hyperparameter_tuning_breaks_under_dis_shifts_experiment_simple},
 which manifests itself in larger variance across different seeds as well.


\end{appendix}

\ifdefined\useorstyle




\else
\newpage
\appendix

\fi


\end{document}